\documentclass[letterpaper,journal]{IEEEtran}

\usepackage{amssymb}
\usepackage{amsmath}
\usepackage{graphicx}
\usepackage{float}
\usepackage[figuresright]{rotating}
\usepackage{booktabs}
\usepackage[caption=false,font=footnotesize]{subfig}
\usepackage{cite}
\usepackage{bibunits}
\usepackage{stfloats}
\usepackage{xurl}
\usepackage[hidelinks]{hyperref}

\newcounter{pipelinestep}
\newcommand{\pipelineparagraph}[1]{%
  \refstepcounter{pipelinestep}%
  \paragraph*{\thepipelinestep) #1}%
}

\usepackage{array}

\begin{document}

\title{Visibility-Aware Reference Generation for Benchmarking Visual 3D Reconstruction in Vegetation-Rich Environments}

\author{Zhipeng Zhong, Matthew Coombes, and Cunjia Liu%
\thanks{The authors are with Department of Aeronautical and Automotive Engineering, Loughborough University, Leicestershire, UK, LE11 3TU.
Email: Z.Zhong@lboro.ac.uk, C.Liu5@lboro.ac.uk}}

\maketitle

\makeatletter
\global\@topnum=0\relax
\makeatother
\begin{figure}[t]
    \centering
    \includegraphics[width=\linewidth]{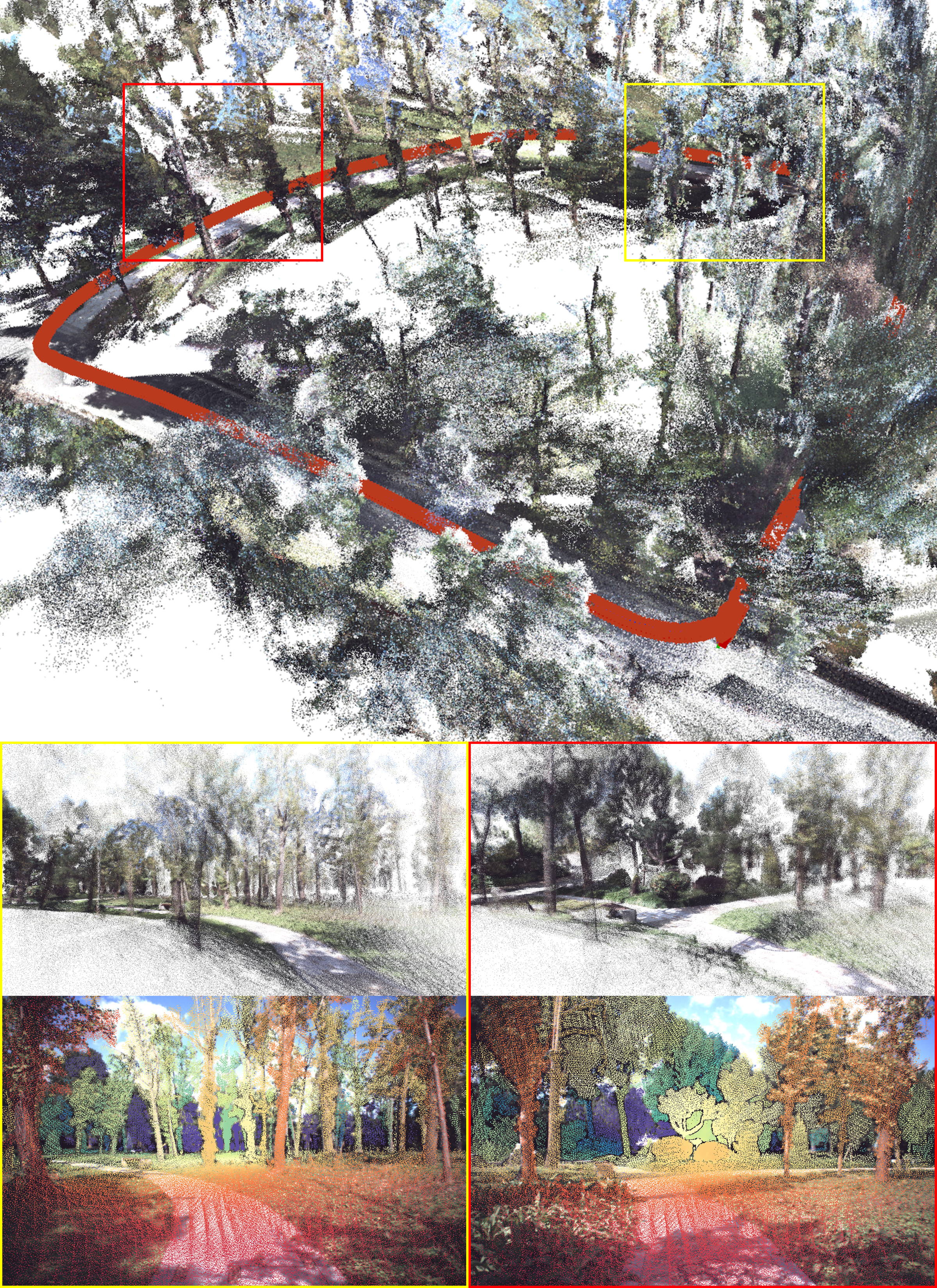}
    \caption{Visualization of the reference data generated 
    by the proposed pipeline in BotanicGarden. 
    Top: the accumulated scene-level 
    reference point cloud and the camera trajectory shown in red. 
    The colored boxes indicate the two individual-view examples below. 
    Bottom: for each example, the upper row shows the corresponding 
    visibility-filtered point cloud in the camera frame, 
    and the lower row shows the projected sparse reference depth image.
    }
    \label{fig:final-colorized-whole-view}
\end{figure}

\begin{abstract}
Reliable evaluation of visual 3D reconstruction and depth estimation in vegetation-rich outdoor environments is limited by the scarcity of geometric ground-truth data and the cost of complete scanner surveys. We present a LiDAR-based reference-generation pipeline for evaluating the zero-shot performance of feed-forward visual-geometry models alongside conventional structure-from-motion (SfM) methods. 
The pipeline reuses camera images, LiDAR scans, and reference poses from existing multi-sensor localization datasets to generate per-view depth maps and visible reference point clouds. To address visibility errors in accumulated LiDAR scans, we introduce the Large-Depth-Gap Visibility Score (LDGVS) for point-wise visibility filtering and automatic scale selection in Hidden Point Removal (HPR). 
Experiments on TartanAirV2 and TartanGround show that, compared with Poisson surface reconstruction, which is an approach suited to completed scan surveys, the combined HPR and LDGVS filtering retains approximately three times the point density at comparable absolute relative error. We also apply the pipeline to the BotanicGarden dataset to systematically compare depth estimation and 3D reconstruction methods. Feed-forward methods produce dense depth predictions efficiently, but their 3D reconstructions can contain duplicated structures. Conventional SfM generally produces more coherent geometry, but its reconstructions remain sparse and more prone to failure in challenging vegetated scenes. The benchmark toolkit and datasets will be released.
\end{abstract}


\section{Introduction}
\label{sec:introduction}

Visual depth estimation and 3D reconstruction are essential capabilities for agricultural and other field robots, providing dense geometric information from lightweight, low-cost cameras. Many studies \cite{A-survey-on-3D-reconstruction-techniques-in-plant-phenotyping, A-Systematic-Review-AI-driven-3D-point-cloud-analysis-in-plant-phenotyping, A-review-fruit-3D-reconstruction, review-Image-Based-3D-reconstruction-for-smart-farming-Applications} 
have sought to improve visual geometric methods in vegetation-rich outdoor environments, including forests, orchards, crop fields, and botanical gardens. 
However, agricultural applications continue to rely predominantly on conventional geometry-based SLAM \cite{greenhouse-vegetables-semantic-3R, apple-orchards-RGBD-SLAM} or SfM \cite{3R-unstructured-citrus-orchards, 3R-trees-SfM}, while the feed-forward methods \cite{3R-trees-SfM, ET-PatchNet, MT-MVSNet} have been less widely adopted. This pattern contrasts with general-purpose 3D-reconstruction and depth-estimation benchmarks \cite{KITTI, DTU, Tanks-and-Temples, ETH3D, Oxford-Spires-Dataset}, where feed-forward models are increasingly prominent\cite{Dust3r, VGGT, VGGT-Omega, DA3, MASt3R, MapAnything, MoGe}. 
This motivates us to evaluate their zero-shot performance in vegetation-rich outdoor environments.

However, benchmarking in these environments remains challenging because publicly available datasets with suitable ground-truth (GT) geometry are scarce and acquiring such data is time-consuming and labor-intensive. A standard approach is to use high-accuracy terrestrial laser scanners to capture complete and dense scans of the environment\cite{ETH3D, Oxford-Spires-Dataset}. This process, however, requires professional surveying expertise and substantial manual effort for data acquisition, scan registration, and outlier removal. These requirements make complete surveys particularly costly in extensive and structurally complex vegetation-rich outdoor environments. A few studies have used laser scanners 
\cite{3R-trees-SfM, Viewing-the-Forest-in-3-D, CaR3DMIC} to 
obtain GT geometry  in such environments, but the resulting datasets are often limited in scale and not fully publicly available.

To address this gap, we investigate repurposing mobile LiDAR measurements from existing multi-sensor localization datasets as geometric references. Mobile LiDAR enables data collection over large areas, and a growing number of datasets provide camera images, LiDAR scans, and reference trajectories in agricultural and other vegetation-rich outdoor environments, as summarized in Table~\ref{tab:multi-sensor-datasets}. Although these datasets contain large-scale observations of complex scenes, they were developed primarily for localization and mapping rather than for benchmarking visual depth estimation or 3D reconstruction. Consequently, they do not directly provide camera-aligned reference depth or visibility-filtered 3D geometry. Previous studies have demonstrated the use of LiDAR-derived references for depth estimation and 3D reconstruction in urban and road environments~\cite{DSEC, KITTI, Waymo, nuScenes, VGGT-Long, SurroundNEXO}, but extending this approach to vegetation-rich environments remains challenging. The difficulty arises because mobile LiDAR measurements are sparse and nonuniform and may be affected by motion distortion. Accumulating multiple scans improves point density but can introduce visibility errors caused by incomplete observations, occlusions, and camera--LiDAR parallax.

These limitations raise a central question: how can mobile LiDAR data be converted into reliable camera-view geometric references for benchmarking visual depth estimation and 3D reconstruction in vegetation-rich environments? We therefore develop a reference-generation pipeline and apply it to the representative BotanicGarden\cite{BotanicGarden} dataset, where the generated references are used to compare feed-forward visual-geometry models with conventional SfM methods.
Our contributions are summarized as follows:
\begin{enumerate}
    \item We propose a pipeline that converts camera, LiDAR, and reference-pose data from existing localization datasets into per-view depth and visible point-cloud references.

    \item We introduce the Large-Depth-Gap Visibility Score (LDGVS) for point-cloud visibility filtering and automatic Hidden Point Removal (HPR) scale selection, providing greater reference density than Poisson surface reconstruction at comparable absolute relative error.

    \item With the proposed pipeline, we construct a BotanicGarden benchmark comparing feed-forward visual-geometry models with conventional SfM methods in geometric quality, metric scale, cross-frame consistency, robustness, and efficiency.
\end{enumerate}

\begin{table}[t]
\centering
\scriptsize
\renewcommand{\arraystretch}{1.15}
\caption{Multi-sensor localization datasets in vegetation-rich outdoor environments.}
\label{tab:multi-sensor-datasets}
\begin{tabular}{@{}p{\dimexpr0.30\columnwidth-2\tabcolsep\relax}p{\dimexpr0.37\columnwidth-2\tabcolsep\relax}p{0.33\columnwidth}@{}}
\toprule
Dataset & GT Trajectory & Data Scale \\
\midrule
SFU-Mountain \cite{SFU-Mountain} & Manually aligned locations & $>$8 h, 150 GB \\
RELLIS-3D \cite{RELLIS-3D} & VoxNet poses & 5 seq., 13.6k scans \\
Wild-Places \cite{Wild-Places} & Wildcat SLAM poses & 8 seq., 63k submaps \\
TreeScope \cite{TreeScope} & Faster-LIO poses & 6 sites, $>$1.8k labels \\
CitrusFarm \cite{CitrusFarm} & GPS-RTK & 7 seq., 1.7 h, 1.3 TB \\
ARD-VO \cite{ARD-VO} & GPS-RTK & 11 sessions, $>$1 TB \\
MAgro \cite{MAgro} & GPS-RTK & Multi-season sequences \\
BLT \cite{BLT} & RTK-GNSS & 11 sessions, 600 GB \\
TAIL \cite{TAIL} & RTK-GNSS & 14 sequences \\
AgriGS-SLAM \cite{ArgiGS-SLAM} & VIO-GNSS-RTK poses & 2 orchards \\
HortiMulti \cite{HortiMulti} & Surveyed GT and AprilTags & 20 seq., $\sim$11 km \\
BotanicGarden \cite{BotanicGarden} & registration in scanner map & 33 seq., 17.1 km \\
\bottomrule
\end{tabular}
\end{table}

\section{Related Work}
\label{sec:related}

\subsection{Agricultural Datasets for 3D Reconstruction and Depth Estimation}

Many studies have investigated 3D reconstruction using agricultural data.
Some studies use RGB-D cameras to obtain GT data in apple orchards \cite{3R-Information-Fusion},
a walnut-tree scene \cite{branches-3R-NeRF}.
some studies evaluate performance through comparisons with SOTA methods \cite{MT-MVSNet, ET-PatchNet},
whereas others rely solely on visualization \cite{greenhouse-vegetables-semantic-3R}.
Several studies evaluate their results using manual measurements
\cite{apple-orchards-RGBD-SLAM, 3D-Reconstruction-and-Information-Fusion-Orchard}.
Laser scanners are more commonly used to provide more accurate and denser GT data for
tree reconstruction \cite{3R-trees-SfM, Viewing-the-Forest-in-3-D, CaR3DMIC}.
To our knowledge, no work has used LiDAR data as GT for 3D reconstruction or depth estimation in this type of environment.
Several LiDAR-based depth-estimation
benchmarks exist for urban and road environments \cite{DSEC, KITTI}. 
And some localization datasets \cite{Waymo, nuScenes} has been 
used for 3D reconstruction evaluation \cite{VGGT-Long, SurroundNEXO}.
Although many studies have explored this topic, 
the availability of datasets for large-scale 3D reconstruction and depth estimation 
in agricultural environment remains limited.

To filt this gap, we propose a LiDAR-based geometric reference-generation
pipeline for data collected directly by lightweight mobile robot platforms.
This eliminates the need for dense, exhaustive scanning, 
thereby enabling the cost-effective creation of large-scale 3D reconstruction
and depth estimation datasets.

\subsection{Point-Cloud Visibility Analysis for Dataset Consturction}
\label{subsec:point cloud visibility analysis}

For fair comparison, we only can compare the point cloud that can be seen 
in image data. Therefore the key problem of datasets for 3D reconstruction and 
depth estimation using point cloud data is to get correct visible the point-cloud. Specifically,
ETH3D and ScanNet++ datasets \cite{ETH3D, SCANNET++} use
Poisson surface reconstruction on laser-scan point
clouds \cite{Poisson-surface-reconstruction} to render depth rather than 
directly using projected point clouds.
More loosely, splatting-based methods \cite{AdaSplats, DOC-Depth}  render a primitive projection.
These approach can work well when dense and complete geometry is available, 
but sparse LiDAR scans in vegetation-rich scenes often contain missing areas and
large depth discontinuities. 
In such cases, surface reconstruction may fail to represent the scene. 
Another approach relies on consistency \cite{KITTI} or
depth differences \cite{Surface-Geometry-Model-for-Depth} 
between estimated depth and projected point-cloud depth. 
These methods can work well in urban and road environments, 
but depth estimates are often inaccurate in vegetation-rich environments.
Consequently, these methods may also filter out too many visible points. 
Hidden Point Removal (HPR) 
\cite{Direct-visibility-of-point-sets} analyzes visibility on the point cloud without first
reconstructing a surface. However, we found that
HPR cannot be applied directly to environments with large depth variations.
\cite{Visibility-Estimation} proposed visibility score can be computed directly in image space.
However, this work only evaluate their method on two manual label points of city scene, can not 
generalize well in our experiments.

Instead, in this work, we propose a modular point-cloud visibility-analysis 
framework that introduces a large-depth-gap visible score
and uses it to adaptively select the HPR radius,
 enabling robust removal of occluded LiDAR points in sparse, 
 unevenly sampled scenes with large depth variations.

\subsection{3D Reconstruction and Depth Estimation in Vegetation-rich and Agricultural Scenes}

Recent reviews \cite{A-survey-on-3D-reconstruction-techniques-in-plant-phenotyping, A-Systematic-Review-AI-driven-3D-point-cloud-analysis-in-plant-phenotyping, A-review-fruit-3D-reconstruction, review-Image-Based-3D-reconstruction-for-smart-farming-Applications} 
indicates that most research on agricultural applications still based on traditional SfM or SLAM.
For examples, some studies integrate existing vSLAM frameworks with semantic-segmentation networks for semantic 3D reconstruction of orchards
\cite{greenhouse-vegetables-semantic-3R, 3R-unstructured-citrus-orchards, apple-orchards-RGBD-SLAM}.
SfM is the most widely used method for 3D reconstruction in plant phenotyping.
\cite{3D-reconstruction-grape-bunches,Monitoring-vertical-SPAD-distribution-SfM-MVS}. 
There is also work on applying the improved SFM to agricultural scenarios, such as detector-free\cite{3R-trees-SfM}
and deep learning matcher\cite{Plant3R, MASt3R-SfM}.

On the other hand, due to the emergence of visual self-supervised large models \cite{DINOv3}
these emerging feed forward methods show a very strong generalization across different 
scenes and benchmarks 
\cite{VGGT, DA3, FoundationStereo, MapAnything, VGGT-Omega}. 
However, These methods have not yet been fully benchmarked in agricultural scenarios against 
traditional SfM methods.

Therefore, we construct a comprehensive
3D-reconstruction and depth estimation benchmark to evaluate the performance of SfM and feed forward methods in large-scale vegetation-rich and agricultural environments.
We also integrate recent state-of-the-art learning-based feature and matching models\cite{LoFTR, RoMav2, LoMa} into an SfM pipeline and compare their performance within this benchmark.

\section{Pipeline for Generating 3D Reconstruction and Depth Estimation Reference Data}
\label{sec:Pipeline for Generating 3D Reconstruction and Depth Estimation reference}

\begin{figure*}[!t]
    \centering
    \includegraphics[width=\textwidth]{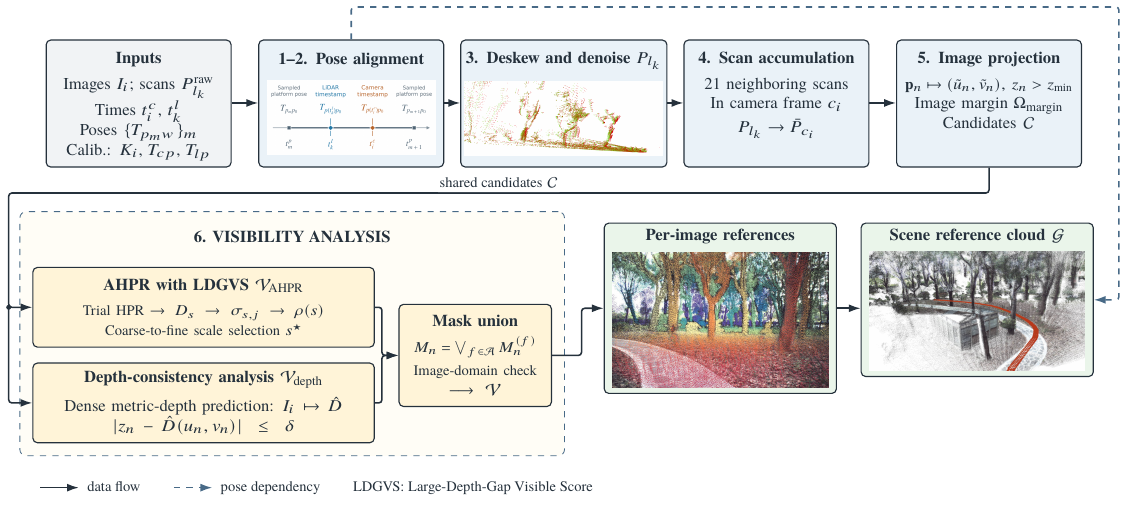}
    \caption{Overview of the LiDAR-based reference-generation pipeline. In the ``Deskew and denoise'' step, the retained point cloud is shown in green and the raw point cloud in red.}
    \label{fig:reference-generation-pipeline}
\end{figure*}

We demonstrate how LiDAR data can be used to generate reference data for 3D
reconstruction and depth estimation. Figure~\ref{fig:reference-generation-pipeline} shows the complete workflow and
its principal mathematical quantities.

We assume a static scene observed by timestamped camera frames
$\{(I_i,t_i^c)\}_i$ and LiDAR scans $\{(P^{\rm raw}_{l_k},t_k^l)\}_k$, with
camera intrinsics $K_i$, nominal sensor extrinsics $T_{cp}$ and $T_{lp}$, and a
platform-pose trajectory $\{T_{p_m p_0}\}_m$ that is either provided by the
dataset or estimated using multi-sensor SLAM\@. From these inputs, the pipeline
computes time-aligned camera and LiDAR poses and generates
visibility-filtered reference point clouds, per-image sparse reference depth
maps, and a scene-level reference point cloud, as illustrated in
Fig.~\ref{fig:final-colorized-whole-view}. Throughout this section, the indices
$i$, $k$, $m$, $n$, and $j$ denote camera frames, LiDAR scans, platform-pose
samples, 3D points, and occupied image pixels, respectively.

\pipelineparagraph{Platform Pose Trajectory $\{T_{p_m p_0}\}_m$}
\label{par:Transformation pose-to-world}

We assume GT poses $T_{p_m w}$ has been given, as shown in
Table~\ref{tab:multi-sensor-datasets}. The GT poses can get from
GPS, survey device, or multi-sensor fusion SLAM.
We then express every $T_{p_m w}$ in the $p_0$ coordinate
frame to obtain the reference poses
$T_{p_m p_0}=T_{p_m w}(T_{p_0 w})^{-1}$.

\pipelineparagraph{LiDAR Pose $T_{l_k p_0}$ and Camera Pose $T_{c_i c_0}$}
\label{par:lidar-camera-pose}

To achieve perfect sensor synchronization,
we use screw linear interpolation (ScLERP)
to obtain a platform pose $T_{p(t')p_0}$ at each sensor query time $t'$.
$T_{cp}$ and $T_{lp}$ map
platform-frame coordinates into the camera and LiDAR frames, respectively.
Given neighboring trajectory samples $T_{p_m p_0}$ and $T_{p_{m+1}p_0}$ at
times $t_m^p\leq t'\leq t_{m+1}^p$,
\begin{equation}
    T_{p(t')p_0}
    =
    T_{p_m p_0}
    \exp\left(
    \alpha
    \log\left(
    T_{p_m p_0}^{-1}
    T_{p_{m+1}p_0}
    \right)
    \right),
\end{equation}
where \(\alpha=(t'-t_m^p)/(t_{m+1}^p-t_m^p)\),
\(\log(\cdot) : \mathrm{SE}(3) \rightarrow \mathfrak{se}(3)\),
and \(\exp(\cdot) : \mathfrak{se}(3) \rightarrow \mathrm{SE}(3)\).

We evaluate this interpolation at LiDAR time $t_k^l$ and camera time $t_i^c$
and apply the extrinsics $T_{lp}$ and $T_{cp}$ to obtain the time-aligned poses
$T_{l_k p_0}=T_{lp}\,T_{p(t_k^l)p_0}$ and
$T_{c_i p_0}=T_{cp}\,T_{p(t_i^c)p_0}$, respectively.
We then rebase every camera pose to $c_0$ using
$T_{c_i c_0}=T_{c_i p_0}\bigl(T_{c_0 p_0}\bigr)^{-1}$,
where $T_{c_0 p_0}=T_{cp}T_{p(t_0^c)p_0}$ is evaluated
at the first camera timestamp.

\pipelineparagraph{Point-Cloud Deskewing and Denoising $P_{l_k}$}
\label{par:Point cloud deskew and denoise}

A scanning LiDAR acquires each scan while the platform is moving, so its points
are expressed at slightly different sensor poses. Online SLAM systems commonly
remove this distortion using IMU-based motion prediction \cite{FAST-LIO}
or a velocity model \cite{KISS-ICP}. In our offline benchmark pipeline,
we instead use the GT trajectory from the preceding step.

Let the raw scan $P^{\rm raw}_{l_k}$ have a nominal time $t_k^l$, and let its
$n$th point $\mathbf p_{k,n}$ be acquired at time $t_{k,n}^l$. Define the
continuous-time LiDAR pose as
$T_{l(t)p_0}=T_{lp}T_{p(t)p_0}$. We transform every point to the LiDAR frame at
the nominal scan time to obtain the deskewed scan
 \begin{equation}
    \hat P_{l_k}
    =
    \left\{
    T_{l_k p_0}\bigl(T_{l(t_{k,n}^l)p_0}\bigr)^{-1}\mathbf p_{k,n}
    \ \middle|\ n=1,\ldots,N_k
    \right\}.
\end{equation}
As reported by \cite{ETH3D}, LiDAR scans can contain outliers caused by weak
returns and multipath reflections. We remove possible noisy samples using a range filter
and standard statistical outlier removal. In the following
steps, $P_{l_k}$ denotes this deskewed and denoised LiDAR scan.

\pipelineparagraph{Accumulated Point Cloud \(\bar{P}_{c_i}\) in Camera Coordinates}
\label{par:culmulated Point cloud in Camera Coordinates}

Using the time-aligned poses $T_{l_k p_0}$, each LiDAR scan
$P_{l_k}$ is transformed into camera frame $c_i$ as
$P_{c_i}^{(k)}=T_{c_i p_0}\bigl(T_{l_k p_0}\bigr)^{-1}P_{l_k}$. 
As a single scan is typically too sparse for point-cloud evaluation, 
we follow the approach in \cite{KITTI} 
but accumulate 21 neighboring LiDAR point clouds $P_{l_k}$ 
in the current camera coordinate frame $c_i$, and denote the 
accumulated point cloud by $\bar P_{c_i}=\bigcup_{k}P_{c_i}^{(k)}$.

\pipelineparagraph{Point-Cloud-to-Image Projection}
\label{par:Point-Cloud-Projection}

We convert the accumulated 3D points cloud $\bar P_{c_i}$ into a sparse depth image. For the current image
$I_i$, let $\bar P_{c_i}=\{\mathbf{p}_n\}_{n=1}^{N}$, where
$\mathbf{p}_n=(x_n,y_n,z_n)^{\mathsf T}$.
For each finite point with $z_n>z_{\min}$, we compute the continuous
image coordinates $\tilde{u}_n=f_x x_n/z_n+c_x$ and
$\tilde{v}_n=f_y y_n/z_n+c_y$. We then assign the point to its nearest
pixel center, $(u_n,v_n)=\operatorname{round}(\tilde{u}_n,\tilde{v}_n)$,
with footprint $\mathcal{F}_n^{\mathrm{round}}=\{(u_n,v_n)\}$.
When multiple points project to the same pixel, we retain the nearest
camera-frame $z$-depth:
$D_{\mathcal{F}}(u,v)=\min_{n:\,(u,v)\in\mathcal{F}_n}z_n$.
Here $\mathcal{F}_n$ denotes the selected projection footprint; for normal situation
it is the singleton round-projection footprint $\mathcal{F}_n^{\mathrm{round}}$. When greater image coverage
is required because the projected point cloud is too sparse for reliable depth
evaluation, We follow \cite{DOC-Depth} and project each point 
onto a circular footprint $\mathcal{F}_n^{\mathrm{circ}}$, 
which covers a circular pixel region rather than a single pixel.
This provides an alternative to round projection and 
produces a much denser reference depth map, but it can also introduce artifacts.

\pipelineparagraph{Visible-Point Index Set $\mathcal{V}$}
\label{par:Visibility Problem and Invisible Point-Cloud Removal}

As discussed in Section~\ref{subsec:point cloud visibility analysis},
identifying visible points is essential for fair evaluation of 3D reconstruction
and depth estimation. We propose a framework to obtain the visible-point index set
$\mathcal{V}$ using modules that produce $\mathcal{V}_{\rm AHPR}$ and
$\mathcal{V}_{\rm depth}$. Finally, we project the selected points and
apply the depth test to generate the corresponding sparse reference depth image.
Further details are provided in Section~\ref{sec:visibility-analysis}.

\section{Point-Cloud Visibility Analysis}
\label{sec:visibility-analysis}

\begin{figure}[t]
    \centering
    \subfloat[{Livox, single scan\label{fig:visible-problem-livox-c0}}]{%
\begin{minipage}[t]{0.24\textwidth}
\centering
        \includegraphics[width=\linewidth]{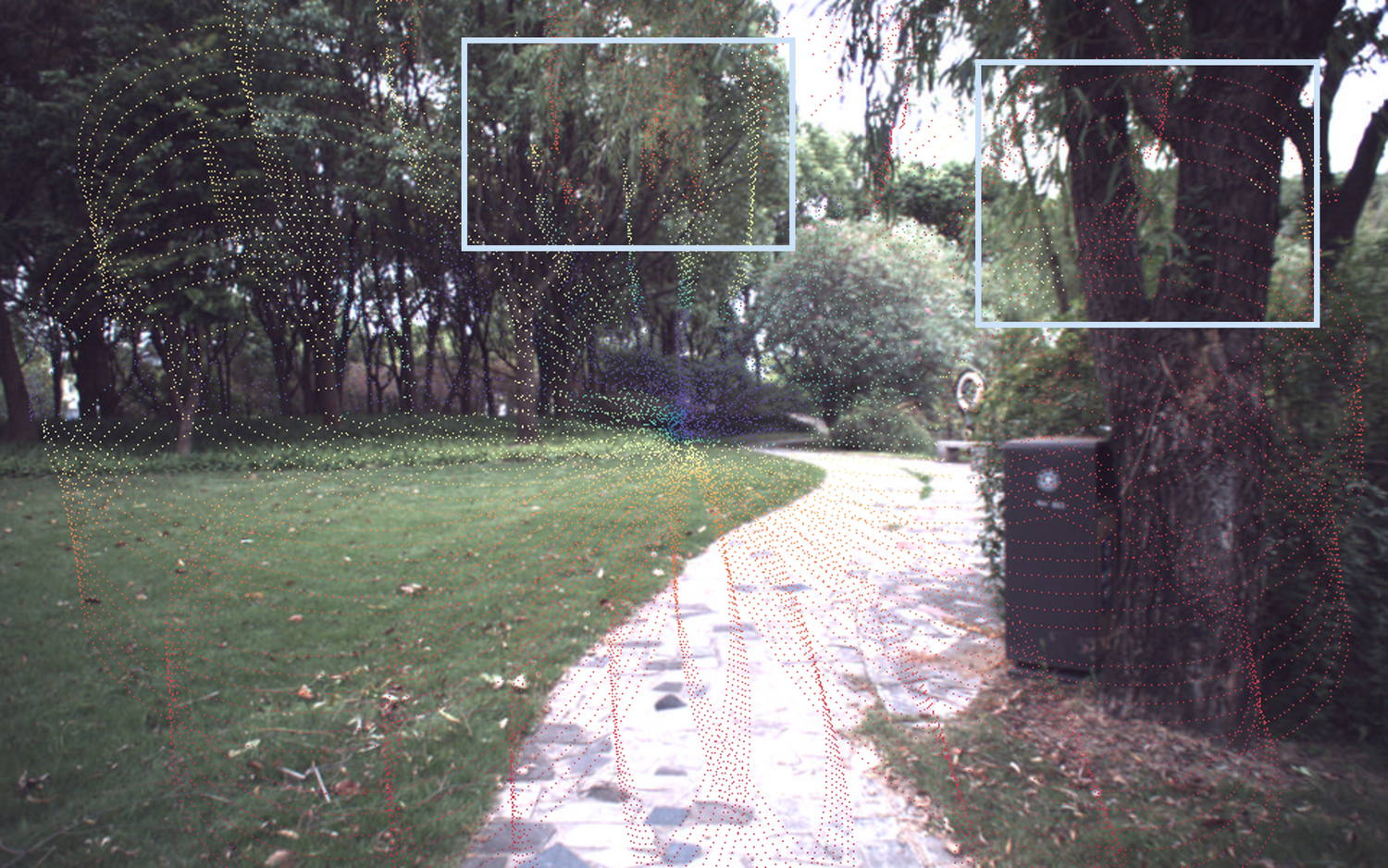}
\end{minipage}}\hfill
    \subfloat[{Livox, 10-scan accumulation\label{fig:visible-problem-livox-c10}}]{%
\begin{minipage}[t]{0.24\textwidth}
\centering
        \includegraphics[width=\linewidth]{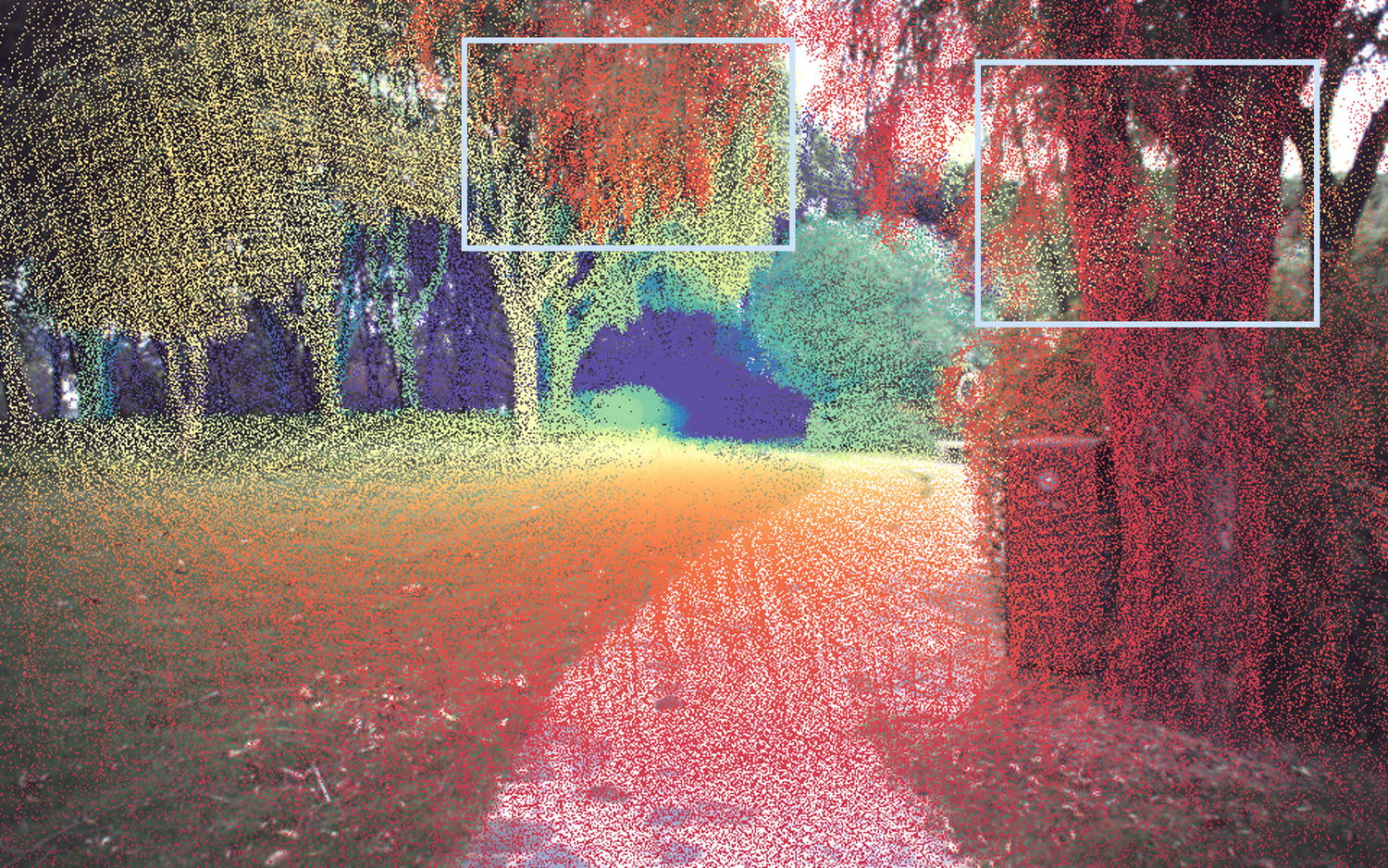}
\end{minipage}}

    \vspace{0.3em}

    \subfloat[{VLP-16, single scan\label{fig:visible-problem-vlp16-c0}}]{%
\begin{minipage}[t]{0.24\textwidth}
\centering
        \includegraphics[width=\linewidth]{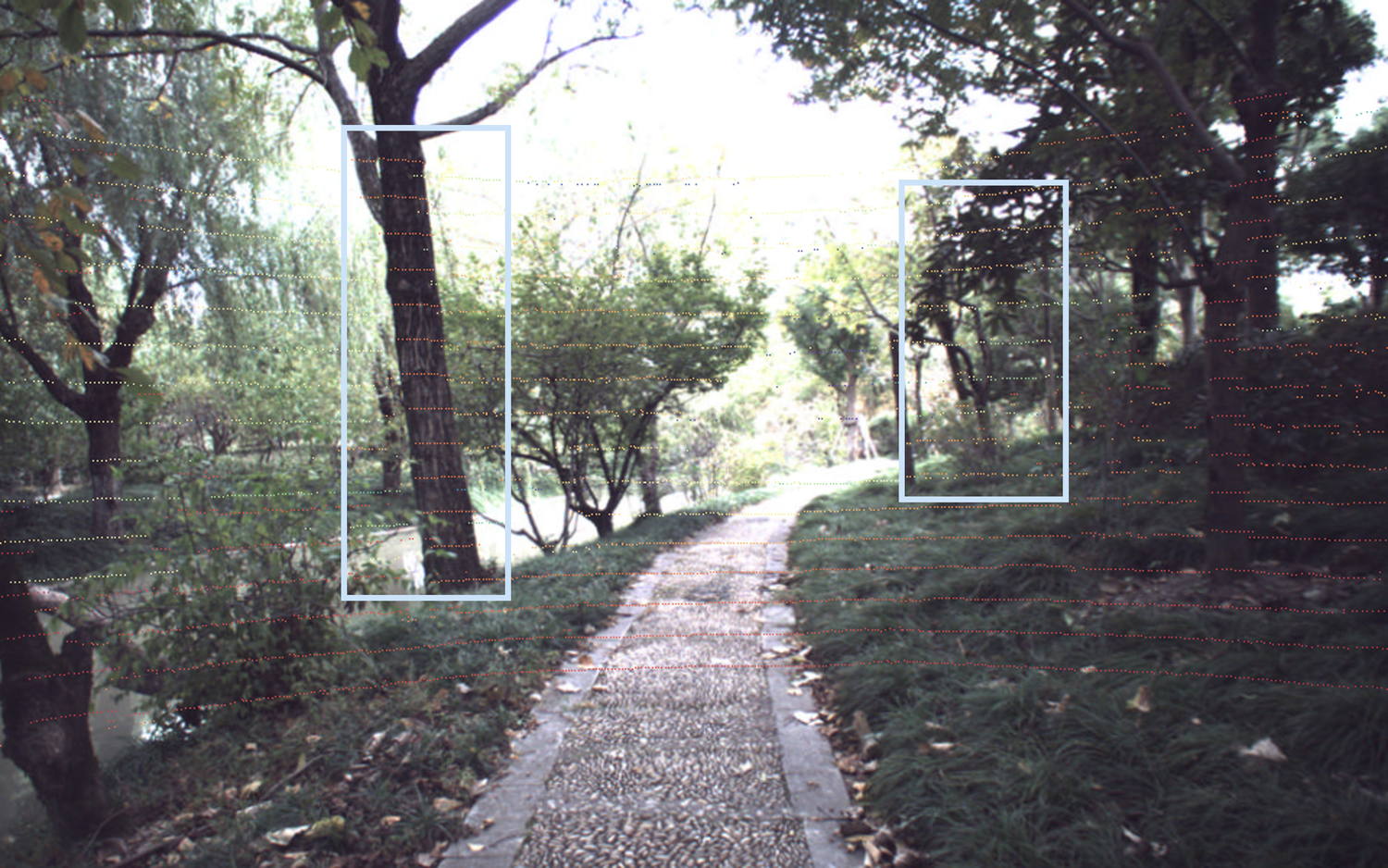}
\end{minipage}}\hfill
    \subfloat[{VLP-16, 10-scan accumulation\label{fig:visible-problem-vlp16-c10}}]{%
\begin{minipage}[t]{0.24\textwidth}
\centering
        \includegraphics[width=\linewidth]{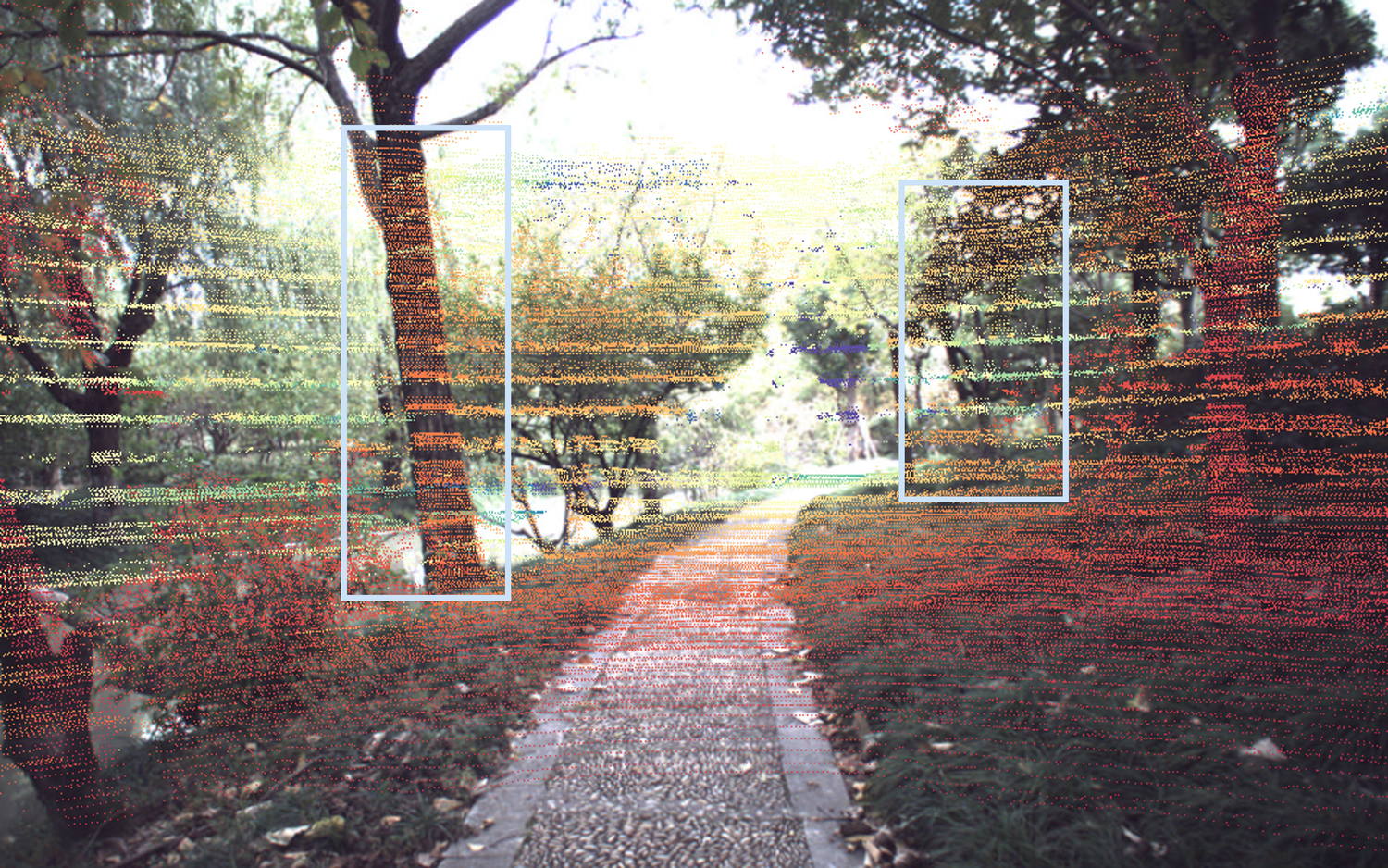}
\end{minipage}}

    \caption{Examples of the visibility problem in single-scan and accumulated
    LiDAR point clouds. Background points are projected onto foreground tree
    regions, producing incorrect occlusion relationships.}
    \label{fig:visible-problem}
\end{figure}

Using the entire accumulated point cloud for comparison is unfair because a
camera can reconstruct only the part of the scene that is visible from the
current view. Ideally, after projection and the depth test,
each image pixel should correspond to the nearest visible surface. However,
sensor parallax, point-cloud accumulation, and sparse or incomplete LiDAR
sampling can still produce incorrect occlusion relationships
\cite{KITTI, Surface-Geometry-Model-for-Depth}. The visibility problem is shown in
Fig.~\ref{fig:visible-problem}.

For all visibility filters, we first define a shared candidate point set:
\begin{equation}
    \mathcal{C} =
    \{n \mid z_n > z_{\min},\ \mathbf{p}_n\ \textrm{is finite},\
    (\tilde{u}_n,\tilde{v}_n)\in \Omega_{\mathrm{margin}}\},
    \label{eq:auto-candidate-set}
\end{equation}
where \(\mathbf{p}_n=(x_n,y_n,z_n)^\top\) denotes point \(n\) in the target
camera frame, \(\mathcal{C}\) is the index set of candidate points used by all
visibility filters, \(z_{\min}\) removes points too close to or behind the
camera, and \(\Omega_{\mathrm{margin}}\) is the enlarged image domain. The
enlarged domain retains a more complete point cloud during visibility analysis.

We treat occluded-point removal as a modular visibility-analysis problem.
Let \(\mathcal{A}=\{f_1,\ldots,f_{N_{\mathrm A}}\}\) denote the selected visibility
analysis modules, each producing a binary point mask
\(M_n^{(f)}\in\{0,1\}\) over the shared candidate
set \(\mathcal{C}\). To favor recall and retain a larger visible point
cloud, we combine these masks by union $M_n=\bigvee_{f\in\mathcal{A}} M_n^{(f)}$, where \( n\in\mathcal{C} \) 
and \(M_n^{(f)}=1\) indicates that module \(f\) considers point \(n\)
visible. The final visible set is
\begin{equation}
    \mathcal{V}
    =
    \{n\in\mathcal{C}
    \mid M_n=1,\ (u_n,v_n)\in\Omega_{\mathrm{image}}\}.
    \label{eq:visibility-final-visible-set}
\end{equation}
For image $I_i$, the resulting visibility-filtered cloud is
$\bar P^{\mathrm{vis}}_{c_i}=\{\mathbf p_n\in\bar P_{c_i}\mid n\in\mathcal V\}$.
Its sparse reference depth map $D_i$ is generated using the projection mode and
depth test defined in
pipeline step~\ref{par:Point-Cloud-Projection}.

A direct combination of existing techniques, such as Poisson surface
reconstruction and SGM-based depth-consistency filtering, can produce reliable
projected depths where the reconstructed geometry is accurate. However, these
methods cannot always faithfully reconstruct complex forest and agricultural
environments. Errors and missing structures in the reconstruction can therefore
cause valid projected depth samples to be incorrectly rejected, as demonstrated
by the experimental results in
Section~\ref{sec:visibility-analysis-experiments}.
To address this problem, we organize occluded-point removal into three
complementary visibility-analysis modules: (i) the Large-Depth-Gap Visible Score,
which evaluates local projected-depth discontinuities; (ii) Automatic Hidden
Point Removal (AHPR), which uses this score to select the HPR scale; and
(iii) Depth-Consistency Visibility Analysis, which checks projected points
against a dense depth estimate.

\subsection{Large-Depth-Gap Visible Score}
\label{par:large-depth-gap-visible-score}

\begin{figure}[t]
    \centering
    \subfloat[{Livox LiDAR\label{fig:local-nn-hist-livox}}]{%
\begin{minipage}[t]{0.49\linewidth}
\centering
        \includegraphics[width=\linewidth]{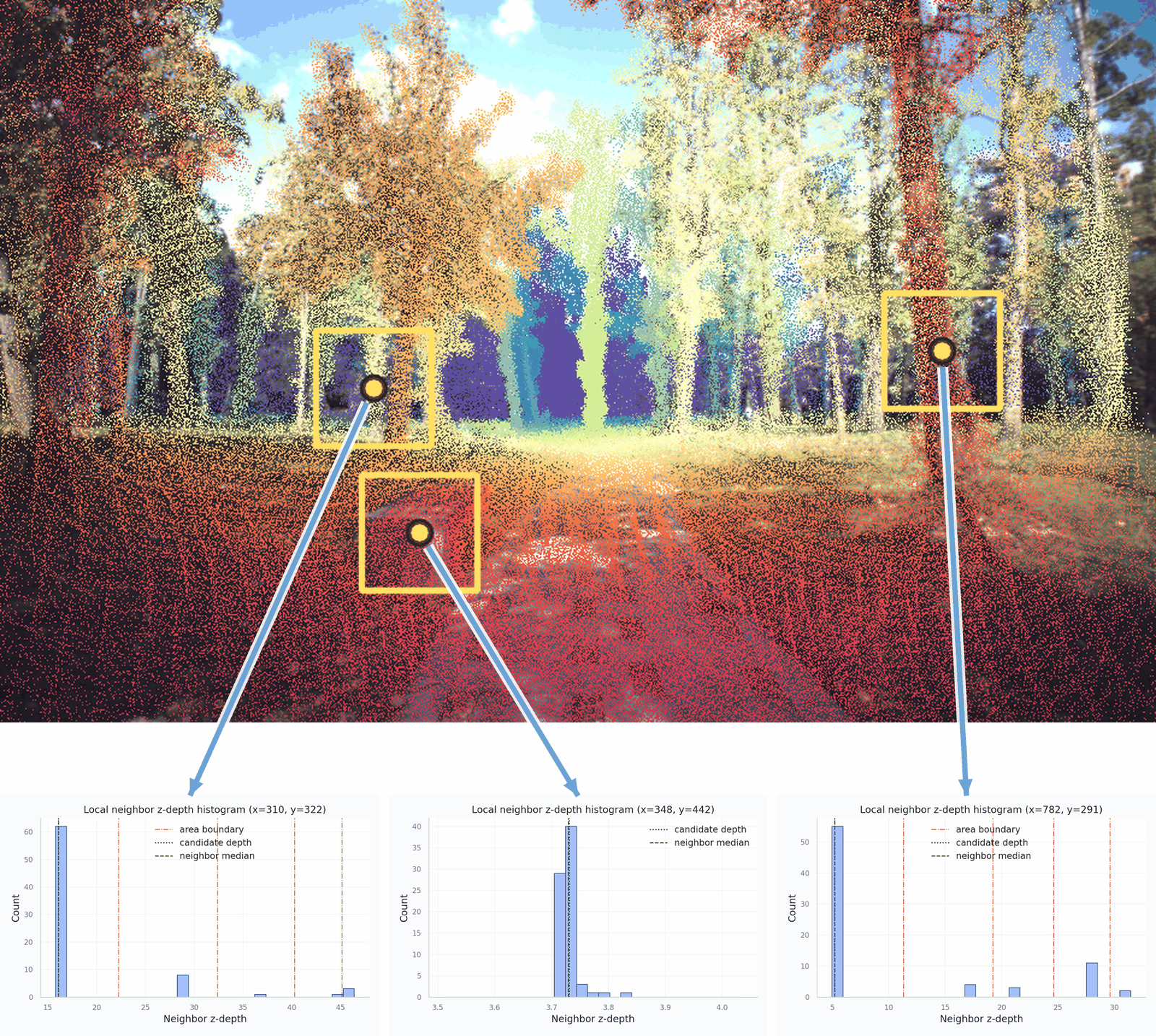}
\end{minipage}}
    \hfill
    \subfloat[{VLP-16 LiDAR\label{fig:local-nn-hist-vlp16}}]{%
\begin{minipage}[t]{0.49\linewidth}
\centering
        \includegraphics[width=\linewidth]{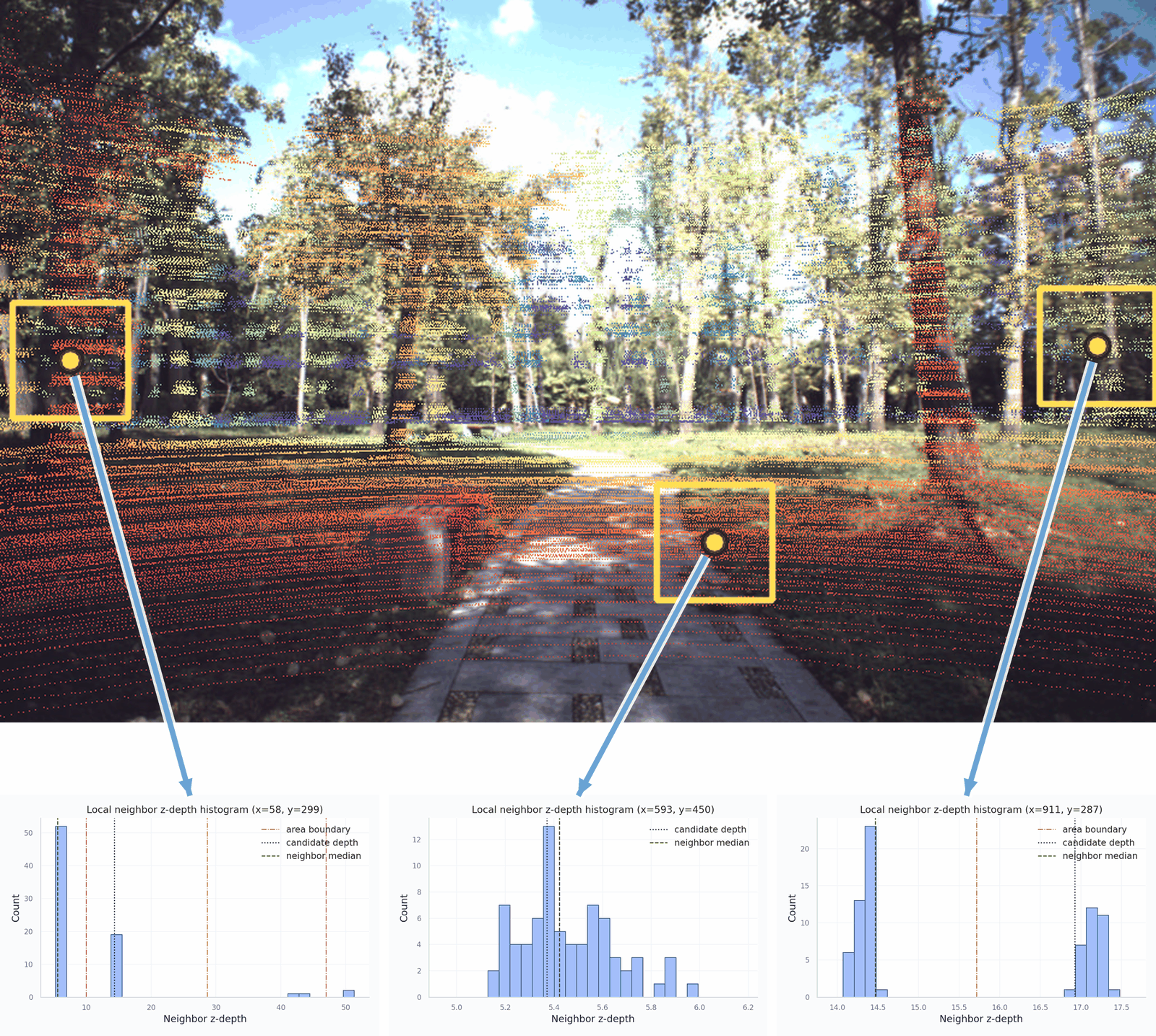}
\end{minipage}}

    \caption{Examples of local depth histograms constructed from the 75 nearest
    projected LiDAR samples. Correct projections, such as the middle example,
    produce no distinct depth gap, whereas incorrect projections, such as the left
    and right examples, produce a clear gap.}
    \label{fig:local-75-nn-hist}
\end{figure}

Visible score \cite{Visibility-Estimation} is a good indicates for a projected depth whether it is visible. 
Motivated by \cite{Visibility-Estimation}, we therefore revisit
the local distributions of neighboring depths. 
we observe that local neighborhoods
containing incorrect depth samples exhibit a large depth gap, as shown in Fig.~\ref{fig:local-75-nn-hist}.
By this insight, we propose the Large-Depth-Gap Visible Score.

We use a local projected-neighborhood score to identify whether a depth sample
belongs to the frontmost visible surface or to a farther surface behind a depth
discontinuity. The input to this score is a sparse depth image \(D\), generated
by projecting the candidate points indexed by \(\mathcal{C}\) in
Eq.~\eqref{eq:auto-candidate-set} and then applying the depth test defined in
pipeline step~\ref{par:Point-Cloud-Projection}. For an occupied
pixel \(j\), let \(d_j=D(u_j,v_j)\) and let
\(\mathcal{N}_j\) be its local set of occupied neighboring pixels, obtained from
the nearest occupied image locations inside a square window. Let
\(m_j=|\mathcal{N}_j|\), and sort the neighbor depths as
\begin{equation}
    d_{j,(1)} \leq d_{j,(2)} \leq \cdots \leq d_{j,(m_j)}.
    \label{eq:ldg-sorted-depths}
\end{equation}
Large adjacent \(z\)-depth gaps create boundaries between local depth areas:
\begin{equation}
    \begin{aligned}
    \mathcal{B}_j &=
    \left\{\ell \mid d_{j,(\ell+1)}-d_{j,(\ell)}
    \geq \Delta_z \right\},
    \\
    b_{j,\ell}&=\frac{d_{j,(\ell)}+d_{j,(\ell+1)}}{2},
    \end{aligned}
    \label{eq:ldg-depth-gap-boundaries}
\end{equation}
where \(\Delta_z\) is the configured large-gap threshold. The local depth-area
index of a depth value \(d\) is
\begin{equation}
    A_j(d) =
    \sum_{\ell\in\mathcal{B}_j}
    \mathbf{1}\!\left[b_{j,\ell}<d\right].
    \label{eq:ldg-area-index}
\end{equation}
The local depth-area index starts from \(0\), with larger indices indicating
surfaces behind more detected depth gaps. For pixel \(j\), we count the
fraction of local neighbors that lie in nearer depth areas:
\begin{equation}
    r_j =
    \frac{1}{m_j}
    \sum_{q\in\mathcal{N}_j}
    \mathbf{1}\!\left[A_j(d_q)<A_j(d_j)\right].
    \label{eq:ldg-near-ratio}
\end{equation}
The large-depth-gap visible score is
\begin{equation}
    \sigma_j = 1-r_j,
    \qquad
    \sigma_j\in[0,1].
    \label{eq:ldg-visible-score}
\end{equation}
Thus, a sample in the nearest local depth area receives a score of \(1\), while a
sample behind a strong local depth discontinuity receives a lower score
proportional to how many nearer samples occupy the same projected neighborhood.
And binary visible mask can be obtained with
\begin{equation}
    M_j = \mathbf{1}\!\left[\sigma_j \geq \tau\right],
    \label{eq:ldg-binary-mask}
\end{equation}
where \(\tau\) denotes the visibility-score threshold.

\subsection{Automatic Hidden Point Removal (AHPR)}
\label{par:auto-direct-visibility}

\begin{figure}[t]
    \centering
    \subfloat[{Radius scale \(s=500\)\label{fig:500-adius-scale-in-close-distance}}]{%
\begin{minipage}[t]{0.24\textwidth}
\centering
        \includegraphics[width=\linewidth,
        height=0.11\textheight,
        ]{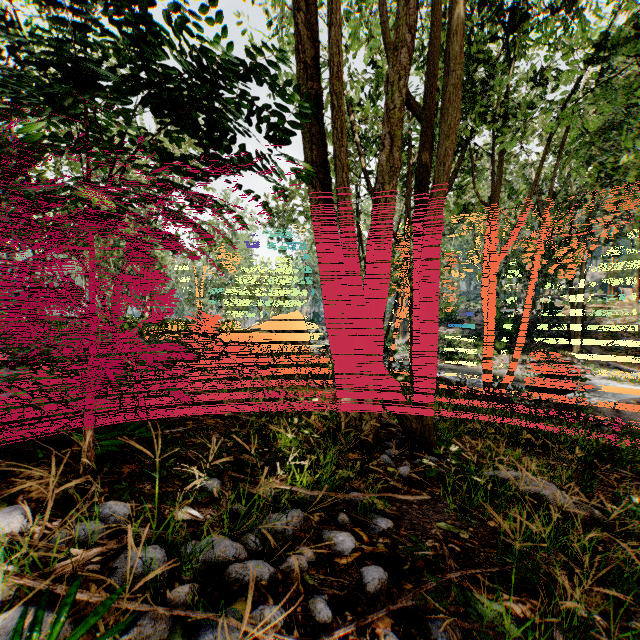}
\end{minipage}}\hfill
    \subfloat[{Radius scale \(s=5000\)\label{fig:5000-radius-scale-in-close-distance}}]{%
\begin{minipage}[t]{0.24\textwidth}
\centering
        \includegraphics[width=\linewidth,
        height=0.11\textheight,
        ]{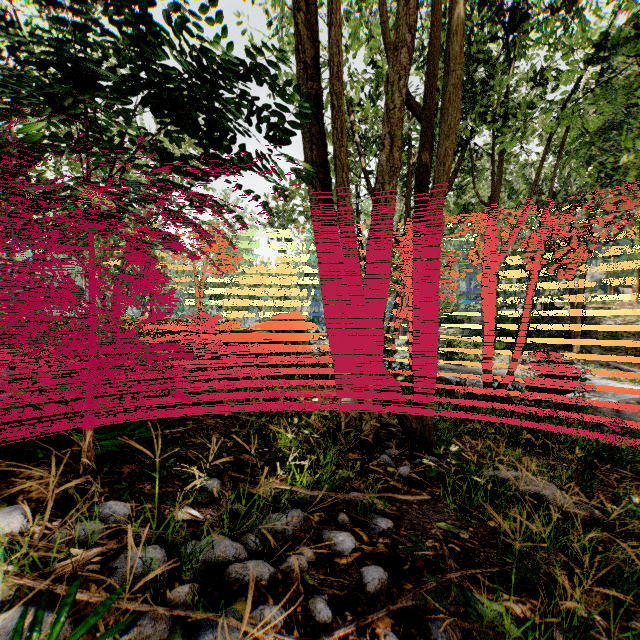}
\end{minipage}}\par\medskip
    \subfloat[{Radius scale \(s=500\)\label{fig:500-radius-scale-in-far-distance}}]{%
\begin{minipage}[t]{0.24\textwidth}
\centering
        \includegraphics[width=\linewidth,
        height=0.11\textheight,
        ]{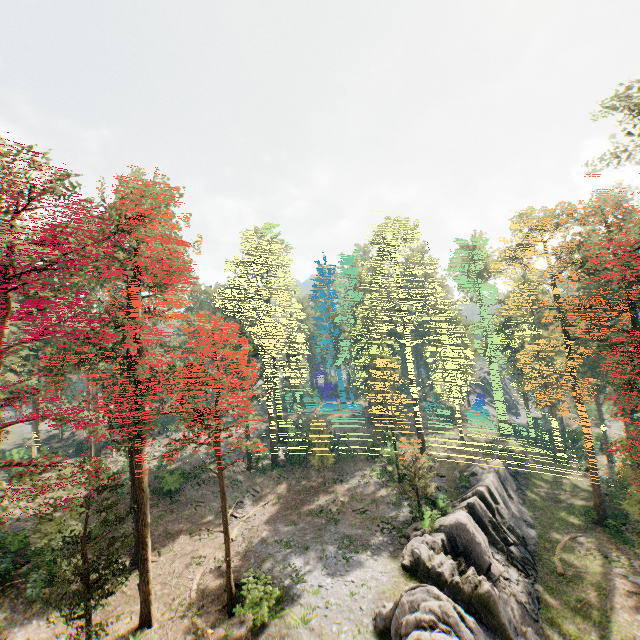}
\end{minipage}}\hfill
        \subfloat[{Radius scale \(s=5000\)\label{fig:5000-radius-scale-in-far-distance}}]{%
\begin{minipage}[t]{0.24\textwidth}
\centering
        \includegraphics[width=\linewidth,
        height=0.11\textheight,
        ]{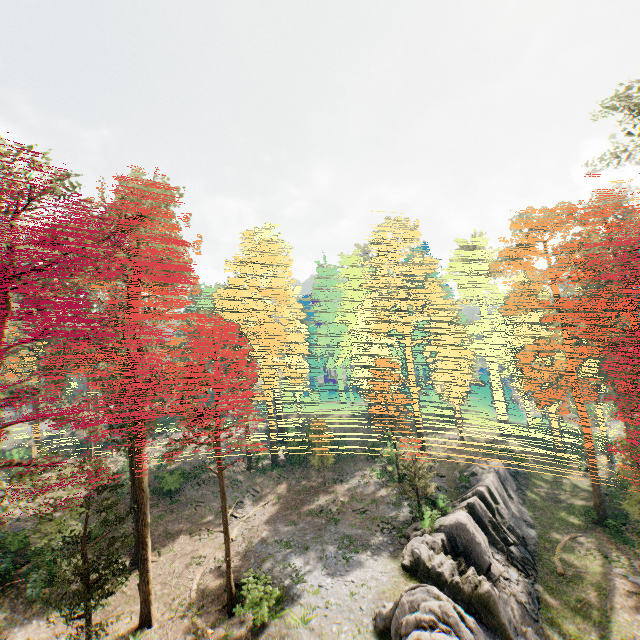}
\end{minipage}}
    \caption{Effect of the HPR radius scale in close-range (top row) and
    far-range (bottom row) scenes. Results are shown for \(s=500\) (left
    column) and \(s=5000\) (right column). A scale that gives an acceptable
    visibility estimate at close range can retain many incorrect depth samples
    at long range, motivating automatic radius scale selection.}
    \label{fig:auto-direct-vis-radius-scale-comparison}
\end{figure}

Hidden point removal (HPR) estimates point visibility directly from a point set
without first reconstructing a surface \cite{Direct-visibility-of-point-sets}.
HPR requires a radius parameter \(R\), which is computed
from the extent of the candidate set \(\mathcal{C}\) defined in
Eq.~\eqref{eq:auto-candidate-set} and a user-specified scale factor \(s\):
\begin{equation}
    R(s)=\max(s r_0,1).
    \label{eq:auto-radius}
\end{equation}
Here \(r_0=\max_{n\in\mathcal{C}}\|\mathbf{p}_n\|_2\) is the maximum distance
of the candidate points from the camera center. Although this computation
adapts the radius to the overall extent of the point cloud, the HPR output
remains sensitive to the chosen scale \(s\). A single fixed scale cannot
consistently accommodate the large variations in geometry and viewing distance
encountered in vegetation-rich and agricultural scenes. As shown in
Fig.~\ref{fig:auto-direct-vis-radius-scale-comparison}, a scale that is too
small may reject valid visible points, whereas a scale that is too large may
retain occluded background points. Consequently, selecting \(s\) manually
would require scene-dependent tuning and is unsuitable for scenes with varying
point-cloud densities and viewing distances.

To enable robust application of HPR in such
environments, we propose Automatic Hidden Point Removal (AHPR),
which uses the large-depth-gap visible score in
Eq.~\eqref{eq:ldg-visible-score} as feedback to select an appropriate scale
\(s\). Let
\(\mathcal{H}_{R}(\mathcal{C})\subseteq\mathcal{C}\) denote the point indices
classified as visible by HPR with radius \(R\). For a trial scale \(s\), the
visible index set is
$\mathcal{V}_{s}^{\rm dir}=\mathcal{H}_{R(s)}(\mathcal{C})$.

For each trial scale \(s\), we project the points indexed by
\(\mathcal{V}_{s}^{\rm dir}\) to get resulting trial
depth image, \(D_s\). And we denote
\(\mathcal{O}_s=\{j:(u_j,v_j)\in\Omega_{\mathrm{image}},\
D_s(u_j,v_j)>0\}\) is its set of occupied pixel indices. Evaluating the score and
binary pixel mask in
Eqs.~\eqref{eq:ldg-visible-score}
and~\eqref{eq:ldg-binary-mask} on \(D_s\) gives \(\sigma_{s,j}\) and
\(M_{s,j}\), respectively, with the score threshold \(\tau\) held fixed across
all trials. We define the fraction of occupied pixels classified as having low
visibility as
\begin{equation}
    \rho(s)
    =
    1 -
    \frac{1}{|\mathcal{O}_s|}
    \sum_{j\in\mathcal{O}_s} M_{s,j}.
    \label{eq:auto-low-score-ratio}
\end{equation}
Rather than minimizing this fraction, we specify a target \(\rho^\star\) to
avoid favoring an overly sparse, conservative HPR output and to control the
accuracy--coverage trade-off. To reduce the search cost, we first evaluate the
coarse logarithmic grid
\begin{equation}
    \begin{aligned}
        \mathcal{S}_c
        &= \operatorname{geomspace}(s_{\min},s_{\max},n_c), \\
        \hat{s}_c
        &= \arg\min_{s\in\mathcal{S}_c}
           \left|\rho(s)-\rho^\star\right|, \\
        \mathcal{S}_f
        &= \operatorname{geomspace}(s_L,s_U,n_f), \\
        \mathcal{S}
        &= \mathcal{S}_c\cup\mathcal{S}_f,
    \end{aligned}
    \label{eq:auto-coarse-to-fine-search}
\end{equation}
where \(s_{\min}\) and \(s_{\max}\) define the coarse search range, \(n_c\) is
the number of coarse candidates, \(s_L\) and \(s_U\) bound the refinement
interval around \(\hat{s}_c\), and \(n_f\) is the number of fine candidates.
The selected radius scale is then
\begin{equation}
    s^\star
    =
    \arg\min_{s\in\mathcal{S}}
    \left|\rho(s)-\rho^\star\right|,
    \label{eq:auto-radius-selection}
\end{equation}
with ties broken in favor of the trial with the larger \(|\mathcal{O}_s|\).

The HPR output at the selected scale is restricted to the image domain:
\begin{equation}
    \mathcal{V}_{\rm AHPR}
    =
    \{n\in\mathcal{V}_{s^\star}^{\rm dir}
    \mid (u_n,v_n)\in\Omega_{\mathrm{image}}\}.
    \label{eq:auto-final-visible-set}
\end{equation}
additionally, the selected points can also be
filtered according to the visibile scores computed from \(D_{s^\star}\):

\begin{equation}
    \begin{aligned}
    \mathcal{V}_{{\rm AHPR},\rm filtered}
    = \{n\in\mathcal{V}_{\rm AHPR}\mid{}&
    \sigma_{s^\star,\pi(n)}\ \textrm{is finite and}\\
    &\sigma_{s^\star,\pi(n)}\geq\tau\},
    \end{aligned}
    \label{eq:auto-optional-score-filter}
\end{equation}
where \(\pi(n)\) is the occupied-pixel index associated with location
\((u_n,v_n)\). The point mask supplied by this module is
\(M_n^{(\mathrm{AHPR})}=\mathbf{1}[n\in\mathcal{V}_{\rm AHPR}]\)

\subsection{Depth-Consistency Visibility Analysis}
\label{par:depth-consistency-visibility-analysis}

Depth-based visibility methods can use depth maps obtained from Poisson surface
reconstruction, SGM, or a metric-depth estimator. Our depth-consistency module
uses a dense metric-depth estimator as an
external visibility reference. Let \(\mathcal{C}\) denote the candidate index
set defined in Eq.~\eqref{eq:auto-candidate-set}. For the target image, the
estimator predicts a metric depth map \(\hat{D}\) at image resolution. A
candidate point is considered visible when its camera-frame \(z\)-depth agrees
with \(\hat{D}\) at the corresponding projected pixel. We first restrict the
candidate indices to
\(\mathcal C_{\mathrm{image}}=\{n\in\mathcal C\mid
(u_n,v_n)\in\Omega_{\mathrm{image}}\}\), and then define

\begin{equation}
    M_n^{\rm depth}
    =
    \mathbf{1}\!\left[
    |z_n-\hat{D}(u_n,v_n)|\leq\delta
    \right],
    \qquad n\in\mathcal C_{\mathrm{image}},
    \label{eq:da-depth-consistency-mask}
\end{equation}
where \((u_n,v_n)\) is the projected pixel of point \(n\). In our experiments,
the depth-consistency tolerance is \(\delta=0.03\,\mathrm{m}\). This test removes
points that are substantially behind the surface predicted by the metric-depth
estimator while retaining points whose depth agrees with the predicted visible
surface.

The visible set produced by the depth-consistency module is
\begin{equation}
    \mathcal{V}_{\rm depth}
    =
    \{n\in\mathcal C_{\mathrm{image}} \mid M_n^{\rm depth}=1\}.
    \label{eq:da-visible-set}
\end{equation}

\section{Experiments}
\label{sec:experiments}

We organize the our experiment into two primary branches. The first evaluates the accuracy
of the generated visible-depth references simulated datasets. 
The second is to benchmarks 3D reconstruction and depth estimation using the generated references 
in a real SLAM dataset. 

\paragraph{Metrics} For point-cloud comparison, we report accuracy (Acc), the
accuracy inlier ratio (Acc~$\tau$), completeness (Comp), the completeness inlier
ratio (Comp~$\tau$), Chamfer distance (CD), F1 score, and reconstruction density.
For depth comparison, we report absolute depth error (Abs), the absolute-error
inlier ratio (Abs~$\tau$, threshold 0.03), absolute relative depth error (AbsRel),
the relative-error inlier ratio (Rel~$\tau$, threshold 1.03), and valid-depth
density.
We also report runtime for efficiency comparison, which denotes the processing time in seconds per image;
Metric definition details are provided in Supplementary Section~\ref{supp:metric-definitions}.

\subsection{Visible-Depth Accuracy Experiment}
\label{sec:visibility-analysis-experiments}
Before benchmarking the downstream tasks, we first evaluate the quality of the
visible-depth references generated by our pipeline and compare them with
existing projection and visibility methods. To our knowledge, no real-world
dataset provides both LiDAR measurements and ground-truth depth. We therefore
use the simulated TartanAirV2 and TartanGround
datasets~\cite{TartanAir,TartanGround}, which
provide LiDAR measurements and ground-truth depth. Because the simulated LiDAR
is aligned with the left camera, we project it into the right camera to
reproduce camera--LiDAR parallax and evaluate all methods against the right-view
ground-truth depth. As a best-case reference for direct LiDAR projection, we
additionally project a single LiDAR scan into the aligned left camera, where no
camera--LiDAR parallax is introduced.

The methods are abbreviated in the tables as follows: circle projection (CP),
Poisson surface reconstruction (PSR), semi-global matching (SGM),
learning-based depth (LD) using DA2, DA3, and MoGe-2, Large-Depth-Gap Visible
Score Filtering (LDGVSF), and Automatic Hidden Point Removal (AHPR).
Visibility estimation (0.7) applies a visibility-score threshold of 0.7.
LDGVSF(0.7) retains samples with a finite Large-Depth-Gap Visible Score of at
least 0.7. AHPR(5\%) selects the HPR radius scale using a target low-visibility
ratio of 5\%; AHPR(10\%) uses a target ratio of 10\%.

AHPR(5\%)+LDGVSF(0.7) denotes sequential filtering: AHPR first selects the
visible point-index set using the 5\% target low-visibility ratio, after which
LDGVSF retains only indices whose selected-scale visibility score is finite
and at least \(0.7\). AHPR(5\%)+LDGVSF(0.7)+LD denotes the union of this
filtered set and the point indices accepted by LD-based depth consistency.

The parameter ablation supporting the selection of
AHPR(5\%)+LDGVSF(0.7) for reference-depth generation is presented in
Supplementary Section~\ref{supp:visible-depth-ablation}.

\begin{table}[t]
\centering
\caption{Evaluation of visible depth on TartanAirV2 Data\_easy and TartanGround
Data\_omni. The best value per metric is shown in bold and the second-best
distinct value is underlined.}
\label{tab:Evaluation-of-visible-depth-in-TartanAirV2-and-TartanGround}
\scriptsize
\setlength{\tabcolsep}{2pt}
\begin{tabular}{@{}lccccc@{}}
\toprule
Method & AbsRel $\downarrow$ & Rel $\tau$ $\uparrow$ & Abs $\downarrow$ & Abs $\tau$ $\uparrow$ & Density $\uparrow$ \\
\midrule
Single scan (left view) & 0.1138 & 0.7782 & 3.7065 & 0.4652 & 0.0232 \\
Single scan (right view) & 0.4024 & 0.7448 & 4.2277 & 0.4030 & 0.0216 \\
\midrule
Cumulation (baseline) & 0.4016 & 0.6384 & 5.4760 & 0.3098 & \underline{0.2524} \\
Cumulation+CP & 0.3214 & 0.6393 & 5.5628 & 0.3176 & \textbf{0.3171} \\
\midrule
PSR & \textbf{0.1386} & \textbf{0.7336} & \underline{3.2805} & \textbf{0.4297} & 0.0582 \\
SGM & 0.3093 & 0.5919 & 3.6819 & 0.2426 & 0.0065 \\
LD & 0.1894 & 0.6220 & \textbf{2.6413} & 0.2799 & 0.0239 \\
Visibility estimation (0.7) & 0.2319 & 0.6499 & 5.0139 & 0.3318 & 0.1828 \\
\midrule
LDGVSF(0.7) & 0.1819 & 0.6918 & 4.3071 & 0.3477 & 0.1944 \\
AHPR(5\%) & 0.1545 & 0.6852 & 4.4641 & 0.3460 & 0.1846 \\
AHPR(5\%)+LDGVSF(0.7) & \underline{0.1399} & \underline{0.6955} & 4.2493 & \underline{0.3601} & 0.1664 \\
AHPR(5\%)+LDGVSF(0.7)+LD & 0.1438 & 0.6909 & 4.2247 & 0.3560 & 0.1705 \\
\bottomrule
\end{tabular}
\end{table}

Table~\ref{tab:Evaluation-of-visible-depth-in-TartanAirV2-and-TartanGround}
shows a clear accuracy--density trade-off. Circle projection (CP) improves 
both reconstruction performance and density compared with direct cumulation, 
achieving the highest coverage among the evaluated methods. 
However, its accuracy gains are limited, 
and a substantial performance gap remains relative to the other methods. 
Therefore, when direct
cumulation already provides sufficient projected LiDAR density, circle
projection is unattractive because the additional coverage can include
unreliable samples without a commensurate improvement in accuracy.

PSR provides the strongest overall accuracy profile, achieving the best AbsRel,
Rel~$\tau$, and Abs~$\tau$ and the second-best Abs, but it retains only sparse
support. In
contrast, AHPR(5\%)+LDGVSF(0.7) and AHPR(5\%)+LDGVSF(0.7)+LD preserve
substantially denser coverage while remaining competitive in accuracy. Adding
LDGVSF(0.7) to AHPR(5\%) improves all four accuracy measures at the cost of a
moderate reduction in density. The additional LD component recovers some
coverage and further improves absolute error, while slightly weakening the
other accuracy measures. These variants therefore offer a more useful balance
between accuracy and spatial coverage than the more conservative
depth-consistency methods.

\begin{figure*}[t]
    \centering
    \subfloat[{Right-view RGB image\label{fig:depth-right-image}}]{%
\begin{minipage}[t]{0.24\textwidth}
\centering
        \includegraphics[width=\linewidth,height=0.11\textheight]{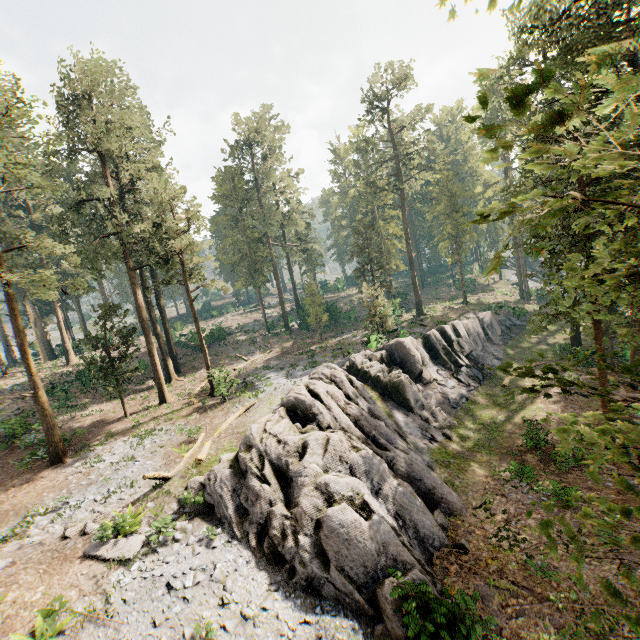}
\end{minipage}}\hfill
    \subfloat[{Ground-truth depth (0--100 m)\label{fig:depth-right-gt}}]{%
\begin{minipage}[t]{0.24\textwidth}
\centering
        \includegraphics[width=\linewidth,height=0.11\textheight]{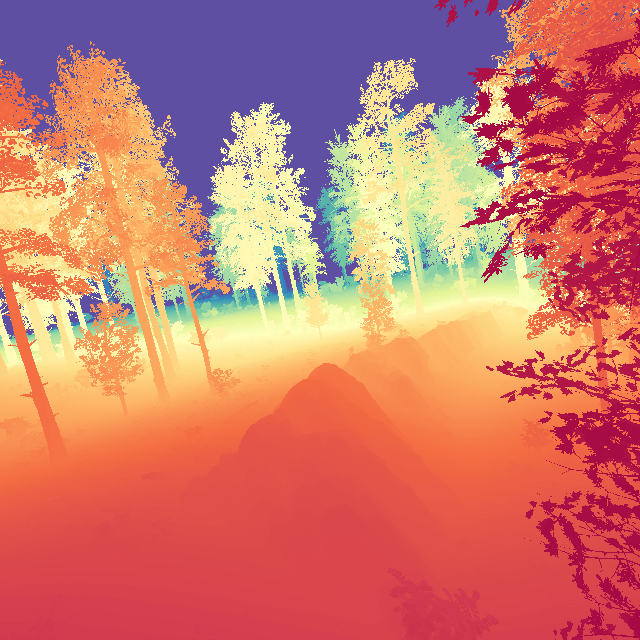}
\end{minipage}}\hfill
    \subfloat[{Cumulation (baseline)\label{fig:depth-from-cumulation}}]{%
\begin{minipage}[t]{0.24\textwidth}
\centering
        \includegraphics[width=\linewidth,height=0.11\textheight]{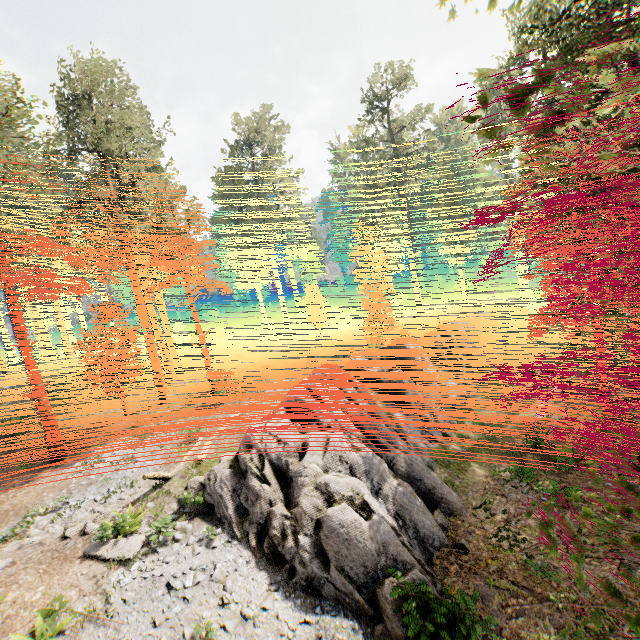}
\end{minipage}}\hfill
    \subfloat[{Cumulation+CP\label{fig:depth-from-cumulation-cp}}]{%
\begin{minipage}[t]{0.24\textwidth}
\centering
        \includegraphics[width=\linewidth,height=0.11\textheight]{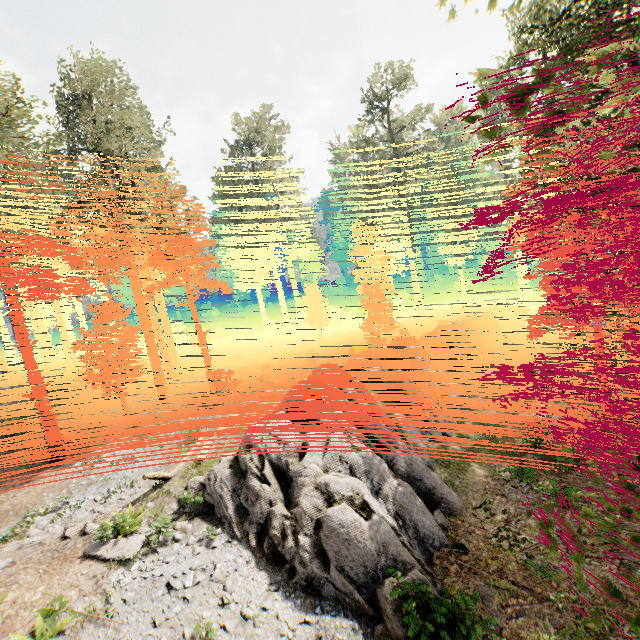}
\end{minipage}}\par\medskip
        \subfloat[{PSR\label{fig:depth-from-psr}}]{%
\begin{minipage}[t]{0.24\textwidth}
\centering
        \includegraphics[width=\linewidth,height=0.11\textheight]{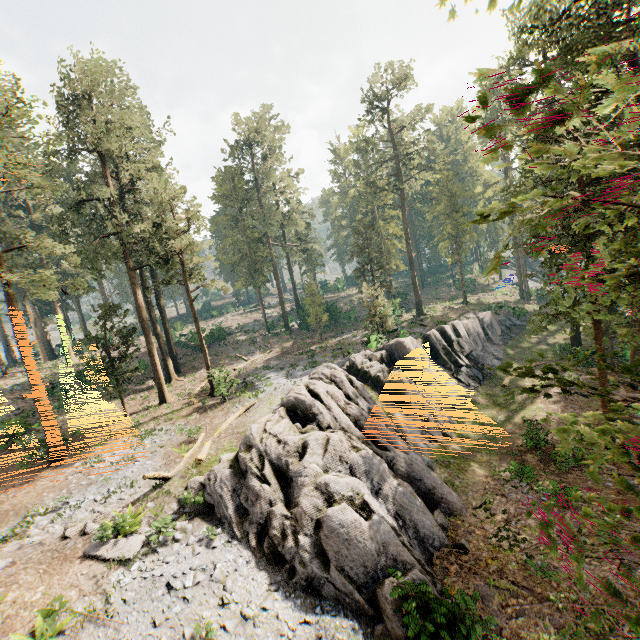}
\end{minipage}}\hfill
    \subfloat[{SGM\label{fig:depth-from-sgm}}]{%
\begin{minipage}[t]{0.24\textwidth}
\centering
        \includegraphics[width=\linewidth,height=0.11\textheight]{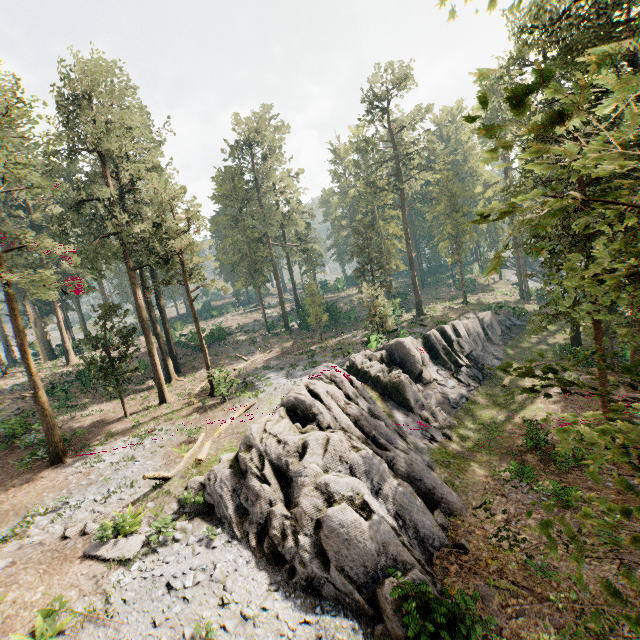}
\end{minipage}}\hfill
    \subfloat[{LD\label{fig:depth-from-ld}}]{%
\begin{minipage}[t]{0.24\textwidth}
\centering
        \includegraphics[width=\linewidth,height=0.11\textheight]{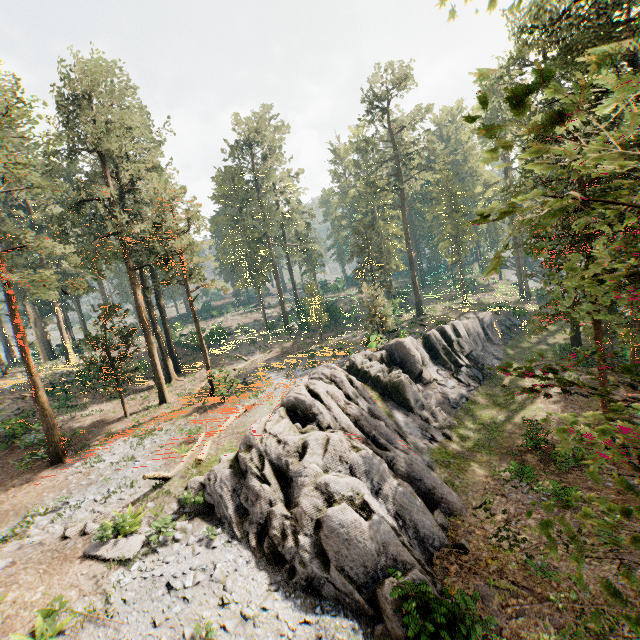}
\end{minipage}}\hfill
    \subfloat[{Visibility estimation (0.7)\label{fig:depth-from-visibility-estimation}}]{%
\begin{minipage}[t]{0.24\textwidth}
\centering
        \includegraphics[width=\linewidth,height=0.11\textheight]{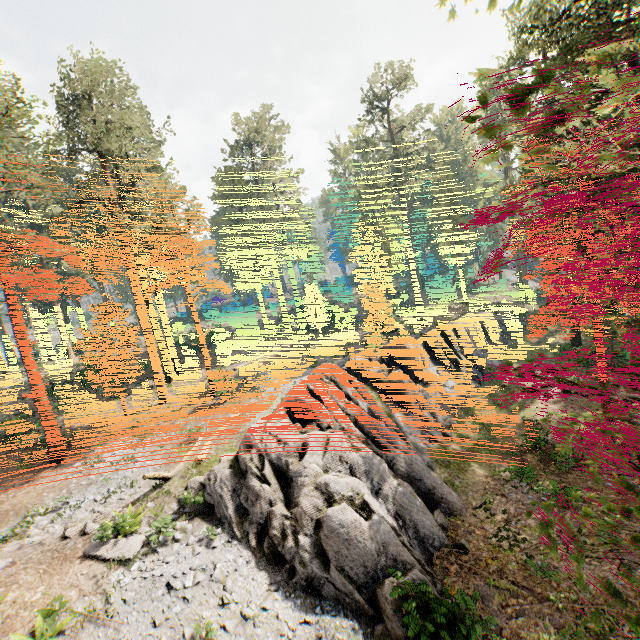}
\end{minipage}}\par\medskip
    \subfloat[{LDGVSF(0.7)\label{fig:depth-from-ldgvsf}}]{%
\begin{minipage}[t]{0.24\textwidth}
\centering
        \includegraphics[width=\linewidth,height=0.11\textheight]{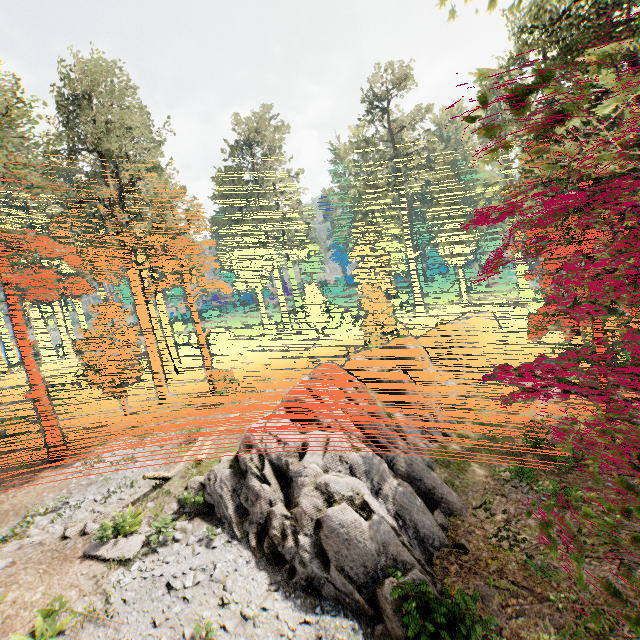}
\end{minipage}}\hfill
    \subfloat[{AHPR(5\%)\label{fig:depth-from-ahpr}}]{%
\begin{minipage}[t]{0.24\textwidth}
\centering
        \includegraphics[width=\linewidth,height=0.11\textheight]{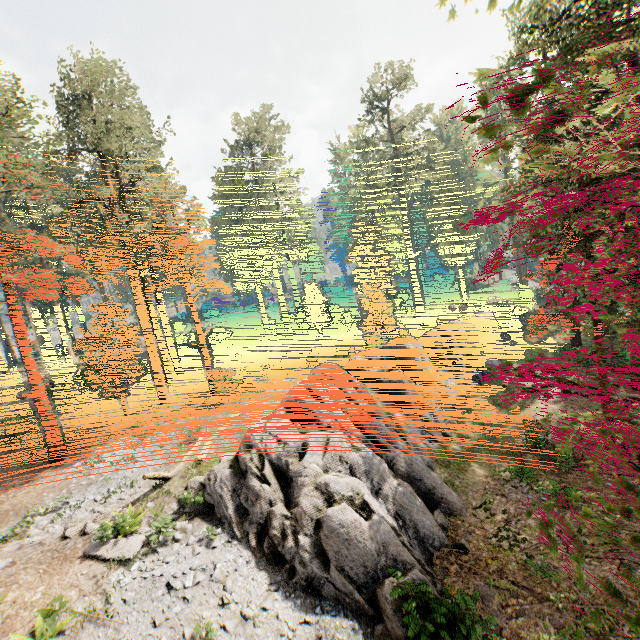}
\end{minipage}}\hfill
    \subfloat[{AHPR(5\%)+LDGVSF(0.7)\label{fig:depth-from-ahpr-ldgvsf}}]{%
\begin{minipage}[t]{0.24\textwidth}
\centering
        \includegraphics[width=\linewidth,height=0.11\textheight]{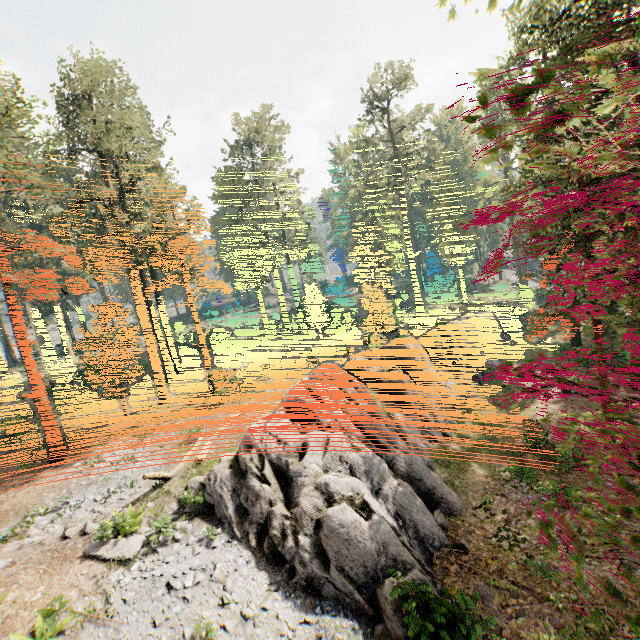}
\end{minipage}}\hfill
    \subfloat[{AHPR(5\%)+LDGVSF(0.7)+LD}]{%
\begin{minipage}[t]{0.24\textwidth}
\centering
        \includegraphics[width=\linewidth,height=0.11\textheight]{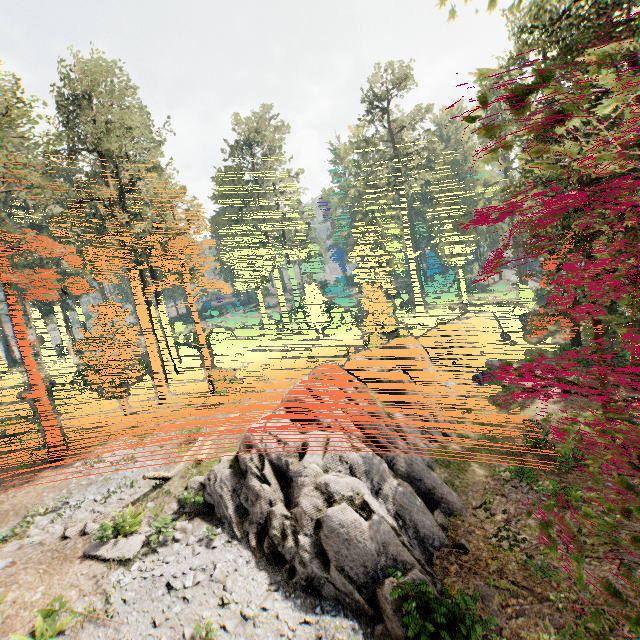}
\end{minipage}}

    \caption{Qualitative comparison of visible-depth projections for a
    TartanAirV2 forest scene. Panels (a) and (b) show the right-view RGB image
    and ground-truth depth, respectively; panels (c)--(l) show the projections
    generated by the selected comparison and proposed methods.}
    \label{fig:cumulated-visible-point-cloud-tartanairv2}
\end{figure*}

The qualitative results in
Figure~\ref{fig:cumulated-visible-point-cloud-tartanairv2} support the quantitative
trade-off. Direct cumulation in (c) retains incorrect depths within objects,
including the tree trunks. Circle projection in (d) increases coverage but also
propagates unreliable depths near discontinuities and into distant regions,
reinforcing its limited value when the baseline cumulation is already
sufficiently dense. The depth-consistency methods in (e)--(g), namely PSR,
SGM, and LD, retain sparse support; in this example, much of that support lies
on low-error object interiors, although SGM remains less accurate in aggregate.
Visibility estimation in (h) leaves some incorrect depths on trunks and foliage
while removing valid boundary samples. By comparison, LDGVSF and the
AHPR-based variants in (i)--(l) preserve substantially denser projections while
maintaining correct depths over most object interiors, providing a better
balance between geometric reliability and spatial coverage.

Additional qualitative comparisons on BotanicGarden and selected depth-error
maps are provided in Supplementary Sections~\ref{supp:botanic-garden-visibility-comparison}
and~\ref{supp:depth-error-analysis}, respectively.

\subsection{Benchmark Experiments}
\label{sec:benchmark-experiments}

\paragraph{Datasets} We apply our reference-generation pipeline, configured 
as AHPR (5\%)+LDGVSF(0.7)+LD, to the BotanicGarden dataset~\cite{BotanicGarden} 
to construct references for the 3D reconstruction and depth benchmarks. 
The pipeline supports complete-sequence reference generation, but
memory and runtime constraints limit the reconstruction benchmark to 200
synchronized stereo frames (400 images), a scale broadly comparable to
established 3D-reconstruction benchmarks such as ETH3D~\cite{ETH3D}.

\paragraph{Common experiment settings}
For the 3D reconstruction comparison, we apply a cross-frame
reprojection-consistency filter to VGGT, VGGT-$\Omega$, DA3, MapAnything, and
MASt3R-SfM. For the depth-estimation comparison, this filter is applied, where enabled,
only to the accumulated 3D visualizations; the quantitative depth metrics and
2D depth maps are obtained before this filtering.
To assess estimation accuracy in high-confidence regions and within a
limited distance range, we include \emph{confidence-P50 mask} and
\emph{range $\leq 10$\,m} variants in the benchmark tables.
The \emph{confidence-P50 mask} retains high-confidence predictions 
by thresholding the model-provided confidence scores at their 50th percentile. 
A \emph{range $\leq 10$\,m} label denotes retaining predictions for which
$0<\widehat d\leq10$\,m.

More details of preprocessing and common settings
are provided in Supplementary Section~\ref{supp:common-experiment-settings}. 

\paragraph{Evaluation protocol}
For both depth and reconstruction evaluations, we compare only pixels with
valid reference depth and a valid prediction. This restricts the comparison to
the part of the scene visible in the camera images and avoids directly
comparing unrestricted global point clouds. Because full-sequence inference
with the multi-view methods exceeds GPU memory,
we divide each sequence into chunks of 16 frames with an 8-frame stride,
yielding a 50\% temporal overlap. Following \cite{VGGT-Long}, adjacent chunks
are registered using a robust confidence-weighted $\mathrm{Sim}(3)$
transformation estimated from pixel-corresponding back-projected points in
their shared frames, as detailed in Supplementary Section~\ref{supp:adjacent-chunk-alignment}. 
After registration, the newer prediction is retained for each duplicated frame
before evaluation and scene fusion. 

\paragraph{Runtime platform} The reference-generation pipeline was run on a
workstation with an NVIDIA GeForce RTX 4070 Ti GPU, a 12-core AMD Ryzen 9 7900X
processor, and 32\,GB of RAM\@. Each benchmark method was run on an HPC node using
one NVIDIA L40S GPU, a 64-core AMD EPYC 9354 processor, and 384\,GB of RAM.

\subsubsection{3D Reconstruction Comparison}

Our traditional geometry-based comparison focuses on incremental SfM, which shares
the core geometric operations of feature-based visual SLAM while
targeting offline scene reconstruction and remains widely used in
agricultural applications. Its modular
correspondence stage also allows different feature matchers to be
evaluated within a common reconstruction framework.
We benchmark SOTA feed-forward methods because they are increasingly
prominent on general reconstruction benchmarks, yet their zero-shot robustness,
metric fidelity, and multi-view consistency in vegetation-rich and agricultural
environments remain insufficiently characterized. Specifically, the benchmark
methods are listed in
Table~\ref{tab:3D-reconstruction-comparison-selected}.

All SfM variants use calibrated intrinsics. For the calibrated-stereo
configurations, we enforce a rigid stereo model and provide strong 
position in the first two frames to anchor the reconstruction to 
the reference coordinate frame.

For reconstruction-to-reference alignment, followed \cite{VGGT-Long}, we fit a
confidence-weighted $\mathrm{Sim}(3)$ transformation to same-frame,
same-pixel 3D correspondences. Each weighted update uses the Umeyama similarity
solution \cite{Umeyama-Algorithm}. The complete procedure is defined in Supplementary Section~\ref{supp:point-cloud-alignment}.

\begin{table}[t]
\centering
\small
\setlength{\tabcolsep}{3pt}
\caption{Quantitative comparison of selected 3D 
reconstruction methods.
Within each method block, the best value per column, except Density, 
is in bold and the second-best distinct value is underlined; 
The Align column indicates whether scene-level point-cloud alignment 
is applied before evaluation. In method labels, 
K denotes the calibrated-intrinsics-only configuration, and S denotes the
calibrated-stereo configuration. All methods in the SfM block use calibrated
intrinsics. The Filter column identifies confidence-P50 masking (P50);
 a dash indicates that this operation is not applied.
The Seq.\ column reports the number of successfully evaluated sequences.}
\label{tab:3D-reconstruction-comparison-selected}
\resizebox{\columnwidth}{!}{%
\begin{tabular}{llcccccccccc}
\toprule
Method & Filter & Align & Seq. & Acc $\downarrow$ & Acc $\tau$ $\uparrow$ & Comp $\downarrow$ & Comp $\tau$ $\uparrow$ & CD $\downarrow$ & F1 $\uparrow$ & Density $\uparrow$ & Runtime $\downarrow$ \\

\midrule
\multicolumn{12}{l}{\textit{Feed-forward dense reconstruction}} \\
VGGT \cite{VGGT} & -- & $\checkmark$ & 10/10 & 0.3136 & 0.2719 & 0.6837 & \underline{0.4106} & 0.9973 & 0.3236 & 1.0000 & 0.0508 \\

VGGT \cite{VGGT} & P50 & $\checkmark$ & 10/10 & 0.1059 & 0.4587 & 1.4361 & 0.2845 & 1.5421 & 0.3486 & 0.5018 & 0.0503 \\

VGGT-$\Omega$ \cite{VGGT-Omega} & -- & $\checkmark$ & 10/10 & 0.2753 & 0.3442 & \textbf{0.2171} & \textbf{0.4848} & \textbf{0.4924} & \underline{0.4015} & 1.0000 & \underline{0.0287} \\

VGGT-$\Omega$ \cite{VGGT-Omega} & P50 & $\checkmark$ & 10/10 & \textbf{0.0639} & \textbf{0.6100} & 1.0890 & 0.3325 & 1.1529 & \textbf{0.4282} & 0.5000 & \textbf{0.0284} \\

DA3 \cite{DA3} & -- & $\checkmark$ & 10/10 & 0.3311 & 0.2462 & 0.3967 & 0.3998 & \underline{0.7278} & 0.3036 & 1.0000 & 0.0539 \\

DA3 \cite{DA3} & P50 & $\checkmark$ & 10/10 & \underline{0.0768} & \underline{0.4976} & 1.4571 & 0.2752 & 1.5339 & 0.3521 & 0.5000 & 0.0579 \\

MapAnything \cite{MapAnything} & -- & $\checkmark$ & 10/10 & 0.3441 & 0.2641 & \underline{0.3965} & 0.3207 & 0.7406 & 0.2886 & 0.6854 & 0.1449 \\

MapAnything \cite{MapAnything} & P50 & $\checkmark$ & 10/10 & 0.1669 & 0.4585 & 0.7268 & 0.3069 & 0.8937 & 0.3660 & 0.4394 & 0.1548 \\

\midrule
\multicolumn{12}{l}{\textit{Structure-from-Motion (SfM)}} \\
MASt3R(K) \cite{MASt3R-SfM} & -- & $\checkmark$ & 10/10 & 0.3333 & 0.2411 & 0.3157 & \underline{0.1756} & 0.6490 & 0.2022 & 0.2184 & 8.2317 \\

MASt3R(K) \cite{MASt3R-SfM} & P50 & $\checkmark$ & 10/10 & 0.1387 & 0.4091 & 1.5183 & 0.1356 & 1.6570 & 0.2018 & 0.1092 & 8.1735 \\

SIFT(K) \cite{COLMAP-SfM} & -- & $\checkmark$ & 10/10 & 0.1976 & 0.4619 & 0.3567 & 0.0381 & 0.5543 & 0.0704 & 0.0020 & \textbf{2.0012} \\

SIFT(S) \cite{COLMAP-SfM} & -- & $\times$ & 10/10 & 0.2123 & 0.3664 & 0.3730 & 0.0223 & 0.5852 & 0.0418 & 0.0020 & 2.2833 \\

SIFT(S) \cite{COLMAP-SfM} & -- & $\checkmark$ & 10/10 & 0.1853 & 0.4786 & 0.3565 & 0.0396 & 0.5418 & 0.0731 & 0.0020 & 2.2833 \\

LoMa(K) \cite{LoMa} & -- & $\checkmark$ & 7/10 & 0.2430 & 0.3785 & 0.2061 & 0.0628 & 0.4491 & 0.1075 & 0.0062 & 6.6940 \\

LoMa(S) \cite{LoMa} & -- & $\times$ & 5/10 & 0.2794 & 0.3148 & 0.2083 & 0.0412 & 0.4877 & 0.0720 & 0.0062 & 6.6476 \\

LoMa(S) \cite{LoMa} & -- & $\checkmark$ & 5/10 & 0.2722 & 0.3587 & 0.2092 & 0.0560 & 0.4813 & 0.0966 & 0.0062 & 6.6476 \\

SuperGlue(K) \cite{SuperGLUE} & -- & $\checkmark$ & 10/10 & 0.2887 & 0.3578 & 0.3571 & 0.0244 & 0.6458 & 0.0456 & 0.0019 & \underline{2.1066} \\

SuperGlue(S) \cite{SuperGLUE} & -- & $\times$ & 10/10 & 0.2862 & 0.2927 & 0.3281 & 0.0142 & 0.6143 & 0.0271 & 0.0019 & 2.1841 \\

SuperGlue(S) \cite{SuperGLUE} & -- & $\checkmark$ & 10/10 & 0.2862 & 0.3646 & 0.3431 & 0.0247 & 0.6293 & 0.0461 & 0.0019 & 2.1841 \\

LoFTR(K) \cite{LoFTR} & -- & $\checkmark$ & 10/10 & 0.1887 & 0.4550 & 0.1993 & 0.1327 & 0.3880 & 0.2052 & 0.0091 & 9.3589 \\

LoFTR(S) \cite{LoFTR} & -- & $\times$ & 10/10 & 0.2330 & 0.3641 & 0.2699 & 0.0960 & 0.5030 & 0.1506 & 0.0090 & 11.1824 \\

LoFTR(S) \cite{LoFTR} & -- & $\checkmark$ & 10/10 & 0.2116 & 0.4633 & 0.2590 & 0.1322 & 0.4706 & 0.2052 & 0.0090 & 11.1824 \\

RoMa-v2(K) \cite{RoMav2} & -- & $\checkmark$ & 6/10 & \underline{0.1025} & \underline{0.5536} & \underline{0.1631} & 0.1599 & \underline{0.2656} & \underline{0.2476} & 0.0086 & 15.5071 \\

RoMa-v2(S) \cite{RoMav2} & -- & $\times$ & 7/10 & 0.1167 & 0.4113 & 0.1860 & 0.0897 & 0.3028 & 0.1466 & 0.0087 & 16.9150 \\

RoMa-v2(S) \cite{RoMav2} & -- & $\checkmark$ & 7/10 & \textbf{0.0901} & \textbf{0.6011} & \textbf{0.1487} & \textbf{0.1851} & \textbf{0.2388} & \textbf{0.2826} & 0.0087 & 16.9150 \\

\bottomrule
\end{tabular}
}
\end{table}

\begin{figure*}[p]
\centering
\scriptsize
\setlength{\tabcolsep}{1pt}
\begin{tabular}{@{}rcccc@{}}
\toprule
\textbf{Method} & \textbf{View 1} & \textbf{View 2}
& \textbf{View 3} & \textbf{View 4} \\
\midrule
\textbf{Reference} & \raisebox{-0.5\height}{\includegraphics[width=0.19\textwidth]{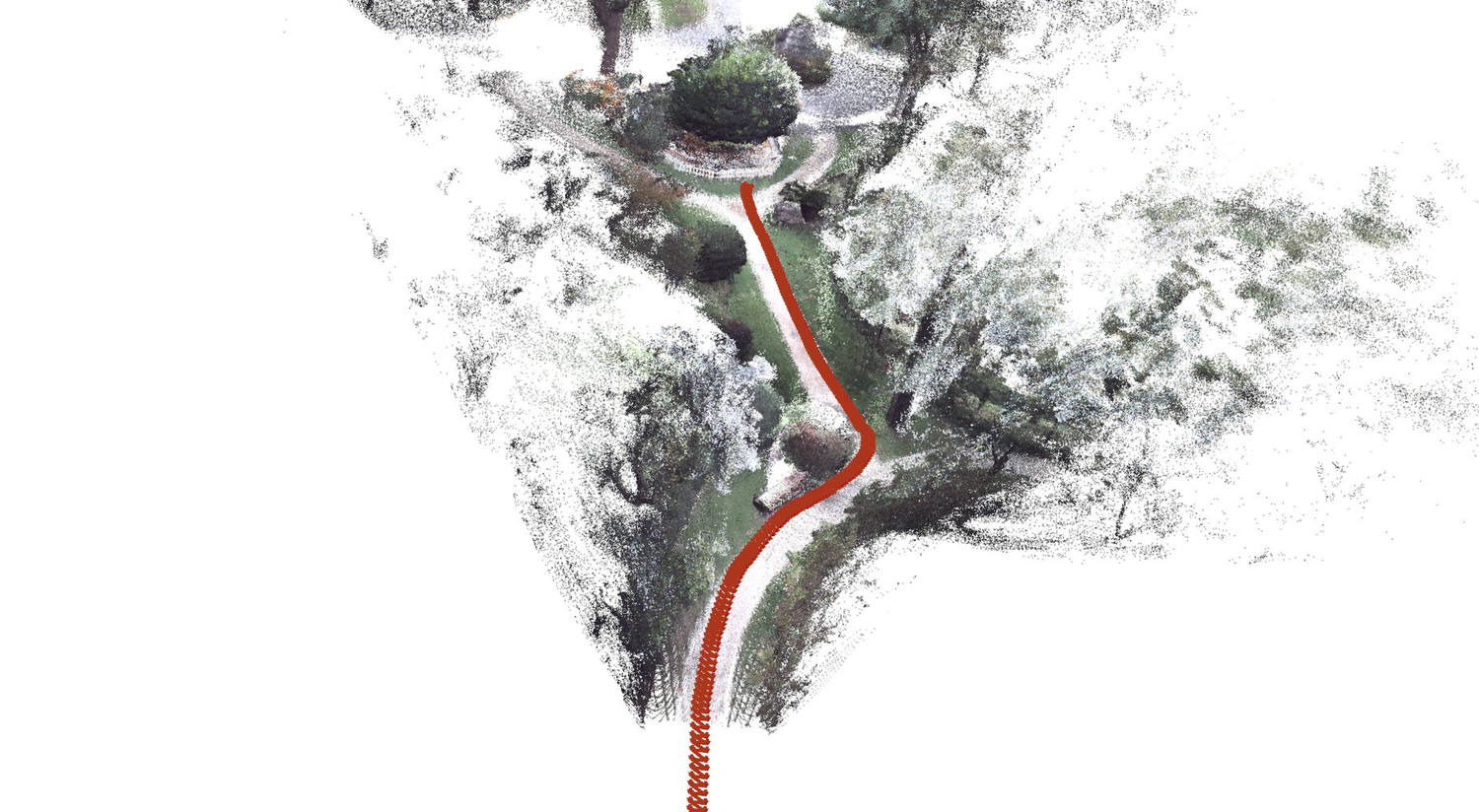}} & \raisebox{-0.5\height}{\includegraphics[width=0.19\textwidth]{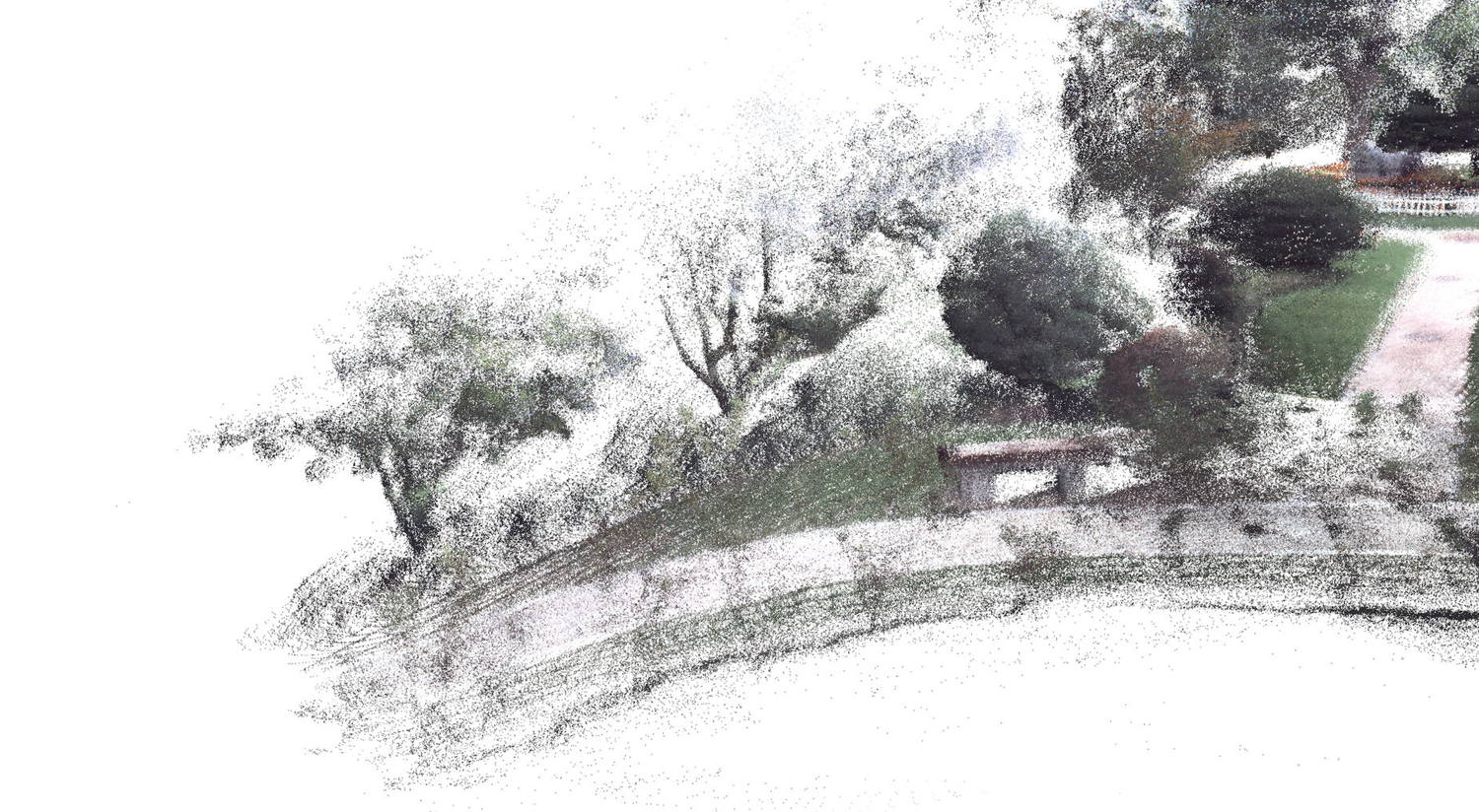}} & \raisebox{-0.5\height}{\includegraphics[width=0.19\textwidth]{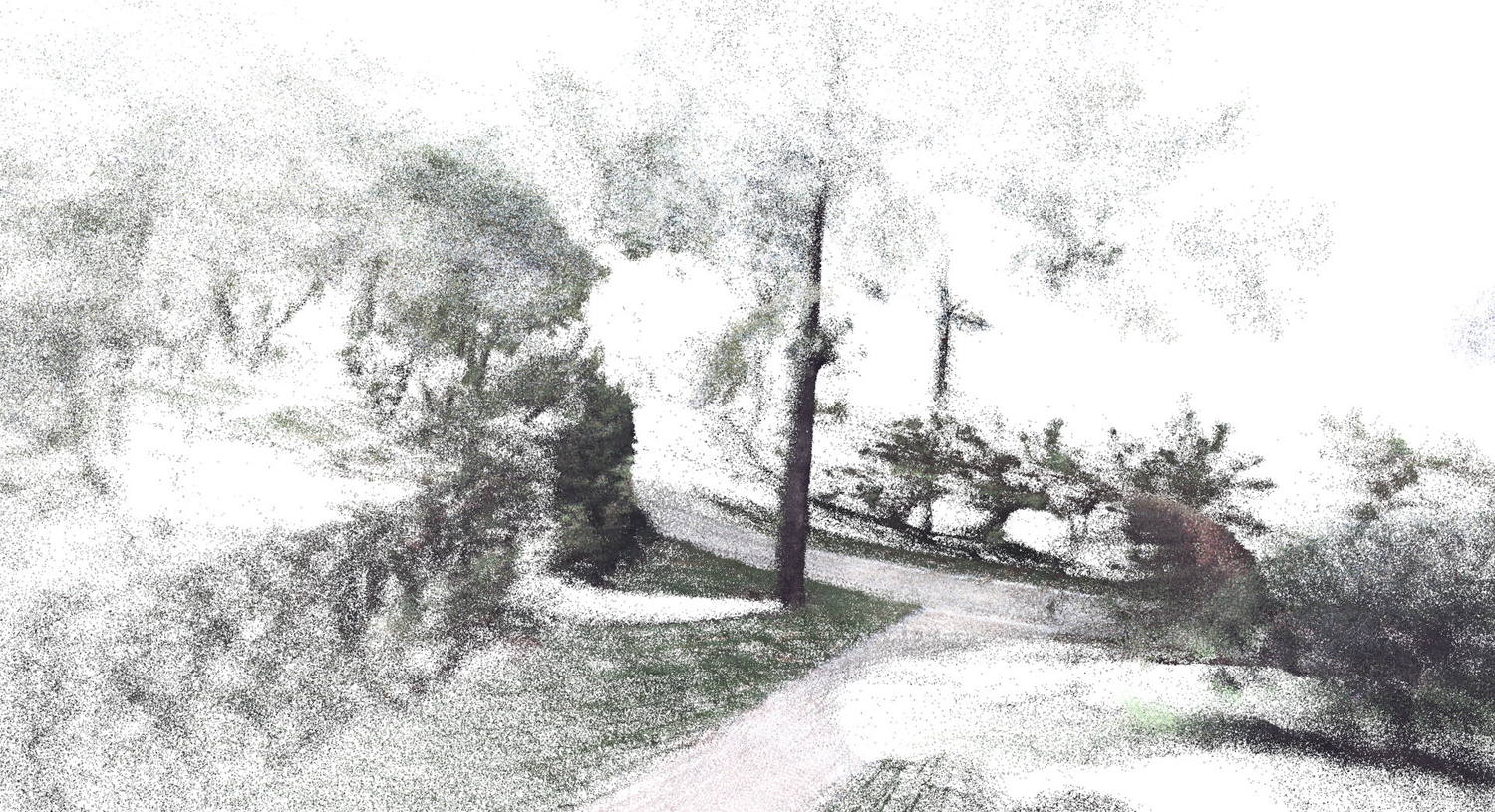}} & \raisebox{-0.5\height}{\includegraphics[width=0.19\textwidth]{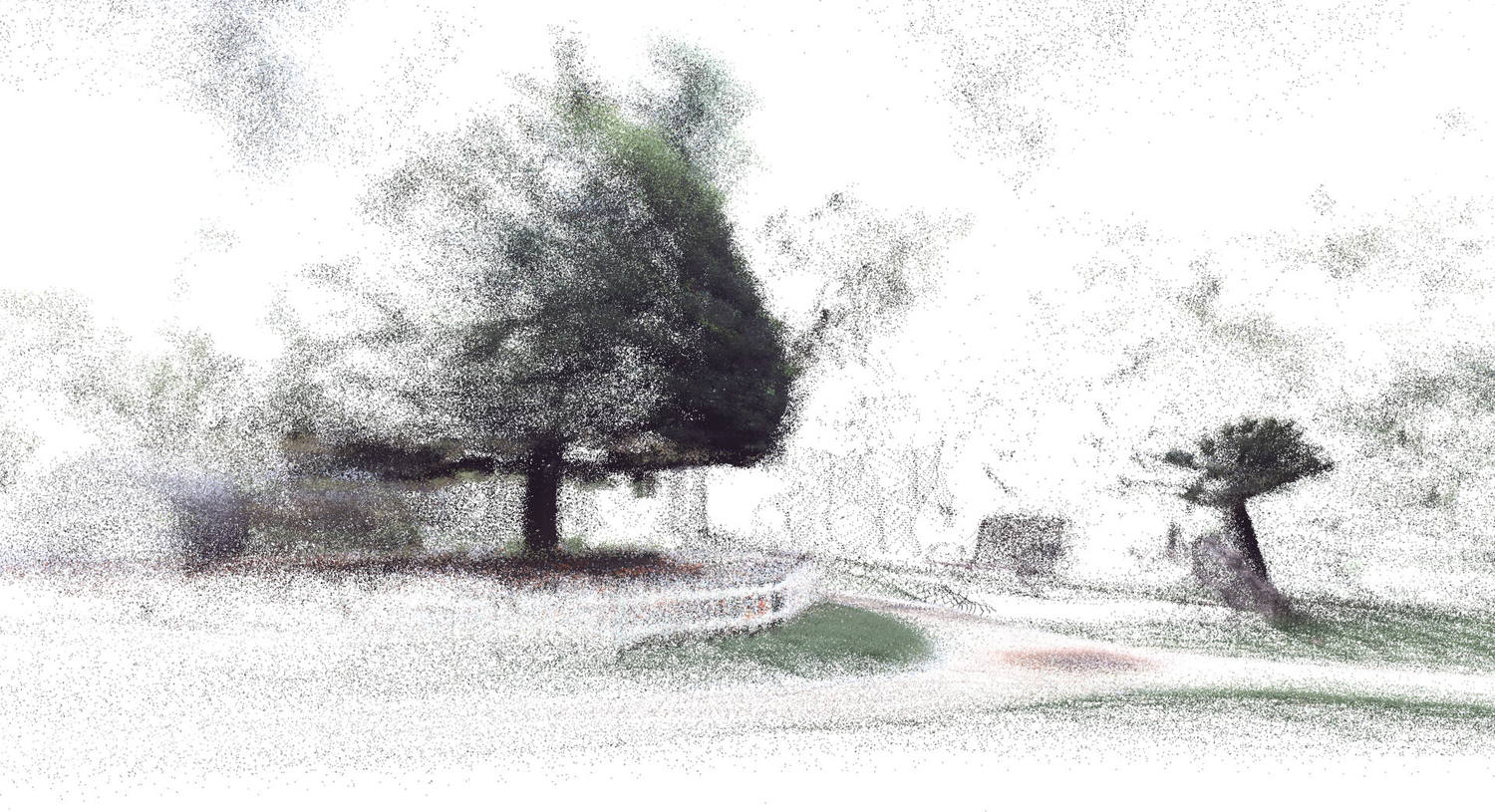}} \\[1pt] 
\textbf{VGGT} & \raisebox{-0.5\height}{\includegraphics[width=0.19\textwidth]{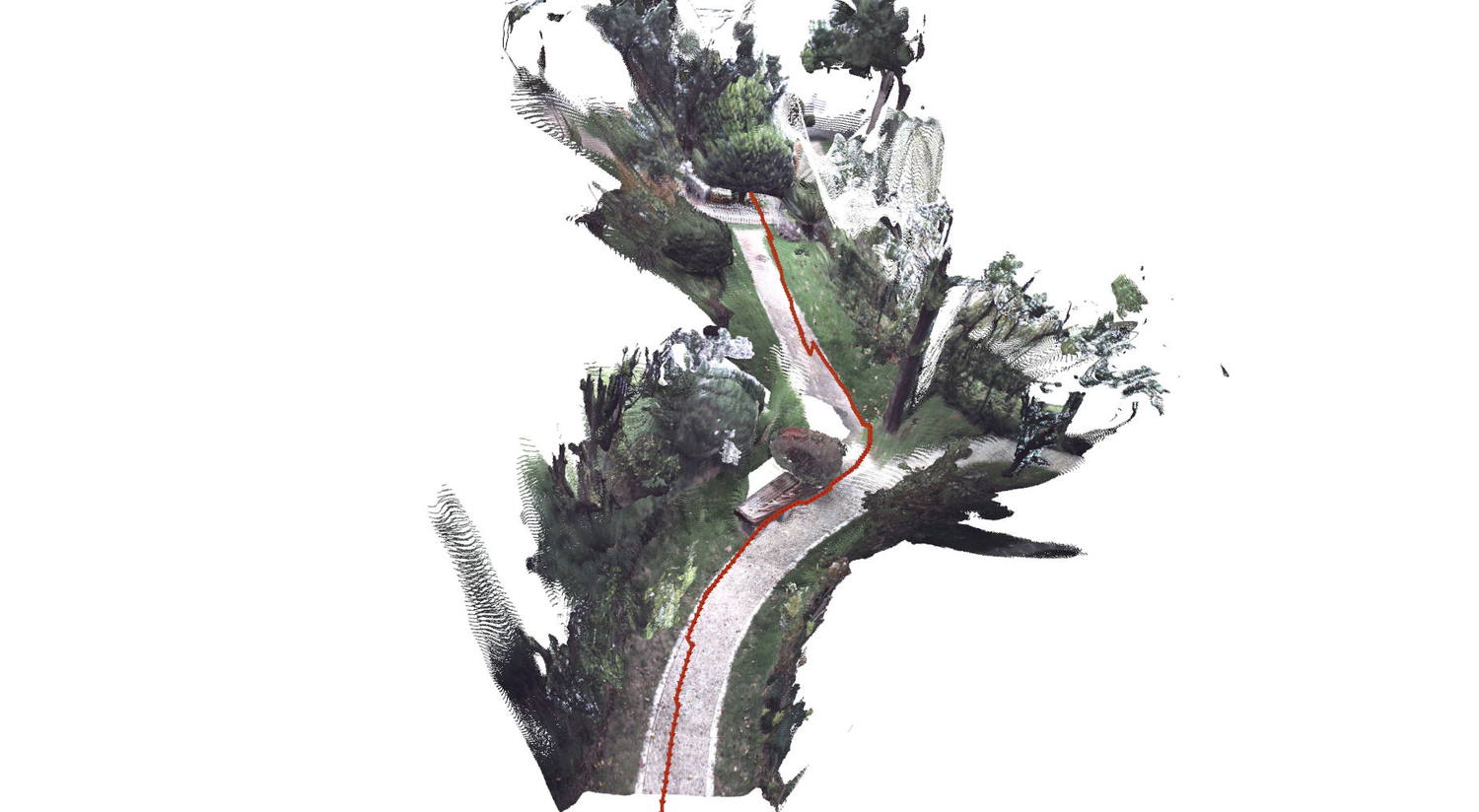}} & \raisebox{-0.5\height}{\includegraphics[width=0.19\textwidth]{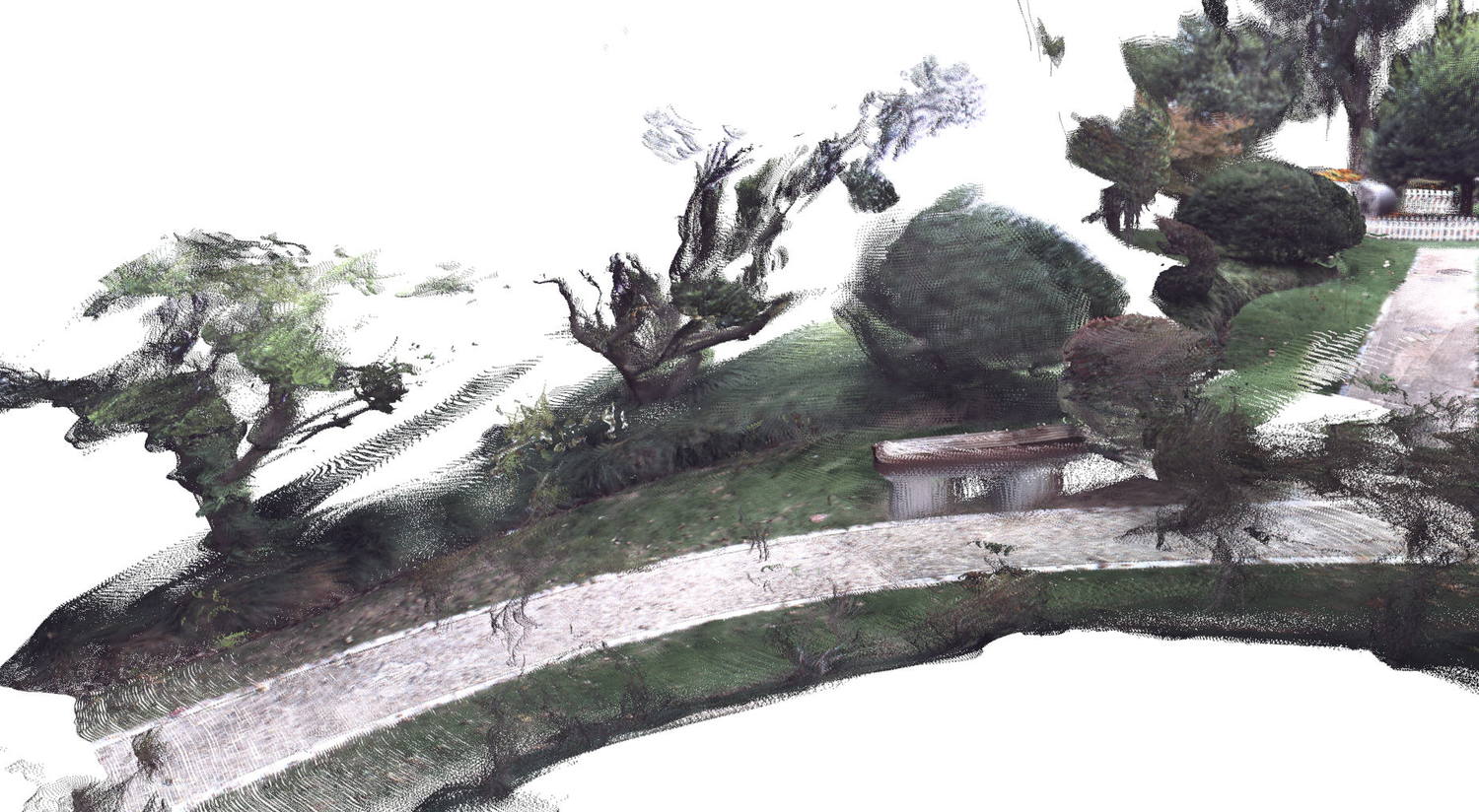}} & \raisebox{-0.5\height}{\includegraphics[width=0.19\textwidth]{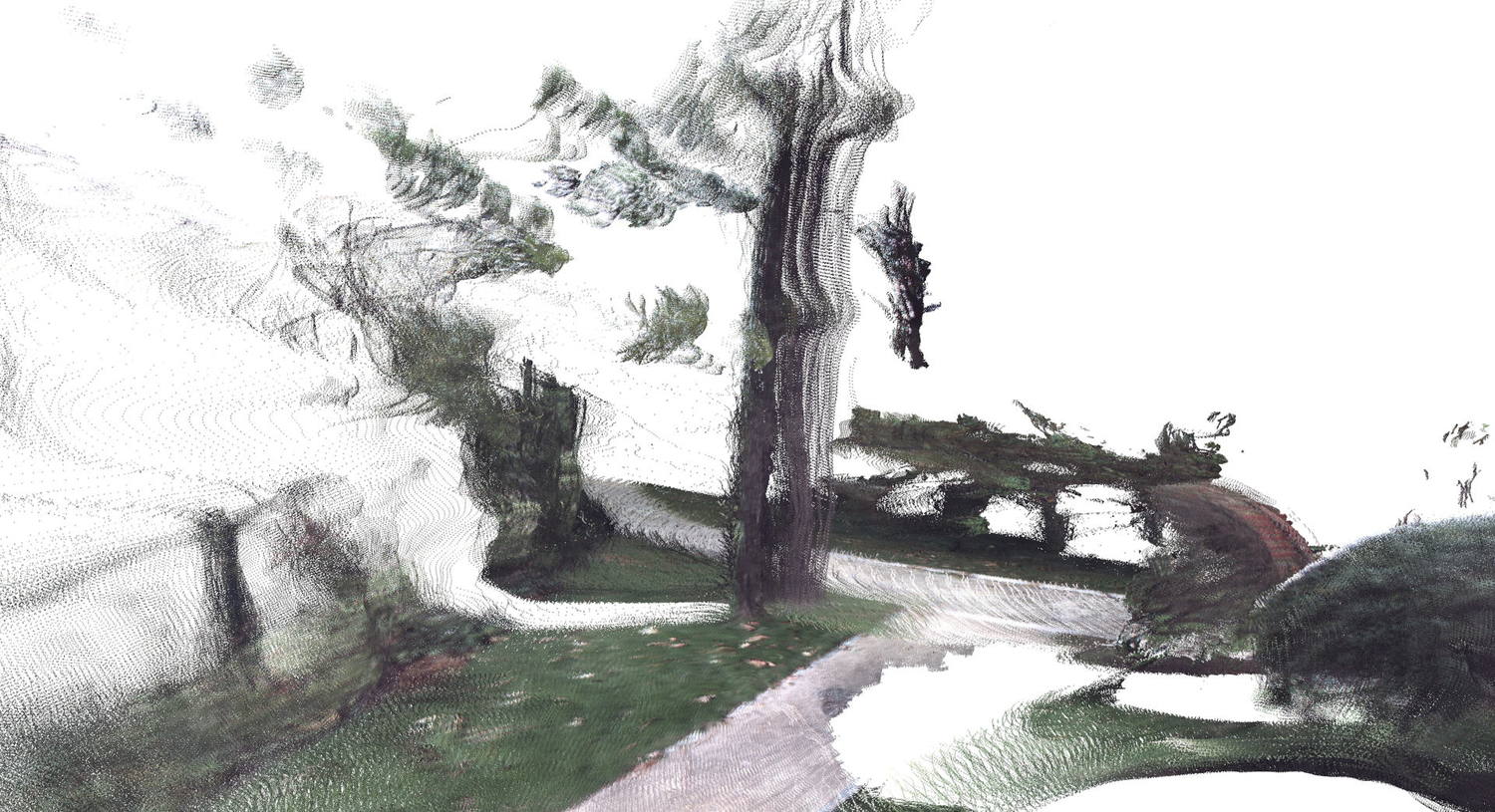}} & \raisebox{-0.5\height}{\includegraphics[width=0.19\textwidth]{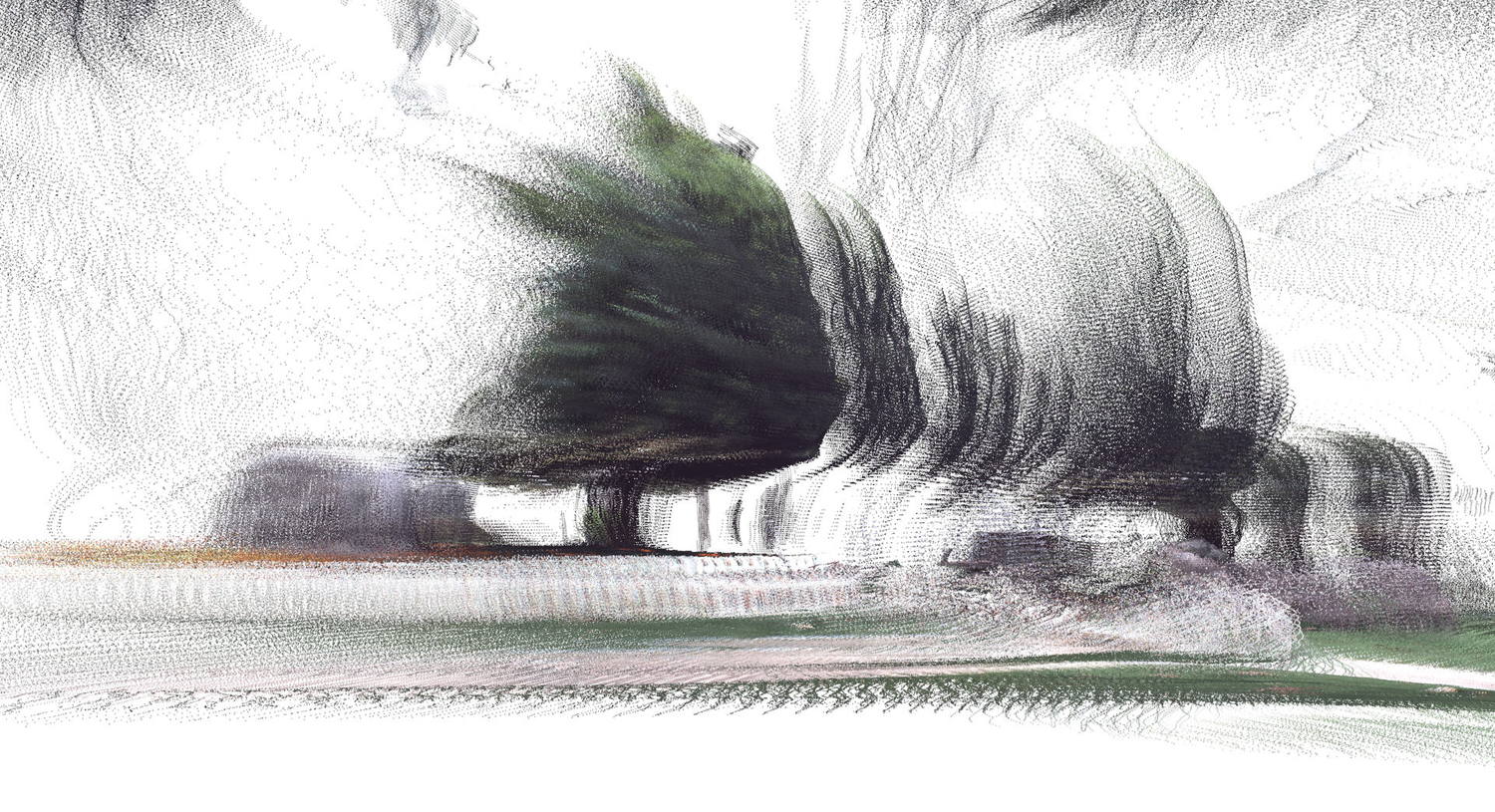}} \\[1pt] 
\textbf{VGGT-$\Omega$} & \raisebox{-0.5\height}{\includegraphics[width=0.19\textwidth]{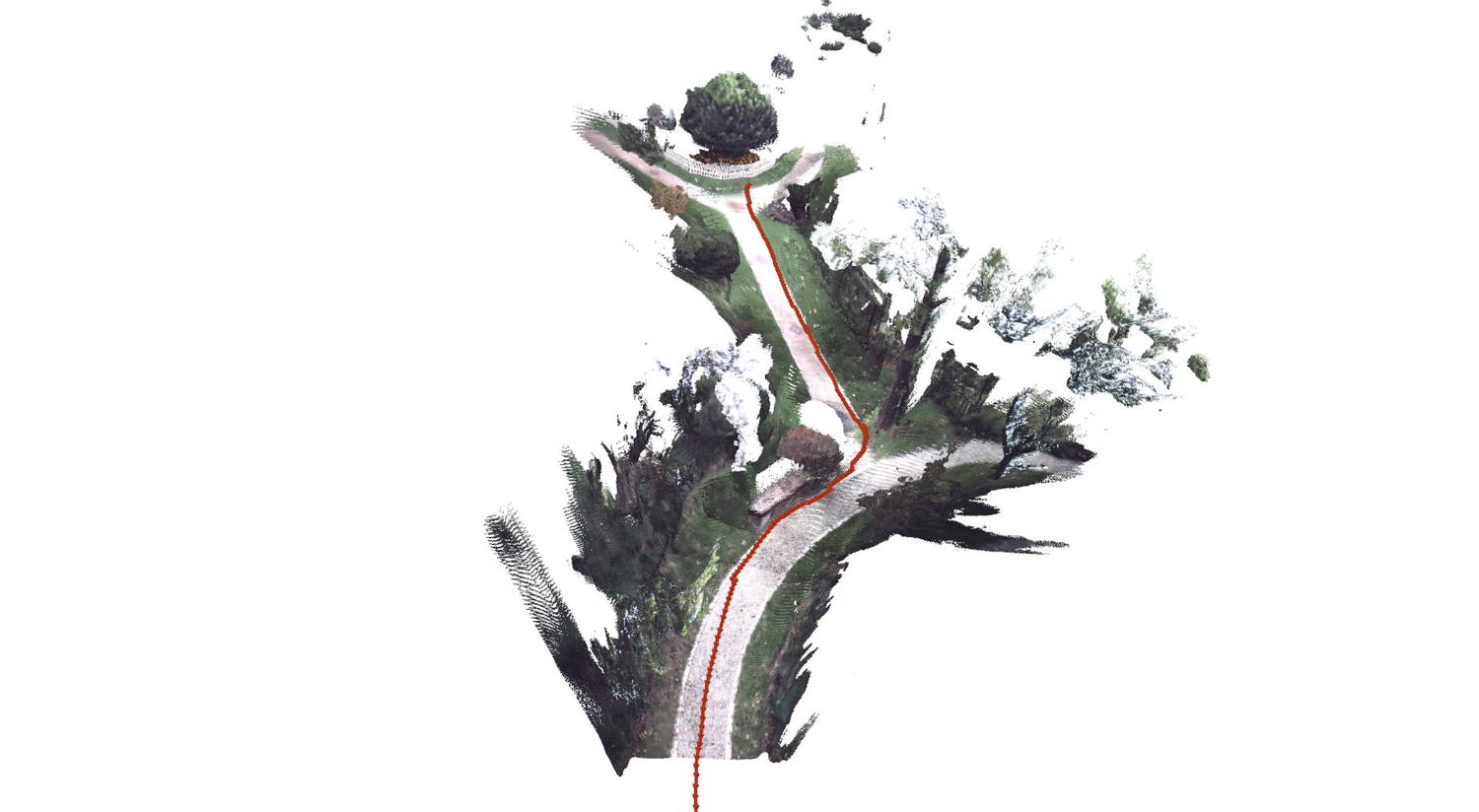}} & \raisebox{-0.5\height}{\includegraphics[width=0.19\textwidth]{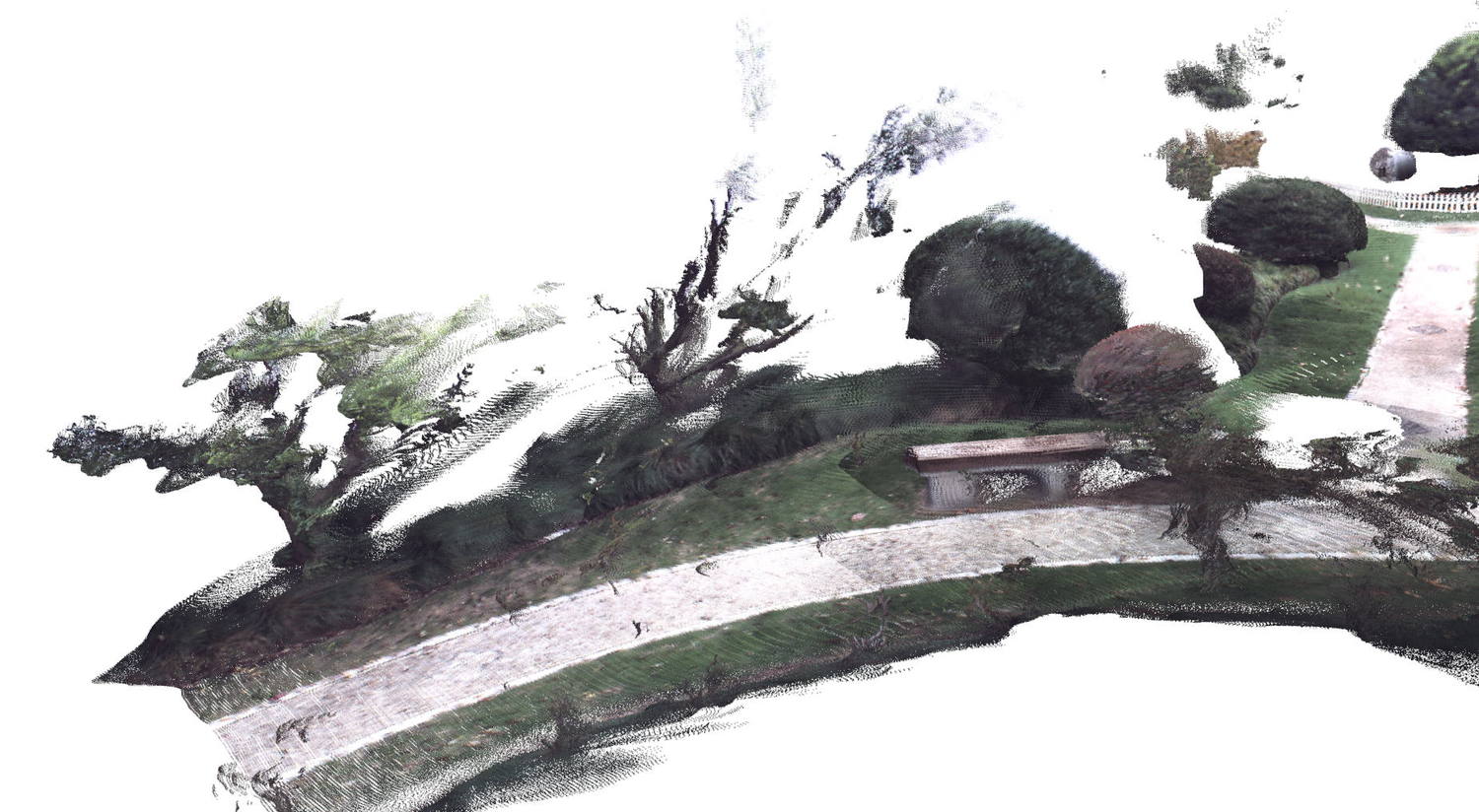}} & \raisebox{-0.5\height}{\includegraphics[width=0.19\textwidth]{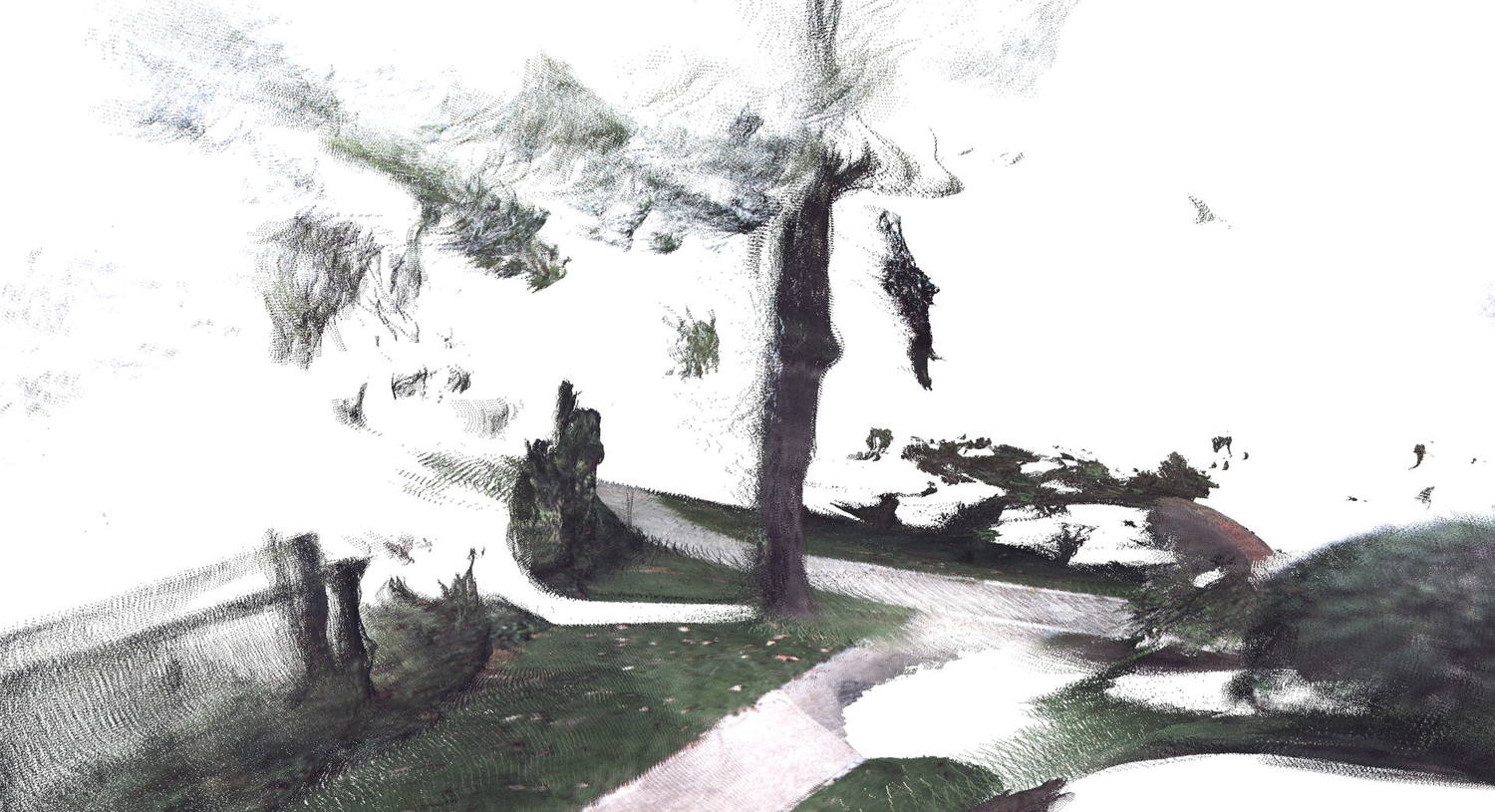}} & \raisebox{-0.5\height}{\includegraphics[width=0.19\textwidth]{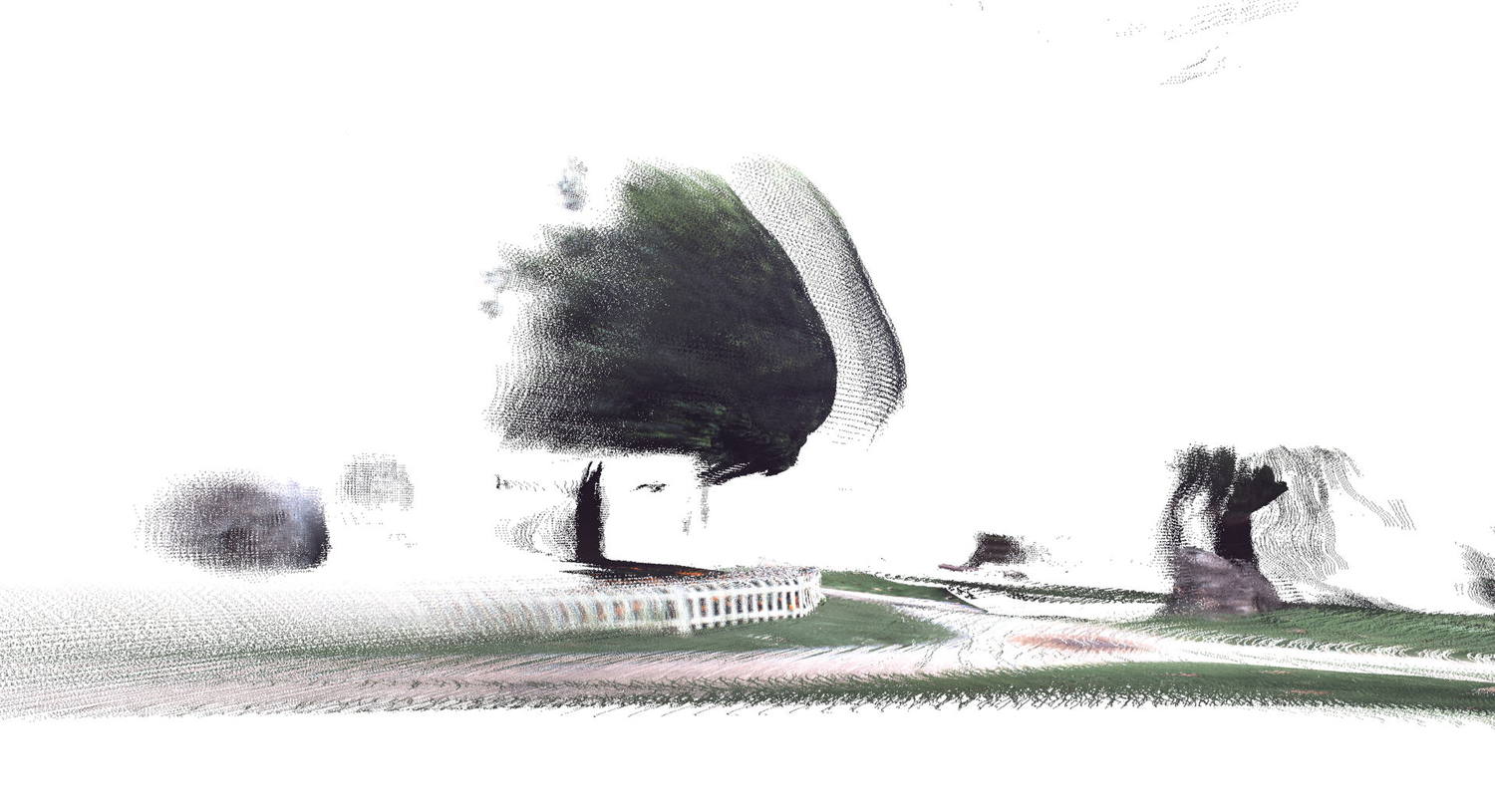}} \\[1pt] 
\textbf{DA3} & \raisebox{-0.5\height}{\includegraphics[width=0.19\textwidth]{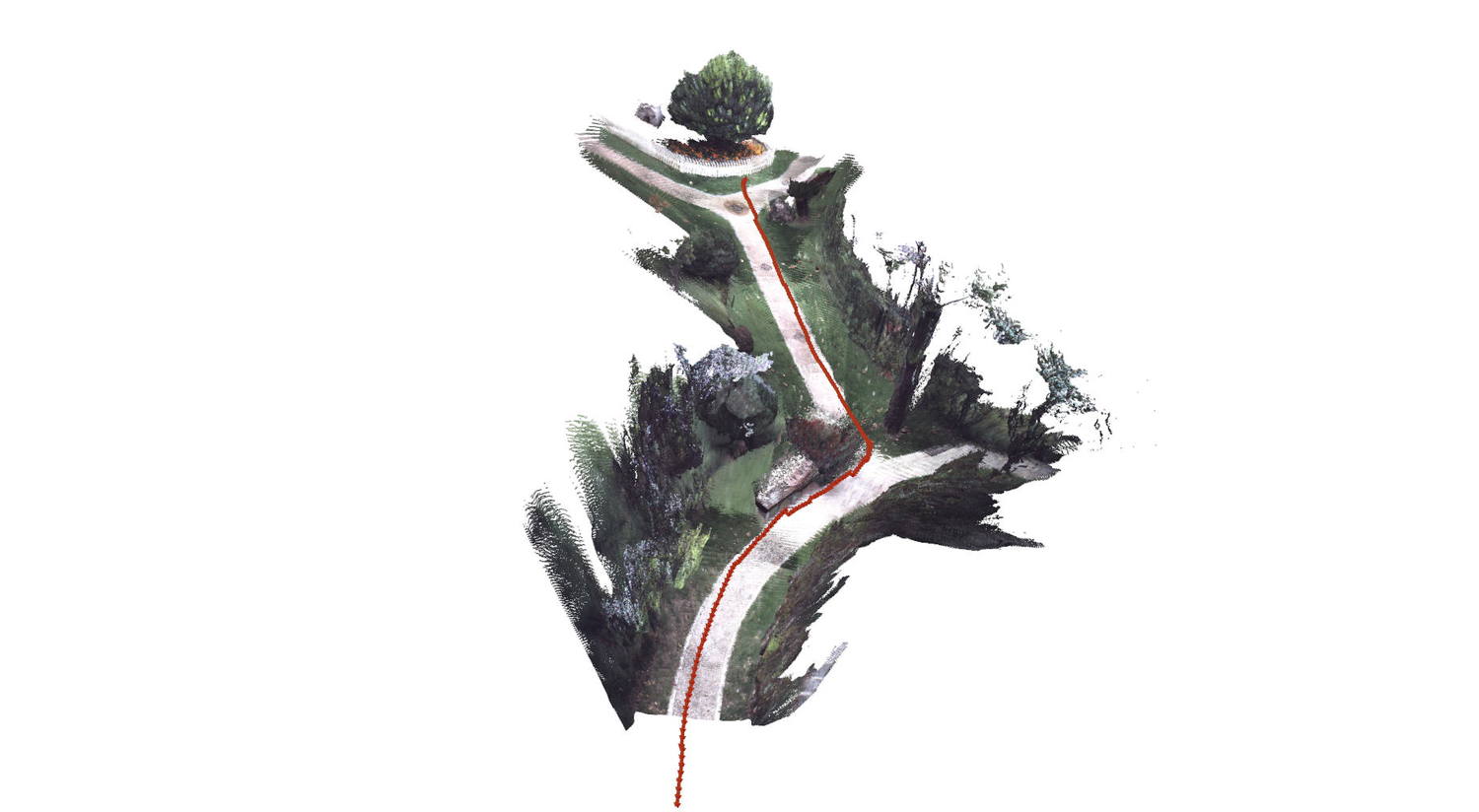}} & \raisebox{-0.5\height}{\includegraphics[width=0.19\textwidth]{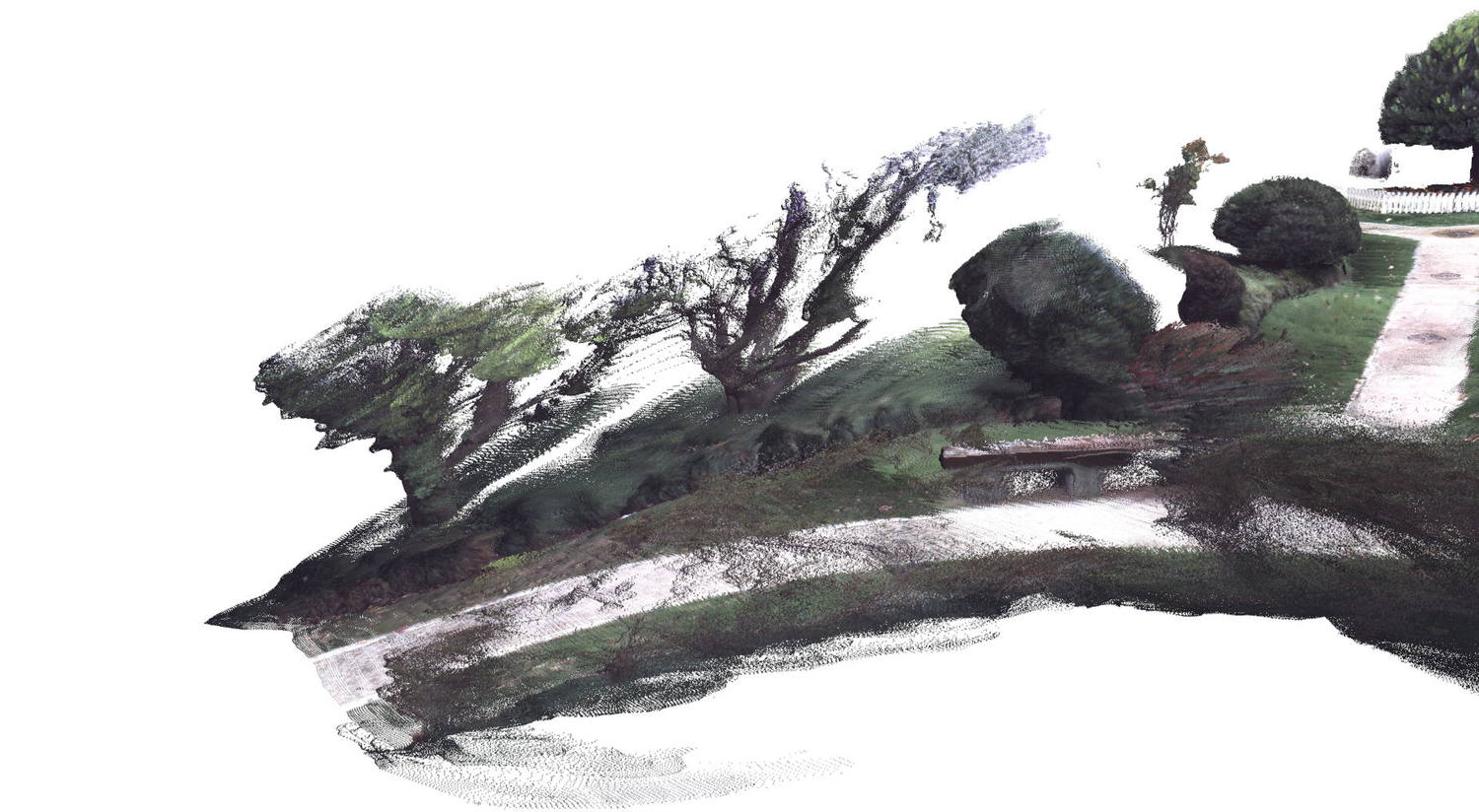}} & \raisebox{-0.5\height}{\includegraphics[width=0.19\textwidth]{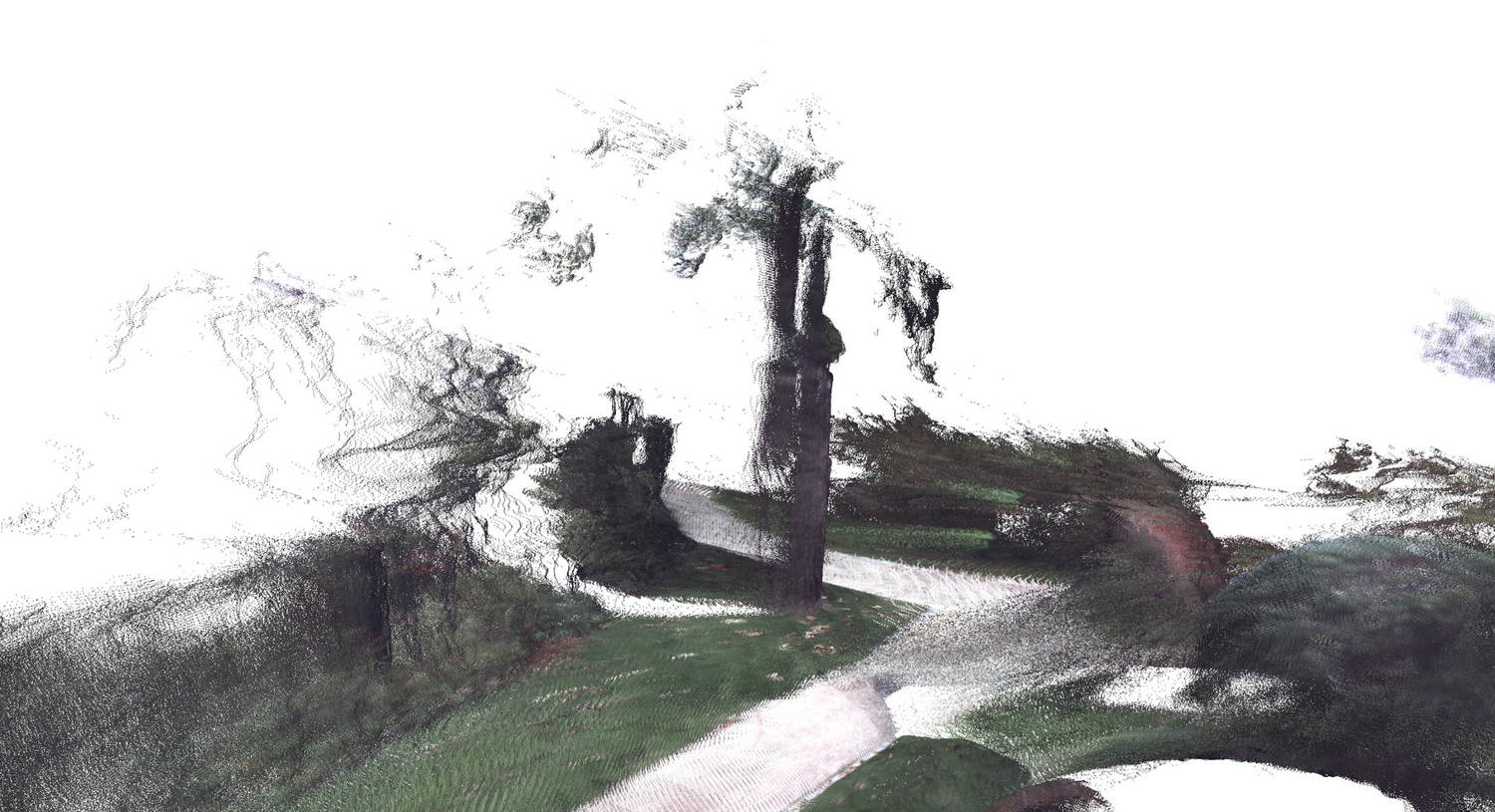}} & \raisebox{-0.5\height}{\includegraphics[width=0.19\textwidth]{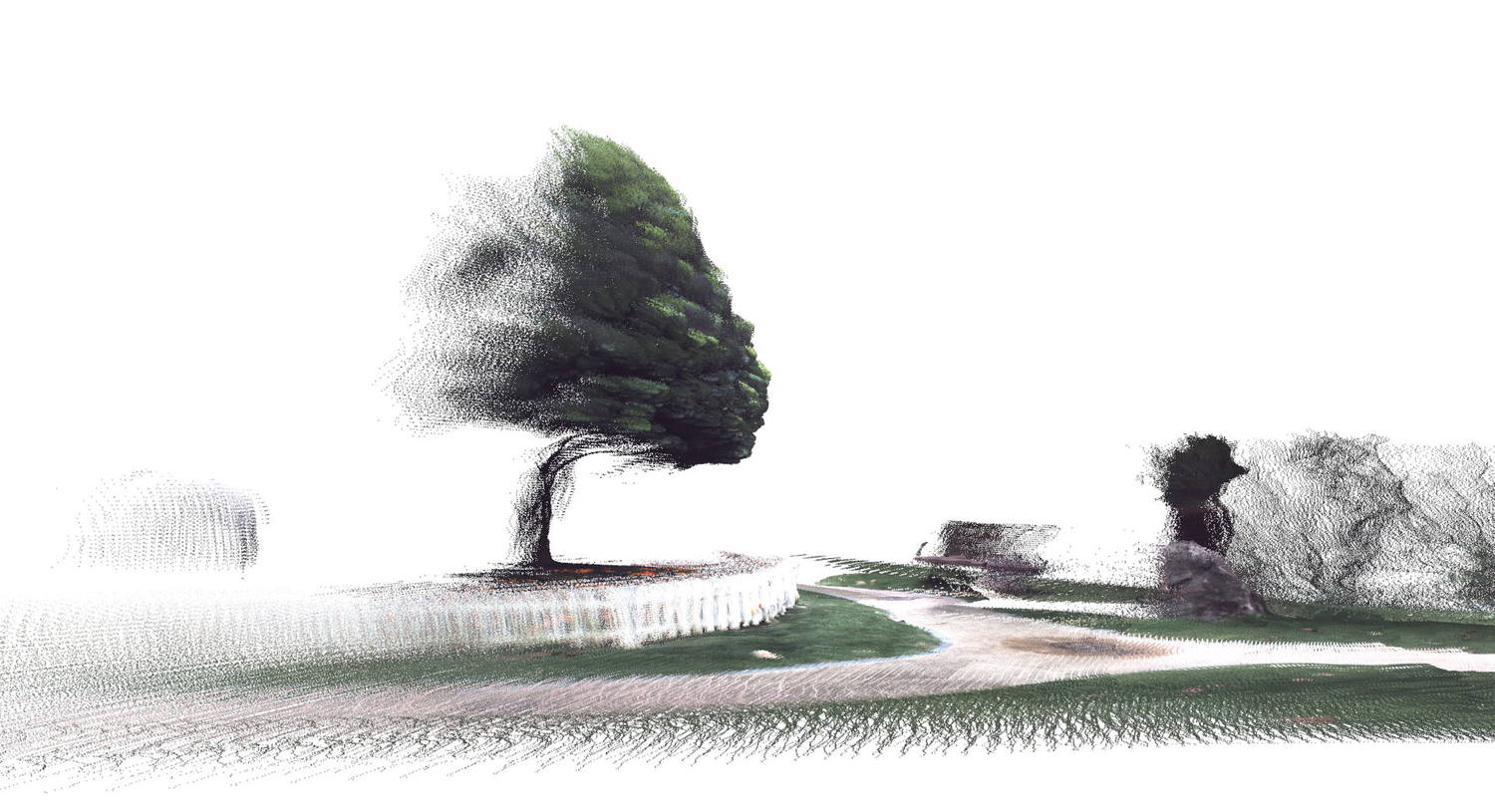}} \\[1pt] 
\textbf{MapAnything} & \raisebox{-0.5\height}{\includegraphics[width=0.19\textwidth]{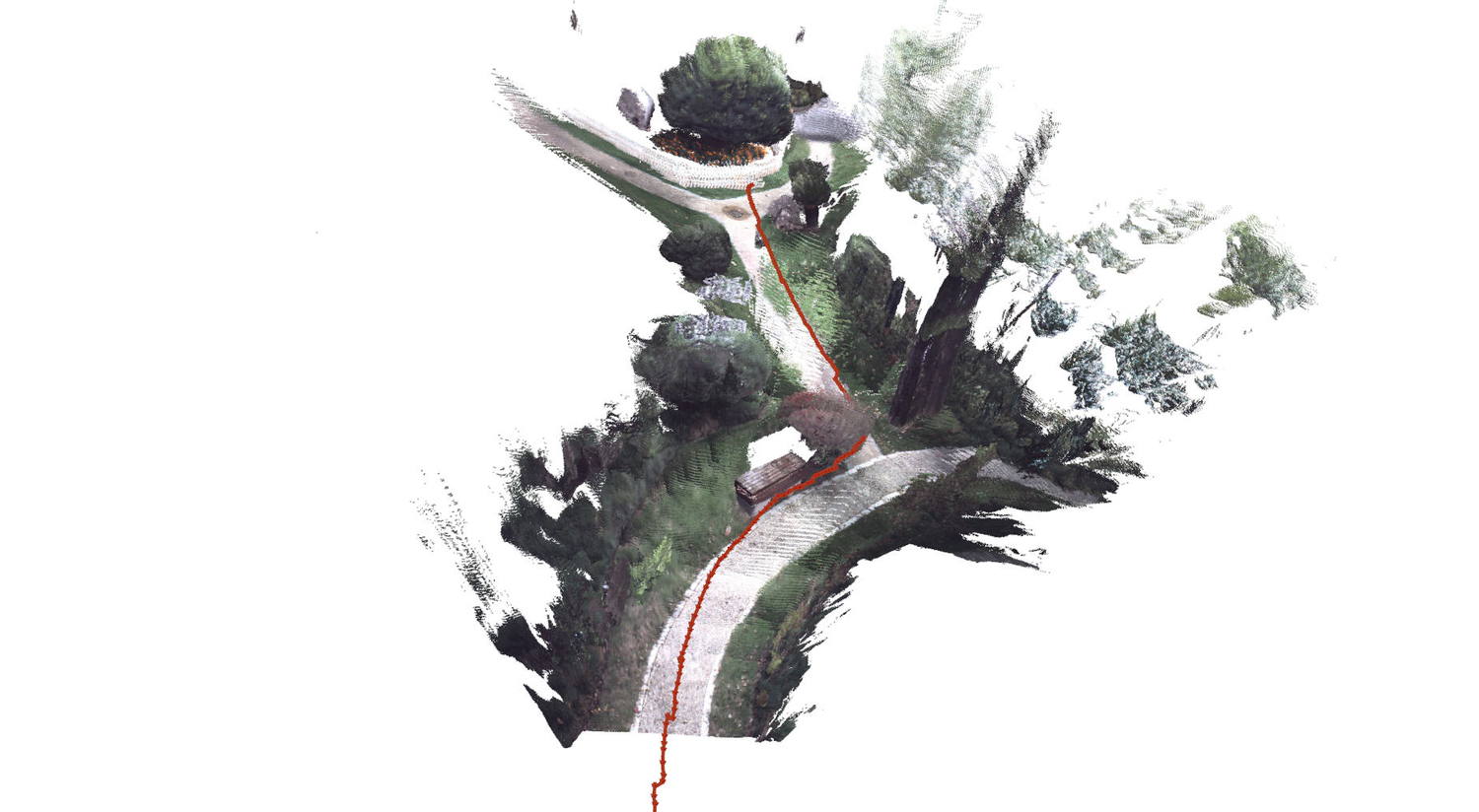}} & \raisebox{-0.5\height}{\includegraphics[width=0.19\textwidth]{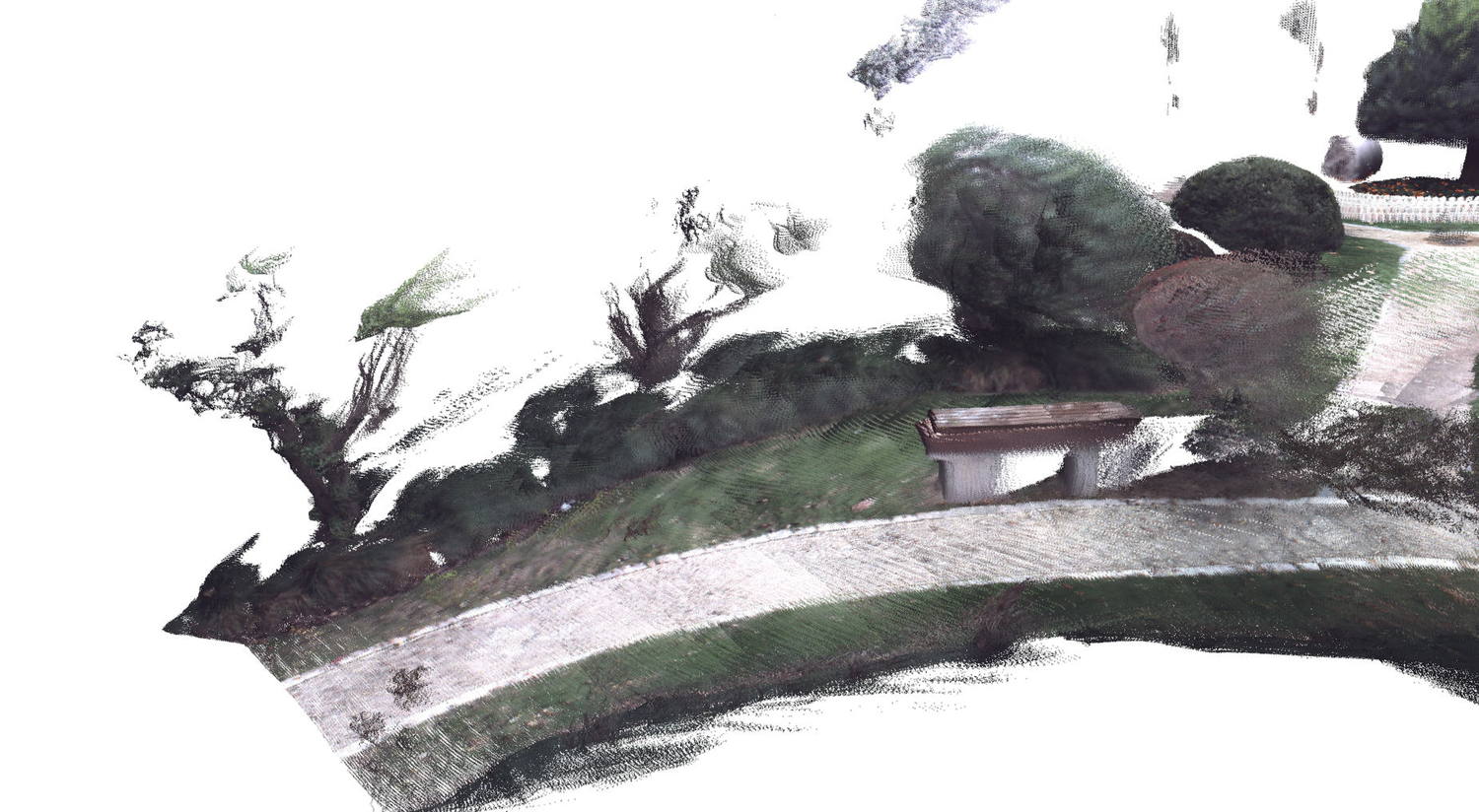}} & \raisebox{-0.5\height}{\includegraphics[width=0.19\textwidth]{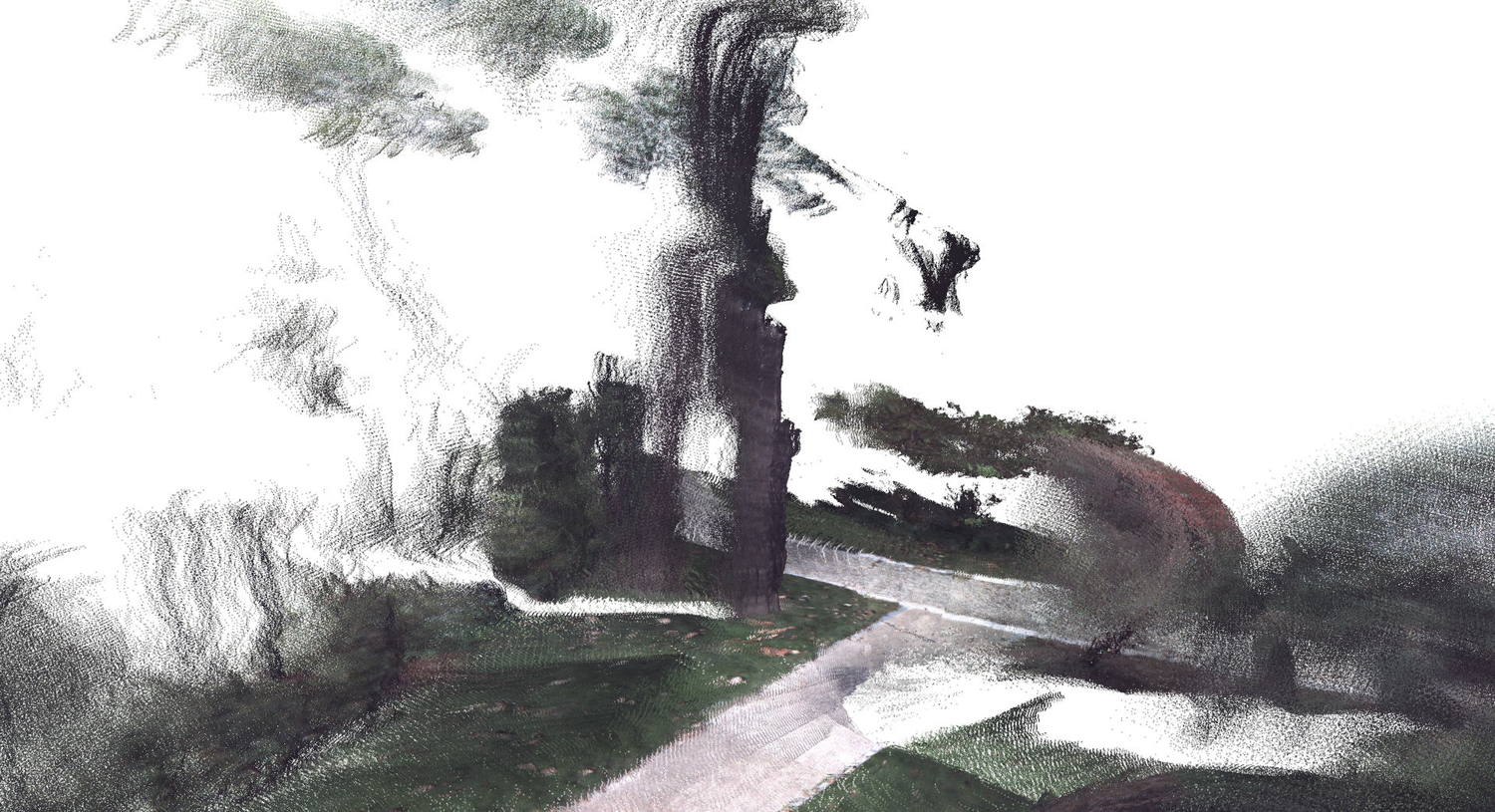}} & \raisebox{-0.5\height}{\includegraphics[width=0.19\textwidth]{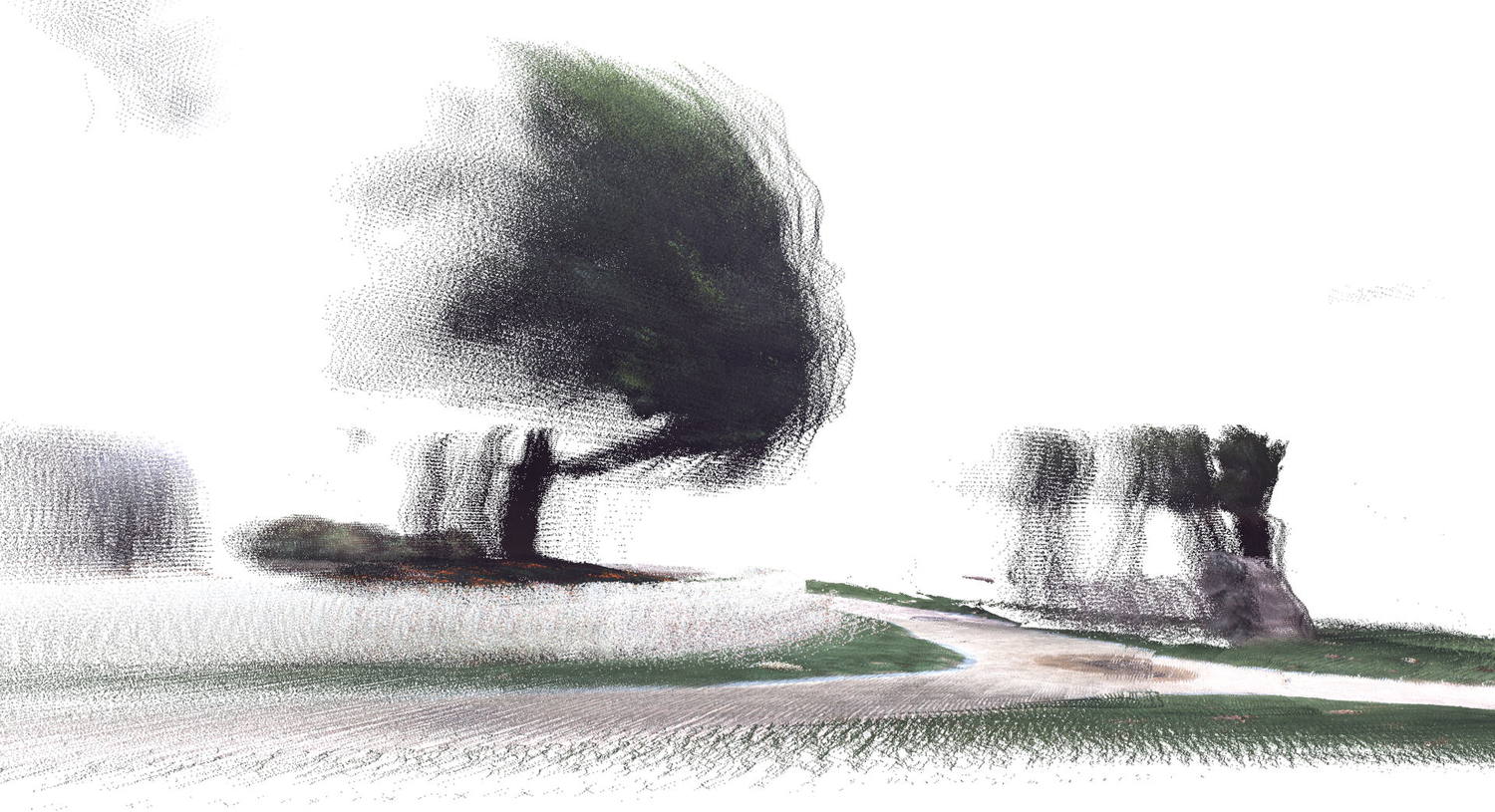}} \\[1pt] 
\textbf{MASt3R-SfM} & \raisebox{-0.5\height}{\includegraphics[width=0.19\textwidth]{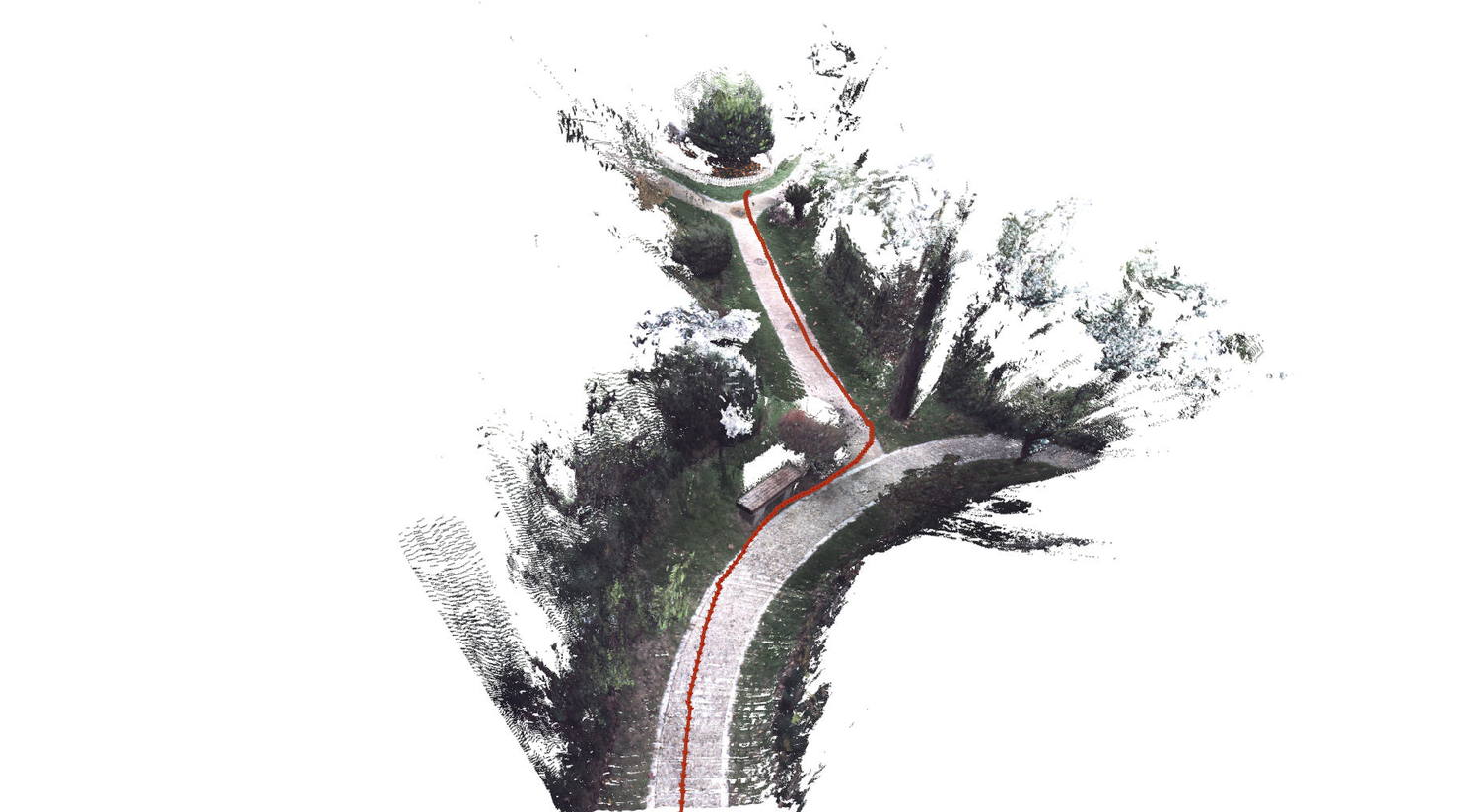}} & \raisebox{-0.5\height}{\includegraphics[width=0.19\textwidth]{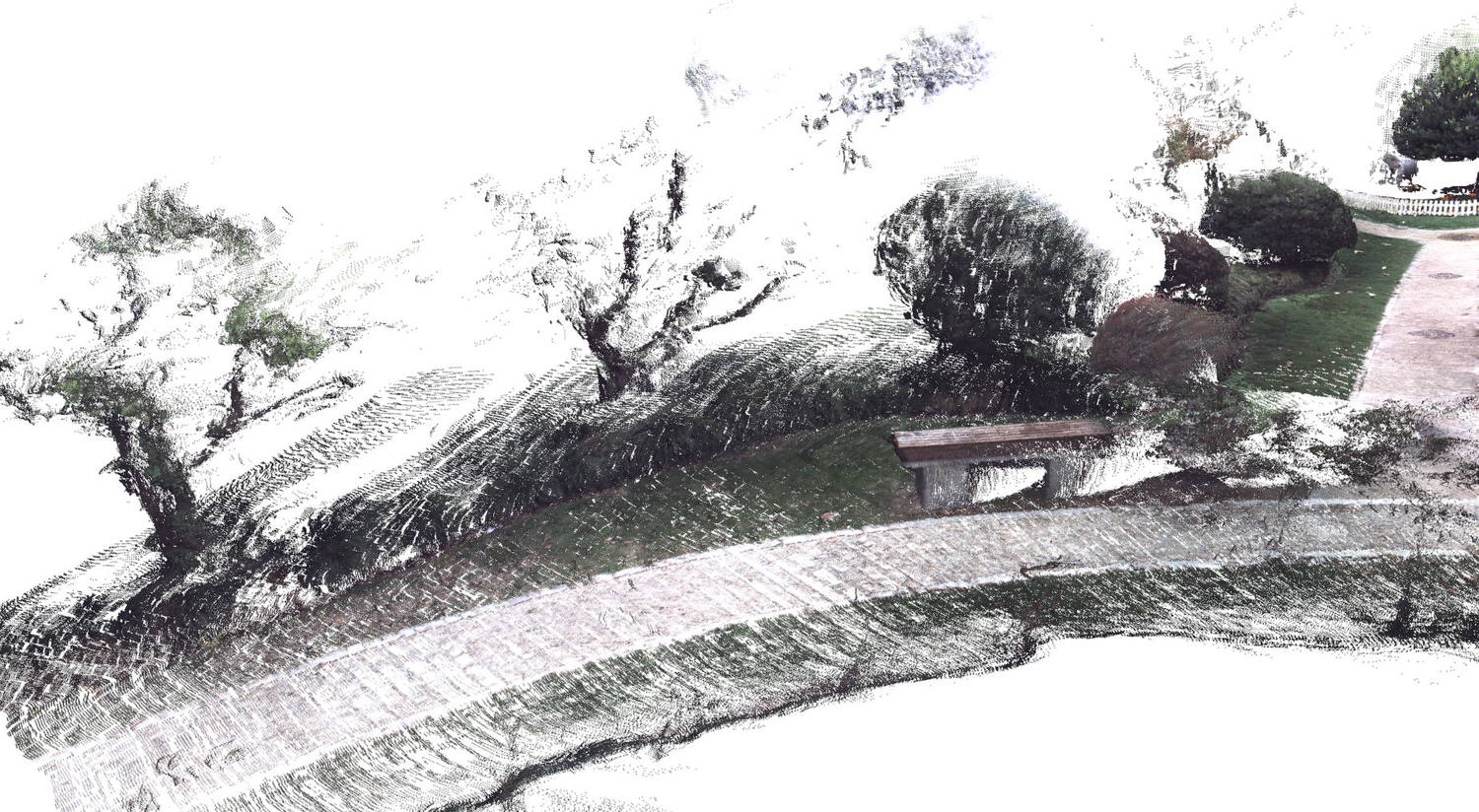}} & \raisebox{-0.5\height}{\includegraphics[width=0.19\textwidth]{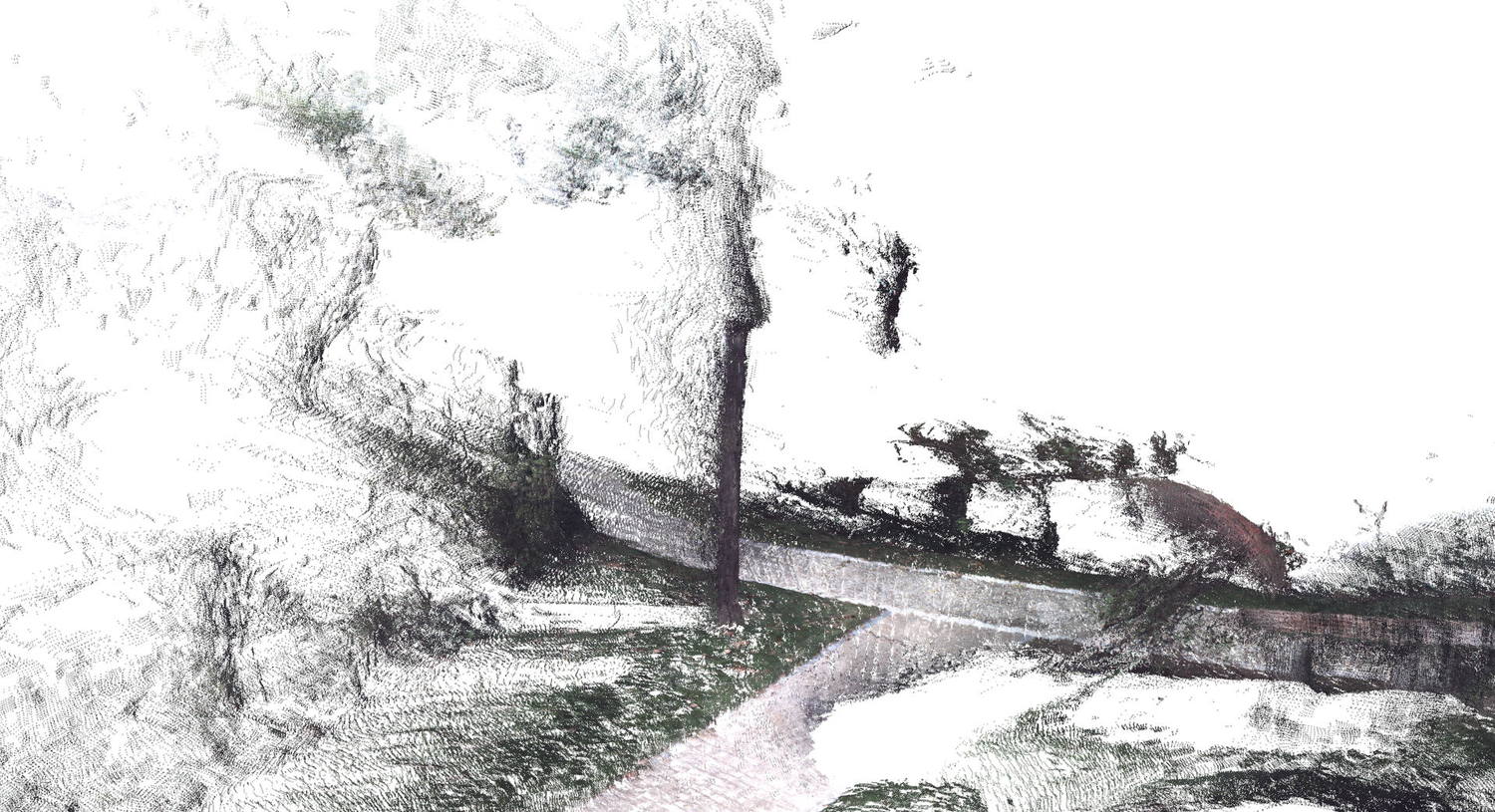}} & \raisebox{-0.5\height}{\includegraphics[width=0.19\textwidth]{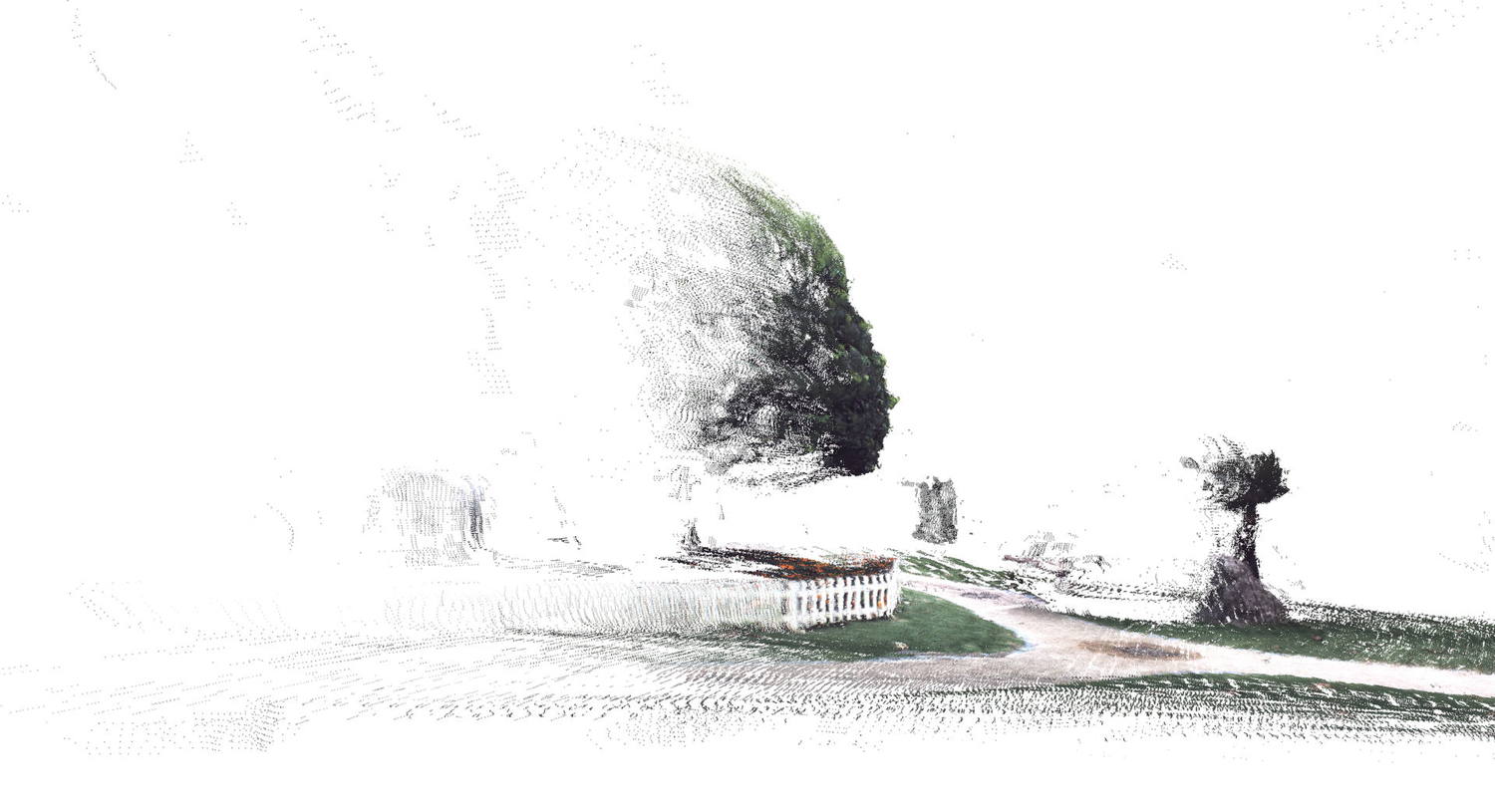}} \\[1pt] 
\textbf{SIFT-SfM} & \raisebox{-0.5\height}{\includegraphics[width=0.19\textwidth]{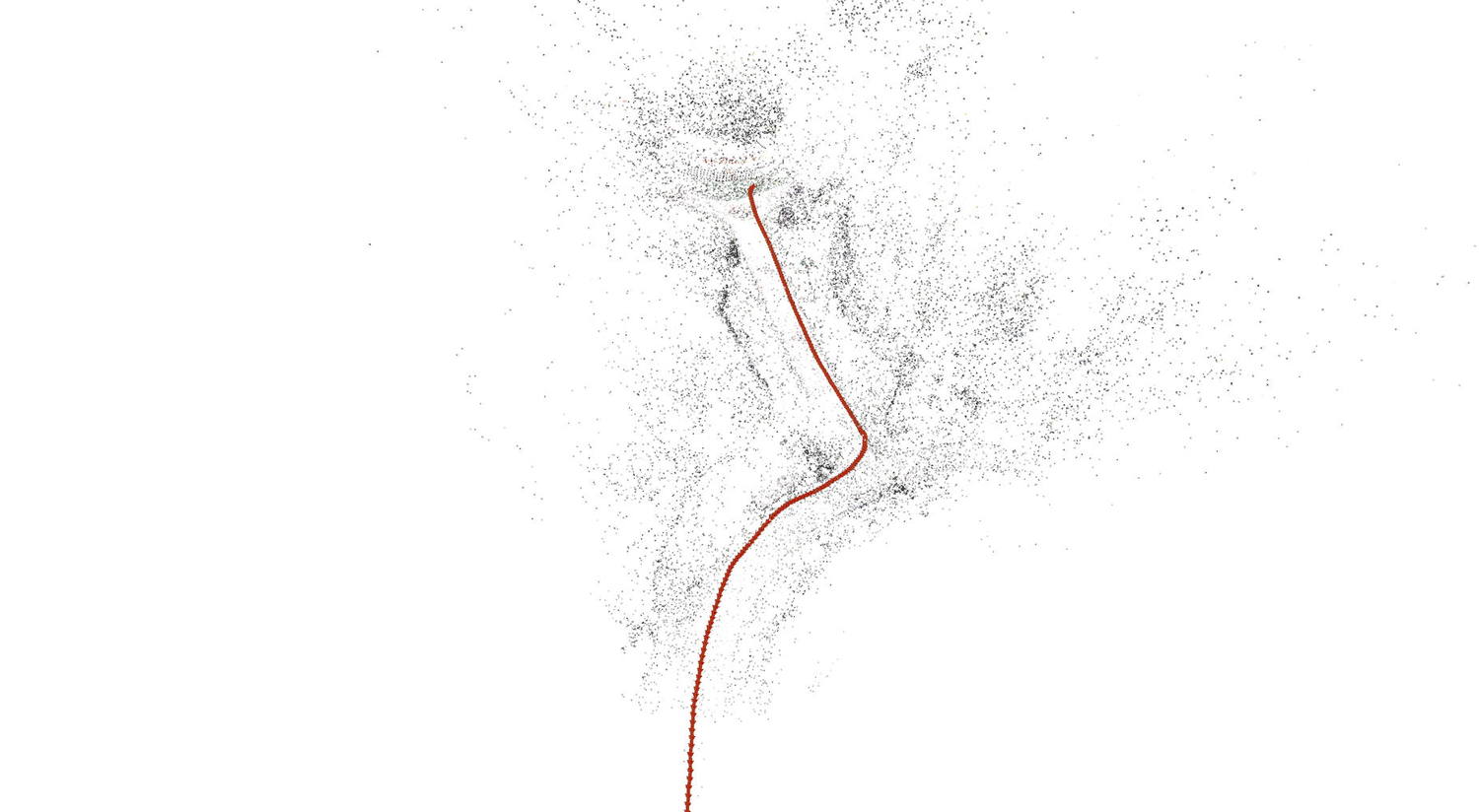}} & \raisebox{-0.5\height}{\includegraphics[width=0.19\textwidth]{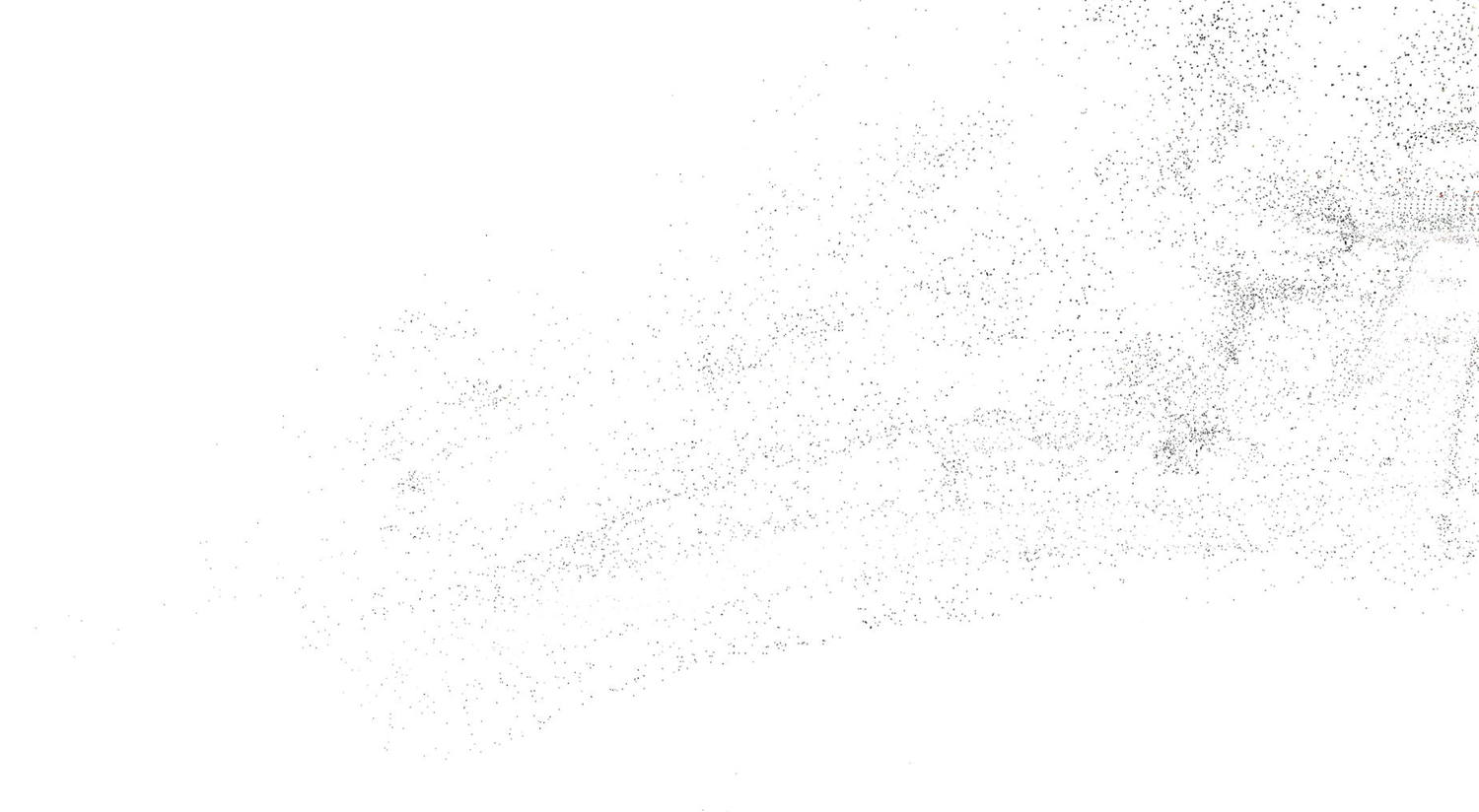}} & \raisebox{-0.5\height}{\includegraphics[width=0.19\textwidth]{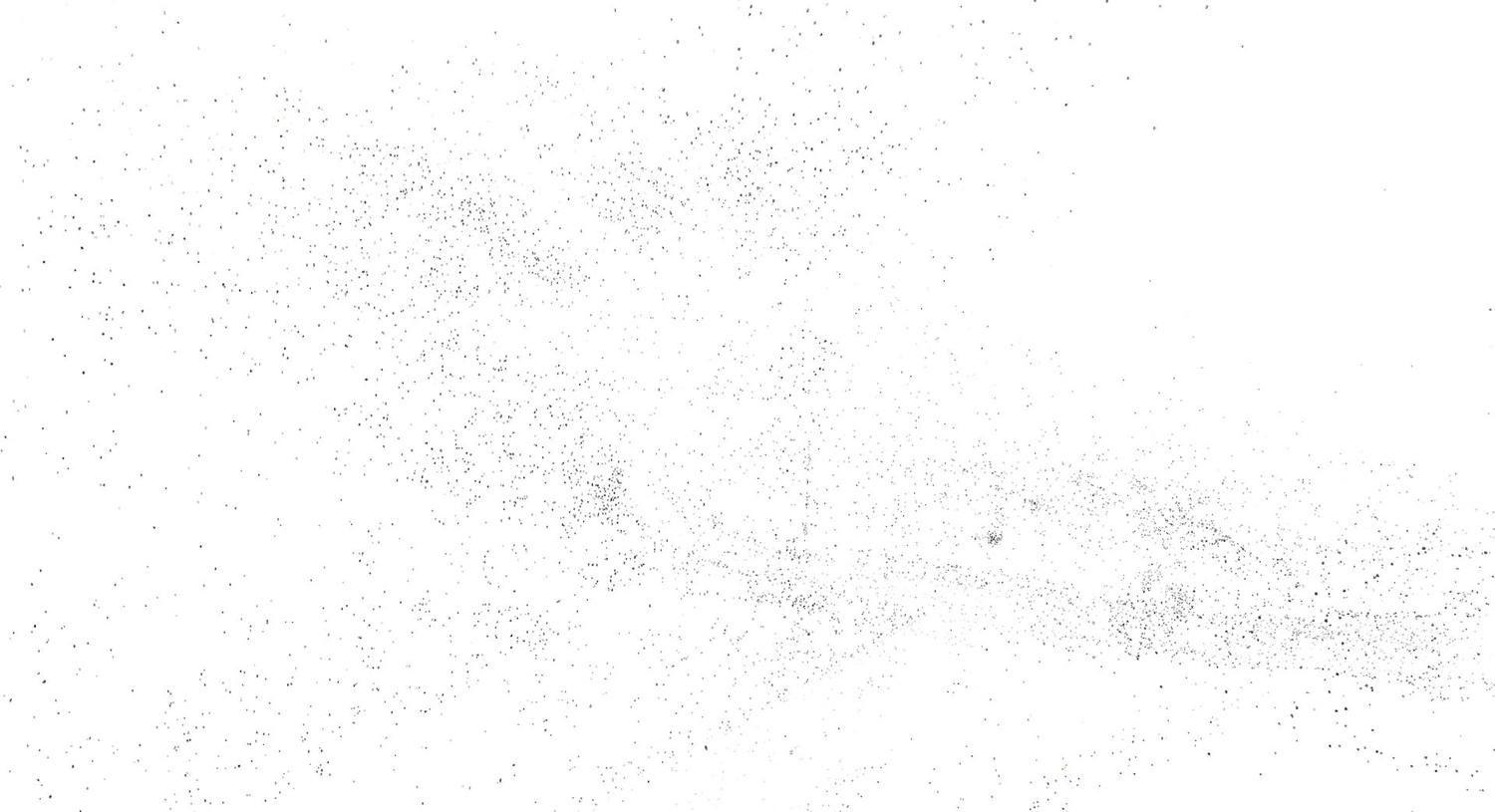}} & \raisebox{-0.5\height}{\includegraphics[width=0.19\textwidth]{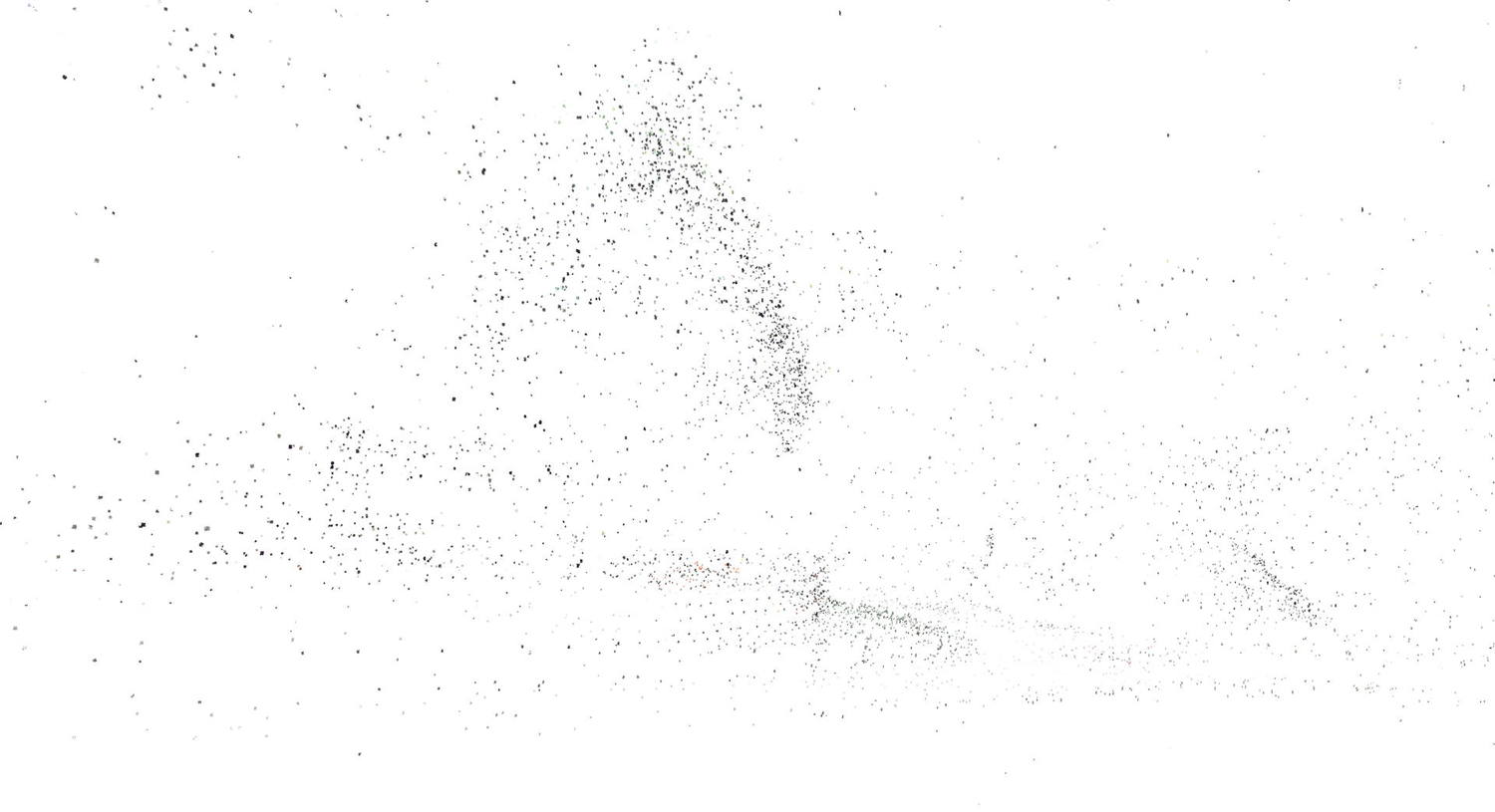}} \\[1pt] 
\textbf{SuperGlue-SfM} & \raisebox{-0.5\height}{\includegraphics[width=0.19\textwidth]{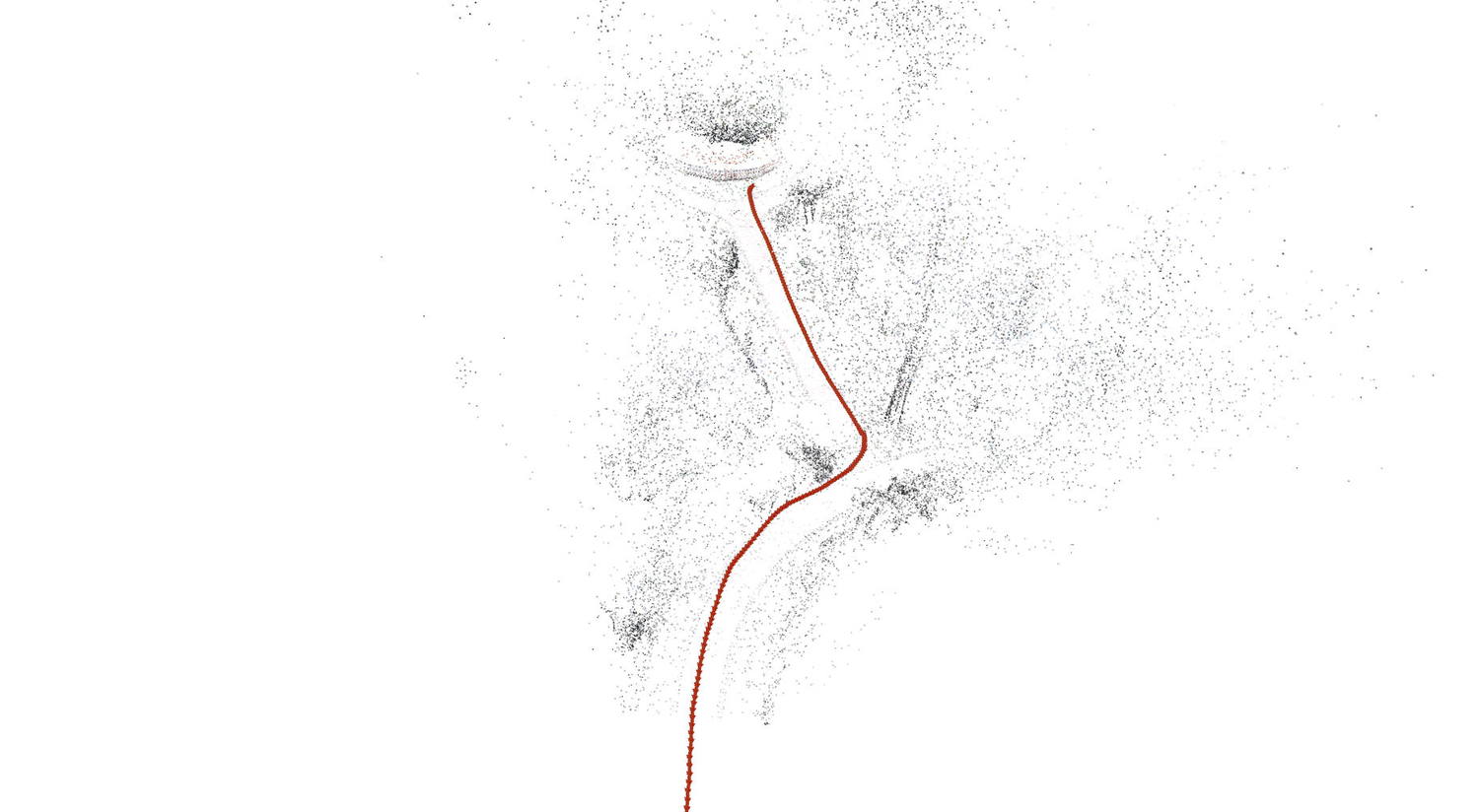}} & \raisebox{-0.5\height}{\includegraphics[width=0.19\textwidth]{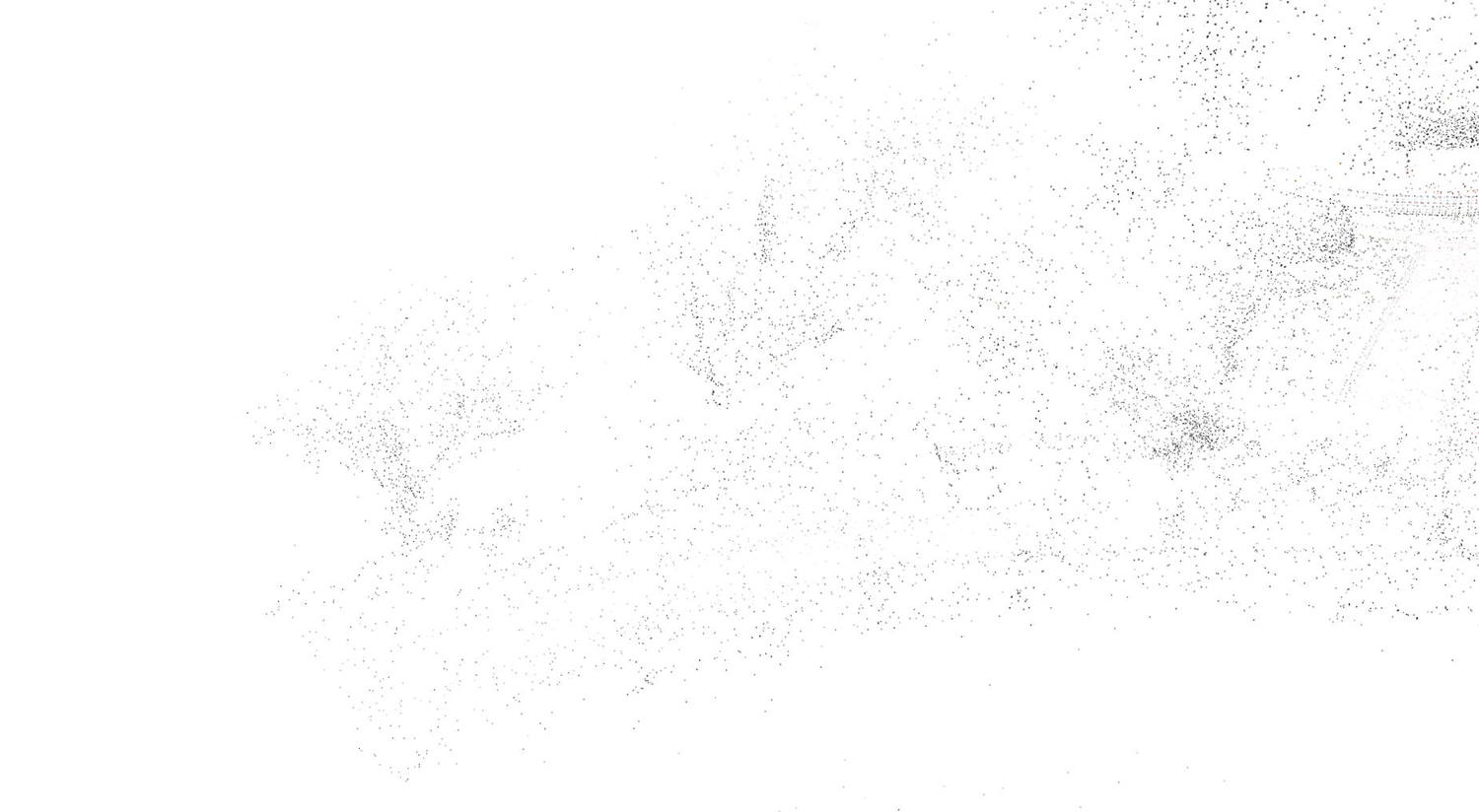}} & \raisebox{-0.5\height}{\includegraphics[width=0.19\textwidth]{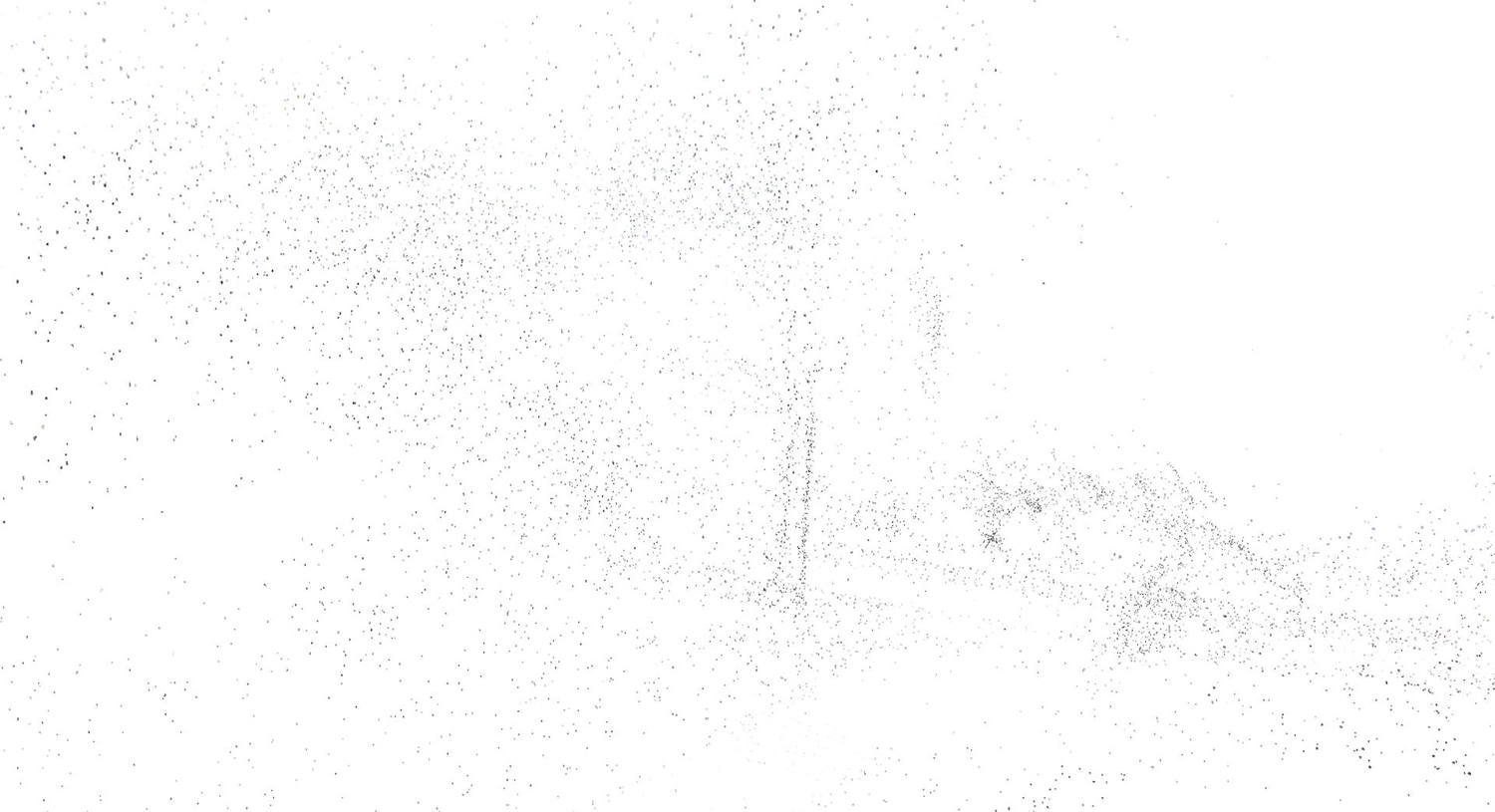}} & \raisebox{-0.5\height}{\includegraphics[width=0.19\textwidth]{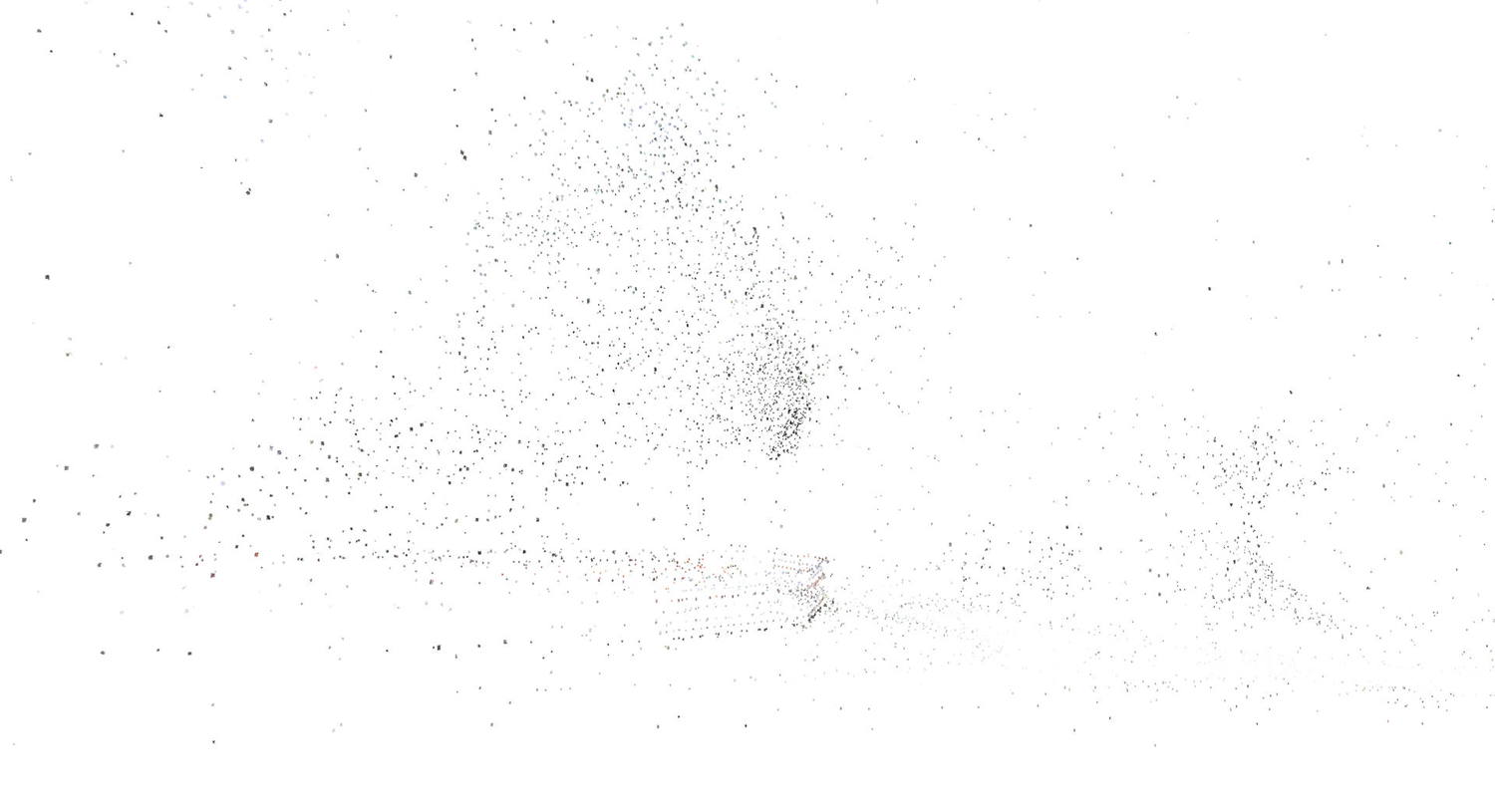}} \\[1pt] 
\textbf{LoFTR-SfM} & \raisebox{-0.5\height}{\includegraphics[width=0.19\textwidth]{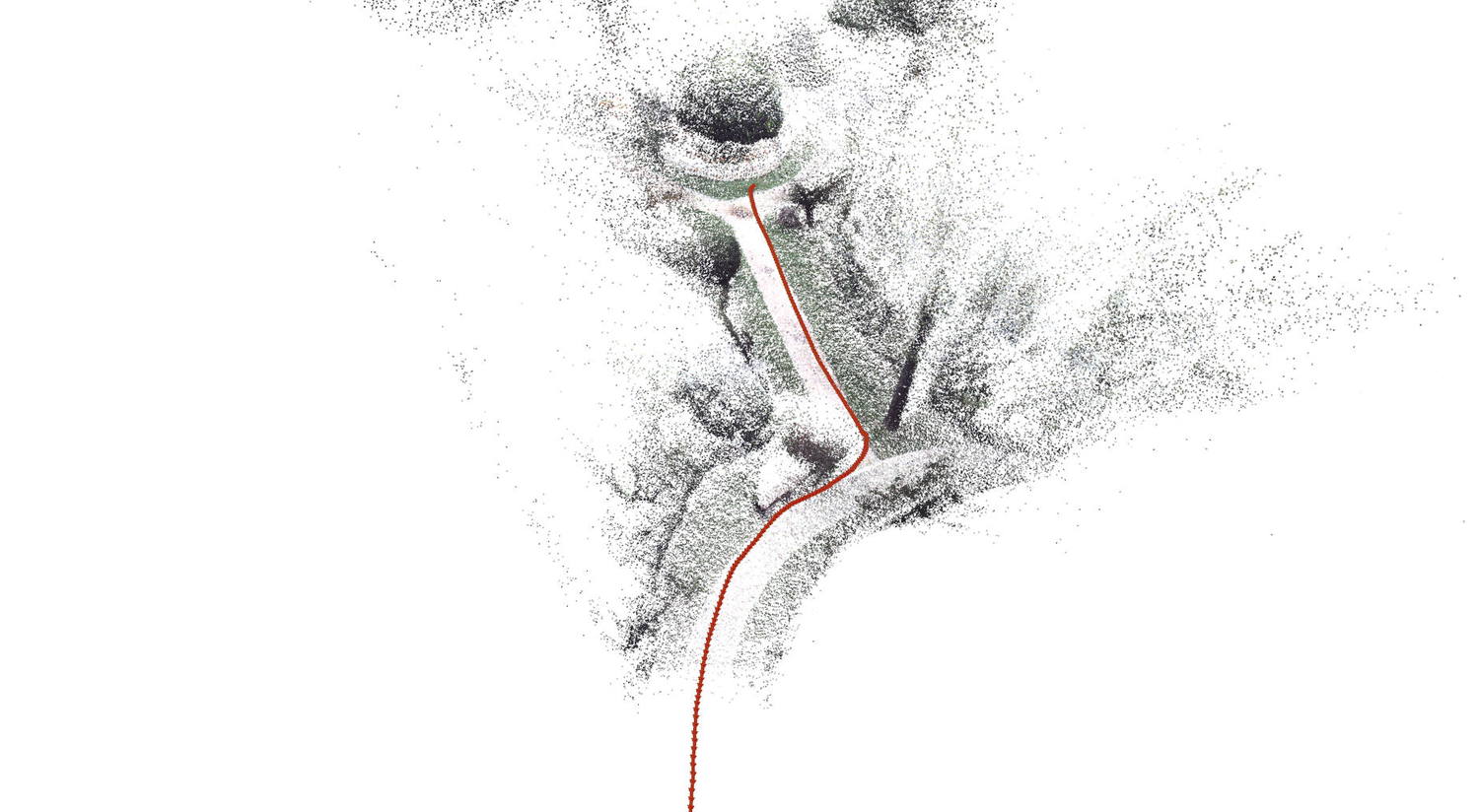}} & \raisebox{-0.5\height}{\includegraphics[width=0.19\textwidth]{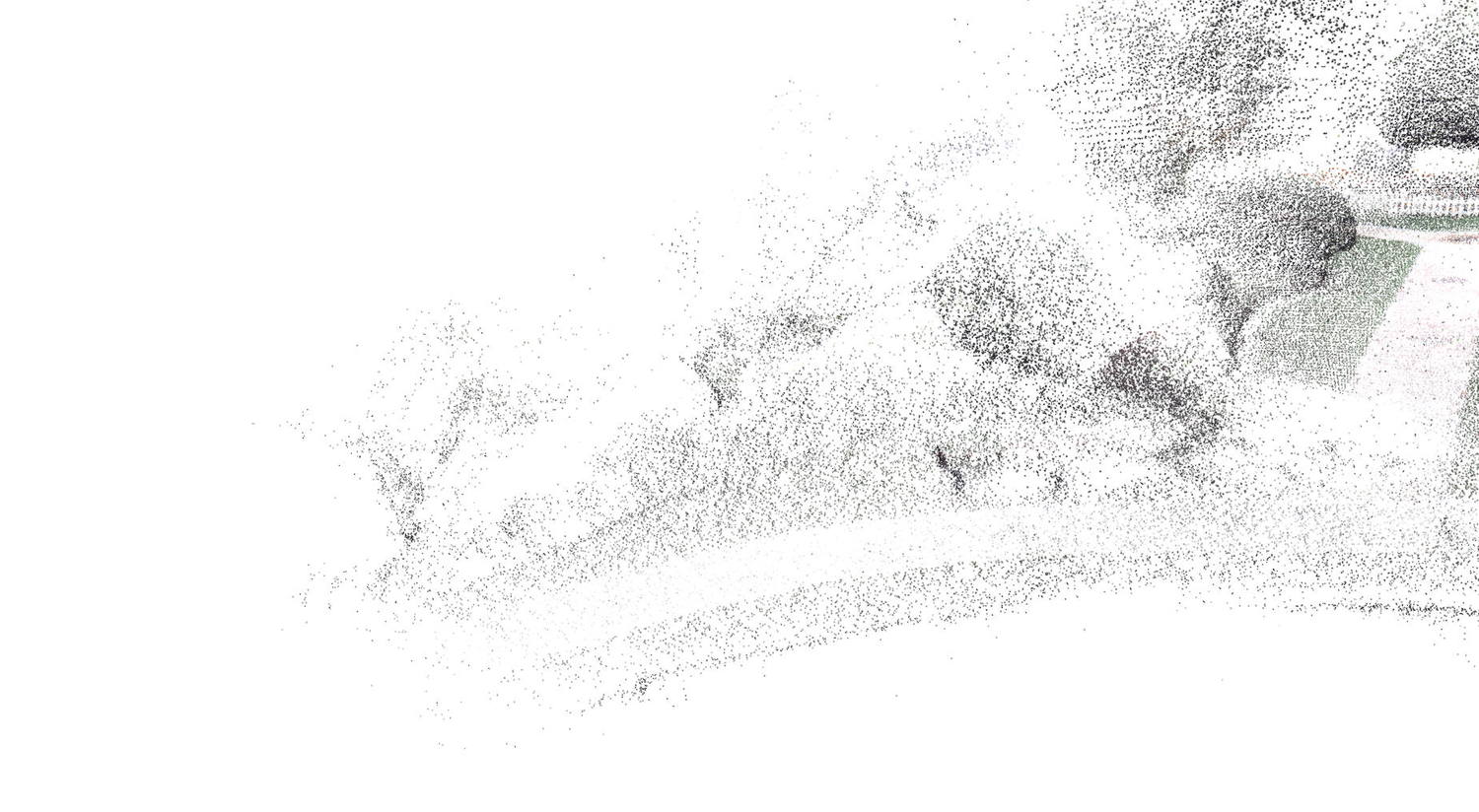}} & \raisebox{-0.5\height}{\includegraphics[width=0.19\textwidth]{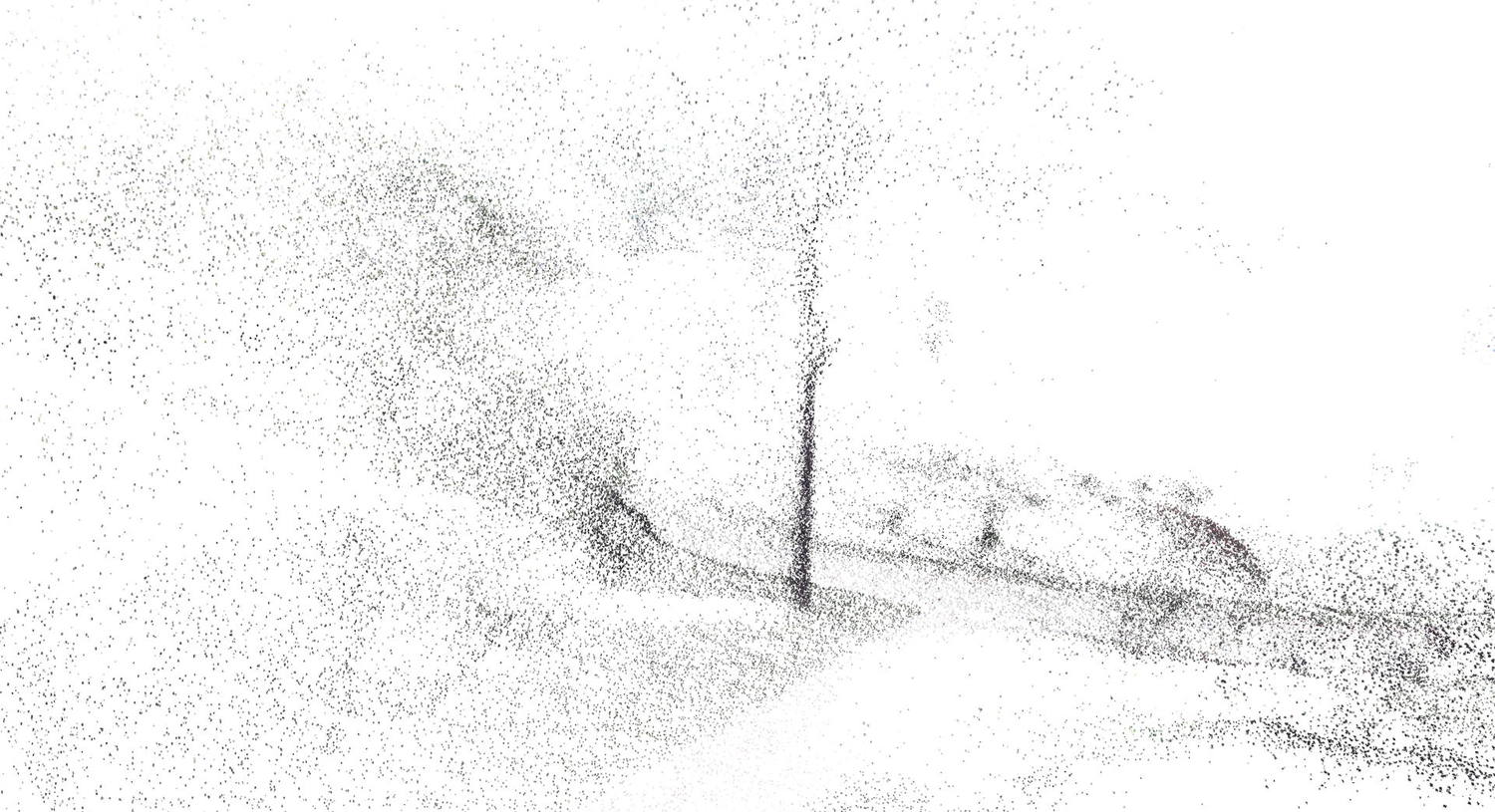}} & \raisebox{-0.5\height}{\includegraphics[width=0.19\textwidth]{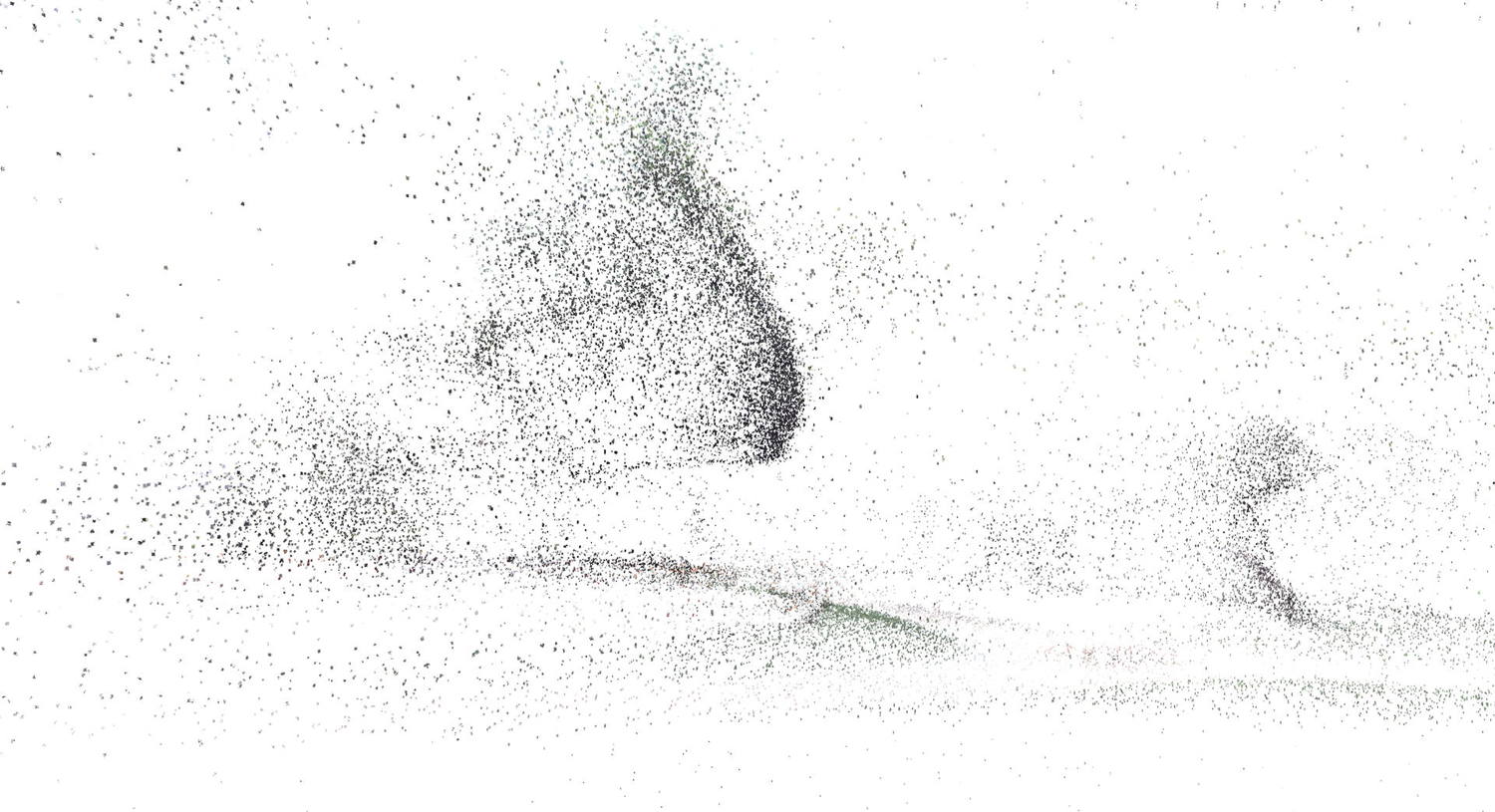}} \\[1pt] 
\textbf{LoMa-SfM} & \raisebox{-0.5\height}{\includegraphics[width=0.19\textwidth]{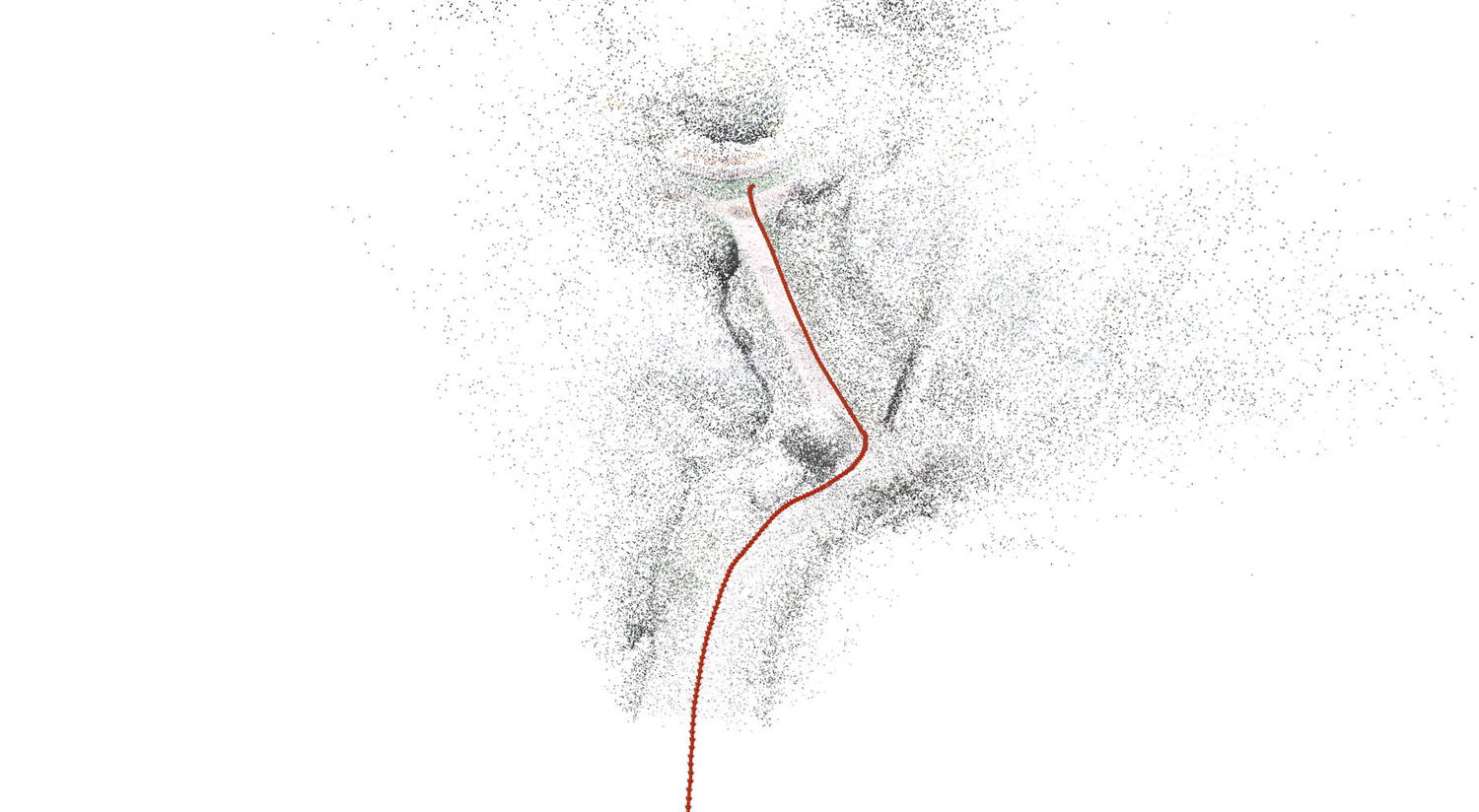}} & \raisebox{-0.5\height}{\includegraphics[width=0.19\textwidth]{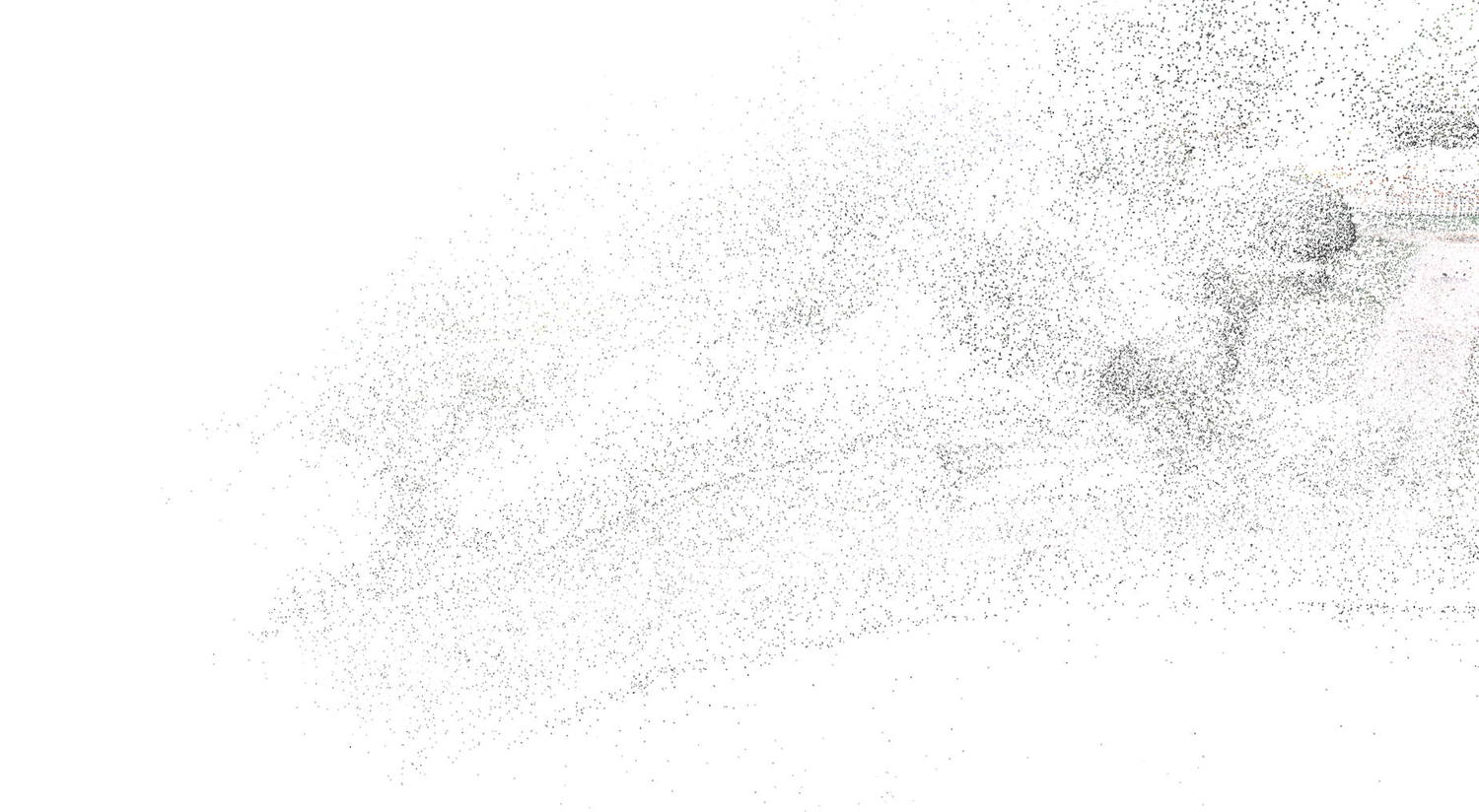}} & \raisebox{-0.5\height}{\includegraphics[width=0.19\textwidth]{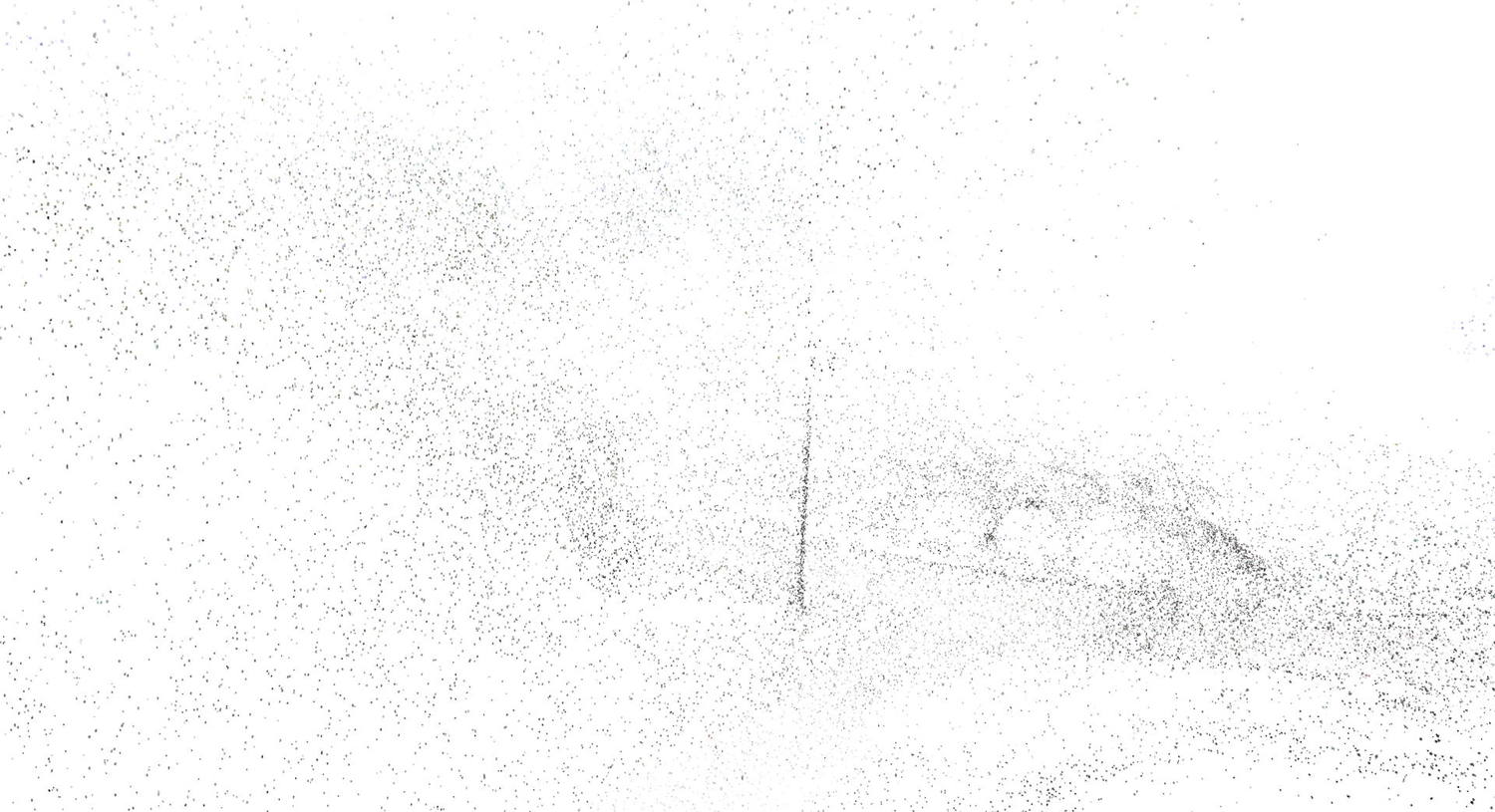}} & \raisebox{-0.5\height}{\includegraphics[width=0.19\textwidth]{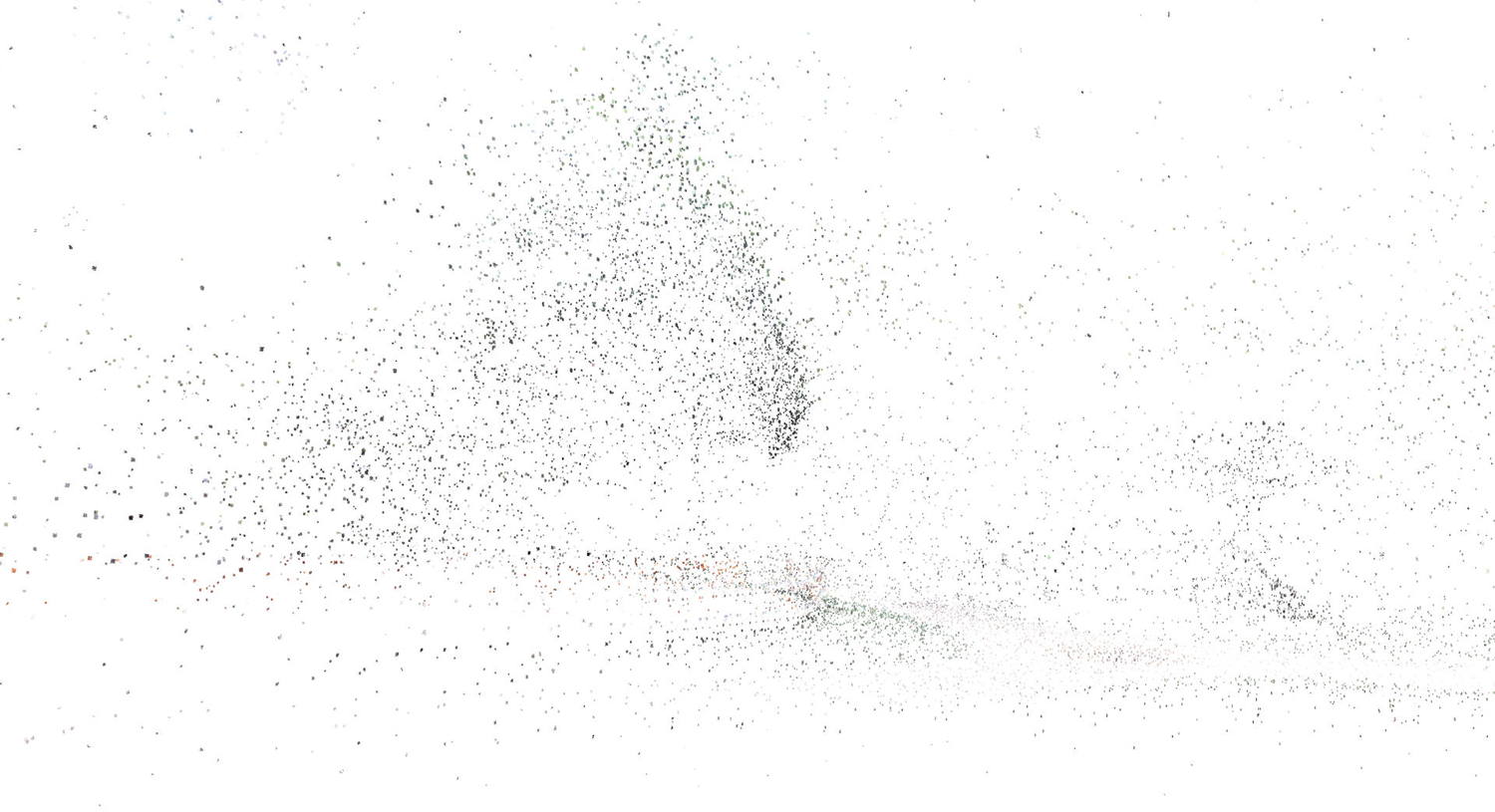}} \\[1pt] 
\textbf{RoMa-v2-SfM} & \raisebox{-0.5\height}{\includegraphics[width=0.19\textwidth]{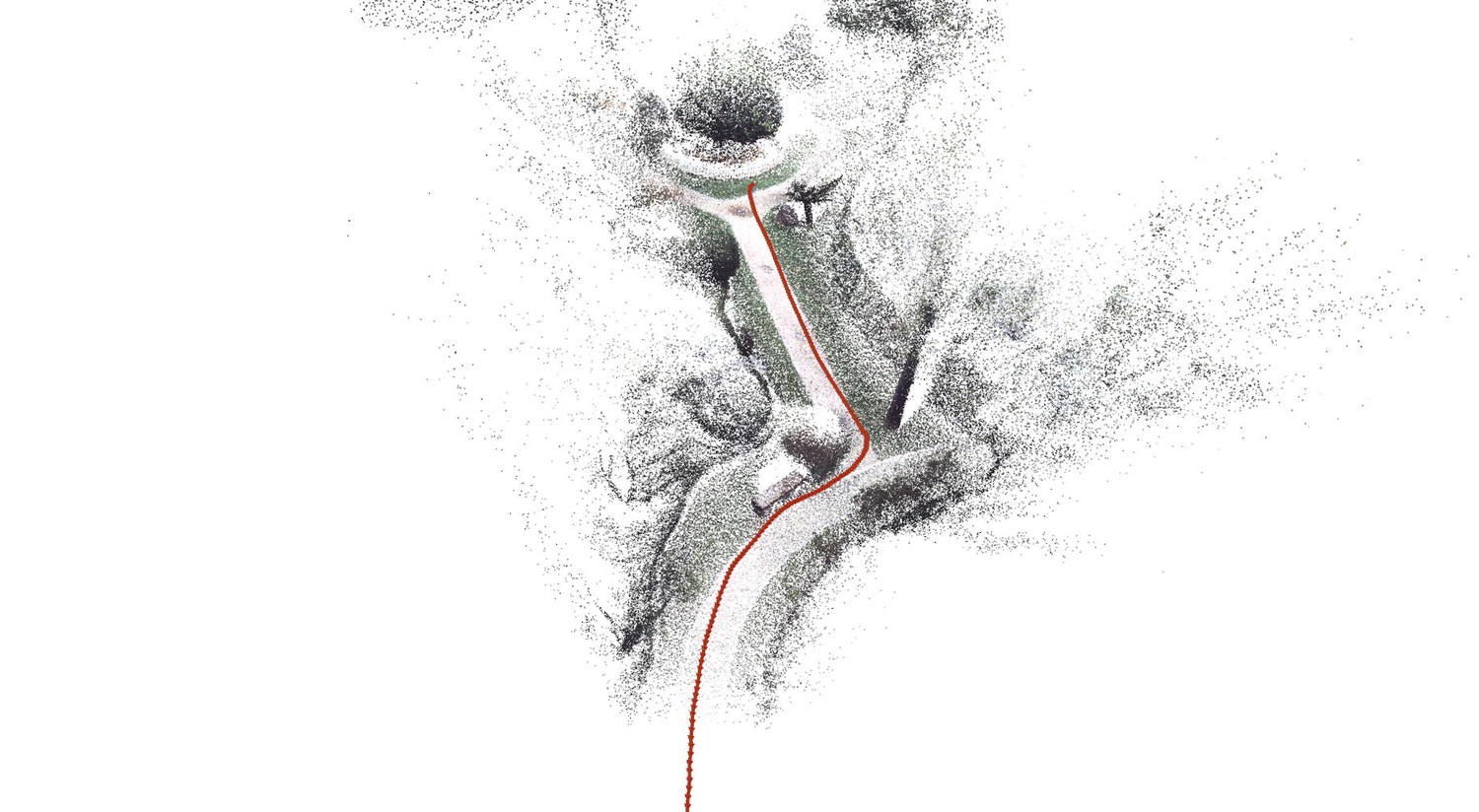}} & \raisebox{-0.5\height}{\includegraphics[width=0.19\textwidth]{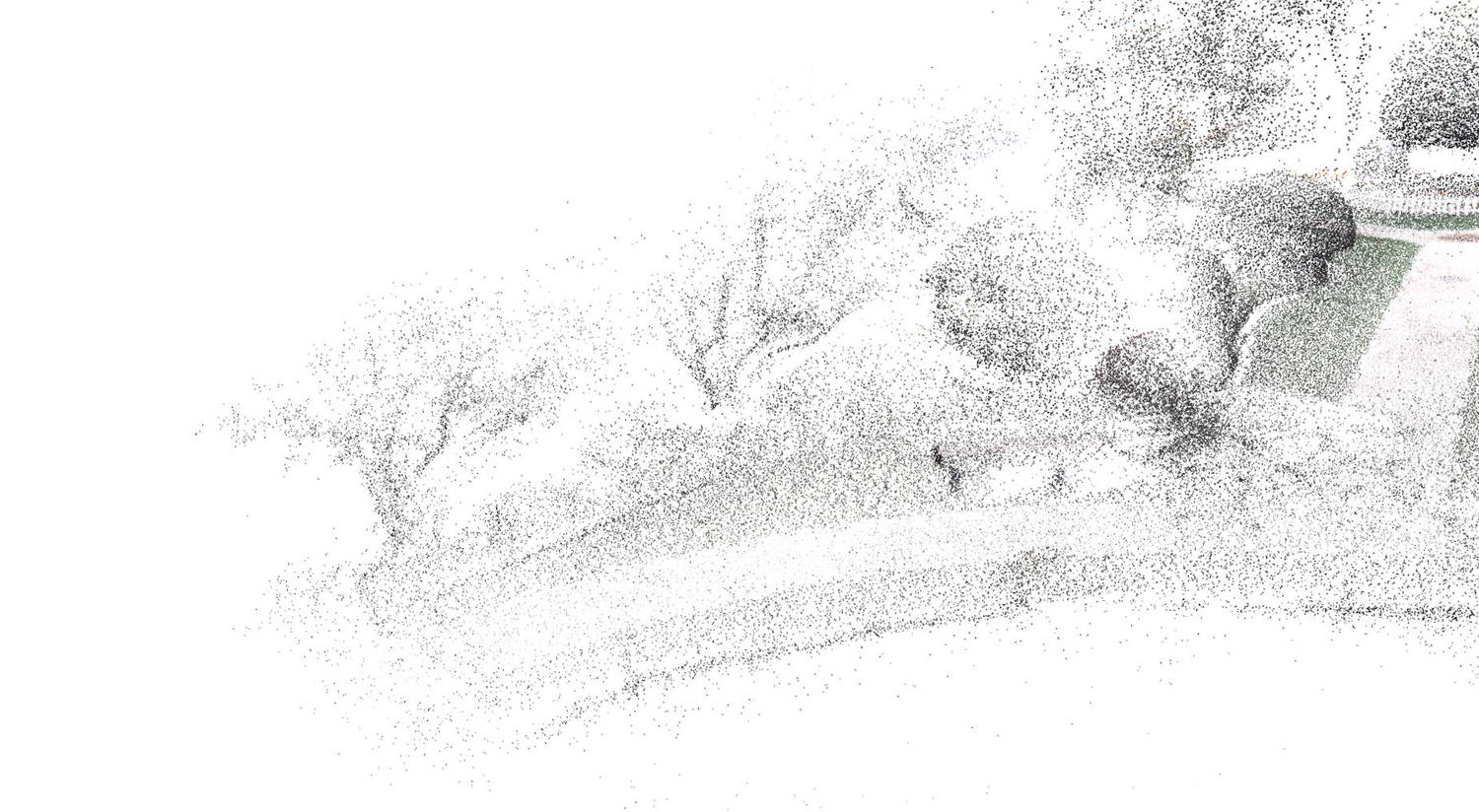}} & \raisebox{-0.5\height}{\includegraphics[width=0.19\textwidth]{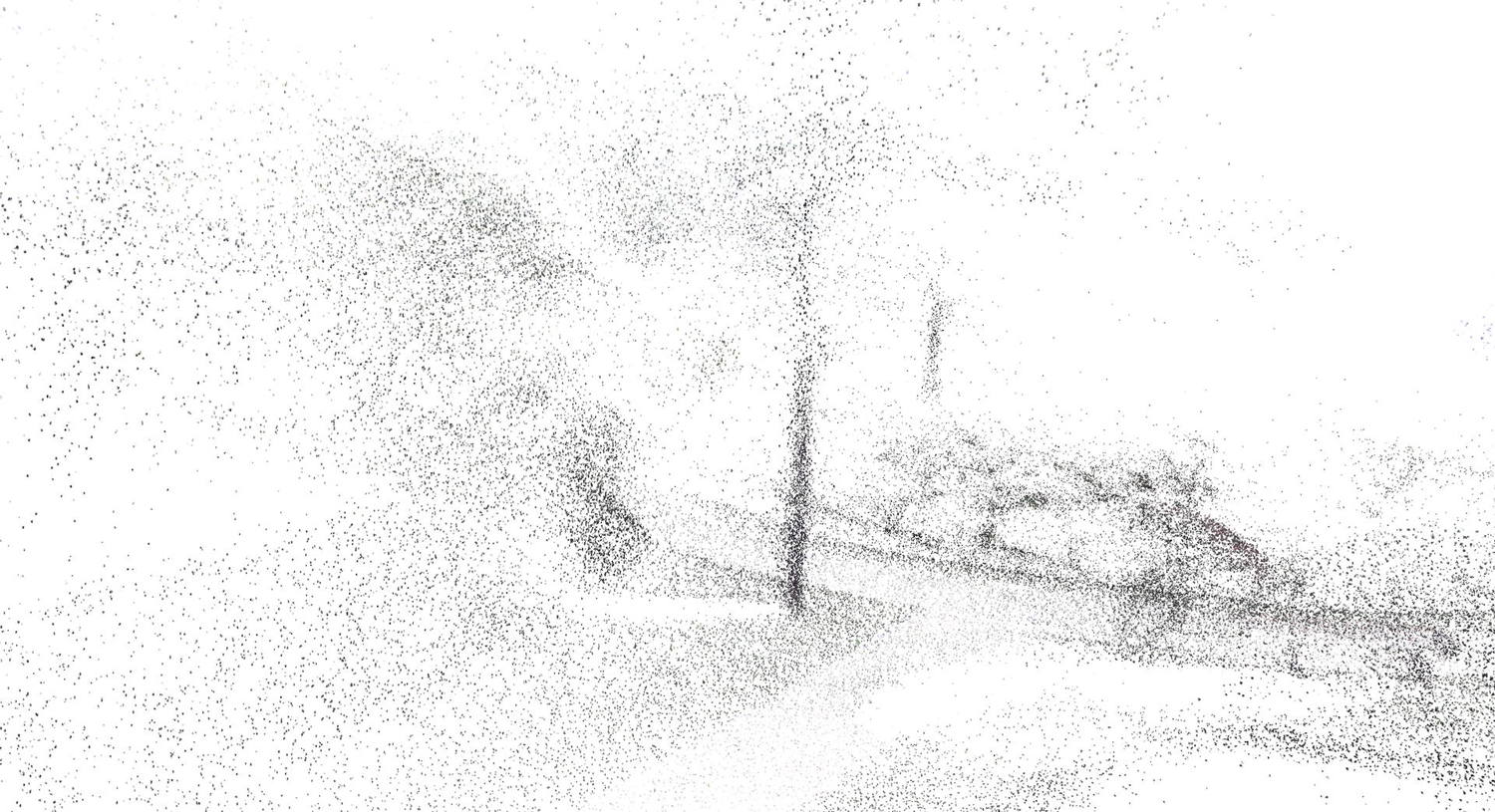}} & \raisebox{-0.5\height}{\includegraphics[width=0.19\textwidth]{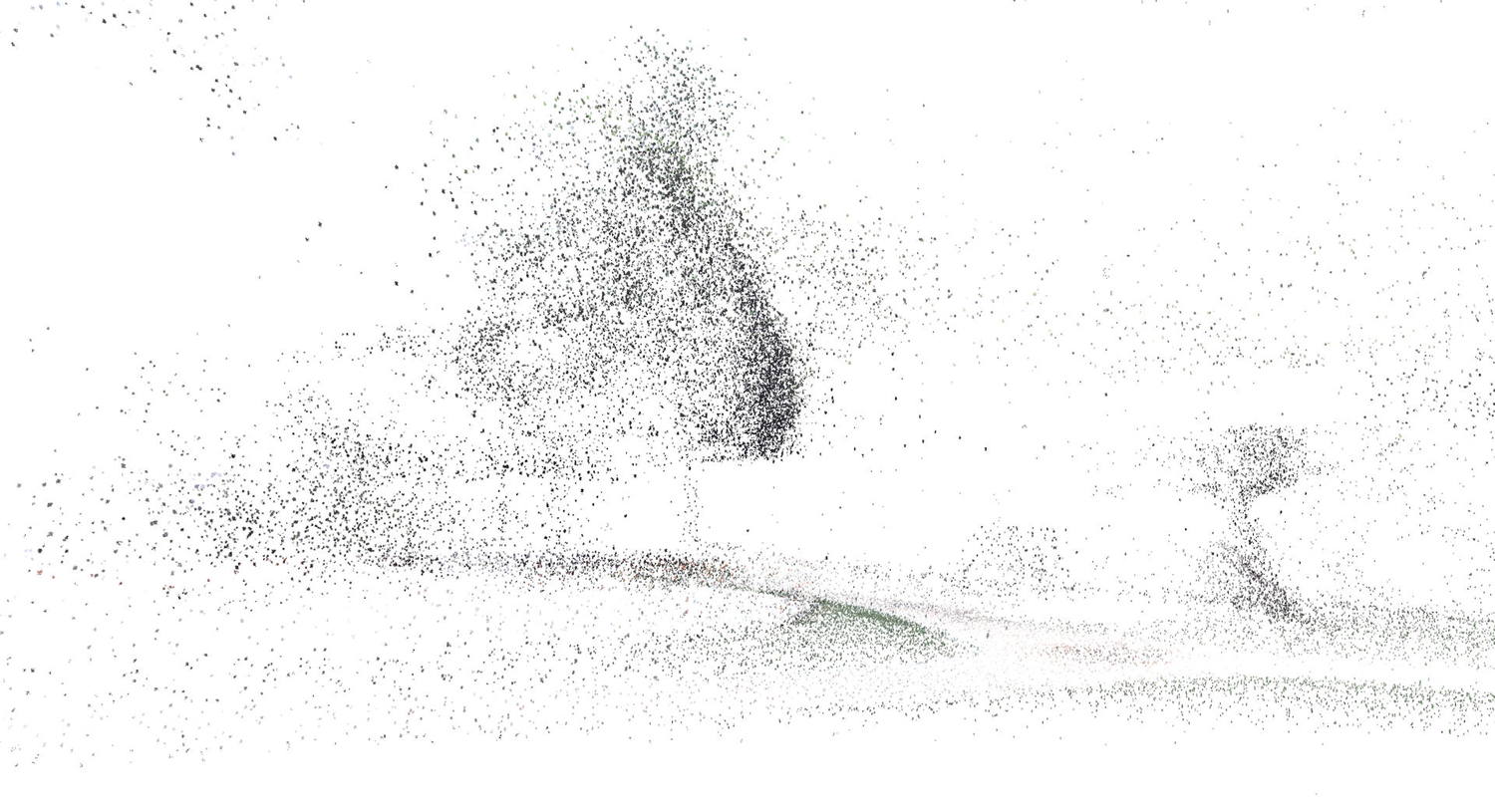}} \\[1pt] 
\bottomrule
\end{tabular}
\caption{Qualitative 3D reconstruction comparison on
\texttt{BotanicGarden\_1006-03}. Columns show four fixed
viewpoints of the same scene. The first row is the reference point cloud,
followed by the feed-forward and SfM reconstruction methods. All methods
use monocular input. VGGT,
VGGT-$\Omega$, DA3, MapAnything, and MASt3R-SfM are shown
with confidence-P50 masking. All predictions
are shown after alignment to the reference coordinate frame.}
\label{fig:reconstruction-qualitative-comparison}
\end{figure*}

Table~\ref{tab:3D-reconstruction-comparison-selected} compares the selected
feed-forward and SfM configurations. Among the aligned feed-forward methods,
VGGT-$\Omega$ provides the best completeness, Chamfer distance, and runtime.
Its confidence-P50 variant gives the lowest retained-point accuracy error and
the highest threshold accuracy and F1 score in this group. Confidence masking
improves retained-point accuracy for all four feed-forward models, but reduces
density and completeness.

For aligned conventional SfM, RoMa-v2(S) achieves the strongest
geometric accuracy, completeness, Chamfer distance, and F1 score, but succeeds
on only seven of the ten sequences. Among conventional SfM configurations
evaluated on all ten sequences, SIFT(S) gives the lowest accuracy error,
whereas LoFTR(K) gives the lowest completeness error and Chamfer distance.
SIFT(K) has the lowest runtime but produces much sparser geometry.
MASt3R(K) retains substantially denser support than conventional SfM; its
confidence mask improves accuracy while reducing density and completeness.

For unaligned calibrated-stereo reconstruction, RoMa-v2(S) again gives the
strongest geometric results with partial sequence coverage. Among the
configurations evaluated on all ten sequences, SIFT(S) has the lowest accuracy
error, whereas LoFTR(S) provides
the best completeness, Chamfer distance, and F1 score. Applying scene-level
alignment improves threshold accuracy and F1 for every calibrated-stereo
method, although not every distance metric improves.

Fig.~\ref{fig:reconstruction-qualitative-comparison} shows complementary
reconstruction failure modes. The feed-forward methods recover dense, continuous
ground, paths, and vegetation, but Views~3 and~4 contain repeated trunks, crowns,
hedges, and path boundaries. These repetitions arise because independently predicted
depth and pose are not fully consistent across frames, so observations of the same
surface do not coincide when fused. Although the applicable cross-frame consistency
filter removes points without sufficient reprojection support, a locally coherent
repeated surface can still satisfy the test. Conventional SfM instead reconstructs only points supported by
multi-view feature tracks and triangulation. Stable correspondences are difficult to
establish in vegetation with repetitive appearance and viewpoint changes, resulting in
sparse geometry; mismatches and unstable triangulation also produce isolated points and
diffuse clusters. Its geometrically verified landmarks nevertheless form a more coherent
scene with fewer duplicated structures. MASt3R-SfM lies between the two families: its
predicted point maps provide denser geometry than conventional triangulation, but their
remaining multi-frame inconsistencies produce repeated or rippled surfaces.

\subsubsection{Dense-Depth Estimation Comparison}

monocular, stereo, and multi-view depth-estimation benchmark comparison are listed in
Table~\ref{tab:depth-estimation-comparison}. Their outputs are standardized
as camera-$z$ depth under the common protocol in Supplementary Section~\ref{supp:common-experiment-settings}.
Following \cite{MiDas,DA2}, we apply per-frame affine depth alignment before
evaluation, as detailed in Supplementary Section~\ref{supp:depth-alignment}.
For the direct comparison of reliable depth, we use the 10-m near-range
outputs of DA2, MoGe-2, SGM, FoundationStereo, 
and the confidence-P50-masked outputs of VGGT, VGGT-$\Omega$, 
and MapAnything.

In fig.~\ref{fig:depth-qualitative-comparison}, we show both Full Depth and Filtered Depth, with the latter
retaining predictions after method-specific filtering. An individual depth map cannot fully reveal
whether a depth-estimation method produces geometrically consistent
reconstructions across a sequence. Motivated by MonoRec \cite{MonoRec}, we
therefore back-project the retained depth maps,
accumulate the resulting point clouds across the sequence using the reference camera poses, 
and visualize the accumulated geometry from
two fixed viewpoints as 3D View 1 and 3D View 2. These views
provide a qualitative assessment of cross-frame depth consistency.

Most methods recover a reasonable relative-depth ordering in the image plane,
while SGM remains sparse and noisy. Their accumulated geometry is considerably
less consistent. Repeated tree trunks and ghost surfaces appear when the
estimated depth of the same object varies across viewpoints or
camera-to-object distances, causing its observations to occupy different 3D
locations. These artifacts persist despite per-frame depth alignment and,
where enabled, cross-frame consistency filtering. Similar artifacts remain
in the multi-view methods that process 16-frame inputs, indicating that
multi-frame context alone does not resolve the problem. DA2 and MapAnything
produce the clearest second 3D view: DA2 retains a more restricted
reconstruction, whereas MapAnything preserves more scene content and shows
the strongest cross-frame depth consistency in this example.

\begin{table}[t]
\centering
\scriptsize
\setlength{\tabcolsep}{1.5pt}
\caption{Quantitative comparison of depth-estimation methods. 
Within each method block, the best value per column, except Density, 
is in bold and the second-best distinct value is underlined; 
The Align column indicates whether per-frame affine depth alignment 
is applied before evaluation. 
In method labels, K denotes camera intrinsics 
and P denotes camera poses; 
The Filter column identifies confidence-P50 masking (P50) 
or the range restriction ($\leq 10$\,m); 
a dash indicates that neither operation is applied.}
\label{tab:depth-estimation-comparison}
\resizebox{\columnwidth}{!}{%
\begin{tabular}{@{}lcccccccc@{}}
\toprule
Method & Filter & Align & Abs $\downarrow$ & Abs $\tau$ $\uparrow$ & AbsRel $\downarrow$ & Rel $\tau$ $\uparrow$ & Density $\uparrow$ & Runtime $\downarrow$ \\

\midrule
\multicolumn{9}{l}{\textit{Monocular depth estimation}} \\
DA2 (Relative) \cite{DA2} & -- & $\checkmark$ & 3.9086 & 0.0090 & 0.4332 & 0.0392 & 0.9410 & 0.0160 \\

DA2 (Metric) \cite{DA2} & -- & $\times$ & 11.3001 & 0.0001 & 2.0257 & 0.0041 & 1.0000 & \textbf{0.0157} \\

DA2 (Metric) \cite{DA2} & $\leq 10$\,m & $\times$ & 5.2689 & 0.0000 & 1.9307 & 0.0002 & 0.2693 & \underline{0.0158} \\

DA2 (Metric) \cite{DA2} & -- & $\checkmark$ & 2.3071 & 0.0234 & 0.2497 & 0.0940 & 1.0000 & \textbf{0.0157} \\

DA2 (Metric) \cite{DA2} & $\leq 10$\,m & $\checkmark$ & \textbf{0.2159} & \textbf{0.1360} & \textbf{0.0761} & \textbf{0.3486} & 0.2693 & \underline{0.0158} \\

MoGe-2 (Metric) \cite{MoGe} & -- & $\times$ & 2.7932 & 0.0062 & 0.4363 & 0.0448 & 0.9433 & 0.0601 \\

MoGe-2 (Metric) \cite{MoGe} & $\leq 10$\,m & $\times$ & 1.5176 & 0.0065 & 0.4053 & 0.0366 & 0.6511 & 0.0598 \\

MoGe-2 (Metric) \cite{MoGe} & -- & $\checkmark$ & 1.7959 & 0.0319 & 0.1916 & 0.1237 & 0.9433 & 0.0601 \\

MoGe-2 (Metric) \cite{MoGe} & $\leq 10$\,m & $\checkmark$ & \underline{0.4708} & \underline{0.0919} & \underline{0.1025} & \underline{0.2965} & 0.6511 & 0.0598 \\

\midrule
\multicolumn{9}{l}{\textit{Stereo depth estimation}} \\
SGM (AD-Census) \cite{SGM} & -- & $\times$ & 3.7364 & 0.0107 & 0.4545 & 0.0430 & 0.5134 & 2.9049 \\

SGM (AD-Census) \cite{SGM} & $\leq 10$\,m & $\times$ & 1.7241 & \underline{0.0141} & 0.2763 & \underline{0.0523} & 0.3505 & 2.8592 \\

FoundationStereo \cite{FoundationStereo} & -- & $\times$ & 39.5402 & 0.0082 & 5.3271 & 0.0327 & 0.9971 & \textbf{0.0806} \\

FoundationStereo \cite{FoundationStereo} & $\leq 10$\,m & $\times$ & \underline{1.1077} & 0.0095 & \underline{0.1943} & 0.0346 & 0.7674 & \underline{0.0812} \\

FoundationStereo \cite{FoundationStereo} & -- & $\checkmark$ & 3.8991 & 0.0123 & 0.4309 & 0.0500 & 0.9971 & \textbf{0.0806} \\

FoundationStereo \cite{FoundationStereo} & $\leq 10$\,m & $\checkmark$ & \textbf{0.8330} & \textbf{0.0740} & \textbf{0.1465} & \textbf{0.2388} & 0.7674 & \underline{0.0812} \\

\midrule
\multicolumn{9}{l}{\textit{Multi-view feed-forward models}} \\
VGGT \cite{VGGT} & -- & $\checkmark$ & 1.7566 & 0.0506 & 0.1669 & 0.1976 & 1.0000 & 0.0508 \\

VGGT \cite{VGGT} & P50 & $\checkmark$ & \underline{0.2767} & \underline{0.1642} & \underline{0.0516} & \underline{0.5072} & 0.5018 & 0.0503 \\

VGGT-$\Omega$ \cite{VGGT-Omega} & -- & $\checkmark$ & 1.5662 & 0.0465 & 0.1697 & 0.1719 & 1.0000 & 0.0287 \\

VGGT-$\Omega$ \cite{VGGT-Omega} & P50 & $\checkmark$ & \textbf{0.2170} & \textbf{0.2001} & \textbf{0.0443} & \textbf{0.5855} & 0.5000 & 0.0284 \\

DA3 (Mono) \cite{DA3} & -- & $\checkmark$ & 2.2931 & 0.0325 & 0.2302 & 0.1233 & 1.0000 & \underline{0.0117} \\

DA3 (Metric+K) \cite{DA3} & -- & $\times$ & 2.8016 & 0.0115 & 0.4206 & 0.0649 & 1.0000 & \textbf{0.0116} \\

DA3 (Metric+K) \cite{DA3} & $\leq 10$\,m & $\times$ & 1.1033 & 0.0153 & 0.2825 & 0.0668 & 0.6570 & 0.0117 \\

DA3 (Metric+K) \cite{DA3} & -- & $\checkmark$ & 2.4133 & 0.0280 & 0.2459 & 0.1060 & 1.0000 & \textbf{0.0116} \\

DA3 (Metric+K) \cite{DA3} & $\leq 10$\,m & $\checkmark$ & 0.5366 & 0.0745 & 0.1194 & 0.2407 & 0.6570 & 0.0117 \\

MapAnything \cite{MapAnything} & -- & $\checkmark$ & 1.8411 & 0.0446 & 0.1852 & 0.1580 & 0.6854 & 0.1449 \\

MapAnything \cite{MapAnything} & P50 & $\checkmark$ & 0.5410 & 0.1150 & 0.0755 & 0.3689 & 0.4394 & 0.1548 \\

MapAnything (K+P) \cite{MapAnything} & -- & $\times$ & 1.6287 & 0.0629 & 0.2166 & 0.2364 & 0.6771 & 0.1468 \\

MapAnything (K+P) \cite{MapAnything} & P50 & $\times$ & 0.4941 & 0.0869 & 0.0814 & 0.3010 & 0.4406 & 0.1547 \\

MapAnything (K+P) \cite{MapAnything} & -- & $\checkmark$ & 1.8260 & 0.0404 & 0.1899 & 0.1441 & 0.6771 & 0.1468 \\

MapAnything (K+P) \cite{MapAnything} & P50 & $\checkmark$ & 0.4548 & 0.1247 & 0.0696 & 0.3956 & 0.4406 & 0.1547 \\

\bottomrule
\end{tabular}
}
\end{table}

\begin{figure*}[p]
\centering
\scriptsize
\setlength{\tabcolsep}{1pt}
\begin{tabular}{@{}rcccc@{}}
\toprule
\textbf{Method} & Full Depth & Filtered Depth
& 3D View 1 & 3D View 2 \\
\midrule
\textbf{Reference} & \raisebox{-0.5\height}{\includegraphics[width=0.19\textwidth]{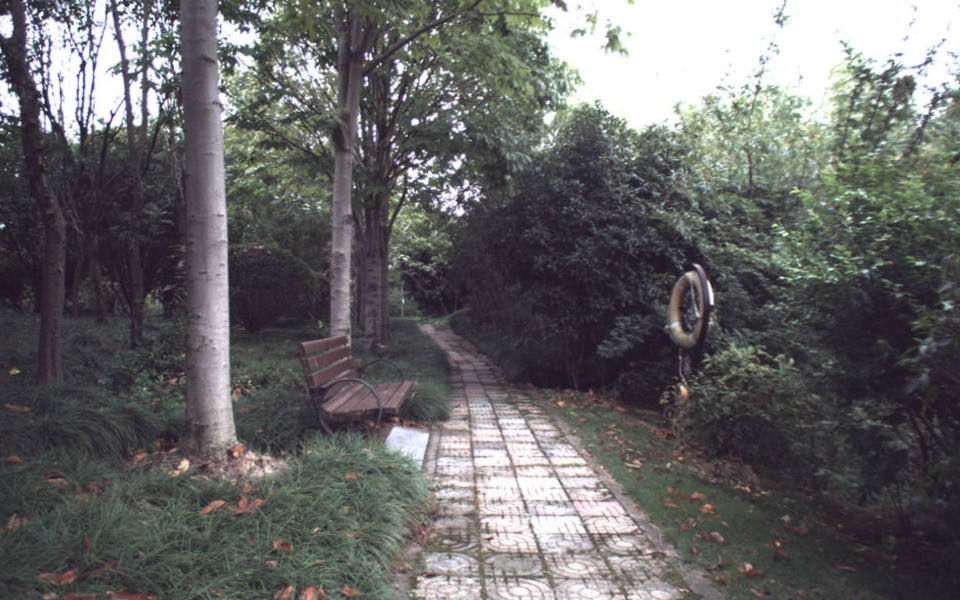}} & \raisebox{-0.5\height}{\includegraphics[width=0.19\textwidth]{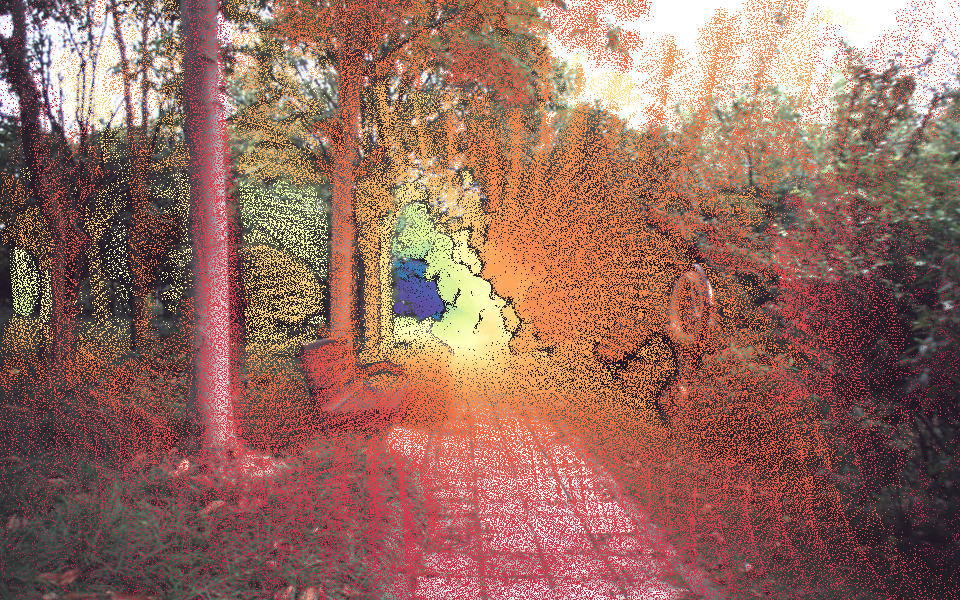}} & \raisebox{-0.5\height}{\includegraphics[width=0.19\textwidth]{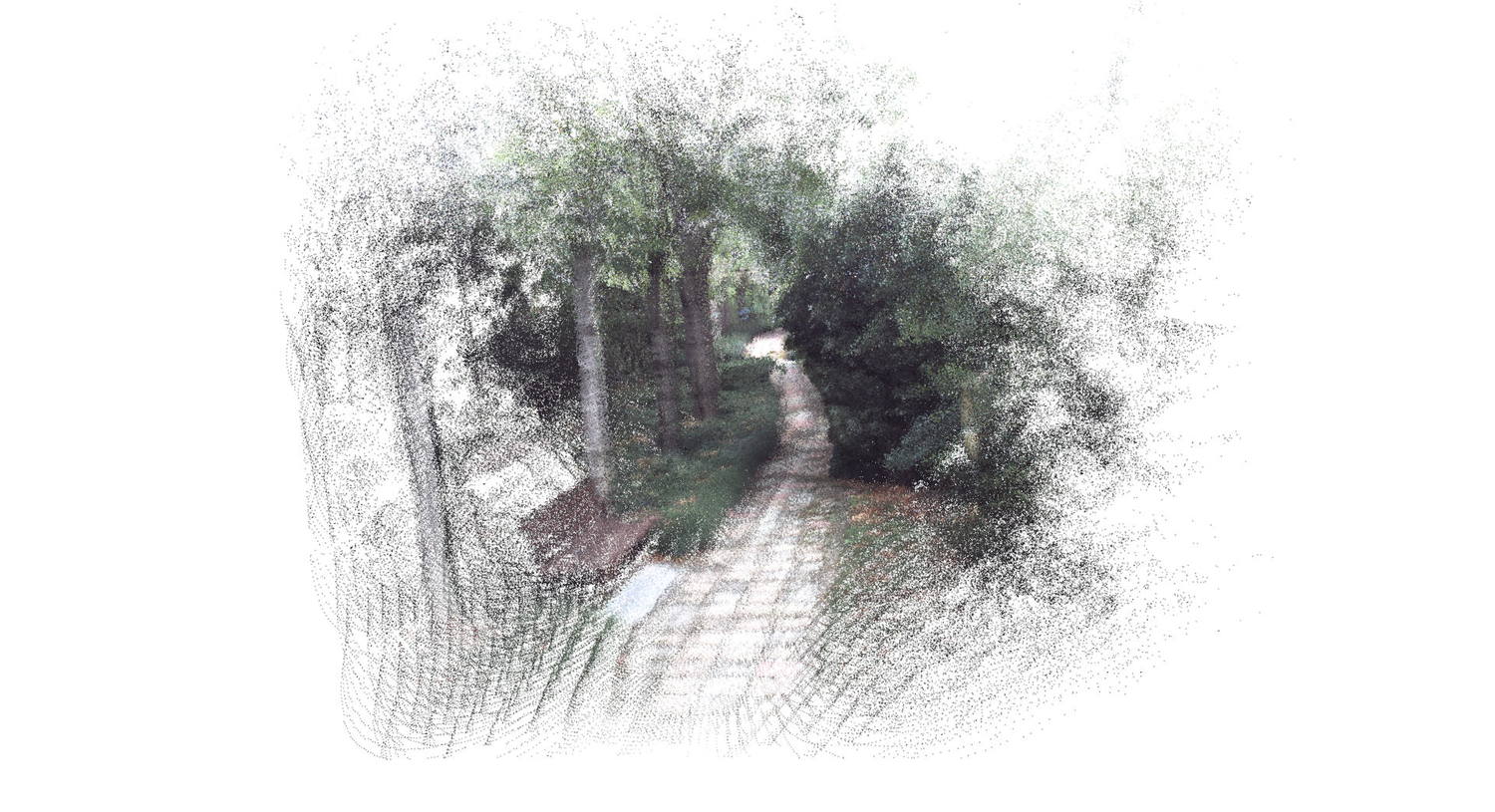}} & \raisebox{-0.5\height}{\includegraphics[width=0.19\textwidth]{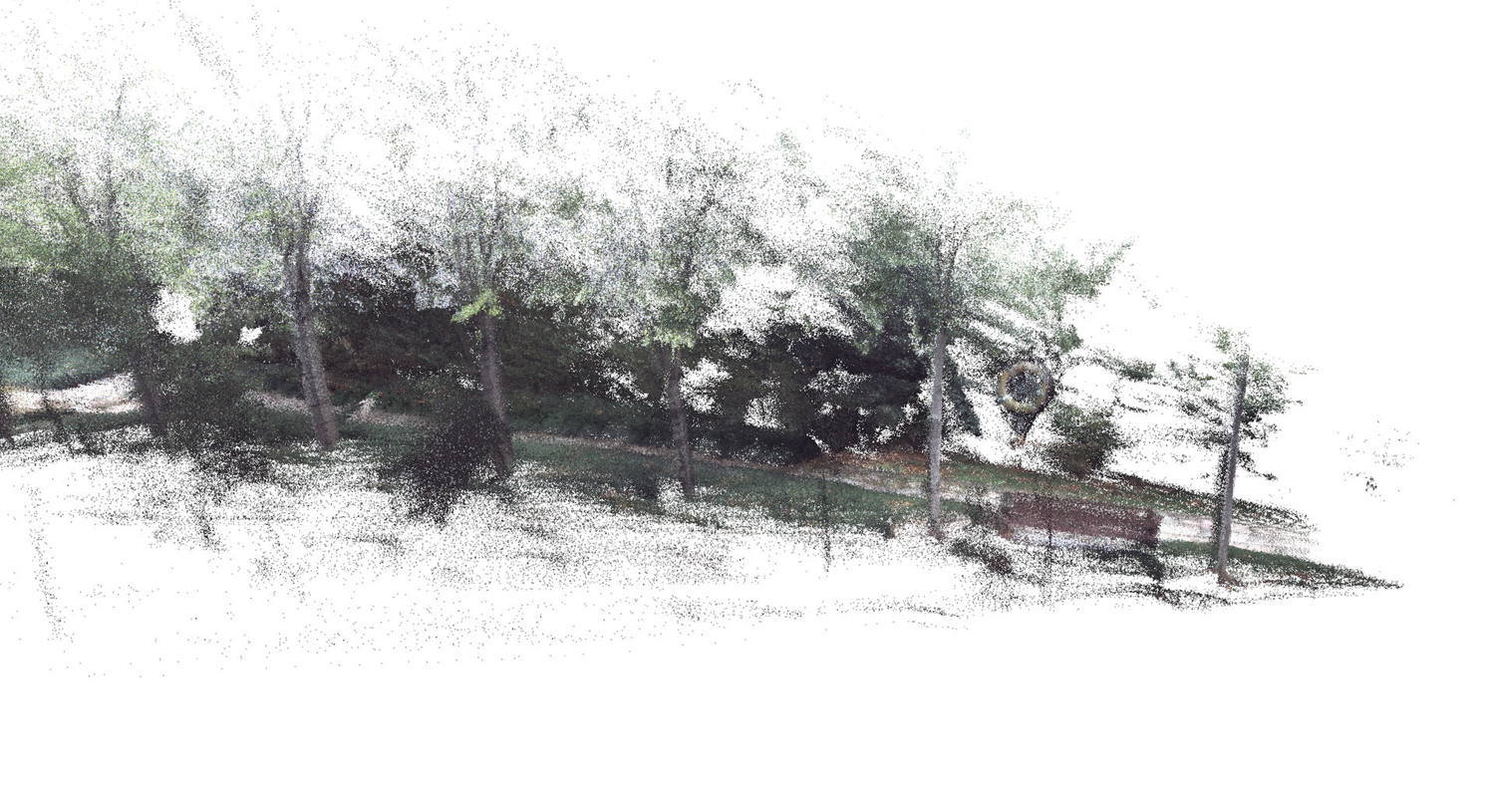}} \\[1pt] \textbf{DA2} & \raisebox{-0.5\height}{\includegraphics[width=0.19\textwidth]{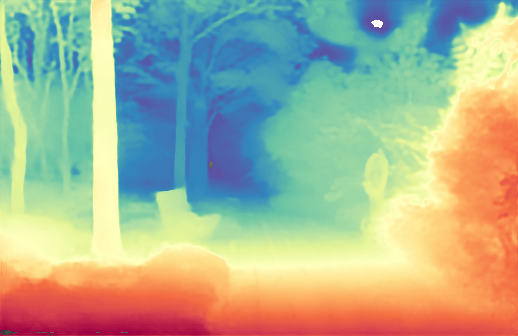}} & \raisebox{-0.5\height}{\includegraphics[width=0.19\textwidth]{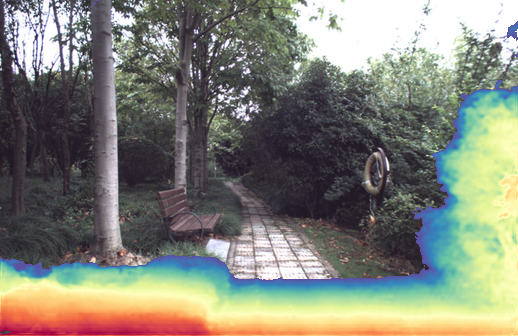}} & \raisebox{-0.5\height}{\includegraphics[width=0.19\textwidth]{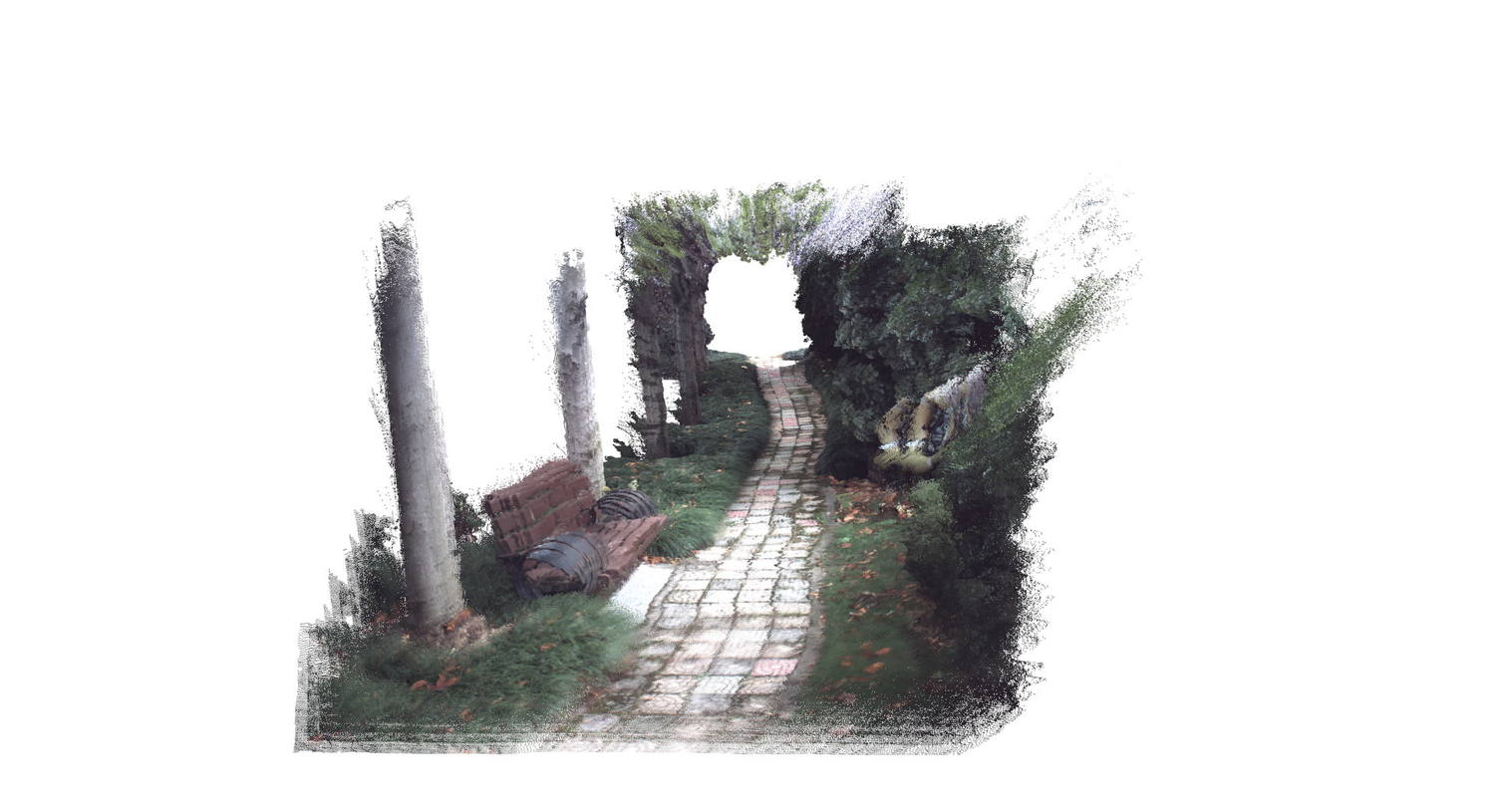}} & \raisebox{-0.5\height}{\includegraphics[width=0.19\textwidth]{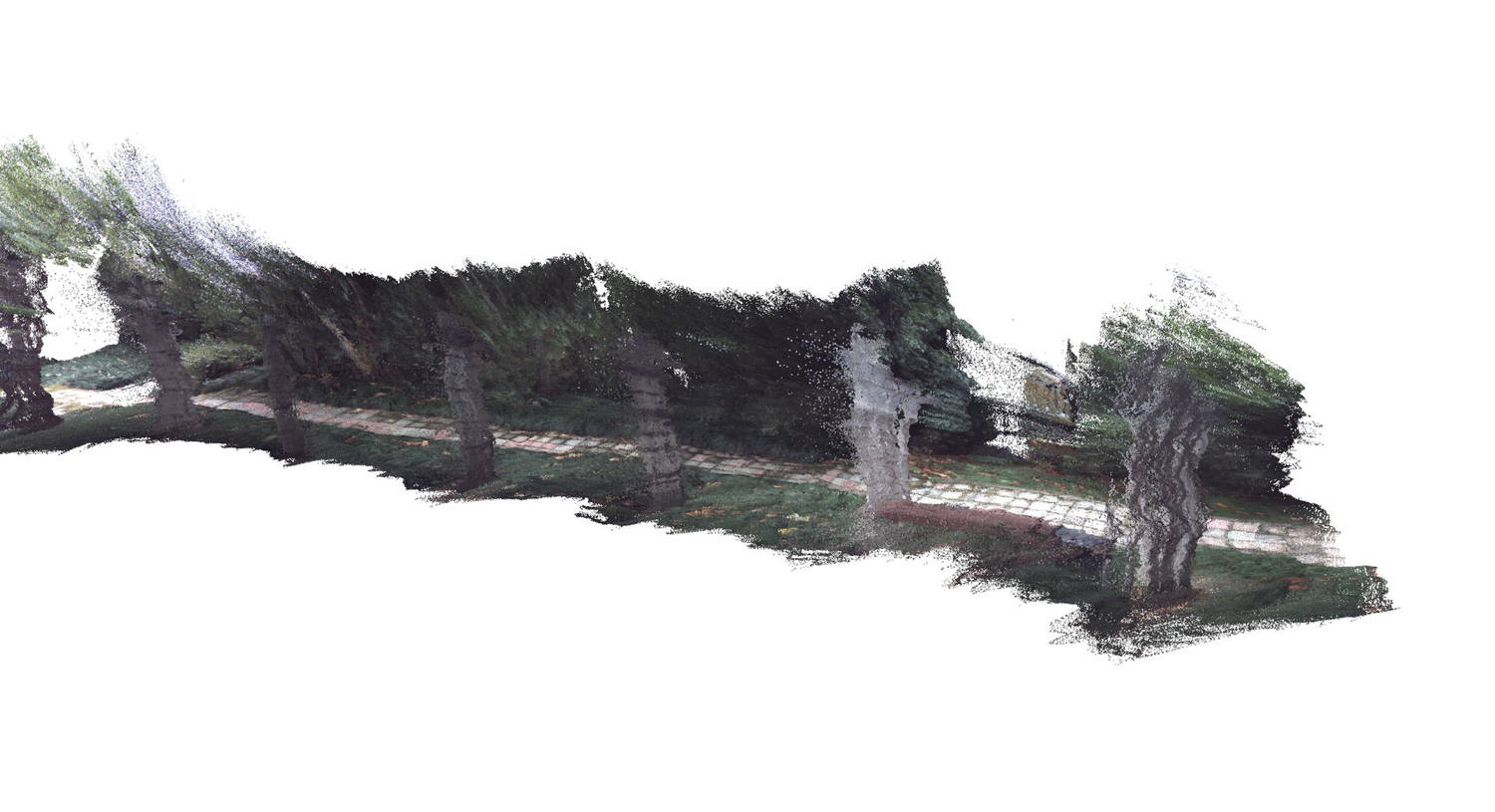}} \\[1pt] 
\textbf{MoGe-2} & \raisebox{-0.5\height}{\includegraphics[width=0.19\textwidth]{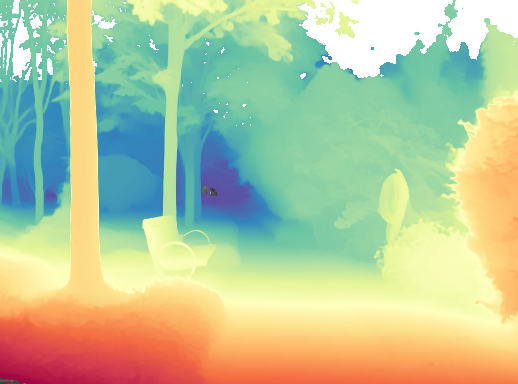}} & \raisebox{-0.5\height}{\includegraphics[width=0.19\textwidth]{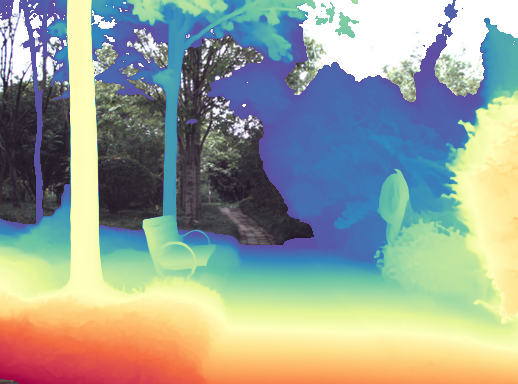}} & \raisebox{-0.5\height}{\includegraphics[width=0.19\textwidth]{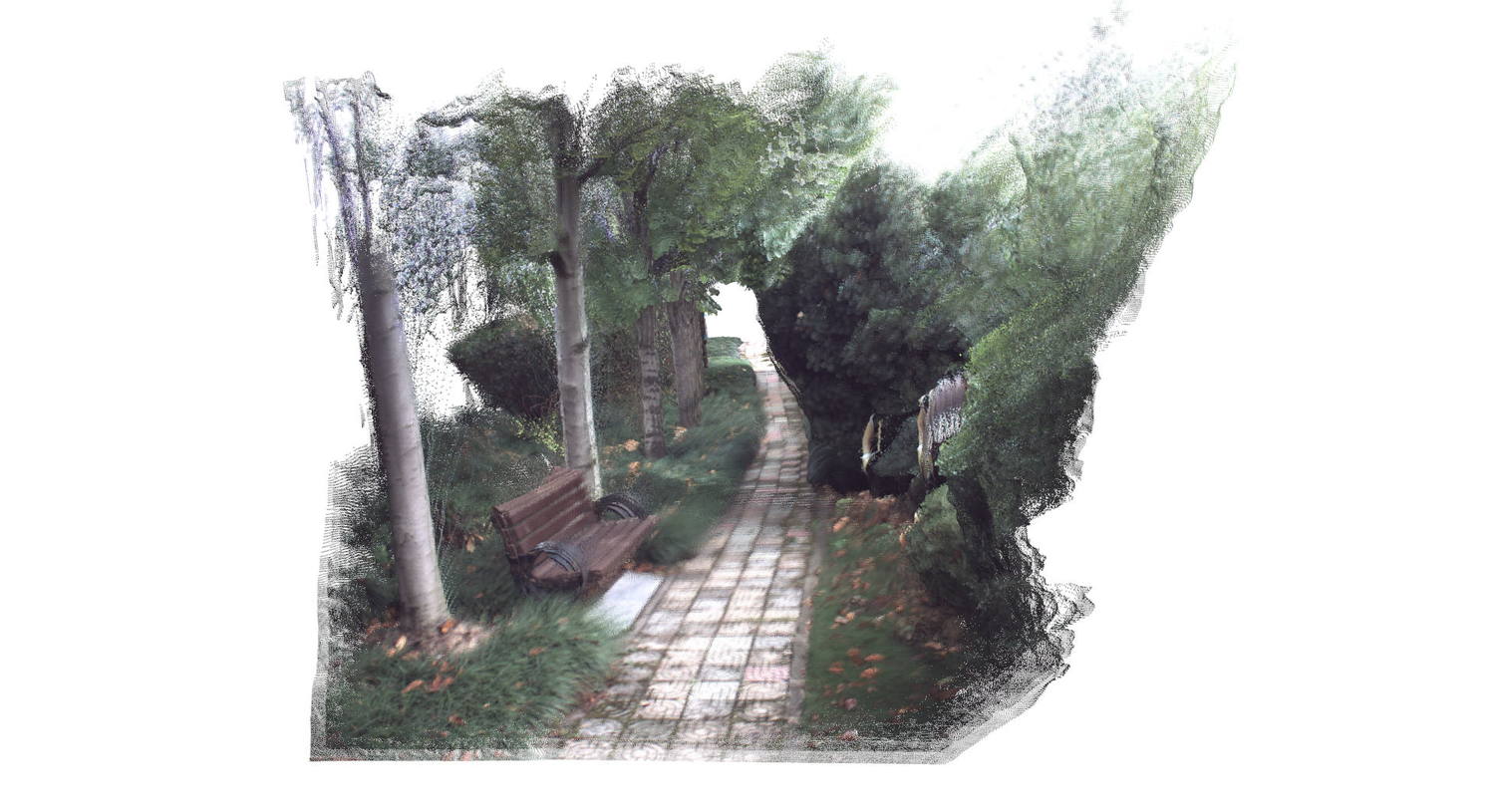}} & \raisebox{-0.5\height}{\includegraphics[width=0.19\textwidth]{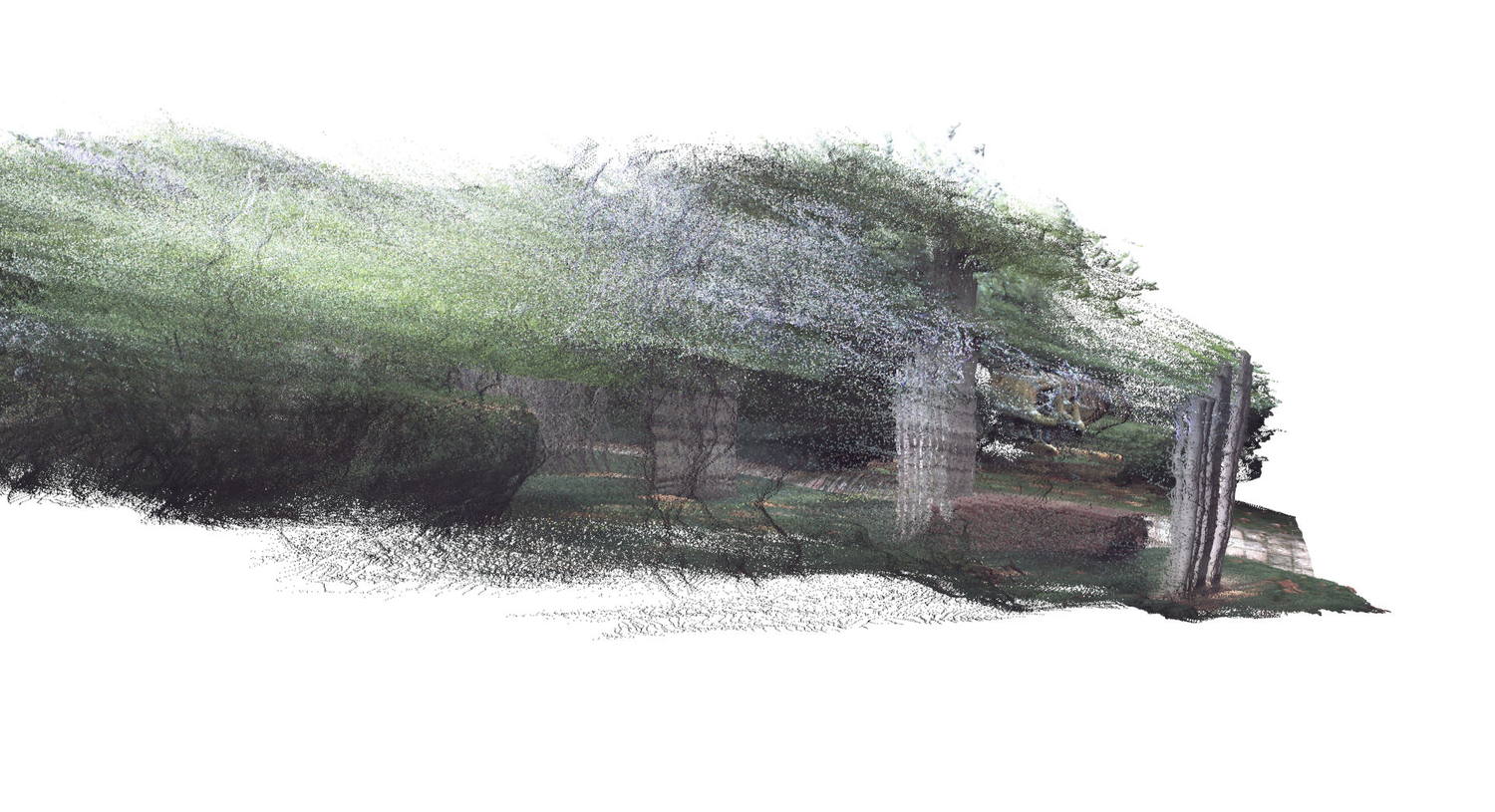}} \\[1pt] 
\textbf{SGM} & \raisebox{-0.5\height}{\includegraphics[width=0.19\textwidth]{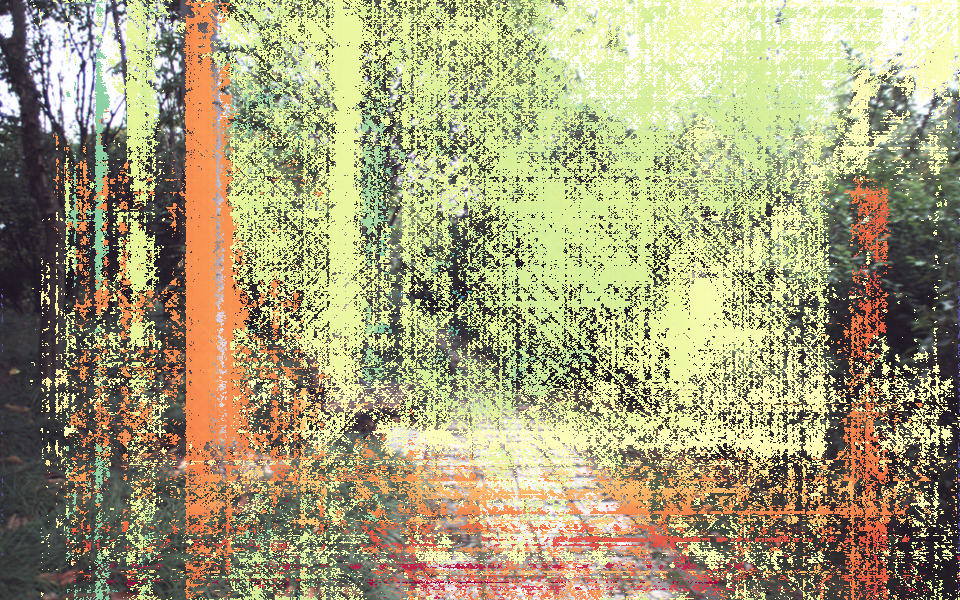}} & \raisebox{-0.5\height}{\includegraphics[width=0.19\textwidth]{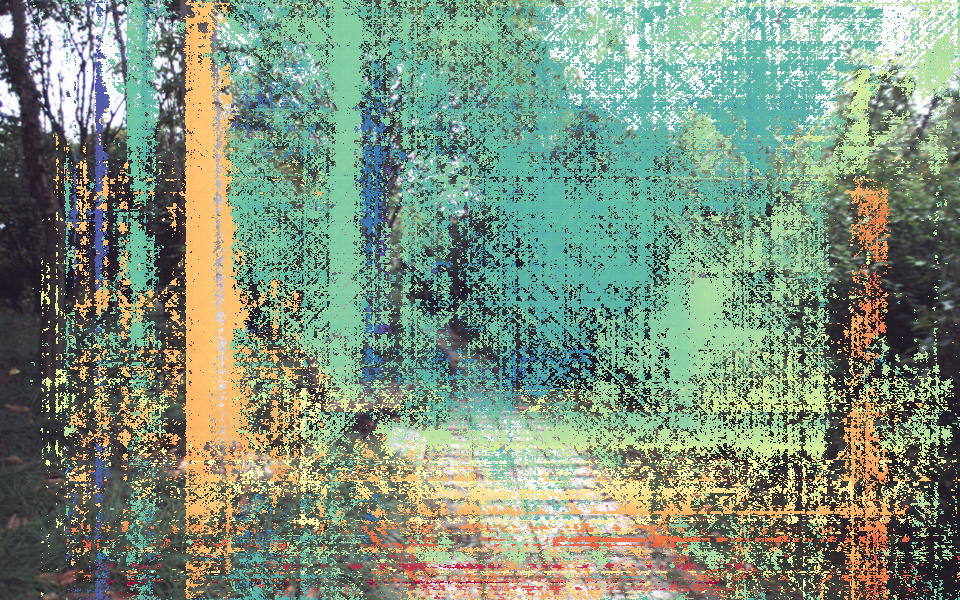}} & \raisebox{-0.5\height}{\includegraphics[width=0.19\textwidth]{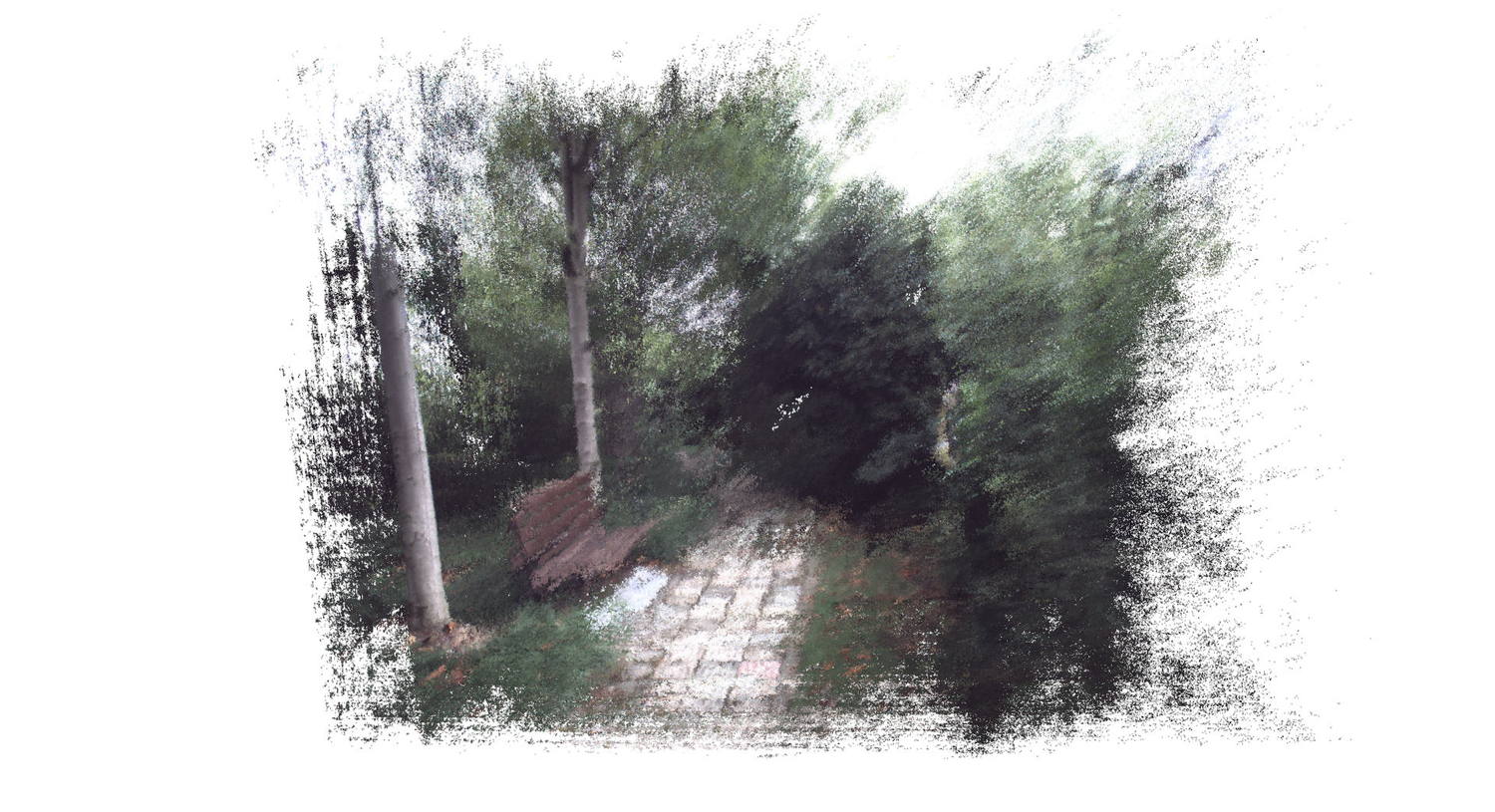}} & \raisebox{-0.5\height}{\includegraphics[width=0.19\textwidth]{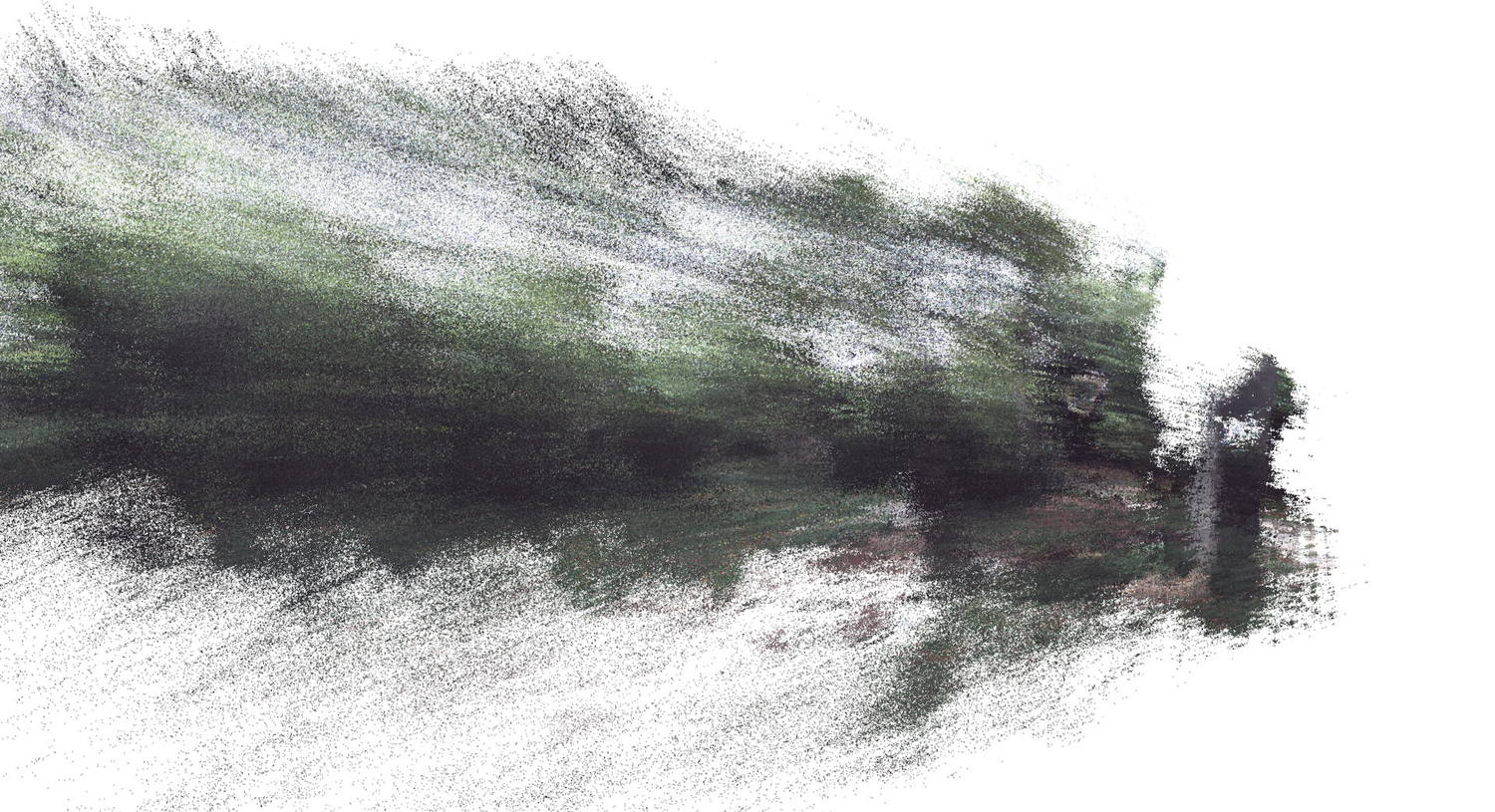}} \\[1pt] 
\textbf{FoundationStereo} & \raisebox{-0.5\height}{\includegraphics[width=0.19\textwidth]{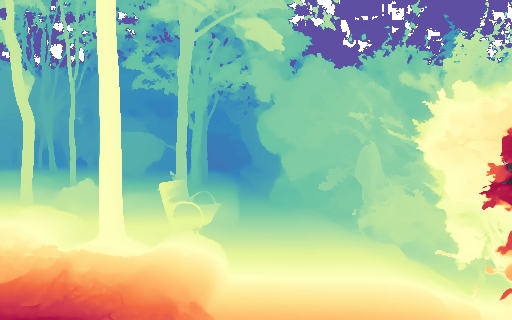}} & \raisebox{-0.5\height}{\includegraphics[width=0.19\textwidth]{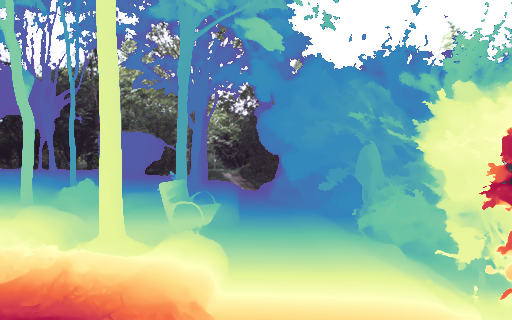}} & \raisebox{-0.5\height}{\includegraphics[width=0.19\textwidth]{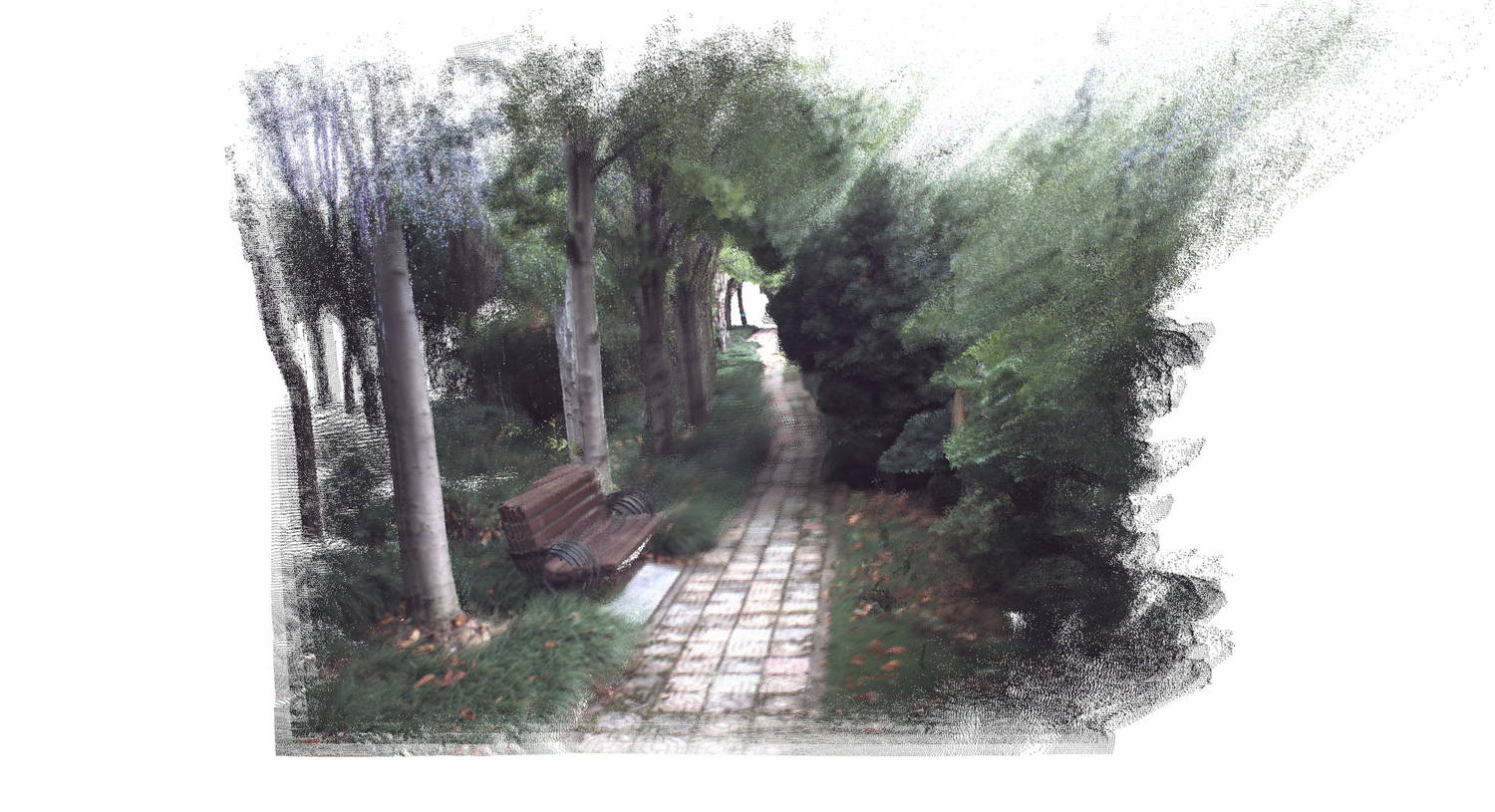}} & \raisebox{-0.5\height}{\includegraphics[width=0.19\textwidth]{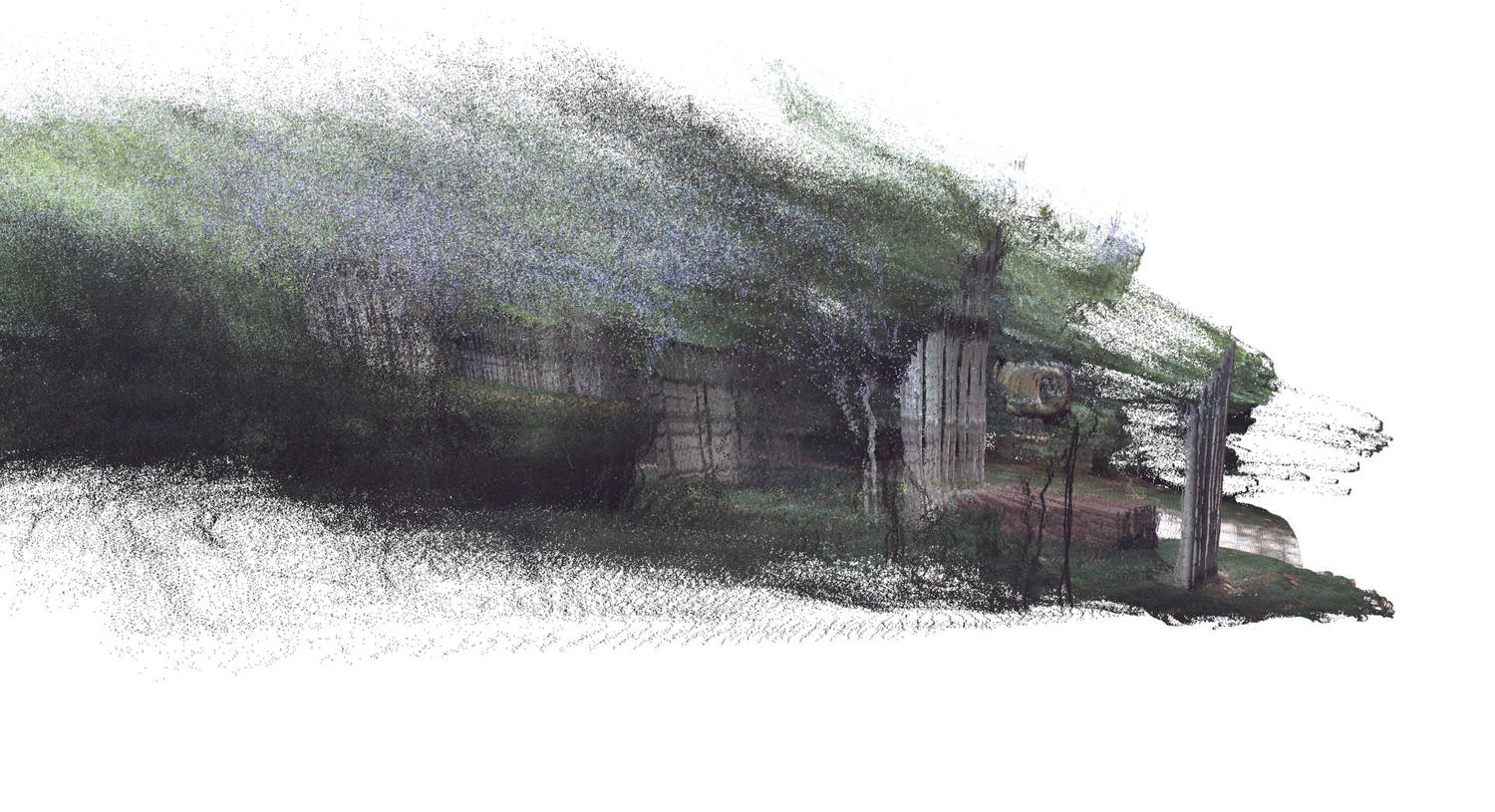}} \\[1pt] 
\textbf{VGGT} & \raisebox{-0.5\height}{\includegraphics[width=0.19\textwidth]{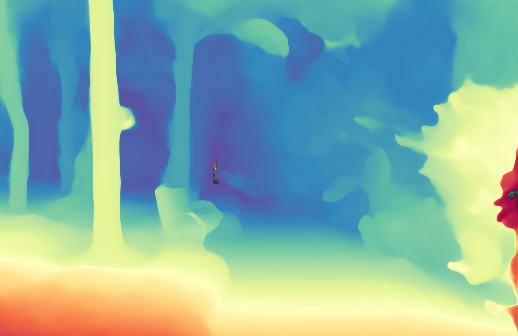}} & \raisebox{-0.5\height}{\includegraphics[width=0.19\textwidth]{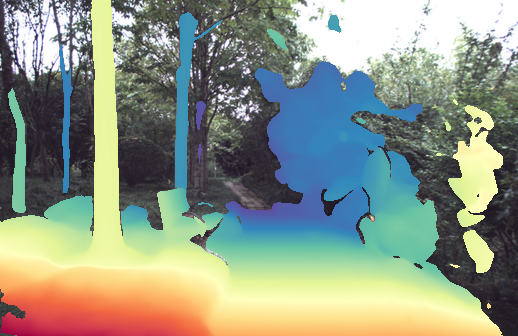}} & \raisebox{-0.5\height}{\includegraphics[width=0.19\textwidth]{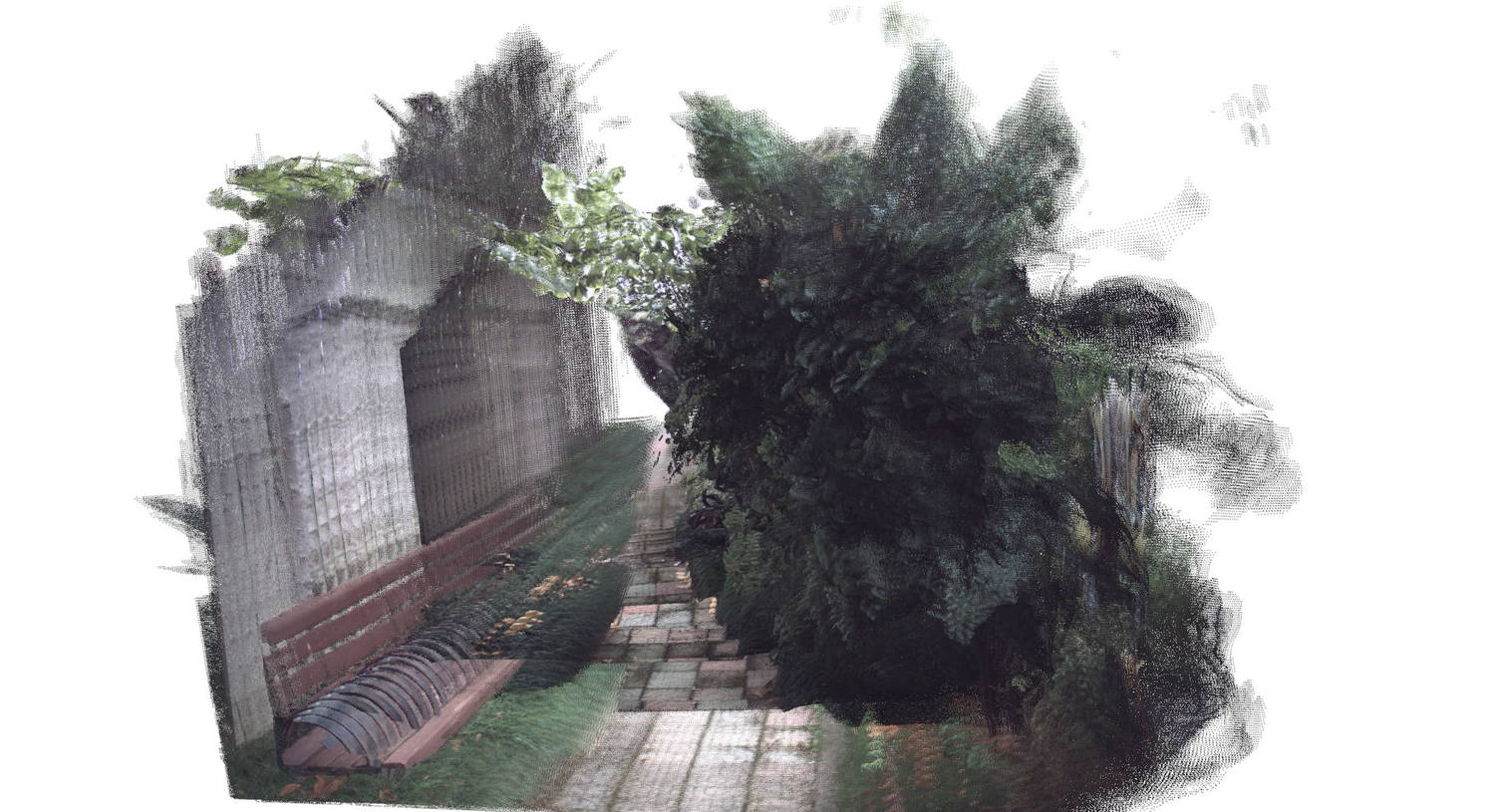}} & \raisebox{-0.5\height}{\includegraphics[width=0.19\textwidth]{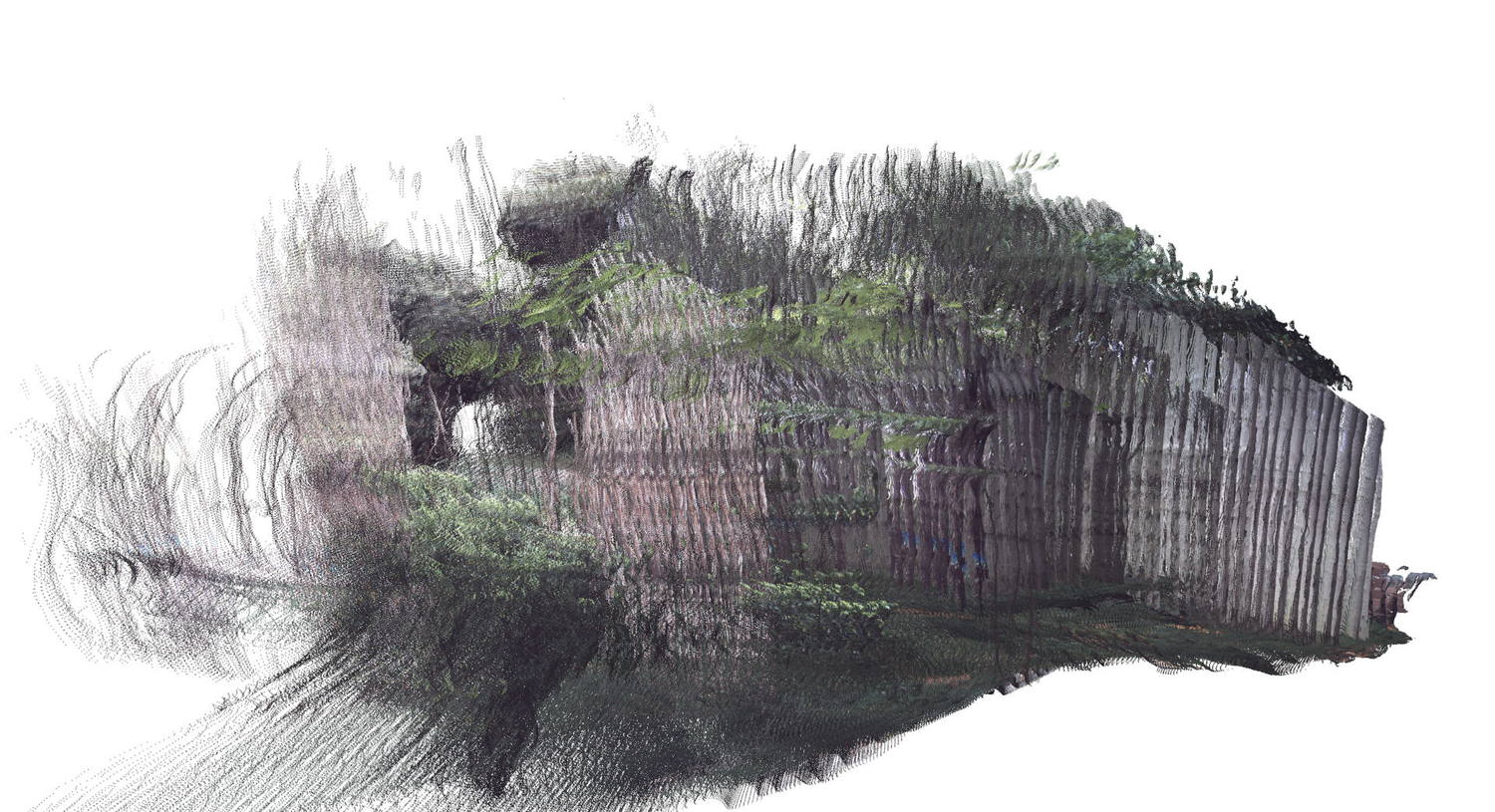}} \\[1pt] 
\textbf{VGGT-$\Omega$} & \raisebox{-0.5\height}{\includegraphics[width=0.19\textwidth]{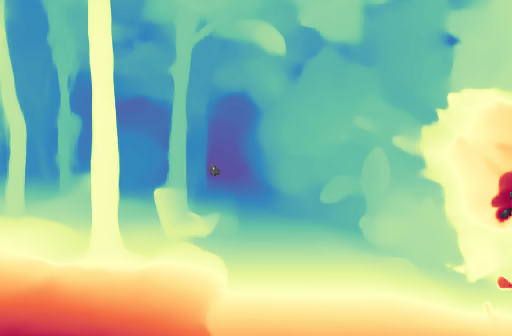}} & \raisebox{-0.5\height}{\includegraphics[width=0.19\textwidth]{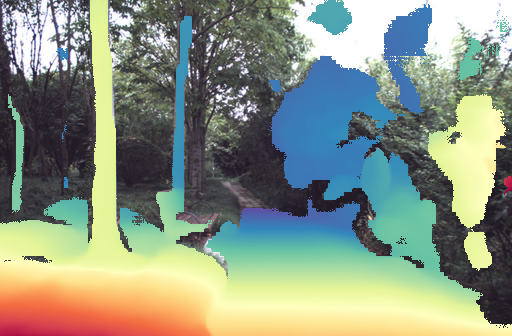}} & \raisebox{-0.5\height}{\includegraphics[width=0.19\textwidth]{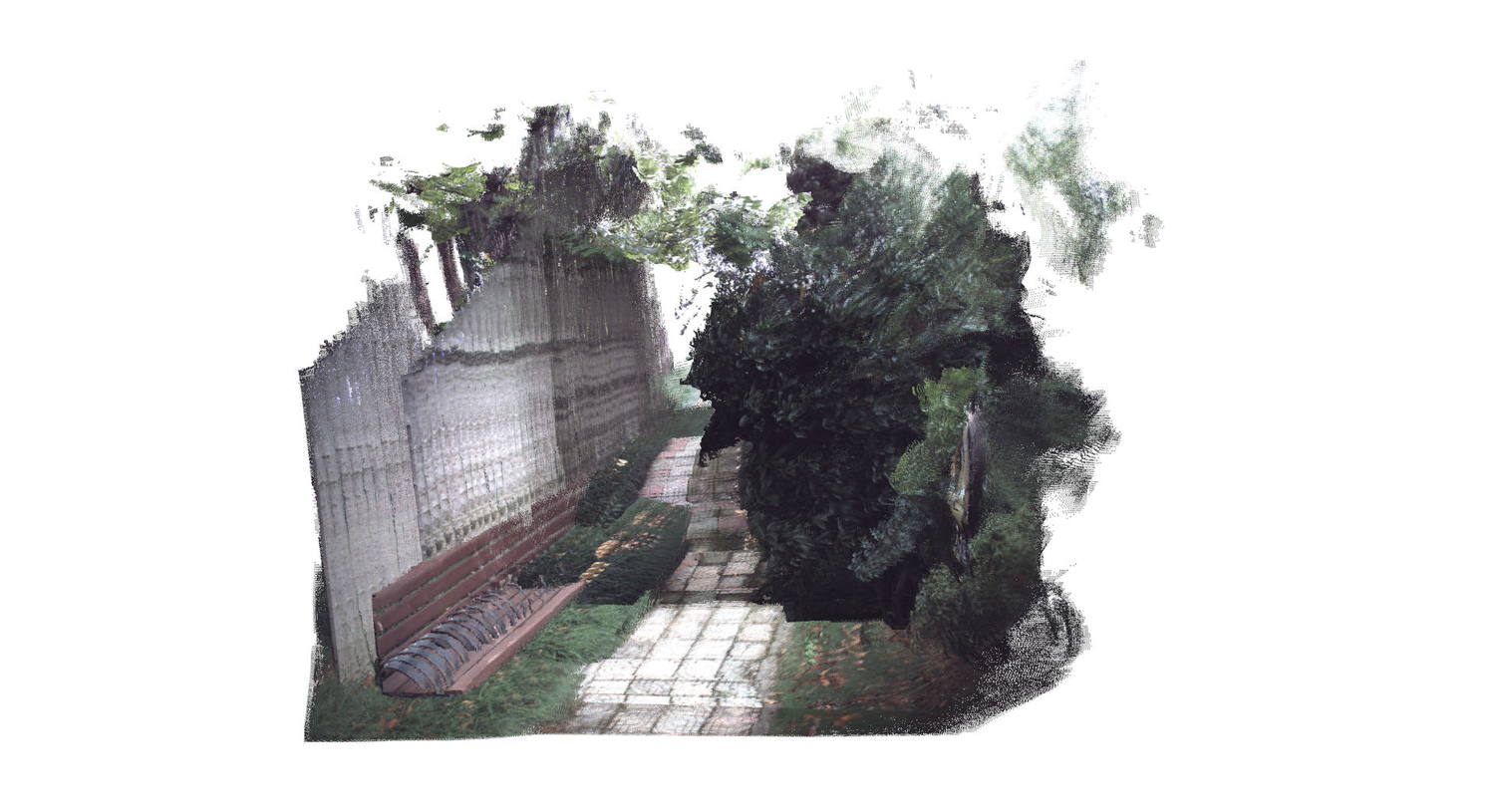}} & \raisebox{-0.5\height}{\includegraphics[width=0.19\textwidth]{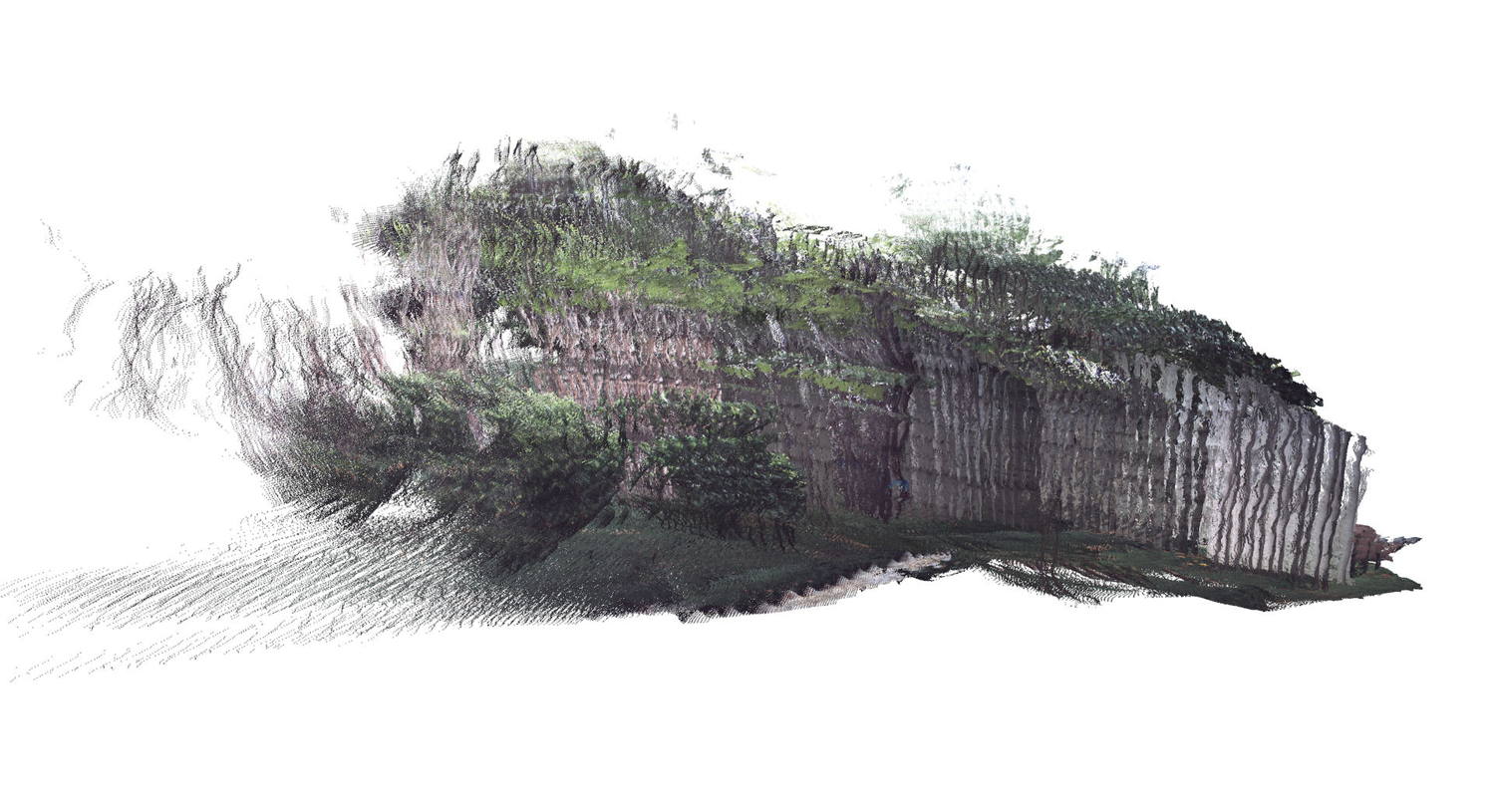}} \\[1pt] 
\textbf{DA3} & \raisebox{-0.5\height}{\includegraphics[width=0.19\textwidth]{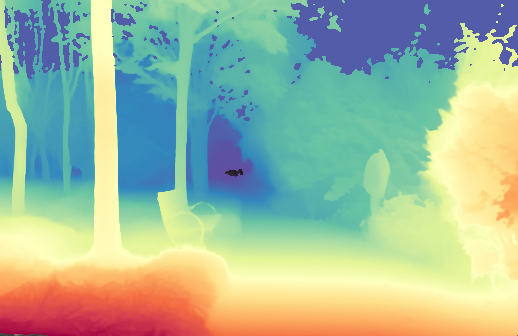}} & \raisebox{-0.5\height}{\includegraphics[width=0.19\textwidth]{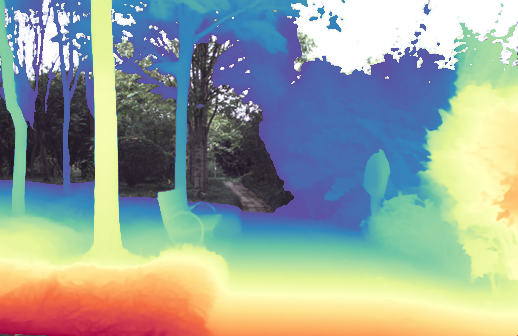}} & \raisebox{-0.5\height}{\includegraphics[width=0.19\textwidth]{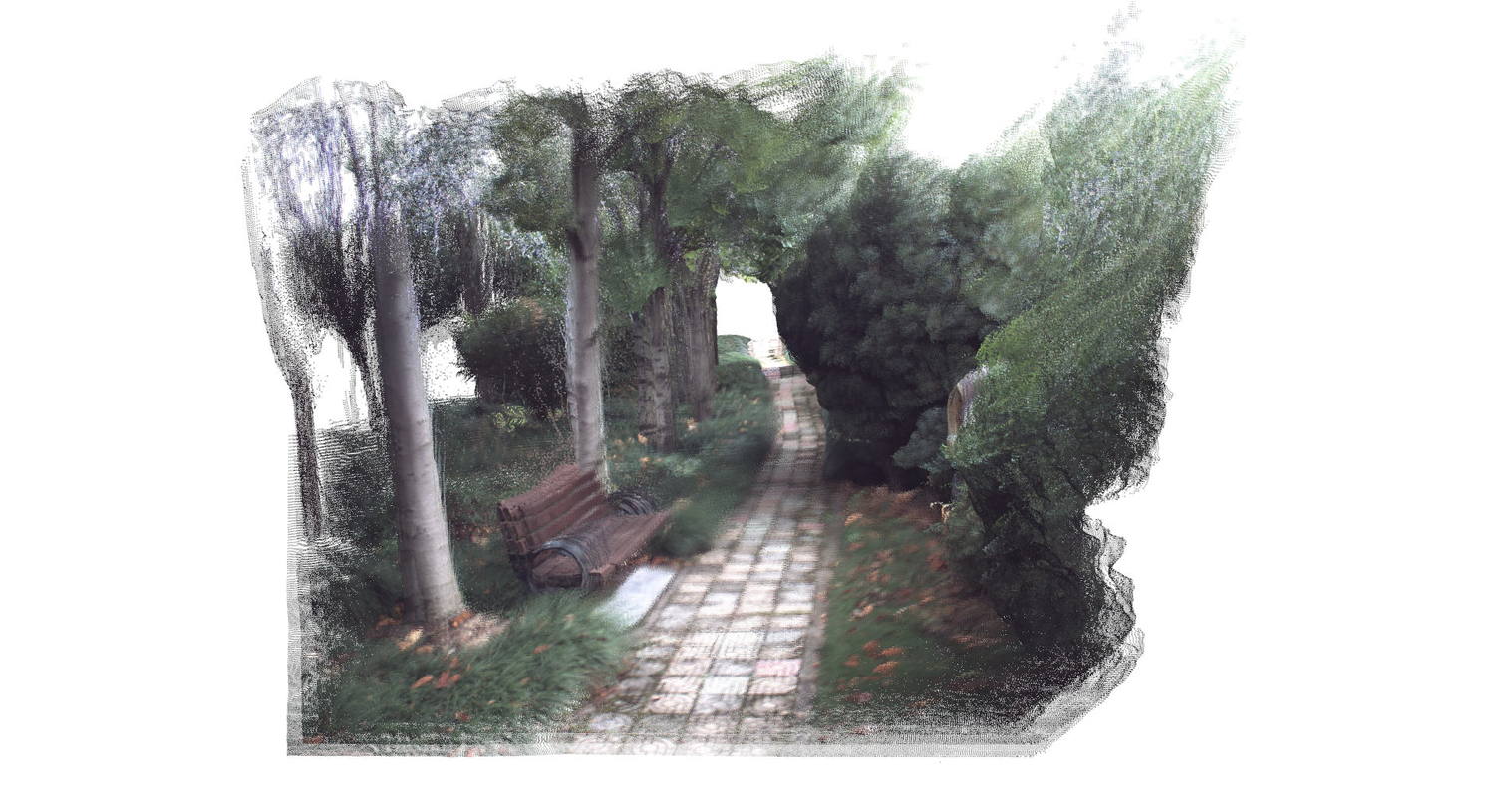}} & \raisebox{-0.5\height}{\includegraphics[width=0.19\textwidth]{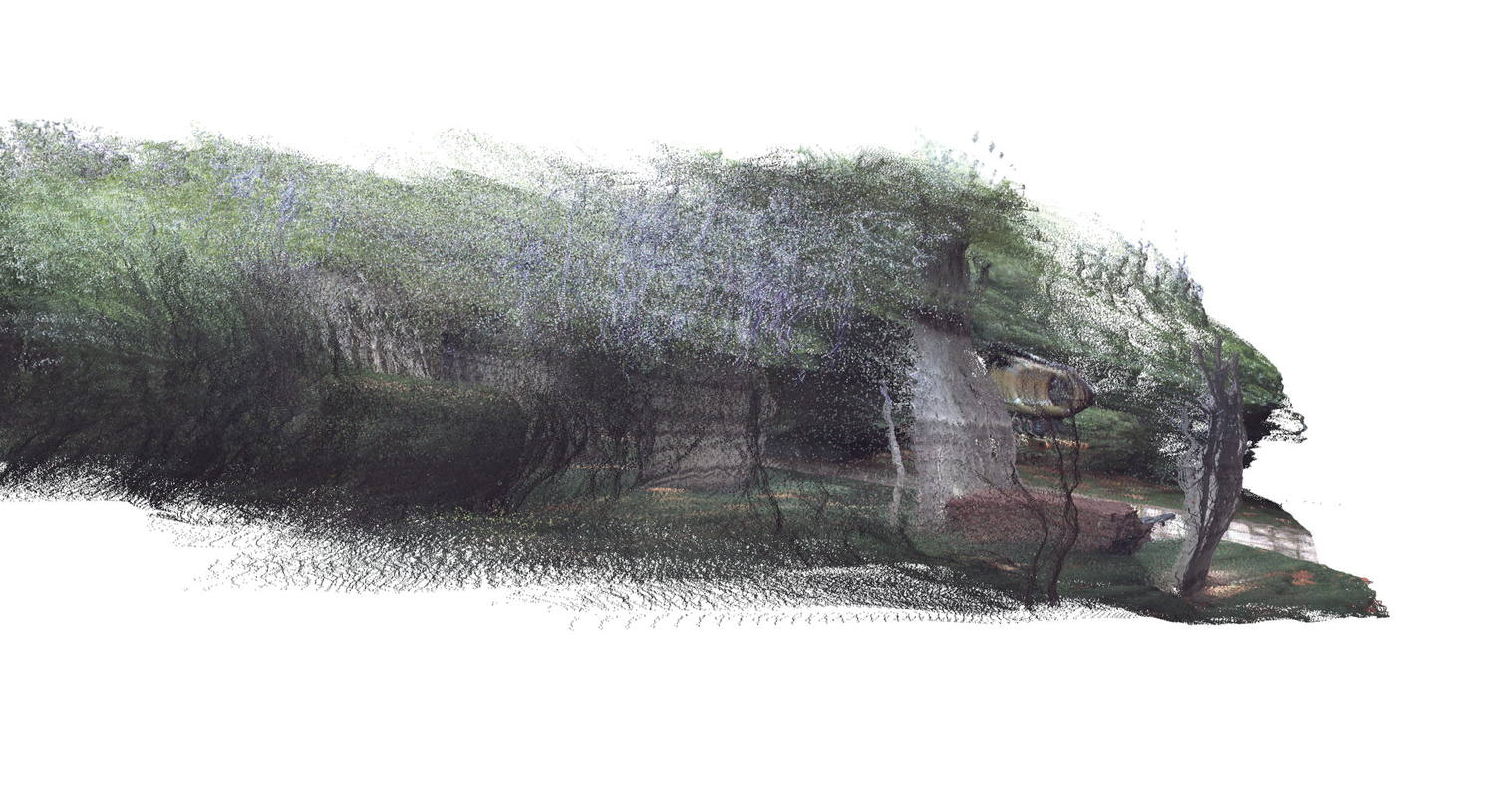}} \\[1pt] 
\textbf{MapAnything} & \raisebox{-0.5\height}{\includegraphics[width=0.19\textwidth]{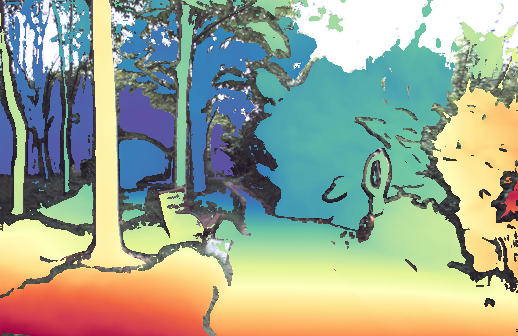}} & \raisebox{-0.5\height}{\includegraphics[width=0.19\textwidth]{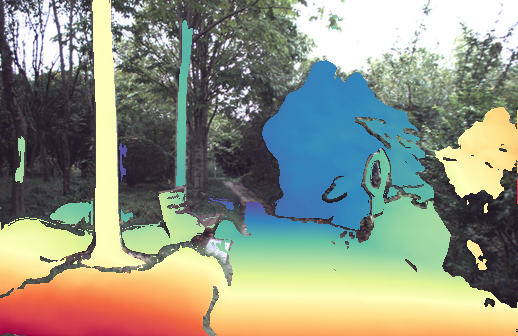}} & \raisebox{-0.5\height}{\includegraphics[width=0.19\textwidth]{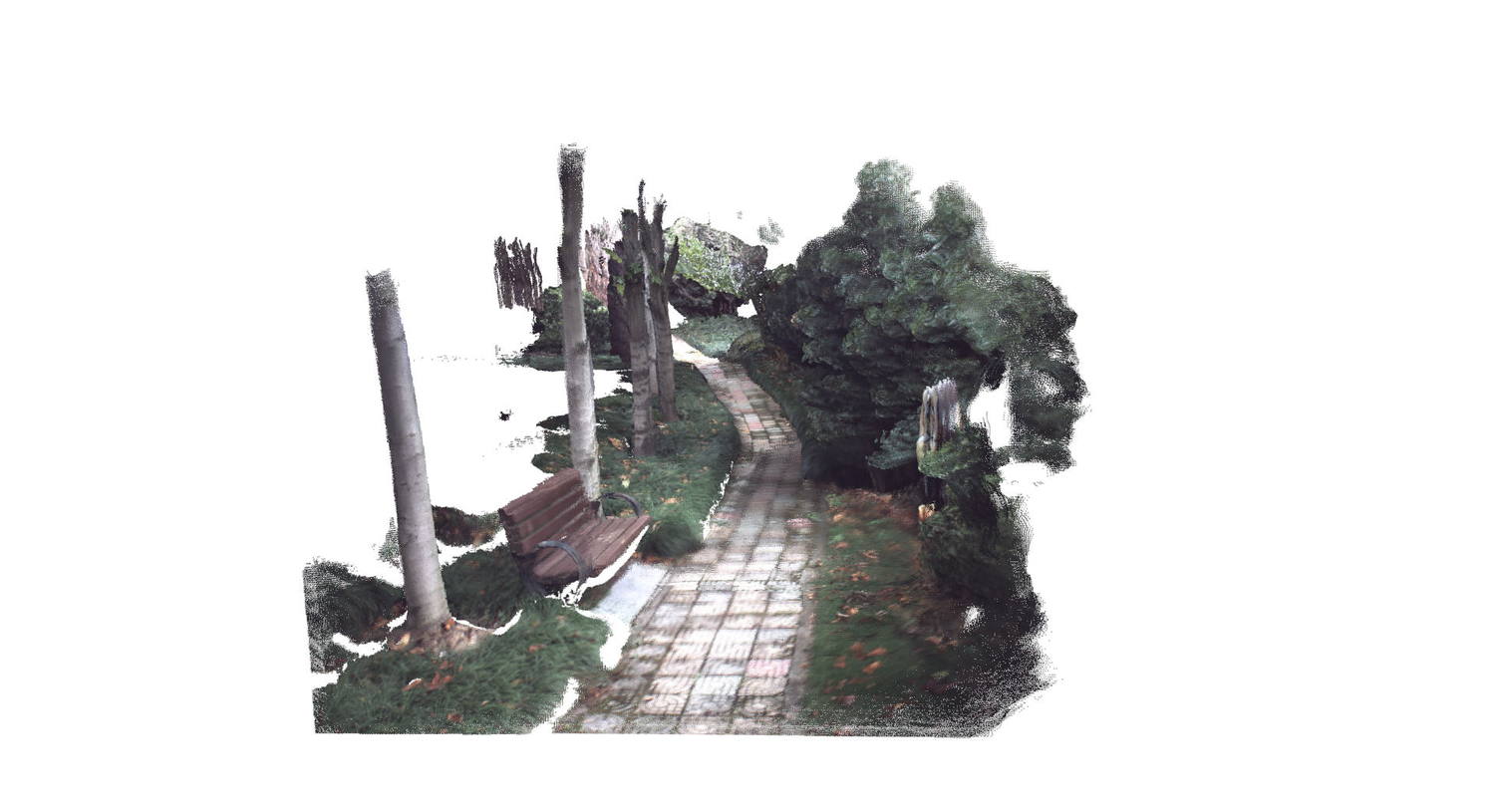}} & \raisebox{-0.5\height}{\includegraphics[width=0.19\textwidth]{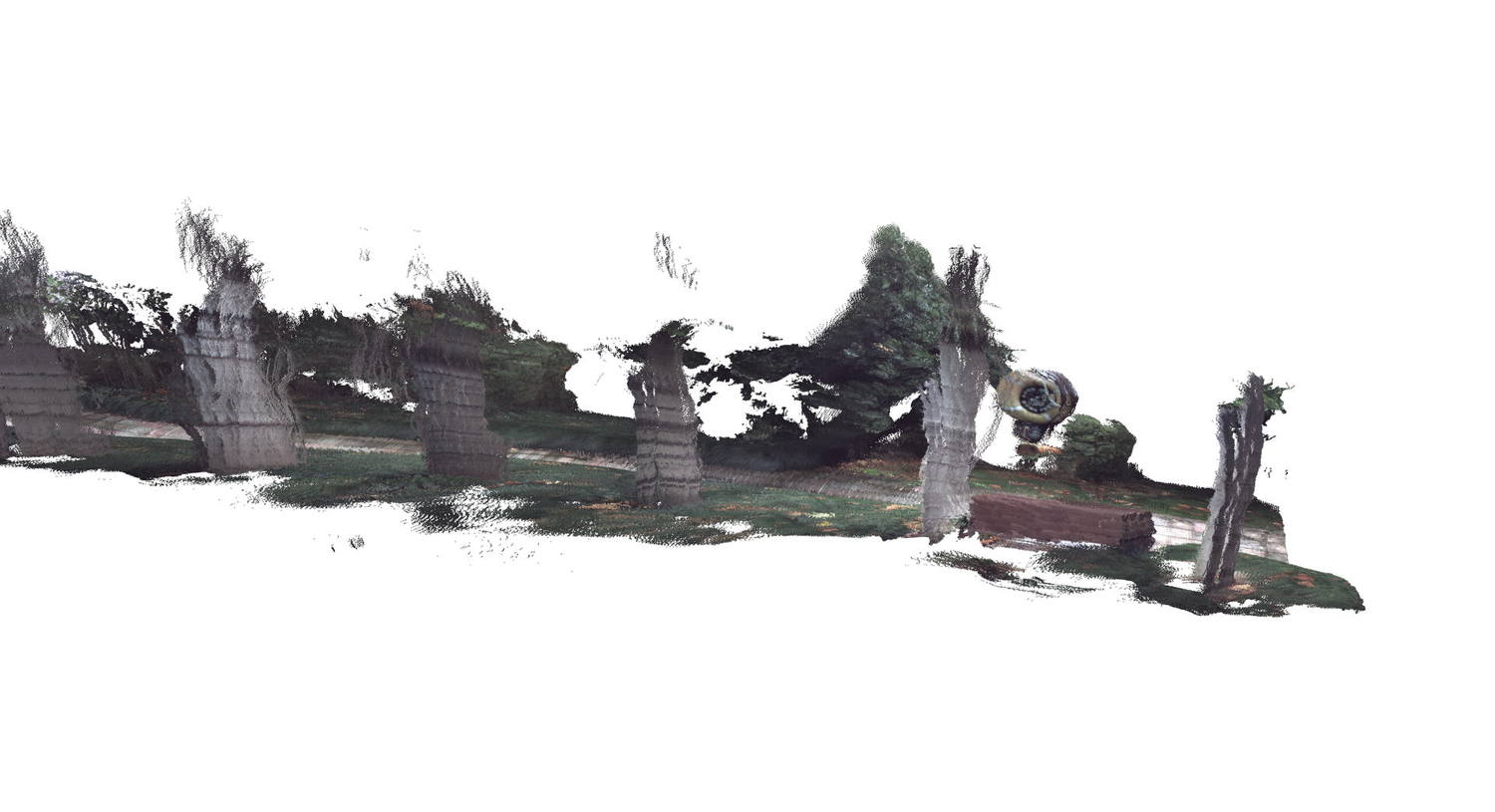}} \\[1pt] 
\bottomrule
\end{tabular}
\caption{Qualitative dense-depth comparison on
\texttt{BotanicGarden\_1006-01}. Full Depth shows the complete
valid prediction for assessing relative-depth discrimination, whereas Filtered
Depth shows the subset retained by method-specific confidence or range
filtering. 3D View 1 and 3D View 2 show two fixed views of the point clouds,
back-projecting the retained depth maps, and accumulating them across the
sequence using the camera poses associated with each configuration. The first
row contains, from left to right, the undistorted input image, reference depth,
and two views of the reference point cloud; the remaining rows show the
monocular, stereo, and multi-view depth-estimation methods.}
\label{fig:depth-qualitative-comparison}
\end{figure*}

In Table~\ref{tab:depth-estimation-comparison}, confidence-filtered
VGGT-$\Omega$ provides the strongest overall depth accuracy, combining the
best relative accuracy and inlier ratios with a competitive absolute error.
Near-distance DA2 achieves a slightly better absolute error, but performs less
strongly in relative accuracy and the inlier measures. Confidence-filtered VGGT
is also competitive, whereas the remaining selected configurations provide a
less balanced combination of these accuracy measures.

For the efficiency comparison, DA3 is the fastest
, followed by DA2, whereas VGGT-$\Omega$ combines a
comparatively short runtime with the strongest aligned accuracy. It therefore
offers the best overall balance between aligned depth quality, coverage, and
efficiency among the selected outputs.

The Align~$\times$ rows evaluate native metric-depth estimation without affine
correction. Pose-conditioned, confidence-filtered MapAnything performs best in
this setting, indicating that its predicted depth is already close to the
reference metric geometry. DA3 provide the strongest
alternatives among the near-distance configurations, whereas DA2 improves most
substantially after alignment and therefore depends more strongly on correcting
its native scale and offset. 

Overall, VGGT-$\Omega$ gives the strongest aligned
accuracy--efficiency balance, MapAnything provides the best native metric-depth
prediction, DA3 is the fastest.

\section{Conclusion}
\label{sec:conclusion}

We presented a LiDAR-based pipeline that converts camera, LiDAR, and reference-pose data from existing multi-sensor localization datasets into per-view depth maps and visible point-cloud references for benchmarking visual depth estimation and 3D reconstruction in vegetation-rich outdoor environments. To address visibility errors in the accumulated LiDAR measurements and enable their reuse as geometric references, we introduced the Large-Depth-Gap Visibility Score (LDGVS) for point-wise filtering and automatic scale selection in Hidden Point Removal (HPR). The resulting pipeline reduces reliance on dedicated laser-scanner surveys. Experiments on TartanAirV2 and TartanGround show that the combined HPR and LDGVS filtering retains approximately three times the reference density of Poisson surface reconstruction at comparable absolute relative error.

We instantiated the pipeline on BotanicGarden to compare the zero-shot performance of representative feed-forward visual-geometry models with conventional SfM in an under-evaluated vegetation-rich outdoor setting. Feed-forward models generate dense geometry efficiently but can exhibit duplicated structures and cross-frame inconsistencies, whereas conventional SfM produces more coherent geometry but remains sparse and prone to failure. These findings demonstrate the need for domain-specific evaluation and the value of the benchmark in exposing trade-offs in density, consistency, robustness, and efficiency beyond geometric accuracy alone.

\bibliographystyle{IEEEtran}
\bibliography{sections/references}

\clearpage
\begingroup
\section*{Supplementary Material}
\setcounter{section}{0}
\renewcommand{\thesection}{S.\Roman{section}}
\renewcommand{\thesectiondis}{\thesection}
\renewcommand{\theHsection}{supplementary.\arabic{section}}
\section{Benchmark Input and Depth Protocol}
\label{supp:benchmark-input-and-depth-protocol}

\subsection{Common Experiment Settings}
\label{supp:common-experiment-settings}

All methods operate on undistorted images. Each image is uniformly rescaled and
center-cropped to the target input resolution of the corresponding method. For
stereo methods, each synchronized left--right image pair is counted as one
stereo frame, and dense predictions are evaluated only in the left-camera view.

For the depth and reconstruction benchmarks, each evaluated view is represented
by camera-$z$ depth and its associated camera pose. DA2 (Metric),
MoGe-2, VGGT, VGGT-$\Omega$, and DA3
 use their native depth predictions. For DA2 (Relative), we obtain a
relative depth proxy by taking the reciprocal of its predicted inverse depth.
Disparities from SGM and FoundationStereo 
are converted to camera-$z$ depth using the calibrated horizontal focal length
and stereo baseline. For MapAnything, camera-$z$ depth is
taken from the $z$-coordinate of its predicted camera-frame point map.
For SfM methods, reconstructed 3D points are projected into the left image. 
Camera parameters estimated by each method are used 
unless calibrated intrinsics are provided as input.

\section{Post-Processing and Alignment Protocols}
\label{supp:reconstruction-postprocessing-and-alignment}
\label{supp:similarity-alignment-protocols}

\subsection{Weighted Umeyama Similarity Alignment}
\label{supp:similarity-alignment}

Adjacent-chunk registration, reconstruction-to-reference alignment, and
pose-trajectory alignment share the same least-squares $\mathrm{Sim}(3)$ core.
Let $(\widehat{\mathbf{x}}_k,\mathbf{x}_k)$ be a source--target 3D
correspondence with a non-negative base weight $q_k$. If the total base weight
is non-positive, the implementation replaces all base weights by one.
Normalizing $\omega_k=q_k/\sum_jq_j$, we estimate
\begin{equation}
(s,\mathbf{R},\mathbf{t})=
\operatorname*{arg\,min}_{\substack{s>0,\,\mathbf{R}\in\mathrm{SO}(3)\\
                                    \mathbf{t}\in\mathbb{R}^{3}}}
\sum_k\omega_k
\left\lVert\mathbf{x}_k-
\left(s\mathbf{R}\widehat{\mathbf{x}}_k+\mathbf{t}\right)
\right\rVert_2^2.
\end{equation}
For at least three retained pairs, define
\begin{align}
\widehat{\boldsymbol{\mu}}
  &=\sum_k\omega_k\widehat{\mathbf{x}}_k,
&
\boldsymbol{\mu}
  &=\sum_k\omega_k\mathbf{x}_k, \\
\mathbf{H}
  &=\sum_k\omega_k
    (\mathbf{x}_k-\boldsymbol{\mu})
    (\widehat{\mathbf{x}}_k-\widehat{\boldsymbol{\mu}})^{\mathsf T}
    =\mathbf{U}\boldsymbol{\Sigma}\mathbf{V}^{\mathsf T}, \\
\widehat{\sigma}^{2}
  &=\sum_k\omega_k
    \lVert\widehat{\mathbf{x}}_k-\widehat{\boldsymbol{\mu}}\rVert_2^2.
\end{align}
The reflection-corrected weighted extension of the Umeyama solution
\cite{Umeyama-Algorithm} is
\begin{align}
\mathbf{D}
  &=\operatorname{diag}\!\left(1,1,
    \operatorname{sign}\det(\mathbf{U}\mathbf{V}^{\mathsf T})\right), \\
\mathbf{R}
  &=\mathbf{U}\mathbf{D}\mathbf{V}^{\mathsf T}, \\
s
  &=\frac{\operatorname{tr}(\boldsymbol{\Sigma}\mathbf{D})}
          {\widehat{\sigma}^{2}}, \\
\mathbf{t}
  &=\boldsymbol{\mu}-s\mathbf{R}\widehat{\boldsymbol{\mu}}.
\end{align}
If the weighted source variance $\widehat{\sigma}^{2}$ is numerically zero, the
implementation uses the degenerate fallback $s=1$ and estimates only the
centroid translation (with the SVD-derived rotation retained).

For adjacent-chunk registration, and in our adaptation of that procedure for
reconstruction-to-reference alignment, we follow the robust fitting strategy of
\cite{VGGT-Long}. The base weights encode prediction confidence, and the
$\mathrm{Sim}(3)$ transform minimizes
\begin{equation}
\sum_k q_k\rho_{\delta}\!\left(r_k\right),
\qquad
r_k=\left\lVert\mathbf{x}_k-
\left(s\mathbf{R}\widehat{\mathbf{x}}_k+\mathbf{t}\right)
\right\rVert_2,
\end{equation}
where
\begin{equation}
\rho_{\delta}(r)=
\begin{cases}
\tfrac{1}{2}r^2, & r\leq\delta,\\
\delta(r-\tfrac{1}{2}\delta), & r>\delta.
\end{cases}
\end{equation}
Starting from the base-weighted solution above, the iteratively
reweighted least squares (IRLS) update at iteration $\ell+1$ uses
\begin{equation}
w_k^{(\ell+1)}=q_k
\begin{cases}
1, & r_k^{(\ell)}\leq\delta,\\
\delta/r_k^{(\ell)}, & r_k^{(\ell)}>\delta,
\end{cases}
\end{equation}
and resolves the weighted Umeyama problem with the normalized
$w_k^{(\ell+1)}$. We use at most five robust updates, with $\delta$ expressed in the
target coordinate units.

\subsection{Adjacent-Chunk Alignment}
\label{supp:adjacent-chunk-alignment}

For the sequential chunk-alignment protocol, the first chunk defines the
accumulated coordinate frame. Each subsequent chunk is registered to the
preceding, already-aligned chunk using jointly valid, same-pixel 3D points from
their shared frames. We retain all jointly valid
candidate correspondences and assign each pair an equal base weight,
$q_k=1$.
The robust solver in Supplementary Section~\ref{supp:similarity-alignment} is then applied with
$\delta=0.1$ in the accumulated target-frame coordinate units. Predictions duplicated
in an overlap are replaced by the newer copy before fusion.

\subsection{Scene-Level Point-Cloud Alignment}
\label{supp:point-cloud-alignment}

Let $\widehat{\mathbf{T}}_f$ and $\mathbf{T}_f$ be the predicted and reference
camera-to-world poses. Each trajectory is first expressed relative to its own
first finite, invertible pose:
\begin{equation}
\bar{\widehat{\mathbf{T}}}_f
  =\widehat{\mathbf{T}}_0^{-1}\widehat{\mathbf{T}}_f,
\qquad
\bar{\mathbf{T}}_f
  =\mathbf{T}_0^{-1}\mathbf{T}_f.
\end{equation}
For homogeneous pixel $\tilde{\mathbf{u}}$ in frame $f$, let
$\widehat{\mathbf{X}}_{f,\mathbf{u}}$ be its predicted camera-frame 3D point,
including any previously accumulated similarity scale. The corresponding
predicted and reference points in their canonical scene frames are
\begin{align}
\widehat{\mathbf{x}}_{f,\mathbf{u}}
  &=\bar{\widehat{\mathbf{R}}}_f
    \widehat{\mathbf{X}}_{f,\mathbf{u}}
    +\bar{\widehat{\mathbf{t}}}_f, \\
\mathbf{x}_{f,\mathbf{u}}
  &=\bar{\mathbf{R}}_f
    \left(d_{f,\mathbf{u}}\mathbf{K}_f^{-1}
          \tilde{\mathbf{u}}\right)
    +\bar{\mathbf{t}}_f.
\end{align}
Because the predicted and reference point maps share the same frame and pixel
indices, each predicted point is paired directly with the reference point at
the corresponding pixel in the same frame. We retain only pairs for
which both the prediction and reference depth masks are valid.

Let $c_k$ be the prediction confidence, with $c_k=1$ when confidence is
unavailable, and let $m_c=\operatorname{median}_k c_k$. We retain
$c_k>0.1m_c$ when $m_c>0$ (and $c_k>0$ otherwise). Because reference
confidence is one, the base weight is
\begin{equation}
q_k=\sqrt{c_k}.
\end{equation}
We adapt the robust procedure in Supplementary Section~\ref{supp:similarity-alignment} with
$\delta=0.10$\,m because the reference is metric. The fitted transform is
applied to reconstructed points and canonically rebased camera-to-world poses:
\begin{equation}
\widehat{\mathbf{x}}_k'
  =s\mathbf{R}\widehat{\mathbf{x}}_k+\mathbf{t},
\qquad
\widehat{\mathbf{R}}_f'
  =\mathbf{R}\bar{\widehat{\mathbf{R}}}_f,
\qquad
\widehat{\mathbf{t}}_f'
  =s\mathbf{R}\bar{\widehat{\mathbf{t}}}_f+\mathbf{t}.
\end{equation}
The aligned reconstruction is stored separately, so the canonically rebased
but otherwise unaligned result remains available.

\subsection{Per-Frame Affine Depth Alignment}
\label{supp:depth-alignment}

For the common valid-pixel set $\mathcal{V}_f$, scale and shift are estimated
independently for each frame by weighted least squares:
\begin{equation}
(a_f,b_f)=
\operatorname*{arg\,min}_{a,b}
\sum_{i\in\mathcal{V}_f}
w_{f,i}\left(a\hat d_{f,i}+b-d_{f,i}\right)^2.
\end{equation}
Inverse-reference-depth weighting is used with power one and
$\epsilon=0.1$\,m:
\begin{equation}
\tilde w_{f,i}=\frac{1}{\max(d_{f,i},\epsilon)},
\qquad
w_{f,i}=\frac{\tilde w_{f,i}}
{\lvert\mathcal{V}_f\rvert^{-1}
 \sum_{j\in\mathcal{V}_f}\tilde w_{f,j}}.
\end{equation}
The fit is defined only when $\lvert\mathcal{V}_f\rvert\geq10$. We constrain
$a_f\geq10^{-6}$ to ensure a finite log-scale error. The corrected depth is
\begin{equation}
d_{f,i}'=a_f\hat d_{f,i}+b_f.
\end{equation}

\section{Metric Definitions}
\label{supp:metric-definitions}

This section defines the reconstruction, depth-estimation, and runtime metrics used in the experiments. The definitions
follow the benchmark evaluation protocol. Let $\mathbb{1}[\cdot]$ denote the
indicator function, $\lVert\cdot\rVert_2$ the Euclidean norm, and
$\operatorname{clip}(x,-1,1)$ the truncation of $x$ to $[-1,1]$. Invalid or
non-finite samples are excluded through the task-specific validity masks
described below.

\subsection{Point-Cloud Reconstruction Metrics}

Let $\mathcal{P}=\{\mathbf{p}_i\}_{i=1}^{N_P}$ be the reconstructed point cloud
and $\mathcal{G}=\{\mathbf{g}_j\}_{j=1}^{N_G}$ the reference point cloud.
Accuracy and completeness, together with their inlier ratios, are
\begin{align}
\mathrm{Acc}
  &=\frac{1}{N_P}\sum_{i=1}^{N_P}d(\mathbf{p}_i,\mathcal{G}), \\
\mathrm{Acc}_{\tau}
  &=\frac{1}{N_P}\sum_{i=1}^{N_P}
    \mathbb{1}\!\left[d(\mathbf{p}_i,\mathcal{G})\leq\tau\right], \\
\mathrm{Comp}
  &=\frac{1}{N_G}\sum_{j=1}^{N_G}d(\mathbf{g}_j,\mathcal{P}), \\
\mathrm{Comp}_{\tau}
  &=\frac{1}{N_G}\sum_{j=1}^{N_G}
    \mathbb{1}\!\left[d(\mathbf{g}_j,\mathcal{P})\leq\tau\right].
\end{align}
Here, $d(\mathbf{x},\mathcal{S})=
\min_{\mathbf{y}\in\mathcal{S}}\lVert\mathbf{x}-\mathbf{y}\rVert_2$
is the point-to-set nearest-neighbor distance, and $\tau=0.03$
in the reported experiments. The Chamfer distance is computed as:
\begin{equation}
\mathrm{CD}=\mathrm{Acc}+\mathrm{Comp}.
\end{equation}
The F1 score is the harmonic mean of the two inlier ratios,
\begin{equation}
\mathrm{F1}=
\frac{2\,\mathrm{Acc}_{\tau}\mathrm{Comp}_{\tau}}
  {\mathrm{Acc}_{\tau}+\mathrm{Comp}_{\tau}}.
\end{equation}

Reconstruction density is evaluated from the valid predicted-depth mask
$m_i^{\mathrm{pred}}$ 
before the additional cross-frame reprojection-consistency filter:
\begin{equation}
\mathrm{Density}=
\frac{1}{HW}\sum_{i=1}^{HW}\mathbb{1}[m_i^{\mathrm{pred}}].
\label{eq:reconstruction-density}
\end{equation}

\subsection{Depth-Estimation Metrics}

Let $d_i>0$ and $\hat d_i$ denote the reference and predicted depths at pixel
$i$, respectively, and let $\mathcal{V}$ contain the $N_V$ pixels for which both
depth masks are valid. Define
\begin{equation}
e_i=\lvert\hat d_i-d_i\rvert,
\qquad
r_i=\max\!\left(\frac{d_i}{\hat d_i},
                       \frac{\hat d_i}{d_i}\right).
\end{equation}
The absolute and relative depth metrics are
\begin{align}
\mathrm{Abs}
  &=\frac{1}{N_V}\sum_{i\in\mathcal{V}}e_i, \\
\mathrm{Abs}_{\tau}
  &=\frac{1}{N_V}\sum_{i\in\mathcal{V}}
    \mathbb{1}[e_i\leq\tau_{\mathrm{abs}}], \\
\mathrm{AbsRel}
  &=\frac{1}{N_V}\sum_{i\in\mathcal{V}}\frac{e_i}{d_i}, \\
\mathrm{Rel}_{\tau}
  &=\frac{1}{N_V}\sum_{i\in\mathcal{V}}
    \mathbb{1}[r_i<\tau_{\mathrm{rel}}].
\end{align}
Here, $\tau_{\mathrm{abs}}=0.03$\,m and
$\tau_{\mathrm{rel}}=1.03$ in the reported experiments.
Valid-depth density is computed using Eq.~\eqref{eq:reconstruction-density}.

\subsection{Runtime Metric}
\label{supp:runtime-metric}

Runtime is measured at the level of unique timed inference calls. If call
$j$ takes $D_j$ seconds for $B_j$ input images, its normalized per-image
runtime is
\begin{equation}
t_j=\frac{D_j}{B_j}.
\end{equation}
For the $N$ recorded calls in a
sequence, the benchmark upper-trims
\begin{equation}
q=\min\!\left(\left\lfloor0.05N\right\rfloor,N-1\right)
\end{equation}
calls with the largest $t_j$. This fixed rule reduces the influence of
initialization or transient system-load spikes while always retaining at least
one call. The reported sequence-level runtime is the 95th percentile of the
remaining call-level values,
\begin{equation}
\mathrm{Runtime}_{95}=\operatorname{P}_{95}
\left(\left\{t_j\mid j\in\mathcal{R}\right\}\right),
\end{equation}
where $\mathcal{R}$ is the retained-call index set.
The comparison tables use

\section{Visible-Depth Reference Ablation Study}
\label{supp:visible-depth-ablation}

We evaluate the effects of the LDGVSF threshold and the AHPR target
low-visibility ratio on the accuracy and density of the generated visible-depth
references.

\begin{table}[t]
\centering
\scriptsize
\setlength{\tabcolsep}{2pt}
\caption{Ablation study of LDGVSF thresholds and AHPR target low-visibility
ratios on TartanAirV2 Data\_easy and TartanGround Data\_omni.
The percentage in AHPR denotes the target low-visibility ratio $\rho^\star$.
Within each ablation block, the best value per metric is shown in bold and
the second-best distinct value is underlined.
}
\label{tab:Evaluation-of-ablation-study}
\resizebox{\columnwidth}{!}{%
\begin{tabular}{@{}lccccc@{}}
\toprule
Method & AbsRel $\downarrow$ & Rel $\tau$ $\uparrow$ & Abs $\downarrow$ & Abs $\tau$ $\uparrow$ & Density $\uparrow$ \\
\midrule
Cumulation (baseline) & 0.4016 & 0.6384 & 5.4760 & 0.3098 & 0.2524 \\
\midrule
\multicolumn{6}{l}{\textit{LDGVSF threshold ablation}} \\
LDGVSF(0.5) & 0.2268 & 0.6837 & 4.4518 & 0.3368 & \textbf{0.2148} \\
LDGVSF(0.7) & \underline{0.1819} & \textbf{0.6918} & \underline{4.3071} & \underline{0.3477} & \underline{0.1944} \\
LDGVSF(0.9) & \textbf{0.1478} & \underline{0.6915} & \textbf{4.2159} & \textbf{0.3569} & 0.1684 \\
\midrule
\multicolumn{6}{l}{\textit{AHPR target low-visibility ratio and LDGVSF threshold ablation}} \\
AHPR(1\%) & 0.1303 & 0.7003 & 4.2484 & 0.4047 & 0.1095 \\
AHPR(1\%)+LDGVSF(0.7) & \underline{0.1280} & \textbf{0.7026} & \underline{4.1538} & \underline{0.4092} & 0.1056 \\
AHPR(1\%)+LDGVSF(0.9) & \textbf{0.1269} & \underline{0.7010} & \textbf{4.1329} & \textbf{0.4118} & 0.1008 \\
AHPR(3\%) & 0.1370 & 0.6892 & 4.4156 & 0.3683 & 0.1487 \\
AHPR(3\%)+LDGVSF(0.7) & 0.1311 & 0.6945 & 4.2572 & 0.3769 & 0.1392 \\
AHPR(3\%)+LDGVSF(0.9) & 0.1292 & 0.6929 & 4.2166 & 0.3809 & 0.1296 \\
AHPR(5\%) & 0.1545 & 0.6852 & 4.4641 & 0.3460 & 0.1846 \\
AHPR(5\%)+LDGVSF(0.7) & 0.1399 & 0.6955 & 4.2493 & 0.3601 & 0.1664 \\
AHPR(5\%)+LDGVSF(0.9) & 0.1358 & 0.6936 & 4.2069 & 0.3656 & 0.1512 \\
AHPR(10\%) & 0.2279 & 0.6652 & 4.6888 & 0.3226 & \textbf{0.2219} \\
AHPR(10\%)+LDGVSF(0.7) & 0.1593 & 0.6926 & 4.2641 & 0.3498 & \underline{0.1862} \\
AHPR(10\%)+LDGVSF(0.9) & 0.1440 & 0.6913 & 4.1993 & 0.3579 & 0.1645 \\
\bottomrule
\end{tabular}
}
\end{table}

\begin{figure*}[t]
    \centering
    \setlength{\tabcolsep}{2pt}
    \renewcommand{\arraystretch}{1.04}
    \scriptsize
    \begin{tabular}{c c c c c}
        & \textbf{BotanicGarden 1005-01} & \textbf{BotanicGarden 1008-01} & \textbf{TartanAirV2 Easy} & \textbf{TartanGround Omni} \\
        \shortstack{\textbf{Cumulation (baseline)}}
        & \includegraphics[width=0.18\textwidth,height=0.09\textheight]{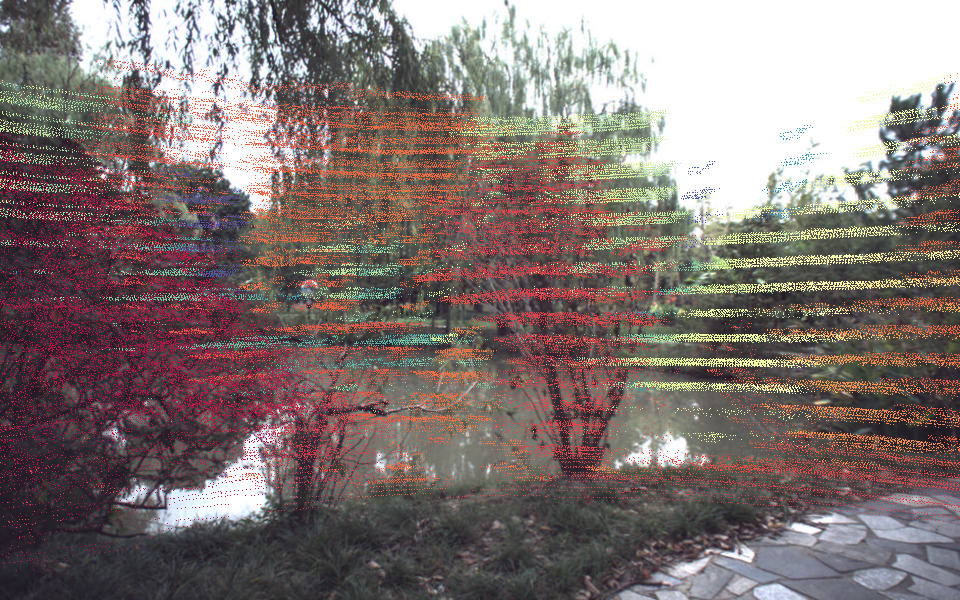}
        & \includegraphics[width=0.18\textwidth,height=0.09\textheight]{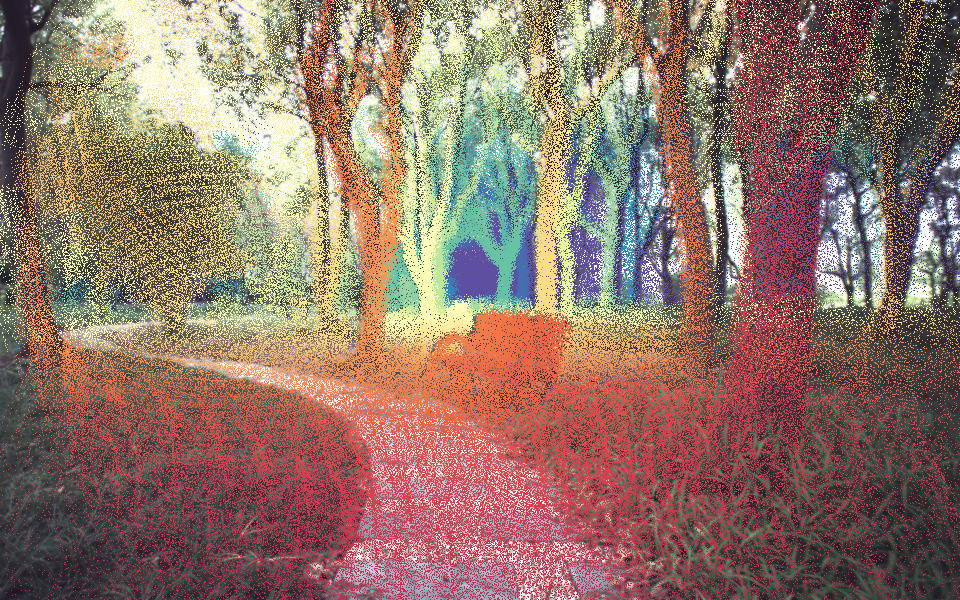}
        & \includegraphics[width=0.18\textwidth,height=0.09\textheight]{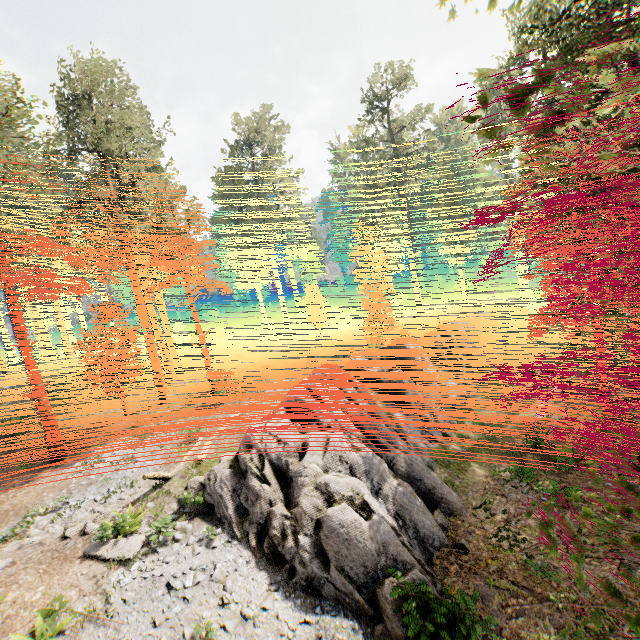}
        & \includegraphics[width=0.18\textwidth,height=0.09\textheight]{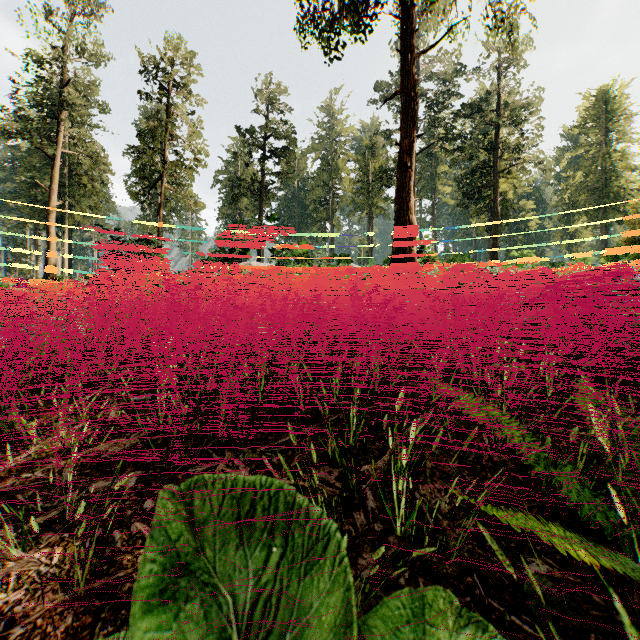} \\
        \shortstack{\textbf{AHPR(1\%)}}
        & \includegraphics[width=0.18\textwidth,height=0.09\textheight]{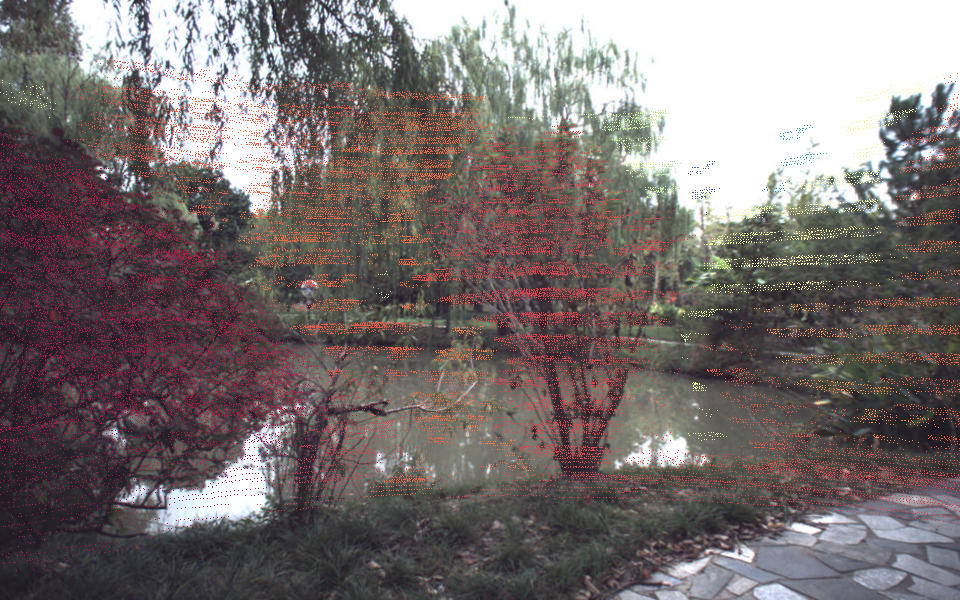}
        & \includegraphics[width=0.18\textwidth,height=0.09\textheight]{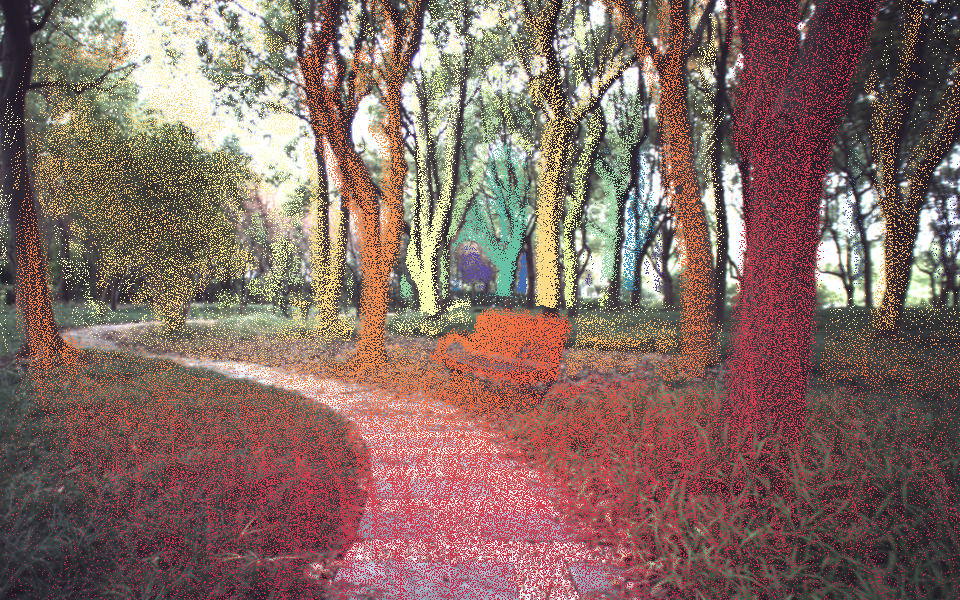}
        & \includegraphics[width=0.18\textwidth,height=0.09\textheight]{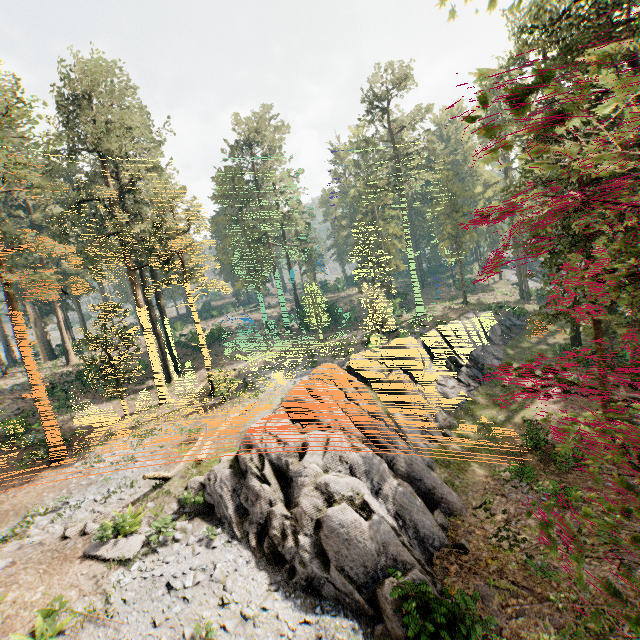}
        & \includegraphics[width=0.18\textwidth,height=0.09\textheight]{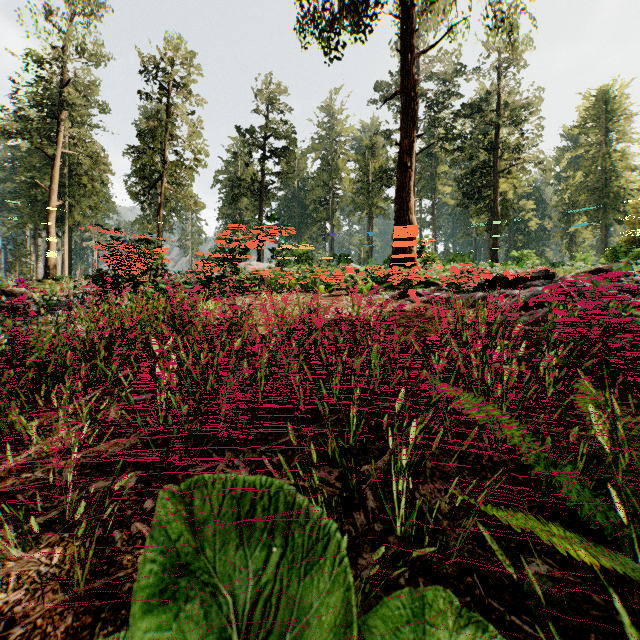} \\
        \shortstack{\textbf{AHPR(1\%)}\\\textbf{LDGVSF(0.7)}}
        & \includegraphics[width=0.18\textwidth,height=0.09\textheight]{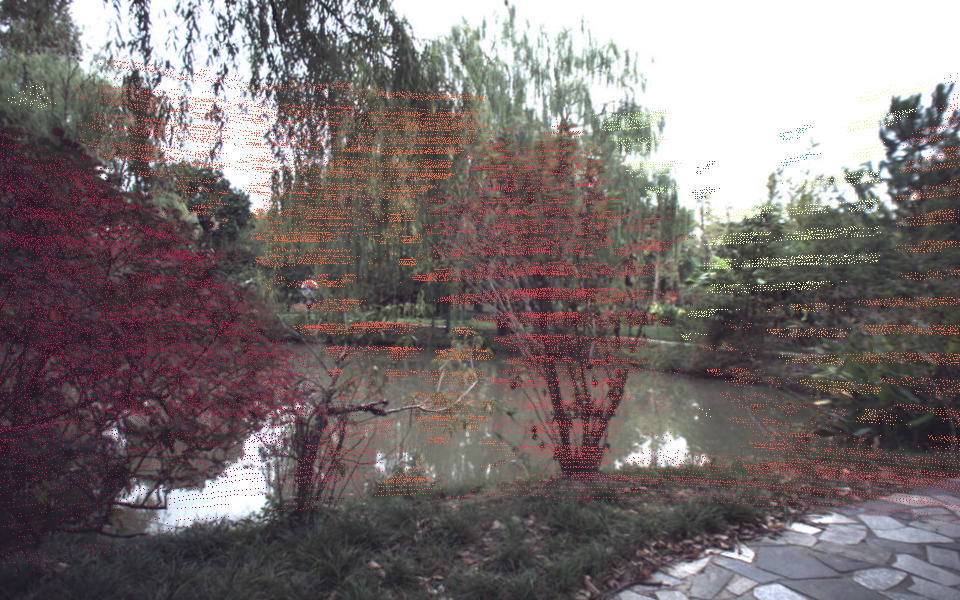}
        & \includegraphics[width=0.18\textwidth,height=0.09\textheight]{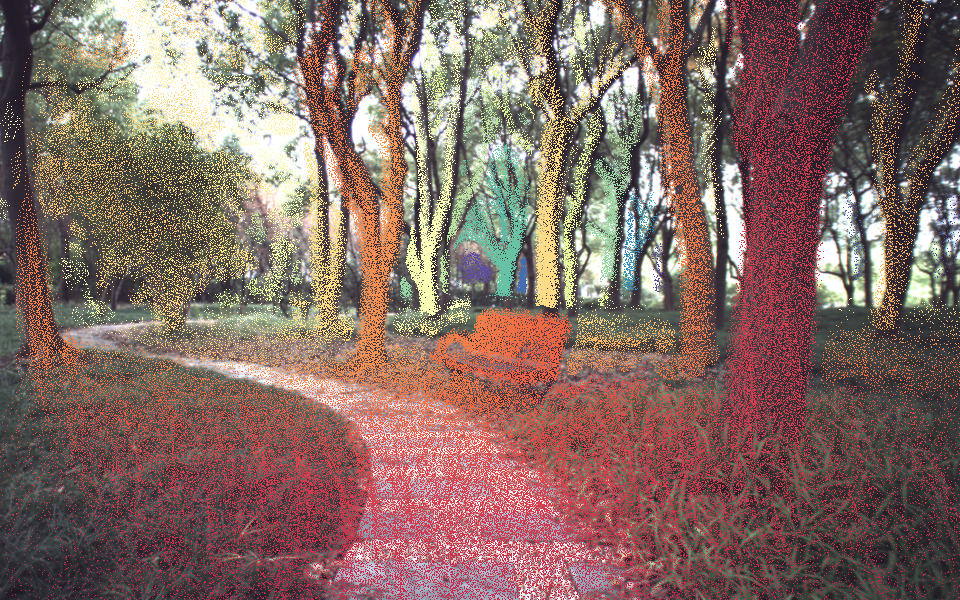}
        & \includegraphics[width=0.18\textwidth,height=0.09\textheight]{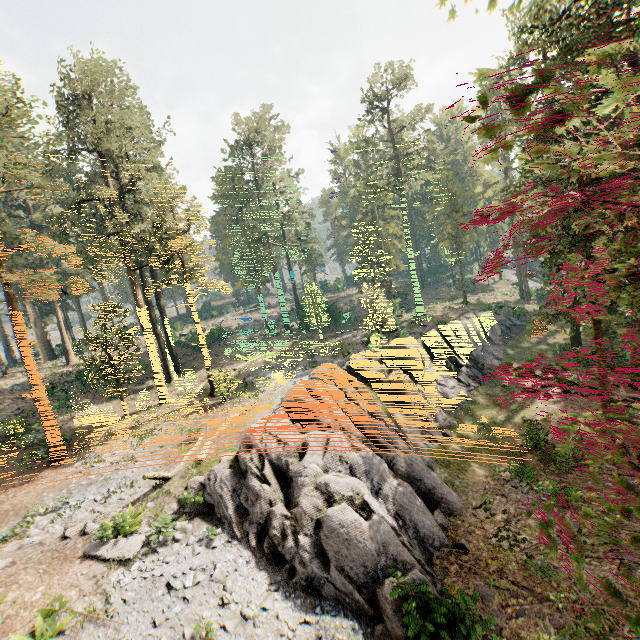}
        & \includegraphics[width=0.18\textwidth,height=0.09\textheight]{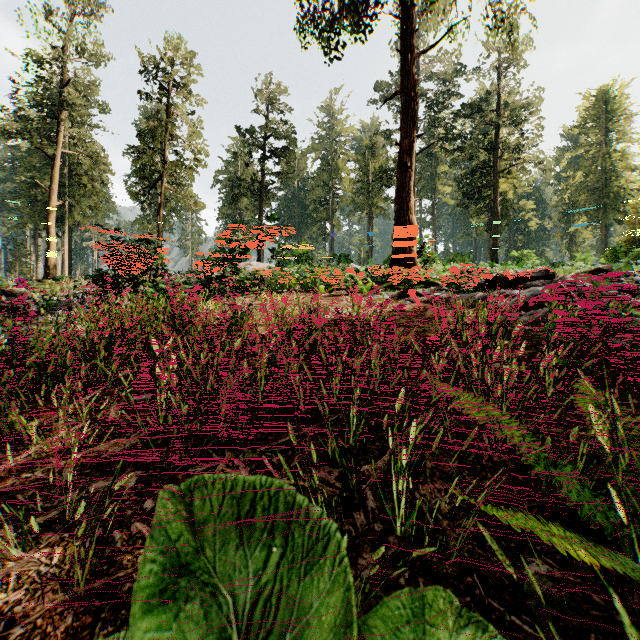} \\
        \shortstack{\textbf{AHPR(1\%)}\\\textbf{LDGVSF(0.9)}}
        & \includegraphics[width=0.18\textwidth,height=0.09\textheight]{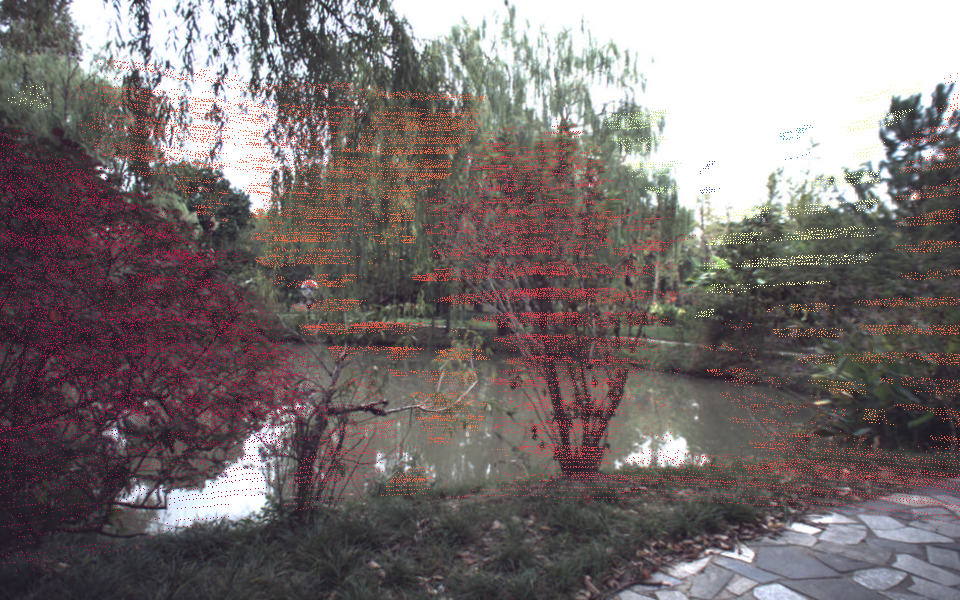}
        & \includegraphics[width=0.18\textwidth,height=0.09\textheight]{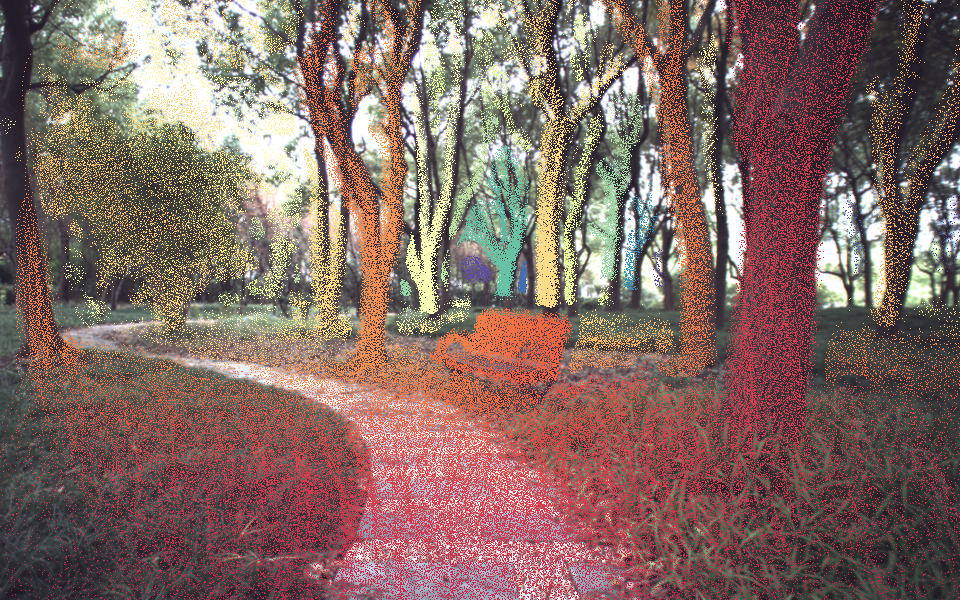}
        & \includegraphics[width=0.18\textwidth,height=0.09\textheight]{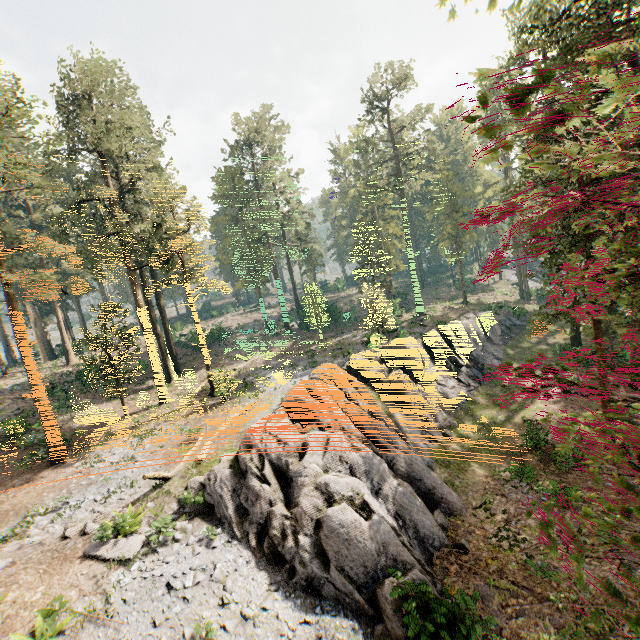}
        & \includegraphics[width=0.18\textwidth,height=0.09\textheight]{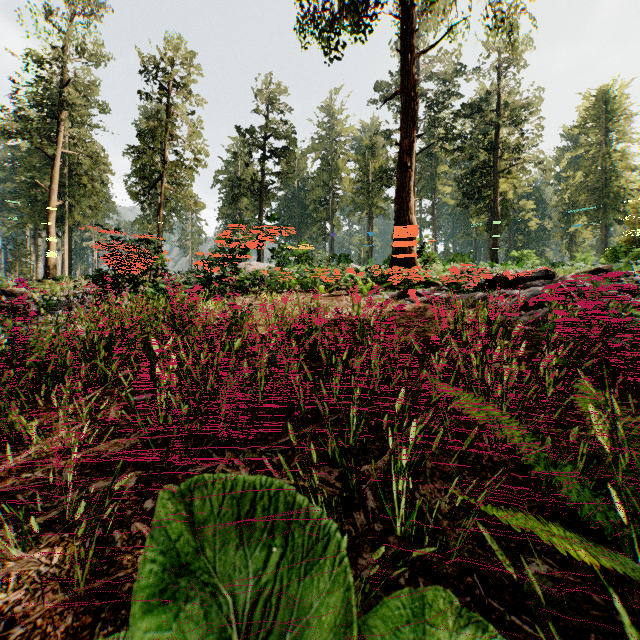} \\
        \shortstack{\textbf{AHPR(5\%)}}
        & \includegraphics[width=0.18\textwidth,height=0.09\textheight]{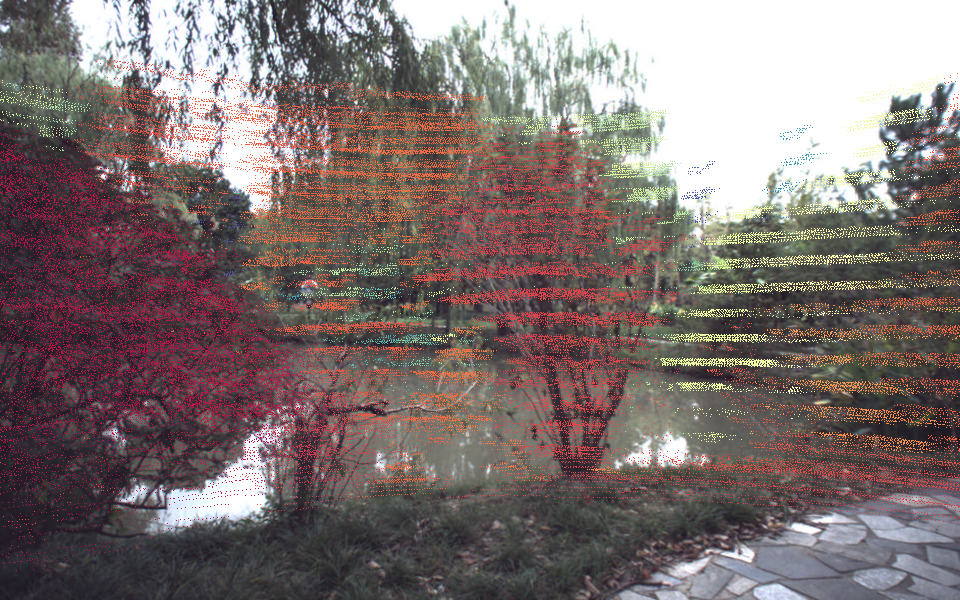}
        & \includegraphics[width=0.18\textwidth,height=0.09\textheight]{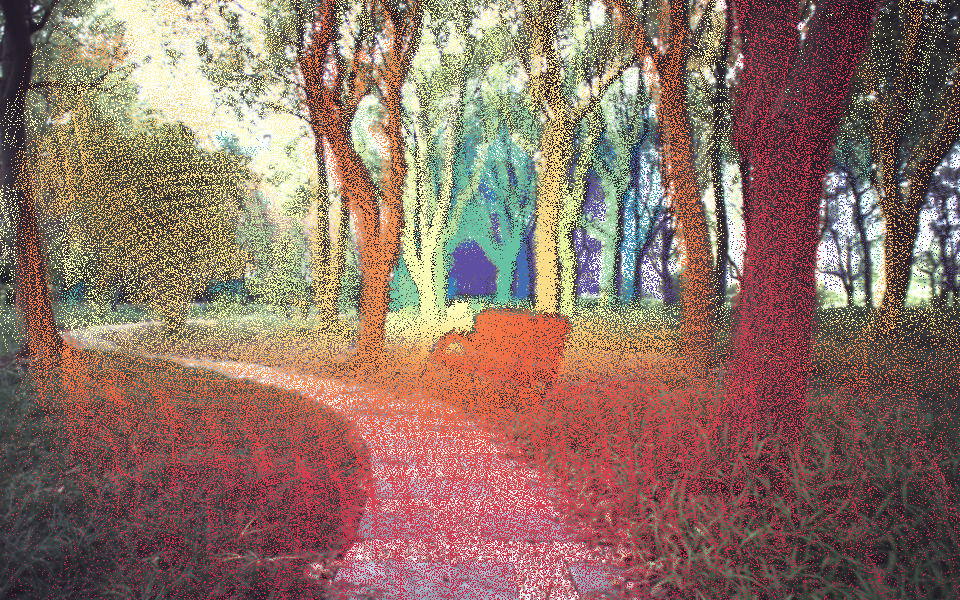}
        & \includegraphics[width=0.18\textwidth,height=0.09\textheight]{figs_low_quality/experiments/depth_ablation/TartanAir_Easy_AHPR_005_no_vis_filter.jpg}
        & \includegraphics[width=0.18\textwidth,height=0.09\textheight]{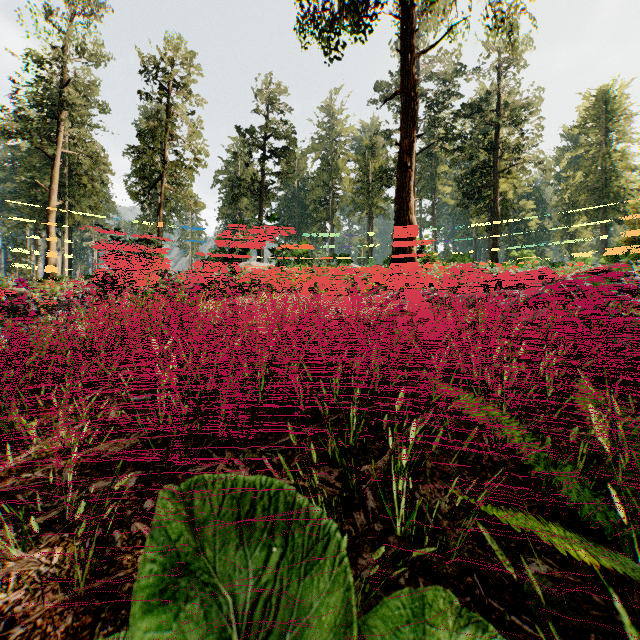} \\
        \shortstack{\textbf{AHPR(5\%)}\\\textbf{LDGVSF(0.7)}}
        & \includegraphics[width=0.18\textwidth,height=0.09\textheight]{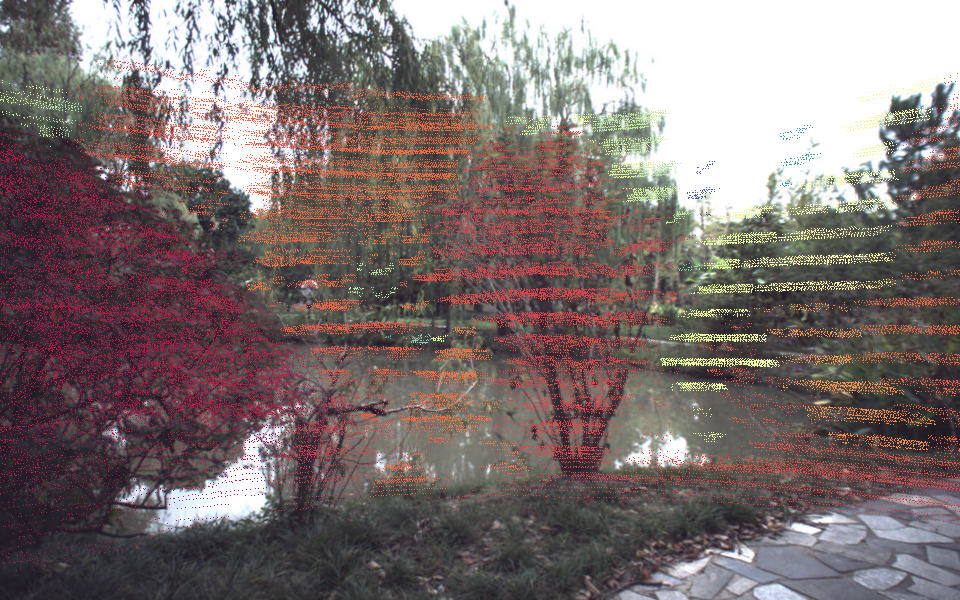}
        & \includegraphics[width=0.18\textwidth,height=0.09\textheight]{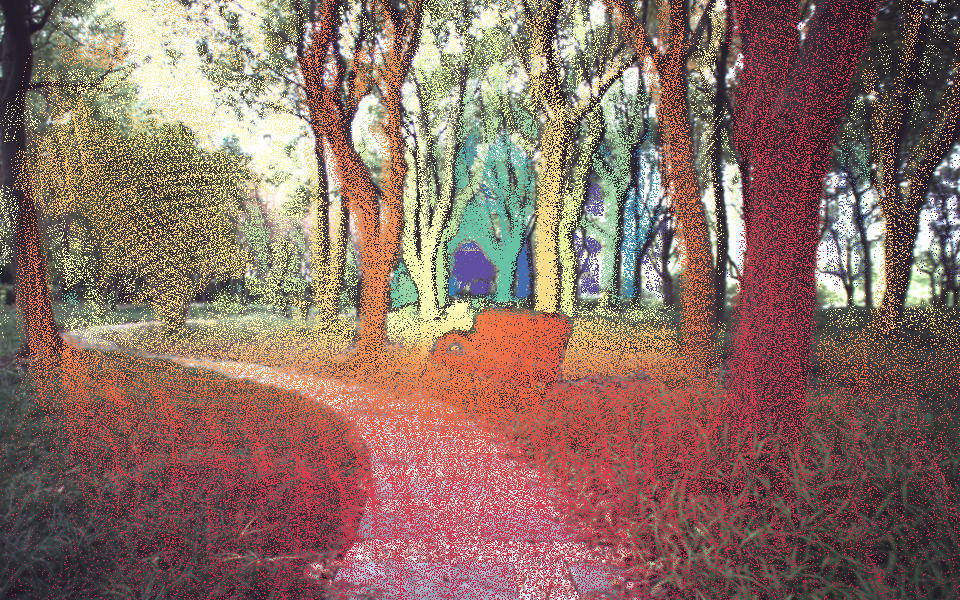}
        & \includegraphics[width=0.18\textwidth,height=0.09\textheight]{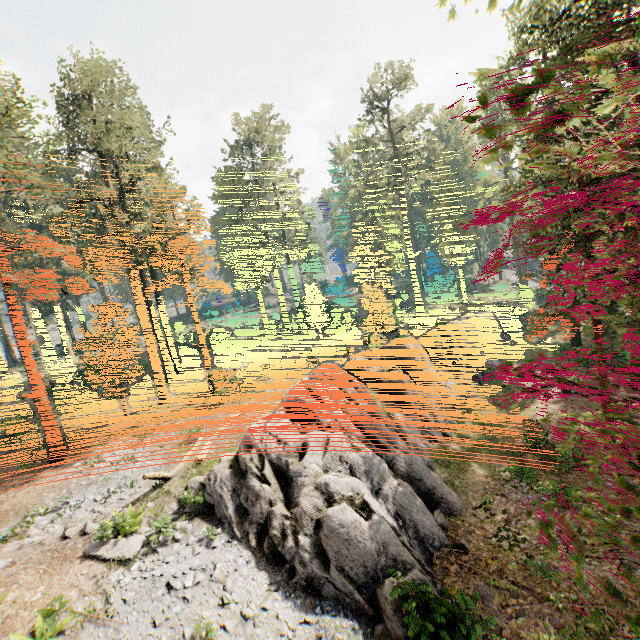}
        & \includegraphics[width=0.18\textwidth,height=0.09\textheight]{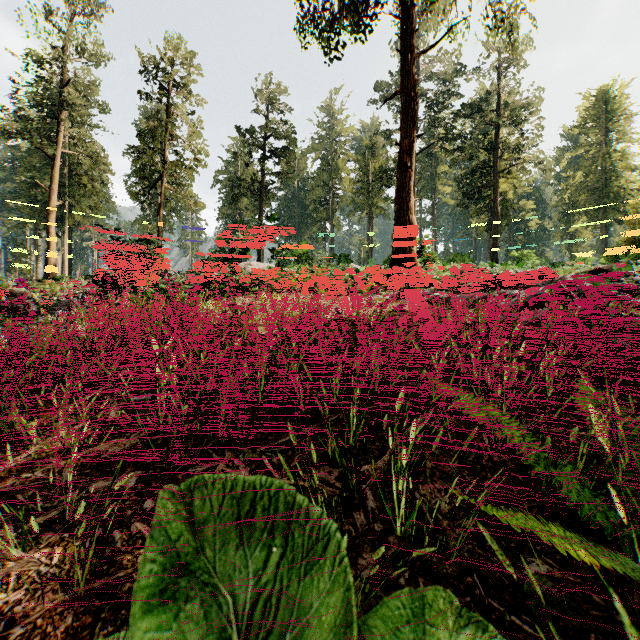} \\
        \shortstack{\textbf{AHPR(5\%)}\\\textbf{LDGVSF(0.9)}}
        & \includegraphics[width=0.18\textwidth,height=0.09\textheight]{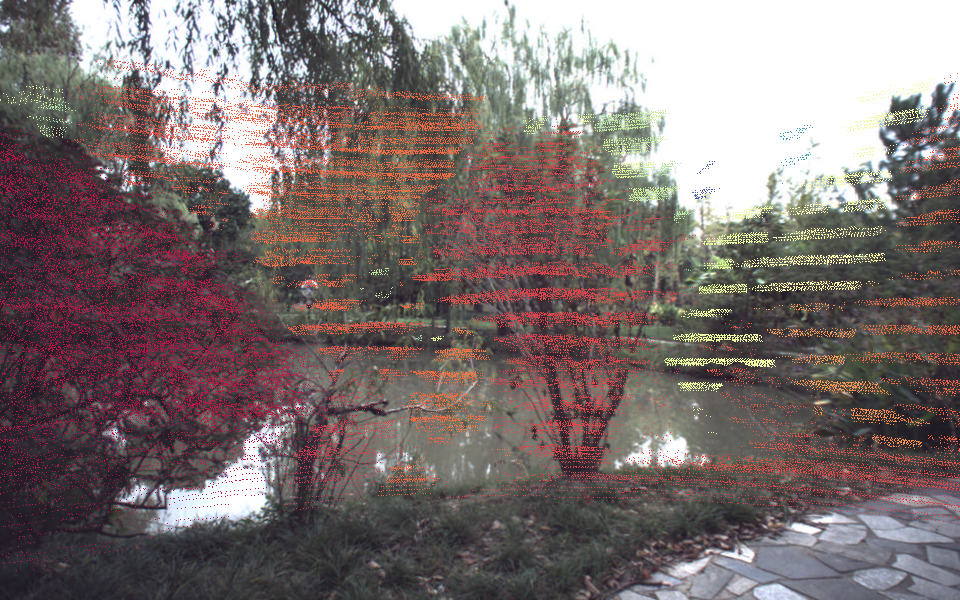}
        & \includegraphics[width=0.18\textwidth,height=0.09\textheight]{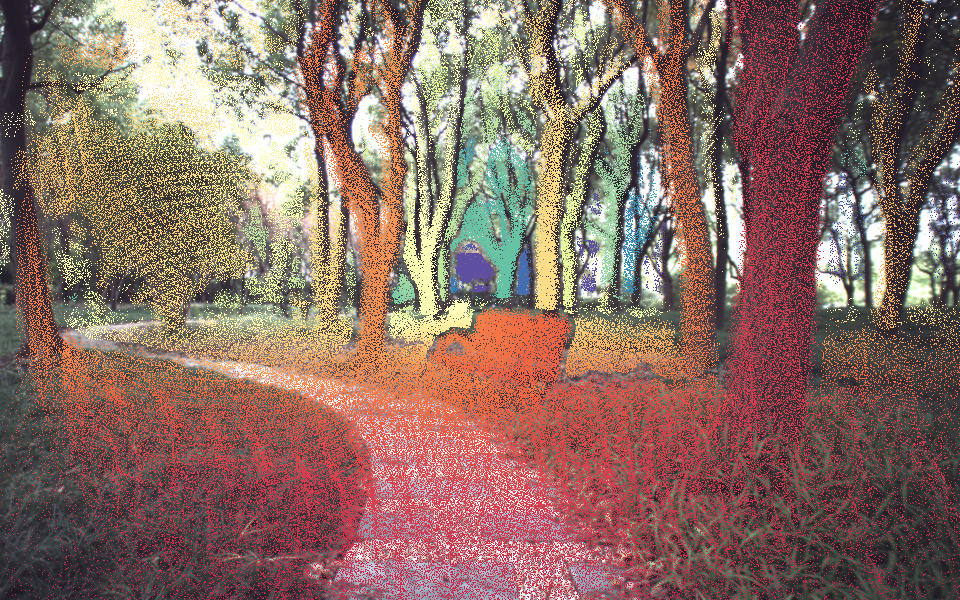}
        & \includegraphics[width=0.18\textwidth,height=0.09\textheight]{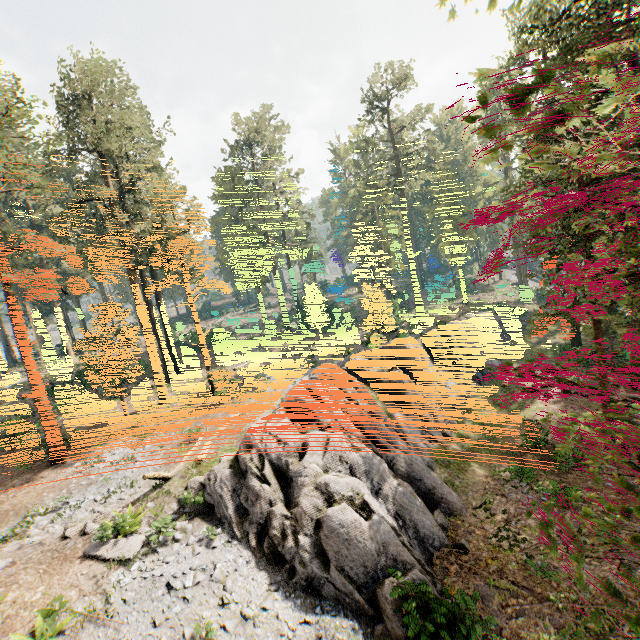}
        & \includegraphics[width=0.18\textwidth,height=0.09\textheight]{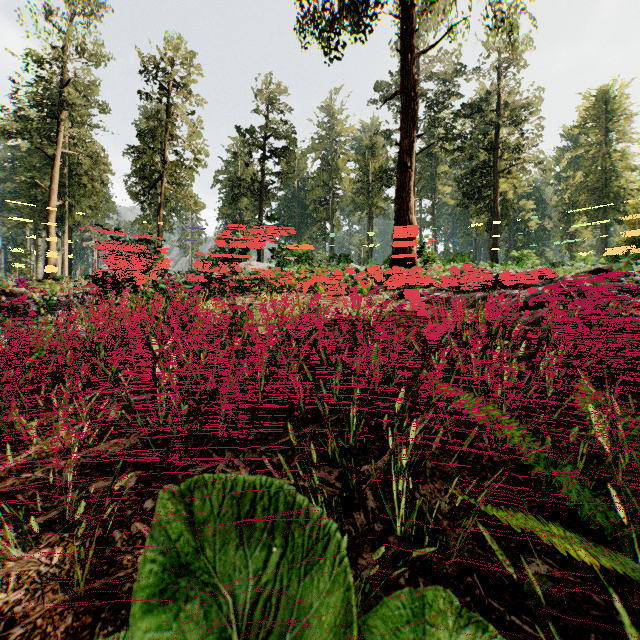} \\
        \shortstack{\textbf{AHPR(10\%)}}
        & \includegraphics[width=0.18\textwidth,height=0.09\textheight]{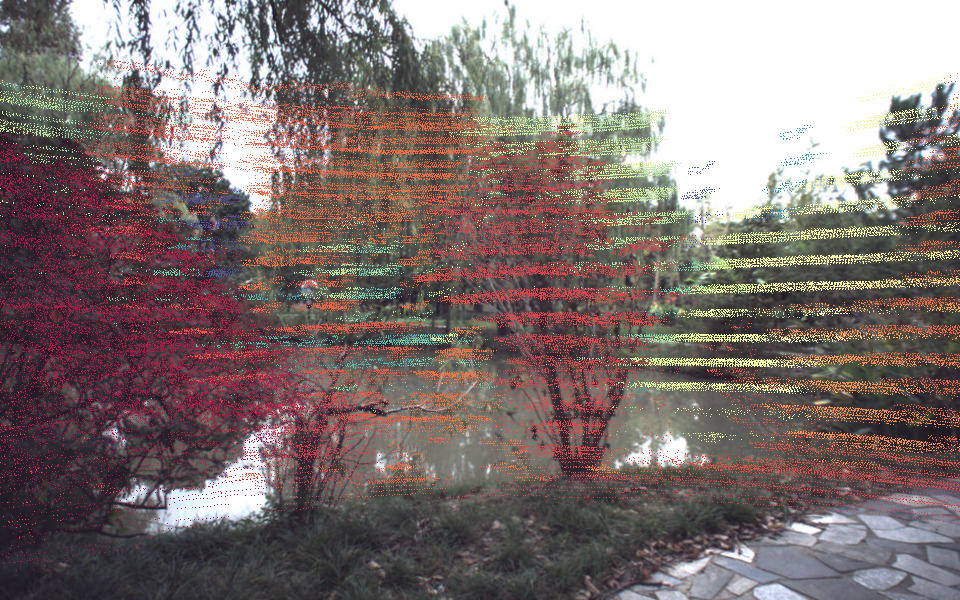}
        & \includegraphics[width=0.18\textwidth,height=0.09\textheight]{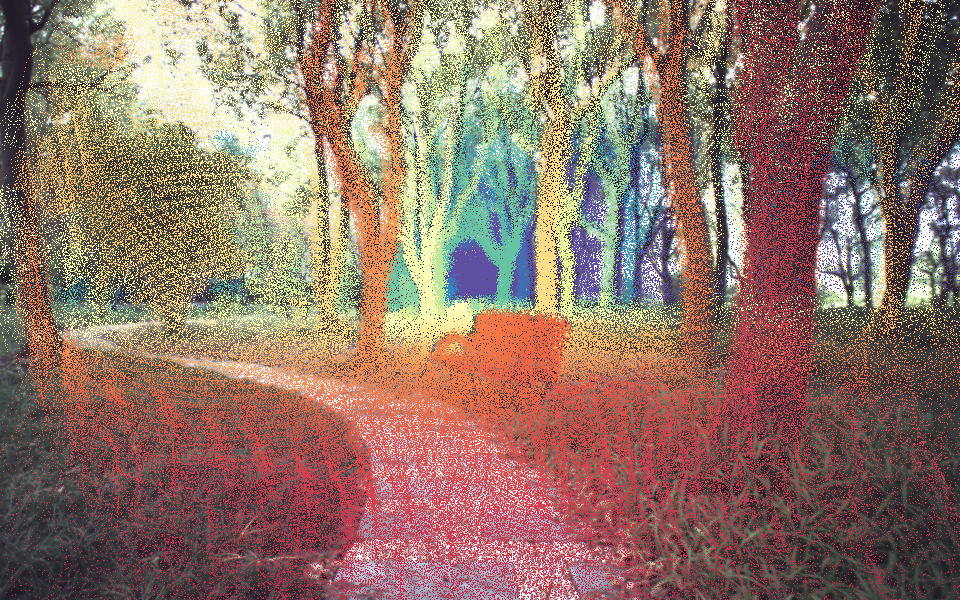}
        & \includegraphics[width=0.18\textwidth,height=0.09\textheight]{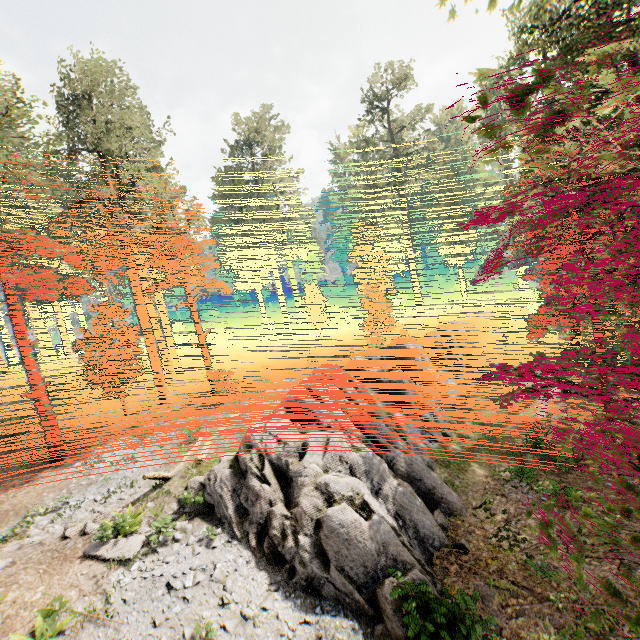}
        & \includegraphics[width=0.18\textwidth,height=0.09\textheight]{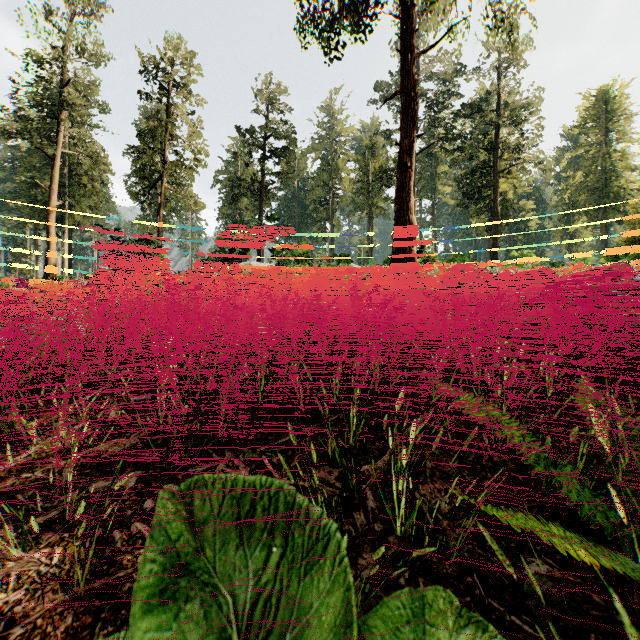} \\
        \shortstack{\textbf{AHPR(10\%)}\\\textbf{LDGVSF(0.7)}}
        & \includegraphics[width=0.18\textwidth,height=0.09\textheight]{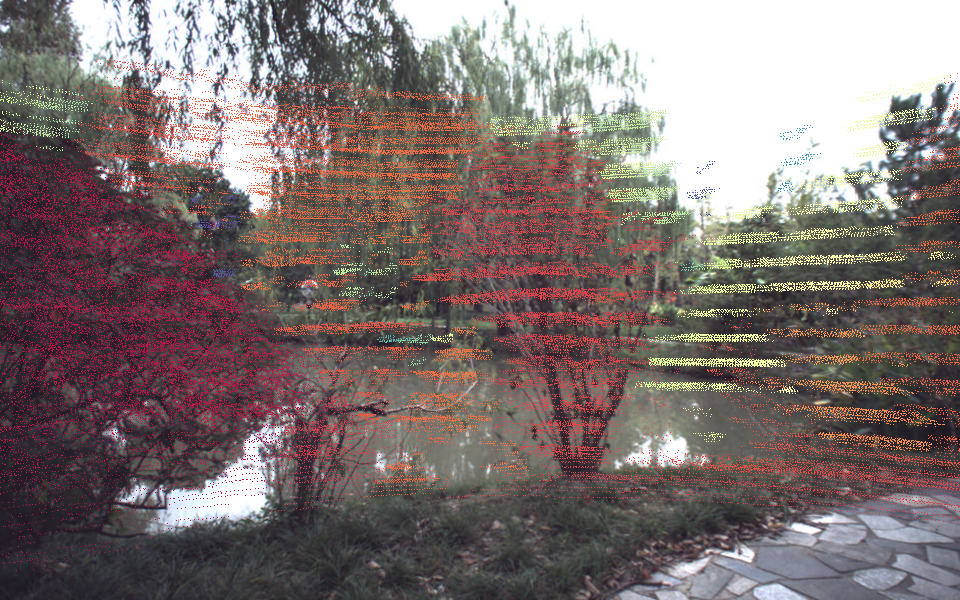}
        & \includegraphics[width=0.18\textwidth,height=0.09\textheight]{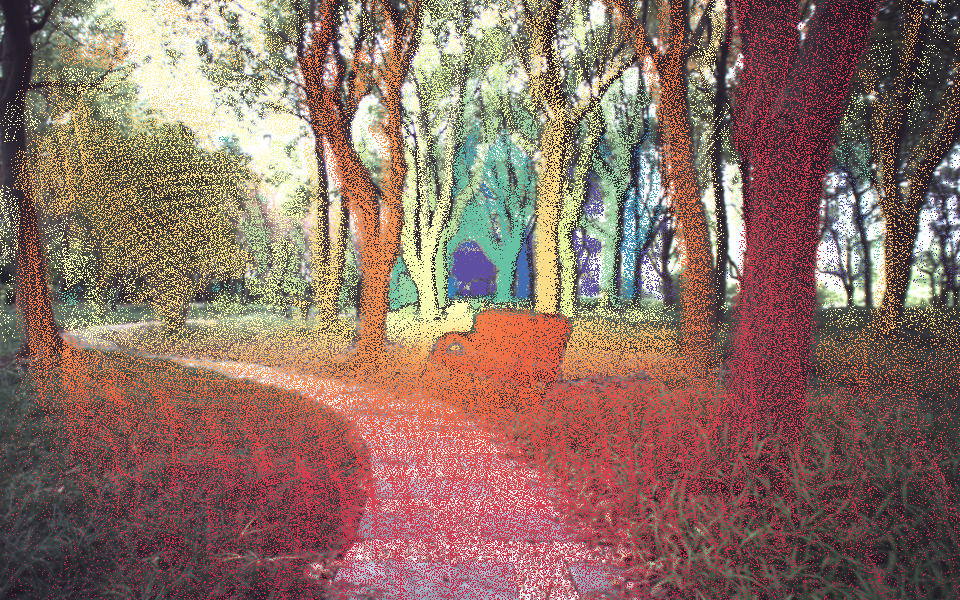}
        & \includegraphics[width=0.18\textwidth,height=0.09\textheight]{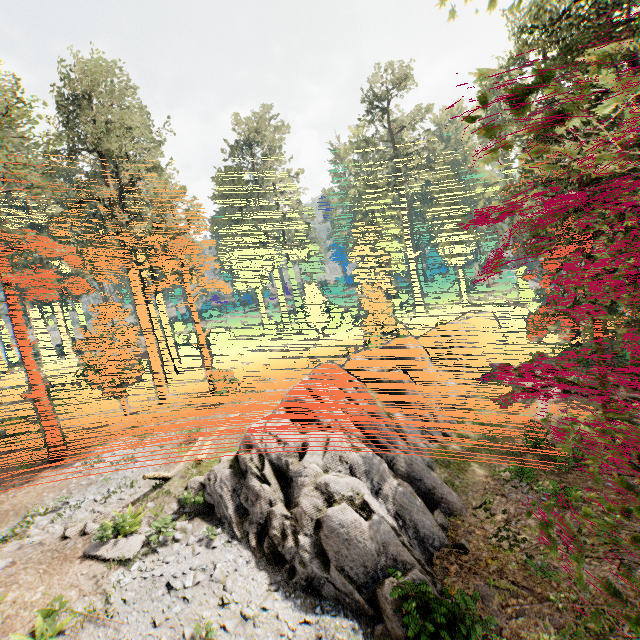}
        & \includegraphics[width=0.18\textwidth,height=0.09\textheight]{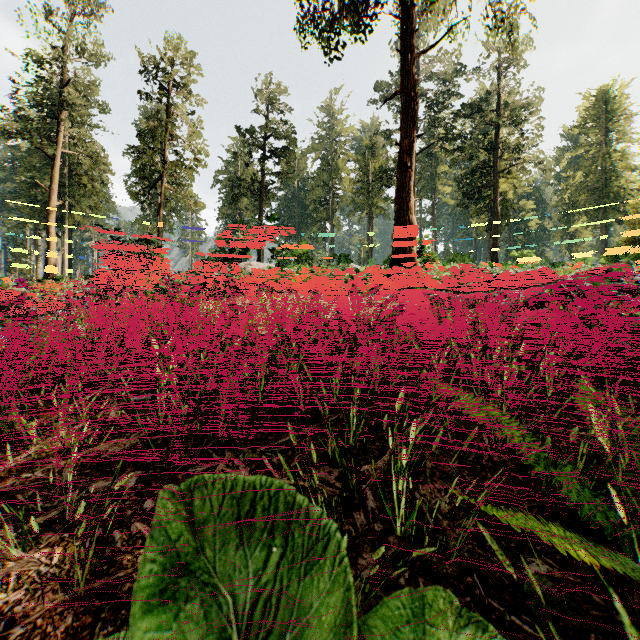} \\
        \shortstack{\textbf{AHPR(10\%)}\\\textbf{LDGVSF(0.9)}}
        & \includegraphics[width=0.18\textwidth,height=0.09\textheight]{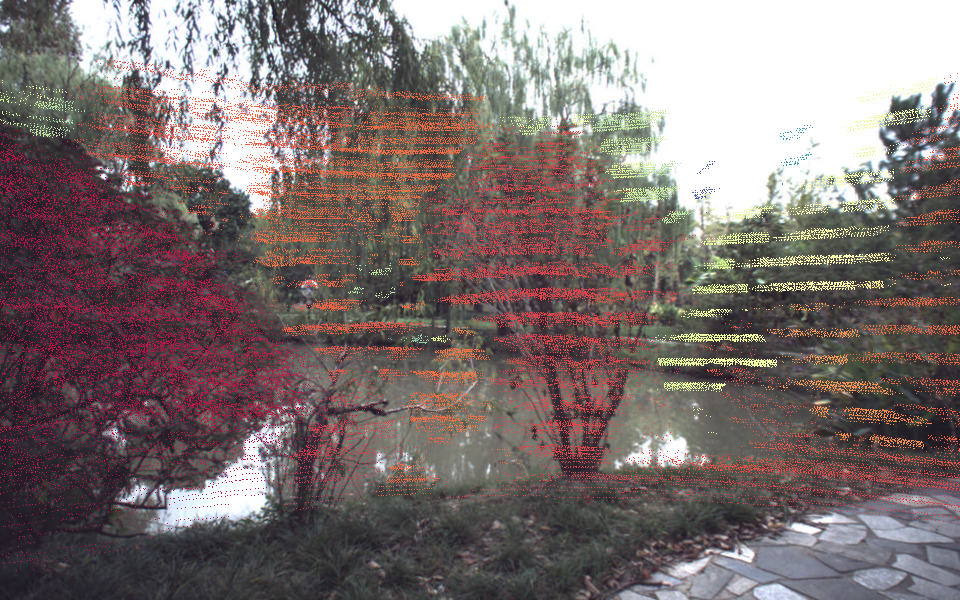}
        & \includegraphics[width=0.18\textwidth,height=0.09\textheight]{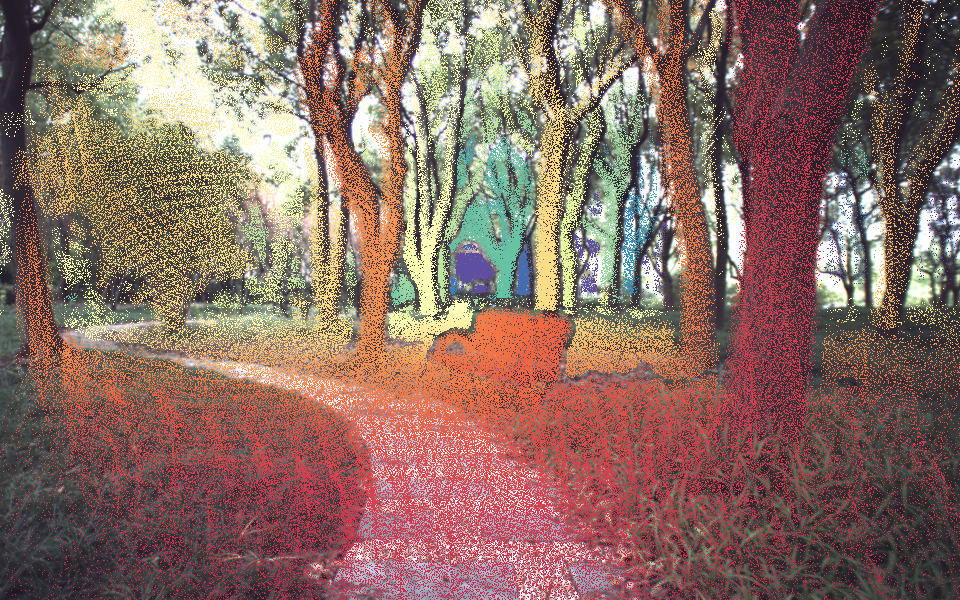}
        & \includegraphics[width=0.18\textwidth,height=0.09\textheight]{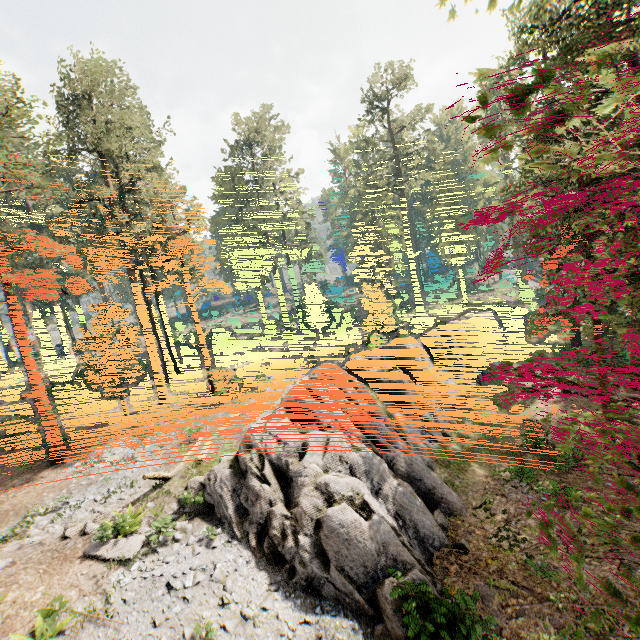}
        & \includegraphics[width=0.18\textwidth,height=0.09\textheight]{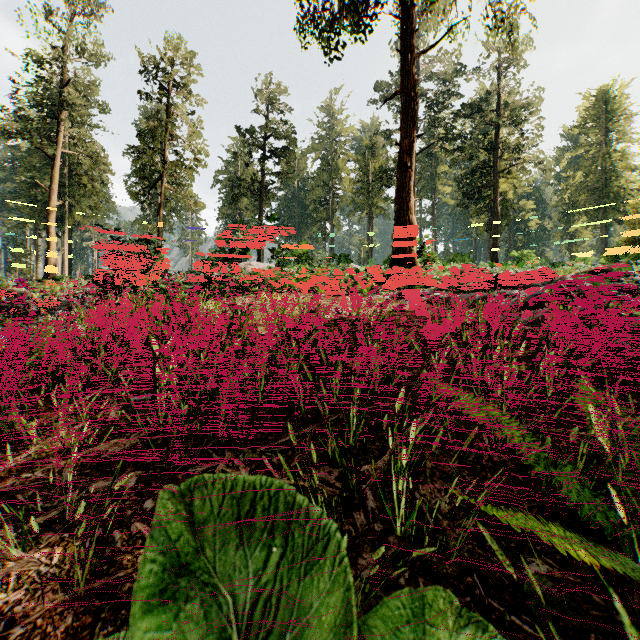} \\
    \end{tabular}

    \caption{Qualitative comparison of the unfiltered cumulation baseline and
    AHPR results with and without LDGVSF-based filtering of
    low-visibility-score points.}
    \label{fig:ahpr-ldgvsf-visualization}
\end{figure*}

Table~\ref{tab:Evaluation-of-ablation-study} and
Figure~\ref{fig:ahpr-ldgvsf-visualization} show a consistent
accuracy--density trade-off. Increasing the AHPR target low-visibility ratio
retains more projected points but generally reduces accuracy. At a fixed AHPR
setting, applying LDGVSF removes low-visibility-score points and generally
improves the error metrics at the cost of lower reference-depth density.
Raising the LDGVSF threshold from 0.7 to 0.9 filters more points, while the
additional changes in accuracy are modest and mixed. The visualization shows
the corresponding coverage trend: the 1\% configurations are sparse, the 10\%
configurations are denser, and the 5\% configurations provide intermediate
coverage; a threshold of 0.7 also retains more support than 0.9.

Several parameter combinations therefore provide a reasonable balance between
accuracy and density. We choose AHPR(5\%)+LDGVSF(0.7) to generate the reference
depth for our pipeline because it retains relatively dense reference-depth
coverage while limiting the retention of erroneous depth values in the
generated references.

\section{Extended Visibility Comparison on BotanicGarden}
\label{supp:botanic-garden-visibility-comparison}

Figure~\ref{fig:cumulated-visible-point-cloud-botanic-garden-full} extends the
qualitative comparison to the challenging real-world BotanicGarden data
acquired with a Livox LiDAR\@. Compared with the simulated example, PSR in (d) retains
a denser visible-point projection in this scene, but it still discards many
points relative to the cumulation baseline in (b). In contrast,
AHPR(5\%)+LDGVSF(0.7) in (j) preserves substantially more visible-point support than
PSR and maintains coverage close to the baseline while filtering likely
occluded points.

\begin{figure*}[t]
    \raggedright
    \captionsetup[subfloat]{justification=centering}
    \subfloat[{Right-view RGB image\label{fig:depth-right-image-botanic-garden}}]{%
\begin{minipage}[t]{0.24\linewidth}
\centering
        \includegraphics[width=\linewidth,height=0.11\textheight]{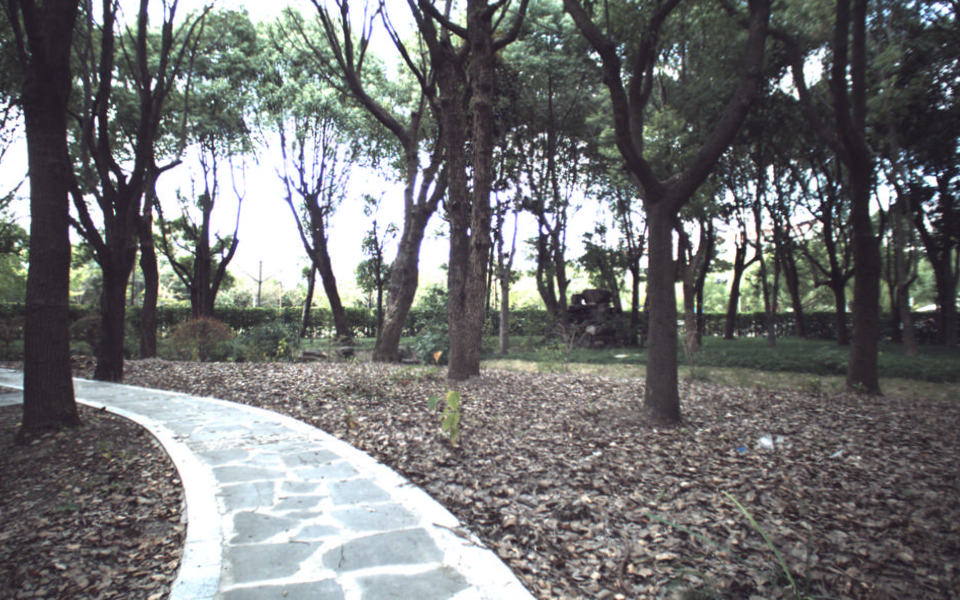}
\end{minipage}}\hspace{0.01\linewidth}%
    \subfloat[{Cumulation (baseline)\label{fig:depth-from-cumulation-botanic-garden}}]{%
\begin{minipage}[t]{0.24\linewidth}
\centering
        \includegraphics[width=\linewidth,height=0.11\textheight]{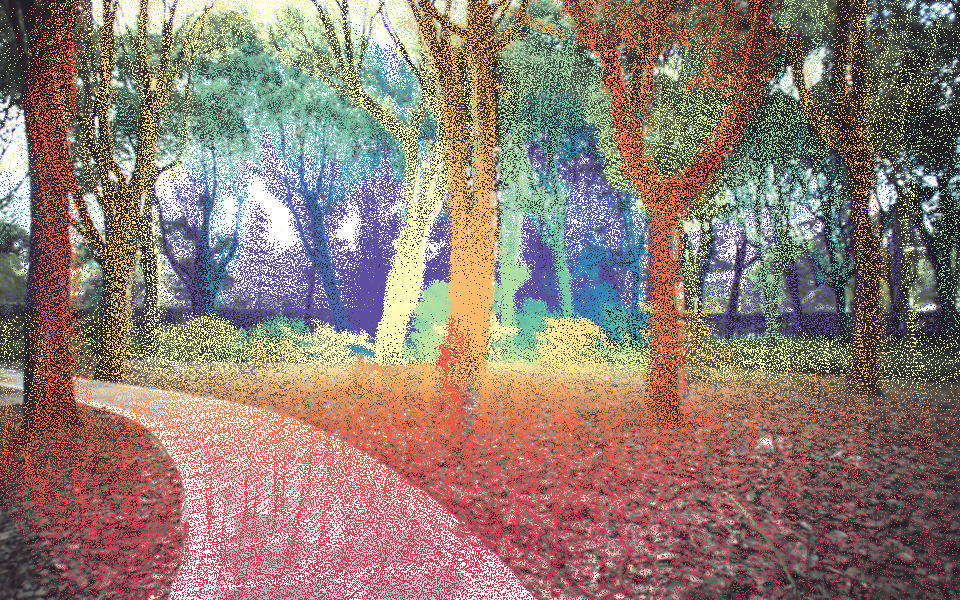}
\end{minipage}}\hspace{0.01\linewidth}%
    \subfloat[{Cumulation+CP\label{fig:depth-from-cumulation-cp-botanic-garden}}]{%
\begin{minipage}[t]{0.24\linewidth}
\centering
        \includegraphics[width=\linewidth,height=0.11\textheight]{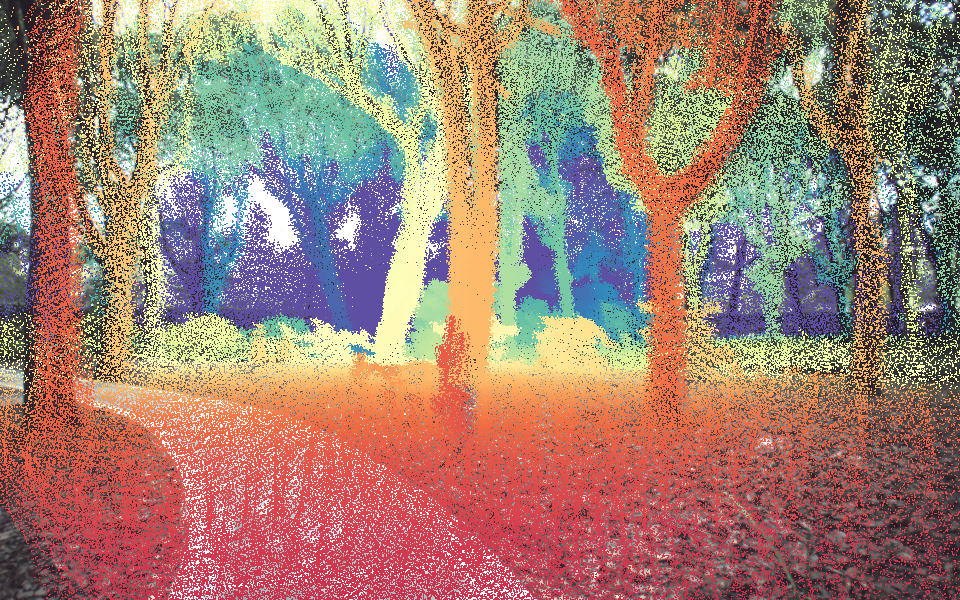}
\end{minipage}}\hspace{0.01\linewidth}%
    \subfloat[{PSR\label{fig:depth-from-psr-botanic-garden}}]{%
\begin{minipage}[t]{0.24\linewidth}
\centering
        \includegraphics[width=\linewidth,height=0.11\textheight]{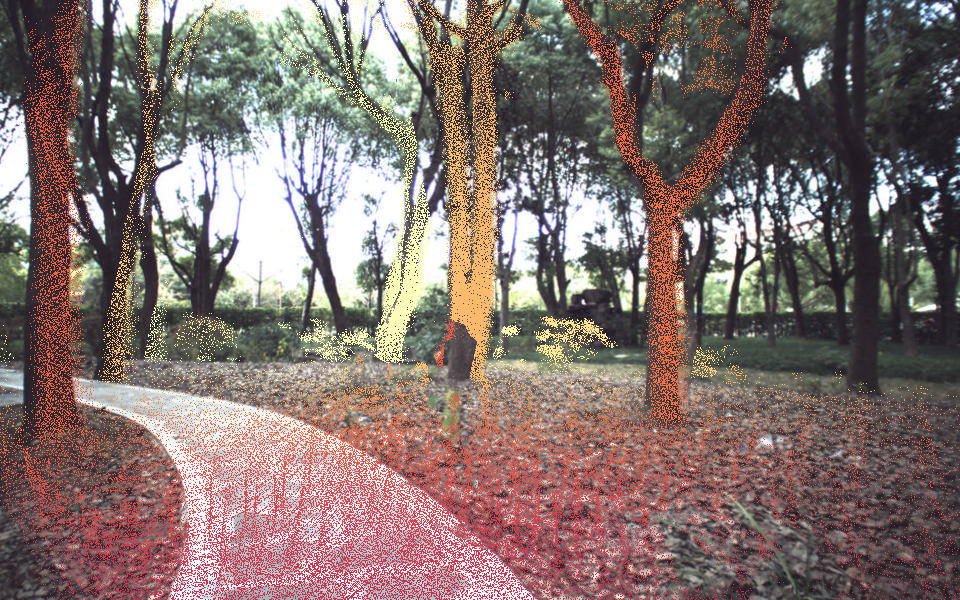}
\end{minipage}}\par\smallskip
    \subfloat[{SGM\label{fig:depth-from-sgm-botanic-garden}}]{%
\begin{minipage}[t]{0.24\linewidth}
\centering
        \includegraphics[width=\linewidth,height=0.11\textheight]{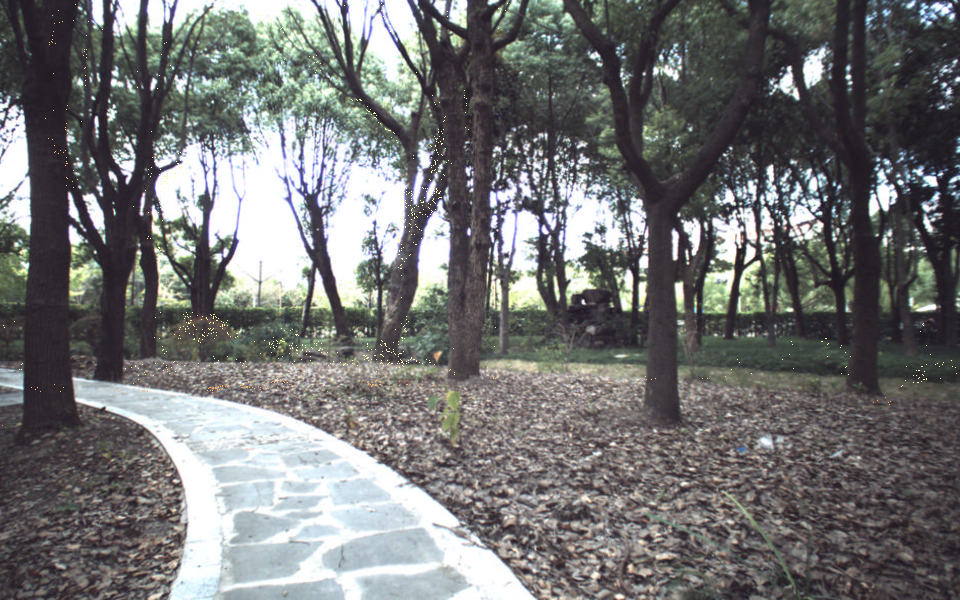}
\end{minipage}}\hspace{0.01\linewidth}%
    \subfloat[{LD\label{fig:depth-from-ld-botanic-garden}}]{%
\begin{minipage}[t]{0.24\linewidth}
\centering
        \includegraphics[width=\linewidth,height=0.11\textheight]{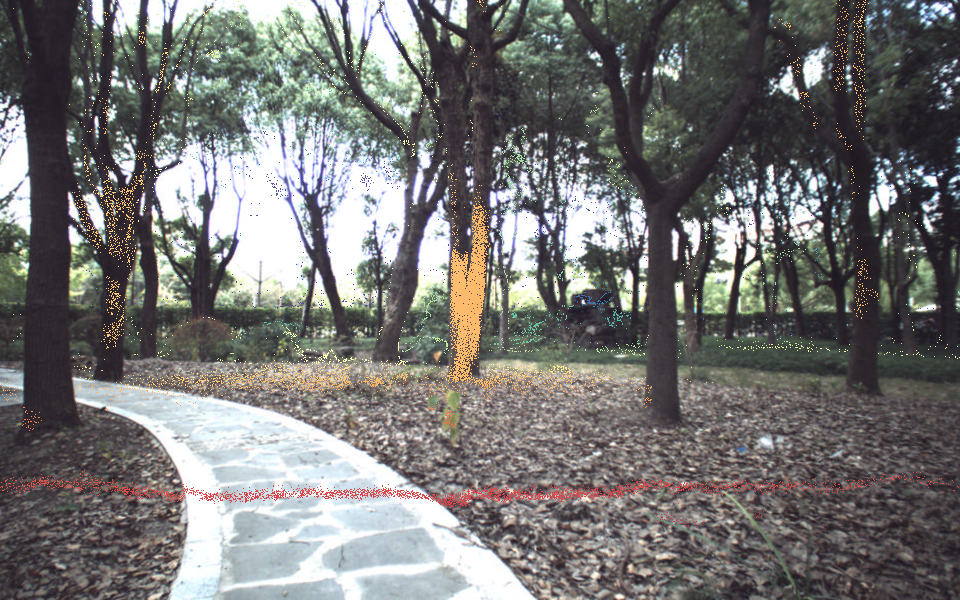}
\end{minipage}}\hspace{0.01\linewidth}%
    \subfloat[{Visibility estimation (0.7)\label{fig:depth-from-visibility-estimation-botanic-garden}}]{%
\begin{minipage}[t]{0.24\linewidth}
\centering
        \includegraphics[width=\linewidth,height=0.11\textheight]{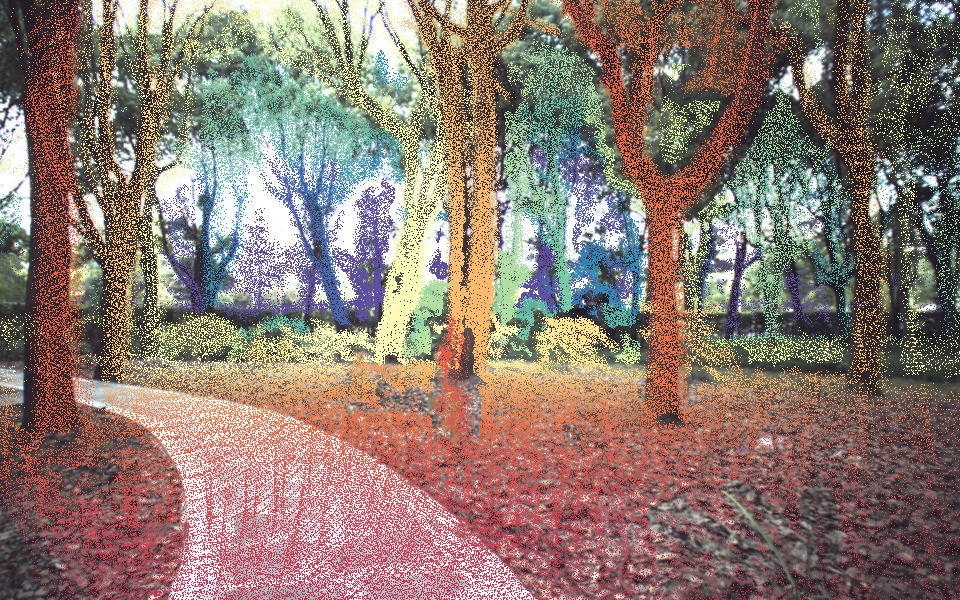}
\end{minipage}}\hspace{0.01\linewidth}%
    \subfloat[{LDGVSF(0.7)\label{fig:depth-from-ldgvsf-botanic-garden}}]{%
\begin{minipage}[t]{0.24\linewidth}
\centering
        \includegraphics[width=\linewidth,height=0.11\textheight]{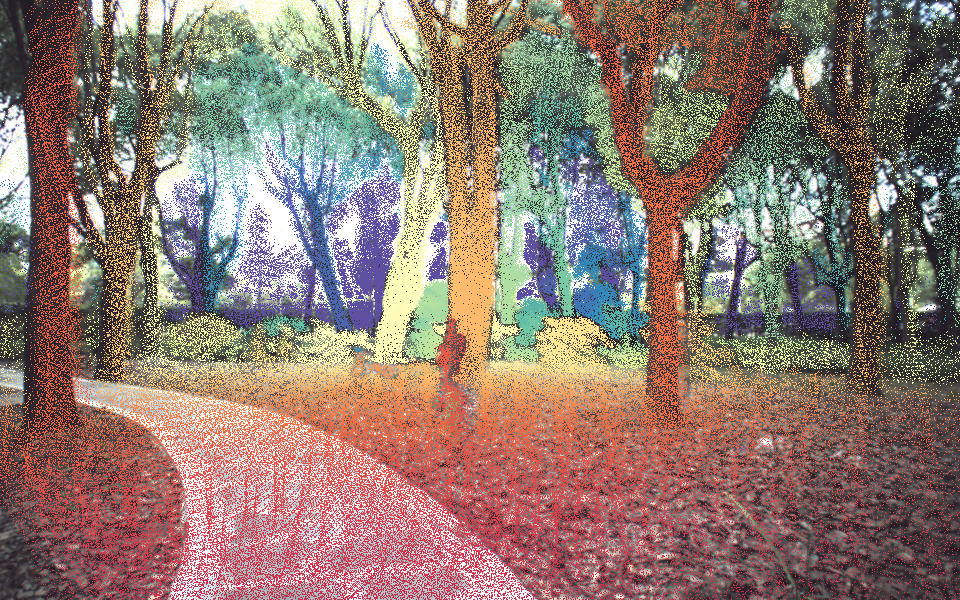}
\end{minipage}}\par\smallskip
    \subfloat[{AHPR(5\%)\label{fig:depth-from-ahpr-botanic-garden}}]{%
\begin{minipage}[t]{0.24\linewidth}
\centering
        \includegraphics[width=\linewidth,height=0.11\textheight]{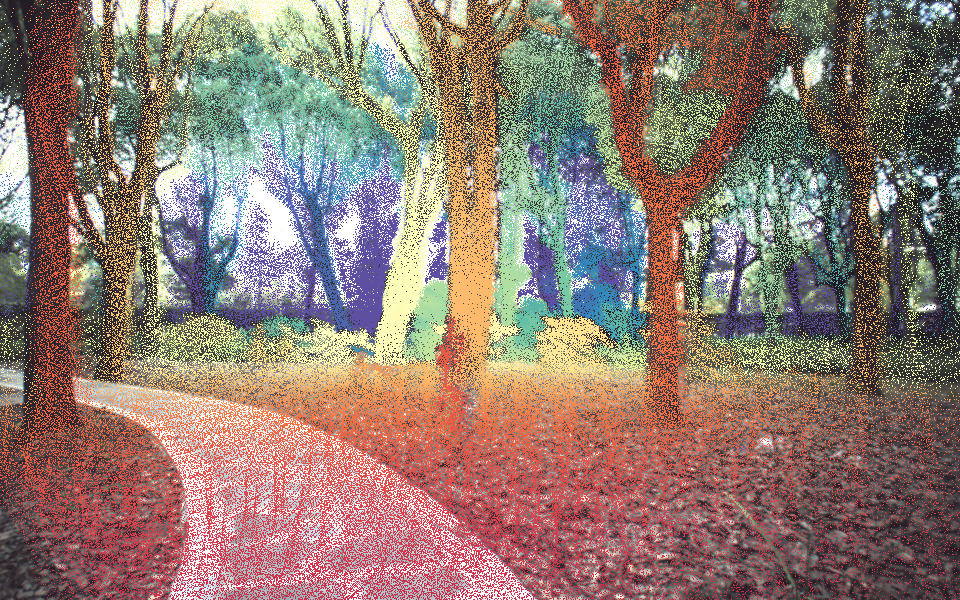}
\end{minipage}}\hspace{0.01\linewidth}%
    \subfloat[{AHPR(5\%)+LDGVSF(0.7)\label{fig:depth-from-ahpr-ldgvsf-botanic-garden}}]{%
\begin{minipage}[t]{0.24\linewidth}
\centering
        \includegraphics[width=\linewidth,height=0.11\textheight]{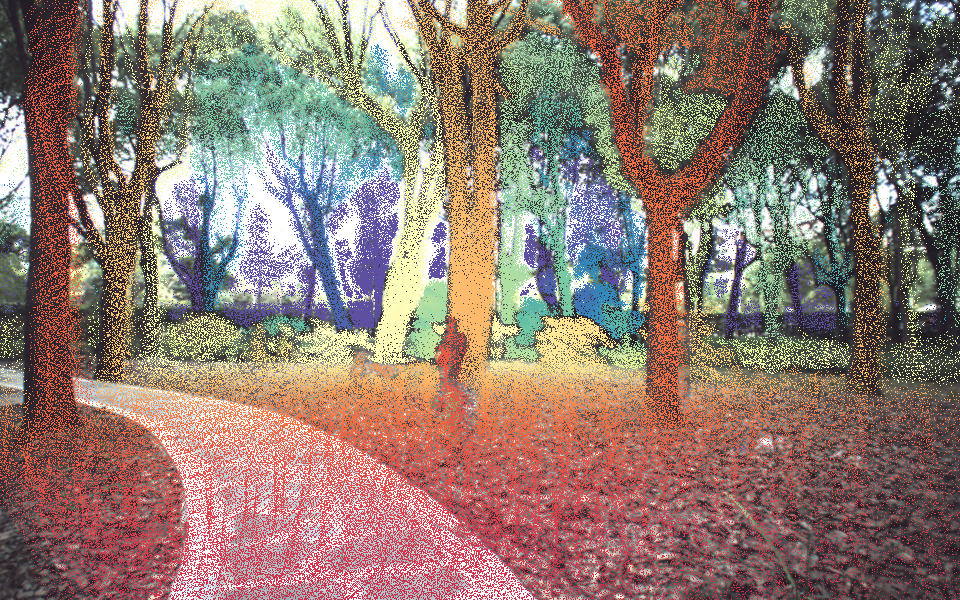}
\end{minipage}}\hspace{0.01\linewidth}%
    \subfloat[{AHPR(5\%)+LDGVSF(0.7)+LD\label{fig:depth-from-ahpr-ldgvsf-ld-botanic-garden}}]{%
\begin{minipage}[t]{0.24\linewidth}
\centering
        \includegraphics[width=\linewidth,height=0.11\textheight]{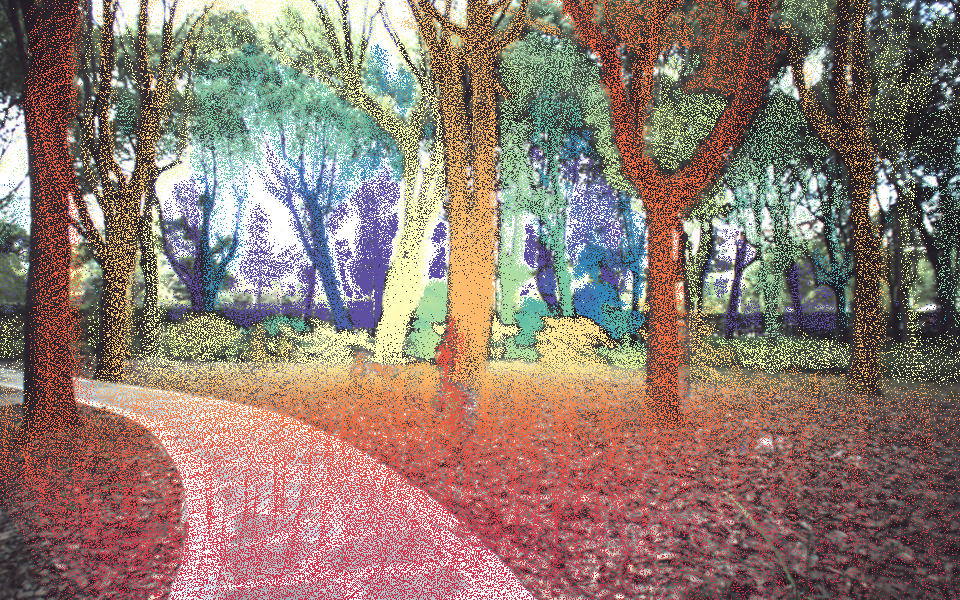}
\end{minipage}}\par
    \caption{Qualitative comparison of visible-depth projections generated
    from Livox LiDAR data for BotanicGarden sequence \texttt{1018-13}. Panel
    (a) shows the right-view RGB image; panels (b)--(k) show the projections
    generated by the selected comparison and proposed methods.}
    \label{fig:cumulated-visible-point-cloud-botanic-garden-full}
\end{figure*}

\section{Qualitative Depth-Error Analysis}
\label{supp:depth-error-analysis}

Figure~\ref{fig:error-images} illustrates the sources of depth error.
Even the aligned left-view single scan in (b) has residual errors at object
boundaries: sparse LiDAR samples and discrete ground-truth pixels can fall on
different surfaces near a depth discontinuity. The right-view cumulation
baseline in (c) also retains incorrect depths within objects because of
camera--LiDAR parallax and accumulated occluded points. PSR in (e) retains
low-error interiors but removes much of the difficult boundary support,
whereas AHPR(5\%)+LDGVSF(0.7) in (k) preserves denser support with residual
errors mainly at edges. Since accuracy is evaluated on each method's retained
pixels, these additional boundary samples help explain the accuracy--coverage
trade-off.

\begin{figure*}[t]
    \centering
    \captionsetup[subfloat]{justification=centering}
    \subfloat[{Right-view RGB image\label{fig:error-right-image}}]{%
\begin{minipage}[t]{0.24\linewidth}
\centering
        \includegraphics[width=\linewidth,height=0.11\textheight]{figs_low_quality/experiments/da/ta/P000_lidar_for_right_000120_rcam_front.jpg}
\end{minipage}}\hfill
    \subfloat[{Single scan (left view)\label{fig:error-single-scan-left}}]{%
\begin{minipage}[t]{0.24\linewidth}
\centering
        \includegraphics[width=\linewidth,height=0.11\textheight]{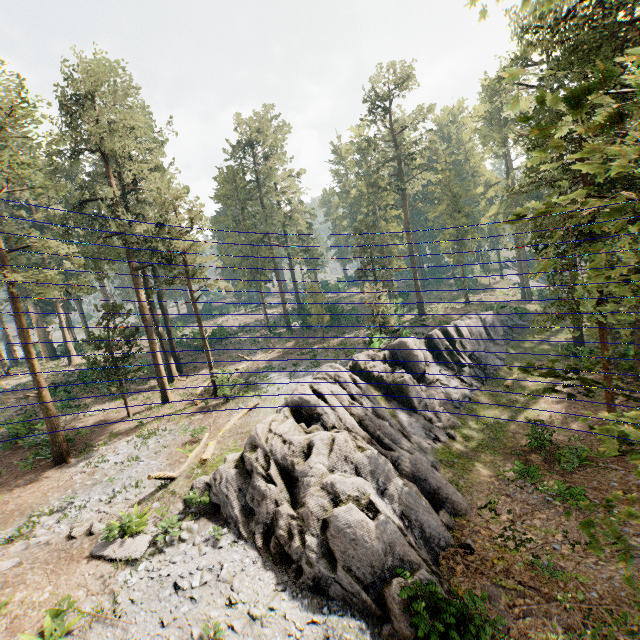}
\end{minipage}}\hfill
    \subfloat[{Cumulation (right view, baseline)\label{fig:error-cumulation}}]{%
\begin{minipage}[t]{0.24\linewidth}
\centering
        \includegraphics[width=\linewidth,height=0.11\textheight]{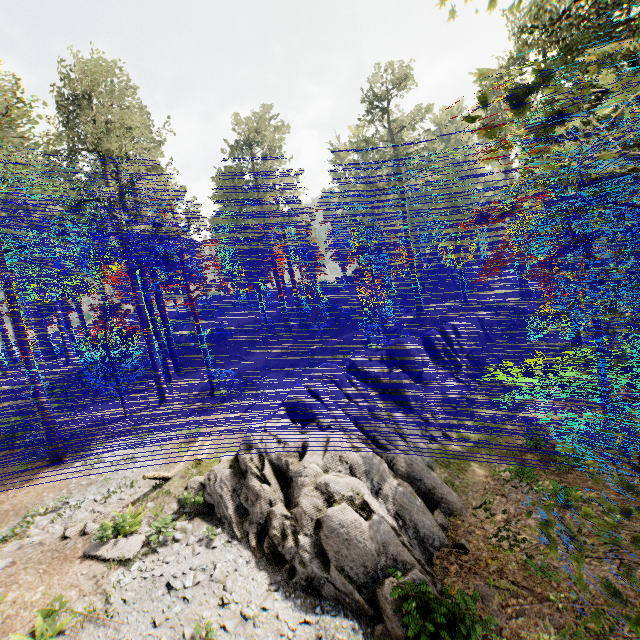}
\end{minipage}}\hfill
    \subfloat[{Cumulation+CS\label{fig:error-cumulation-circle-splatting}}]{%
\begin{minipage}[t]{0.24\linewidth}
\centering
        \includegraphics[width=\linewidth,height=0.11\textheight]{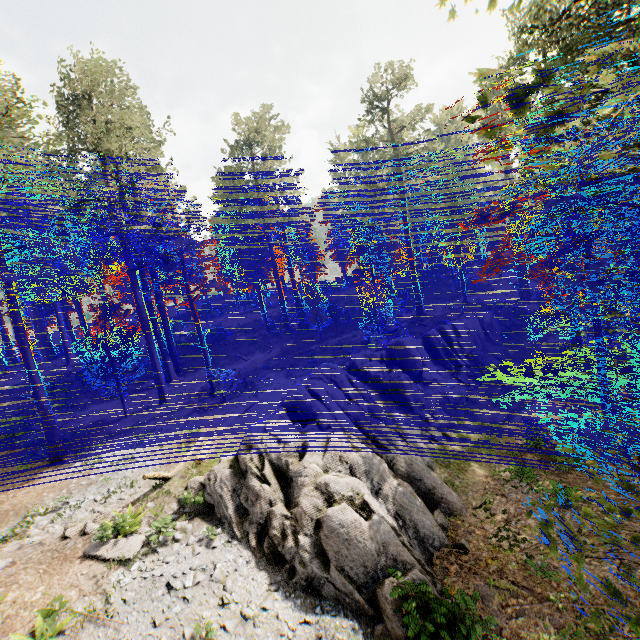}
\end{minipage}}\par\medskip
    \subfloat[{PSR\label{fig:error-psr}}]{%
\begin{minipage}[t]{0.24\linewidth}
\centering
        \includegraphics[width=\linewidth,height=0.11\textheight]{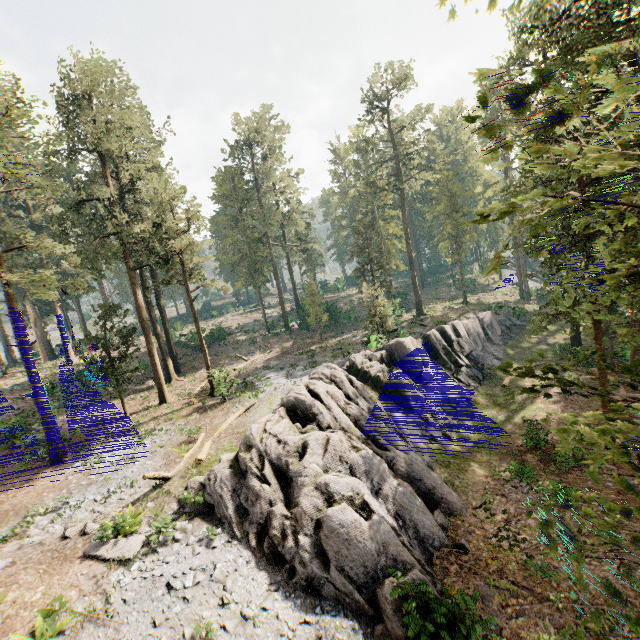}
\end{minipage}}\hfill
    \subfloat[{SGM\label{fig:error-sgm}}]{%
\begin{minipage}[t]{0.24\linewidth}
\centering
        \includegraphics[width=\linewidth,height=0.11\textheight]{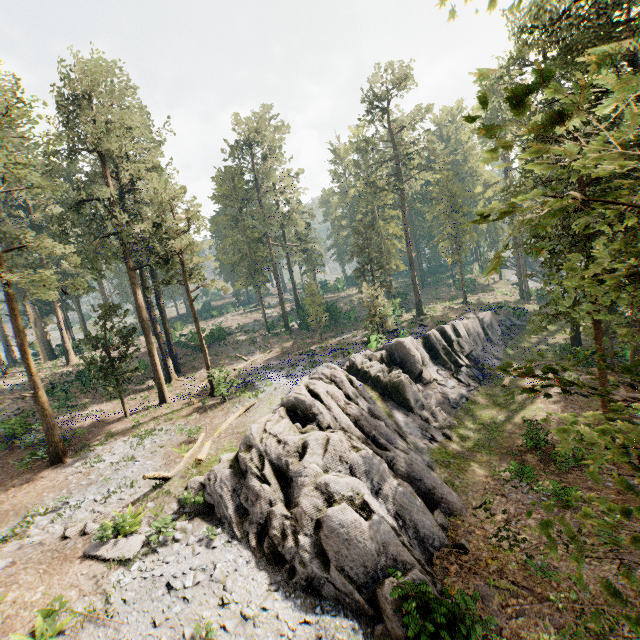}
\end{minipage}}\hfill
    \subfloat[{LD\label{fig:error-ld}}]{%
\begin{minipage}[t]{0.24\linewidth}
\centering
        \includegraphics[width=\linewidth,height=0.11\textheight]{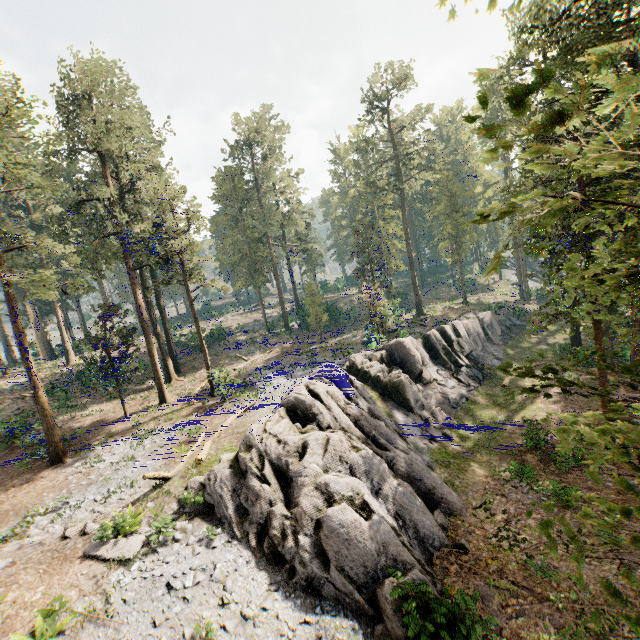}
\end{minipage}}\hfill
    \subfloat[{Visibility estimation (0.7)\label{fig:error-visibility-estimation}}]{%
\begin{minipage}[t]{0.24\linewidth}
\centering
        \includegraphics[width=\linewidth,height=0.11\textheight]{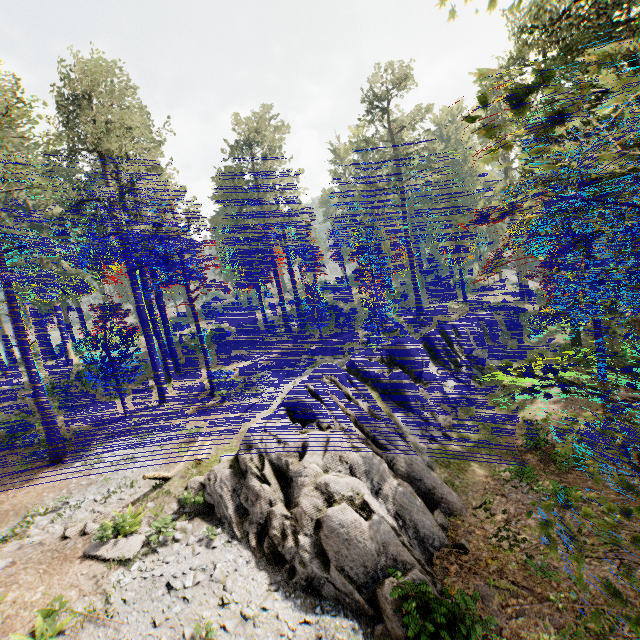}
\end{minipage}}\par\medskip
    \subfloat[{LDGVSF(0.7)\label{fig:error-ldgvsf}}]{%
\begin{minipage}[t]{0.24\linewidth}
\centering
        \includegraphics[width=\linewidth,height=0.11\textheight]{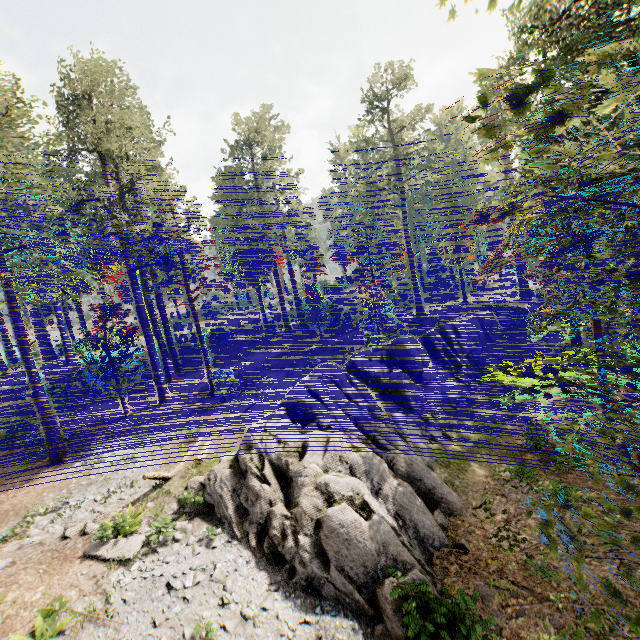}
\end{minipage}}\hfill
    \subfloat[{AHPR(10\%)\label{fig:error-ahpr}}]{%
\begin{minipage}[t]{0.24\linewidth}
\centering
        \includegraphics[width=\linewidth,height=0.11\textheight]{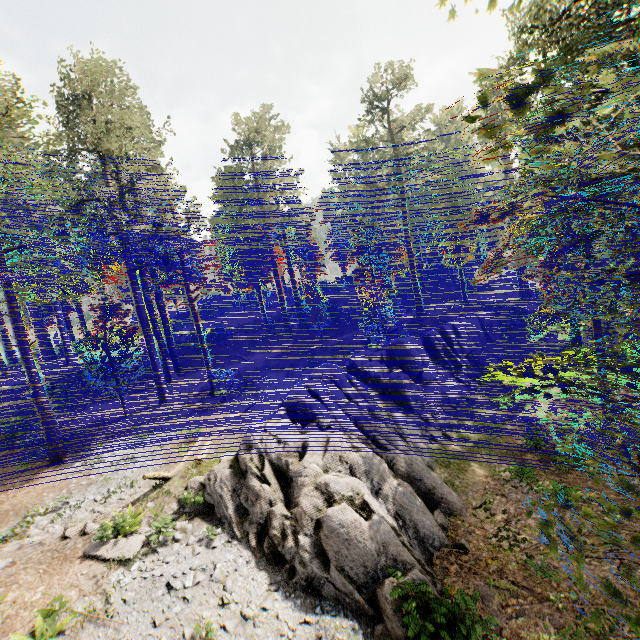}
\end{minipage}}\hfill
    \subfloat[{AHPR(5\%)+LDGVSF(0.7)\label{fig:error-ahpr-ldgvsf}}]{%
\begin{minipage}[t]{0.24\linewidth}
\centering
        \includegraphics[width=\linewidth,height=0.11\textheight]{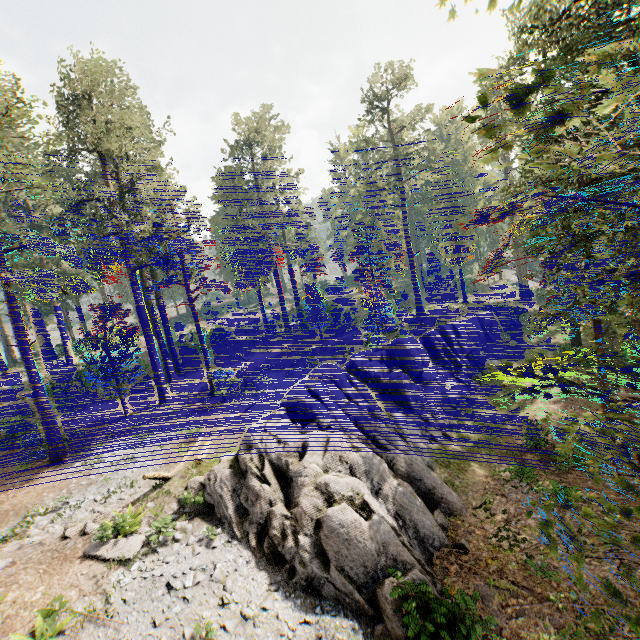}
\end{minipage}}\hfill
    \subfloat[{AHPR(5\%)+LDGVSF(0.7)+LD\label{fig:error-ahpr-ldgvsf-ld}}]{%
\begin{minipage}[t]{0.24\linewidth}
\centering
        \includegraphics[width=\linewidth,height=0.11\textheight]{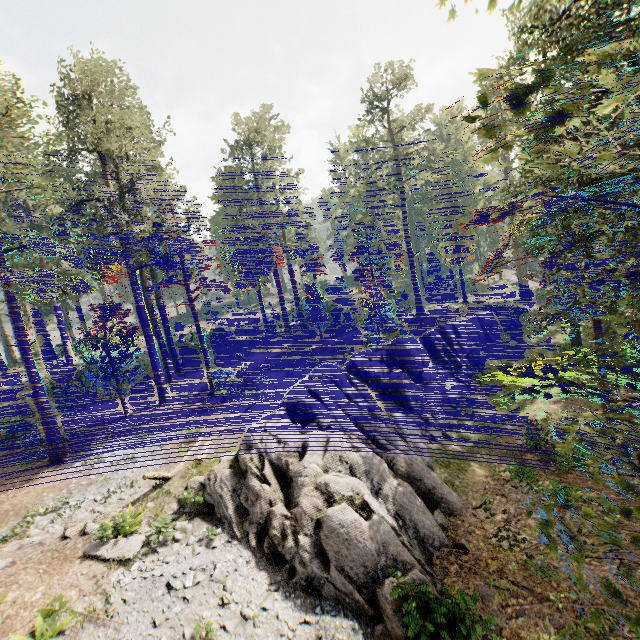}
\end{minipage}}

    \caption{Reference image and per-pixel depth-error maps for the TartanAirV2
    forest scene. Panel (a) shows the right-view RGB image, and panel (b) shows
    the aligned left-view single-scan error. Panels (c)--(l) show the right-view errors
    for the ten projections in
    Fig.~\ref{fig:cumulated-visible-point-cloud-tartanairv2}(c)--(l),
    in the same order. Errors are displayed with a jet color map (blue: low
    error; red: high error).}
    \label{fig:error-images}
\end{figure*}

\section{Reference Point-Cloud and Depth Visualizations}
\label{supp:reference-point-cloud-and-depth-visualizations}

To show more examples for the generated reference, fig.~\ref{fig:supp-reference-point-cloud-and-depth-examples} presents additional
reference outputs for all ten BotanicGarden benchmark subsequences.

\begin{figure*}[t]
\centering
\begingroup
\scriptsize
\setlength{\tabcolsep}{2pt}
\renewcommand{\arraystretch}{1.04}

\begin{tabular}{c c c c c}
\textbf{Scene}
& \textbf{3D View~1}
& \textbf{3D View~2}
& \textbf{Depth View~1}
& \textbf{Depth View~2} \\
\shortstack{\textbf{1006-01}}& \includegraphics[width=0.18\textwidth,height=0.09\textheight]{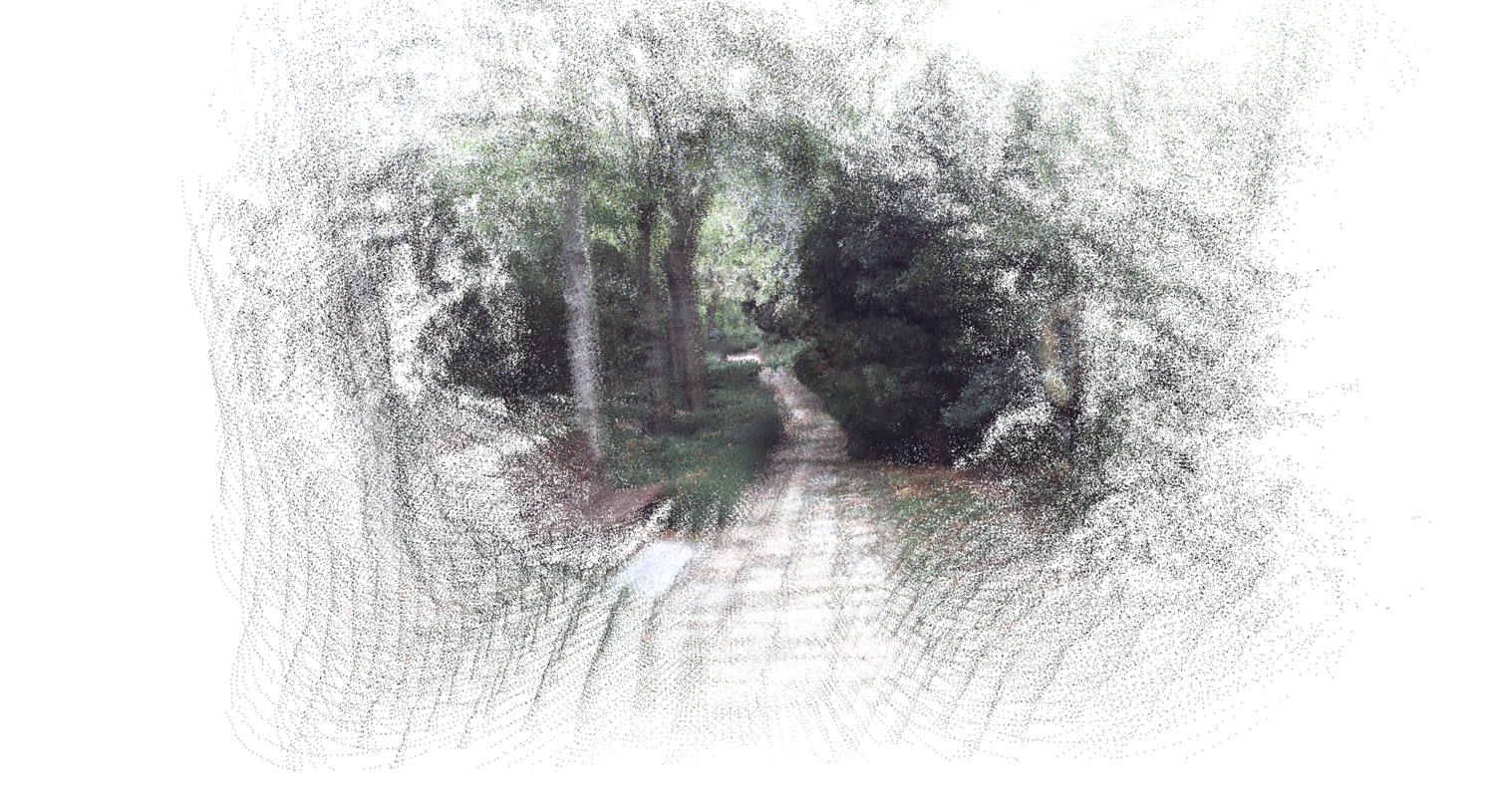}& \includegraphics[width=0.18\textwidth,height=0.09\textheight]{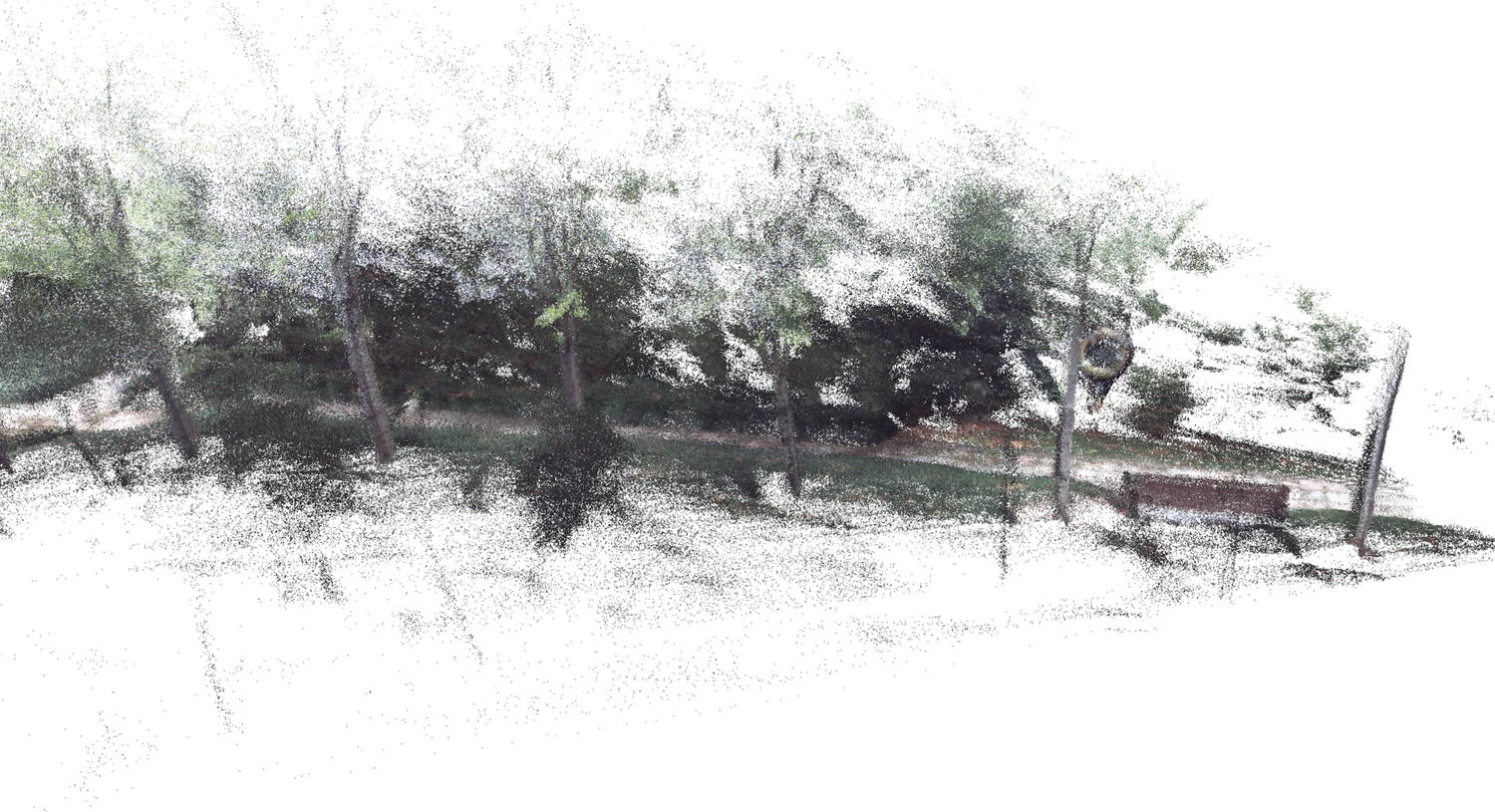}& \includegraphics[width=0.18\textwidth,height=0.09\textheight]{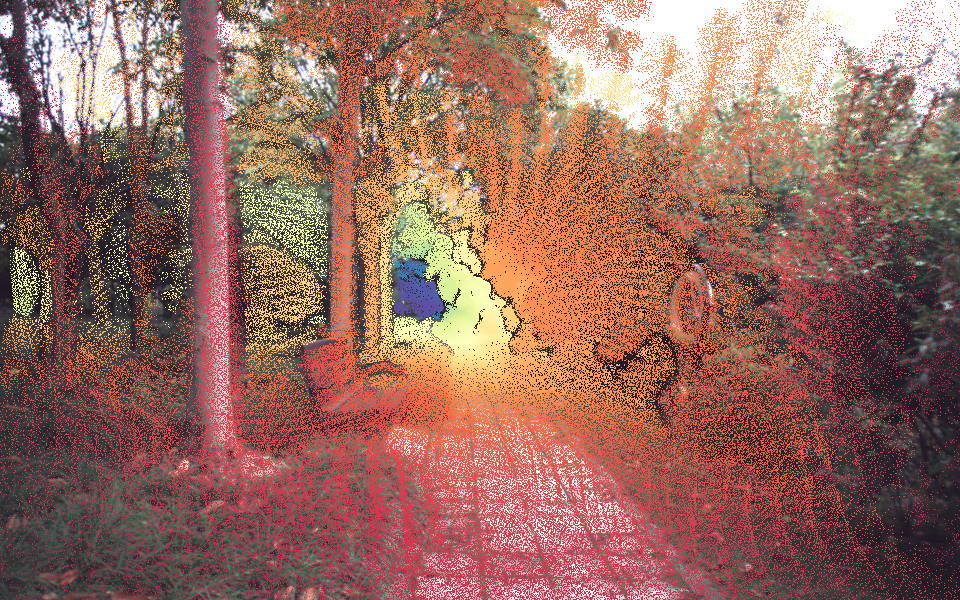}& \includegraphics[width=0.18\textwidth,height=0.09\textheight]{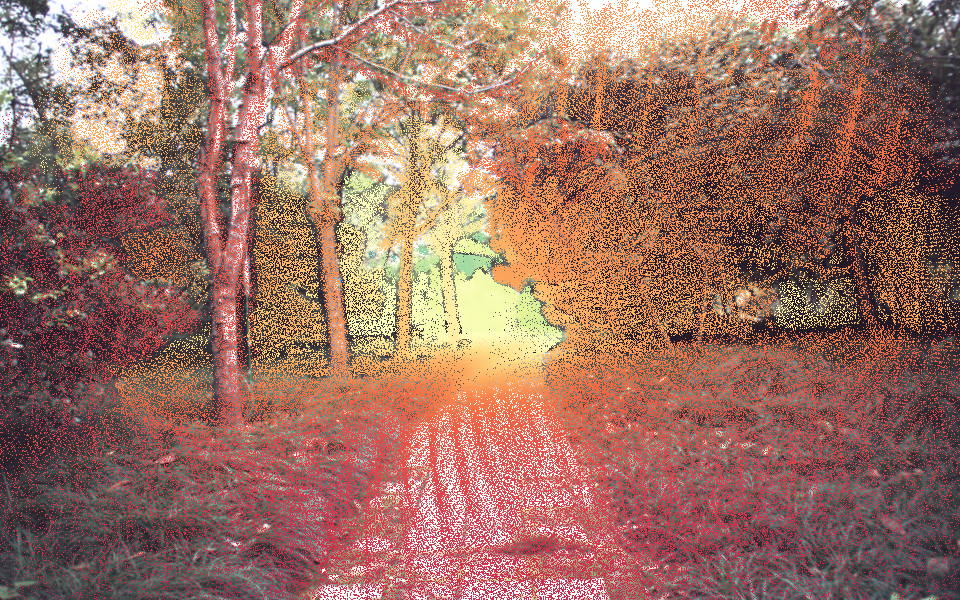} \\ 
\shortstack{\textbf{1006-03}}& \includegraphics[width=0.18\textwidth,height=0.09\textheight]{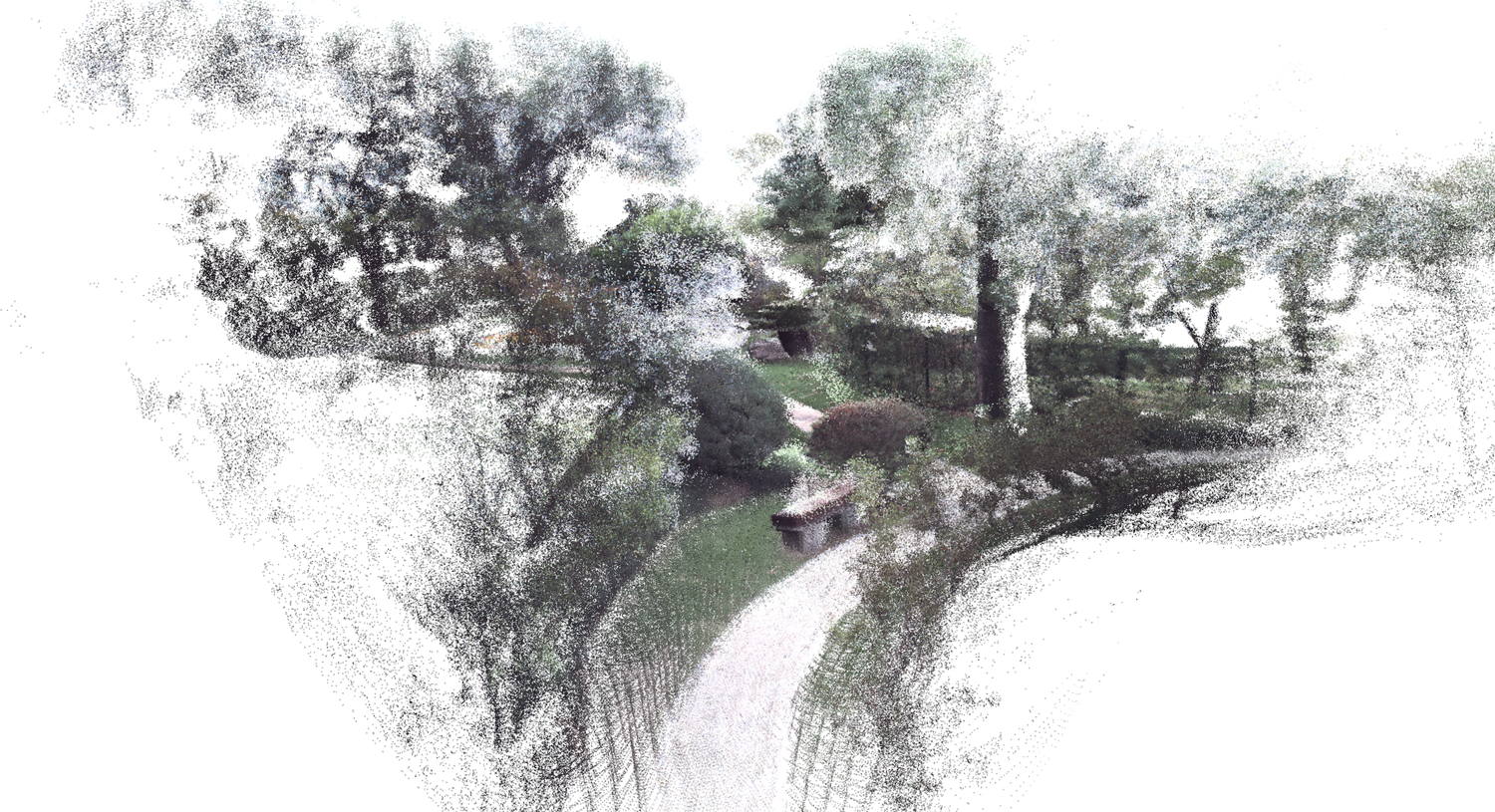}& \includegraphics[width=0.18\textwidth,height=0.09\textheight]{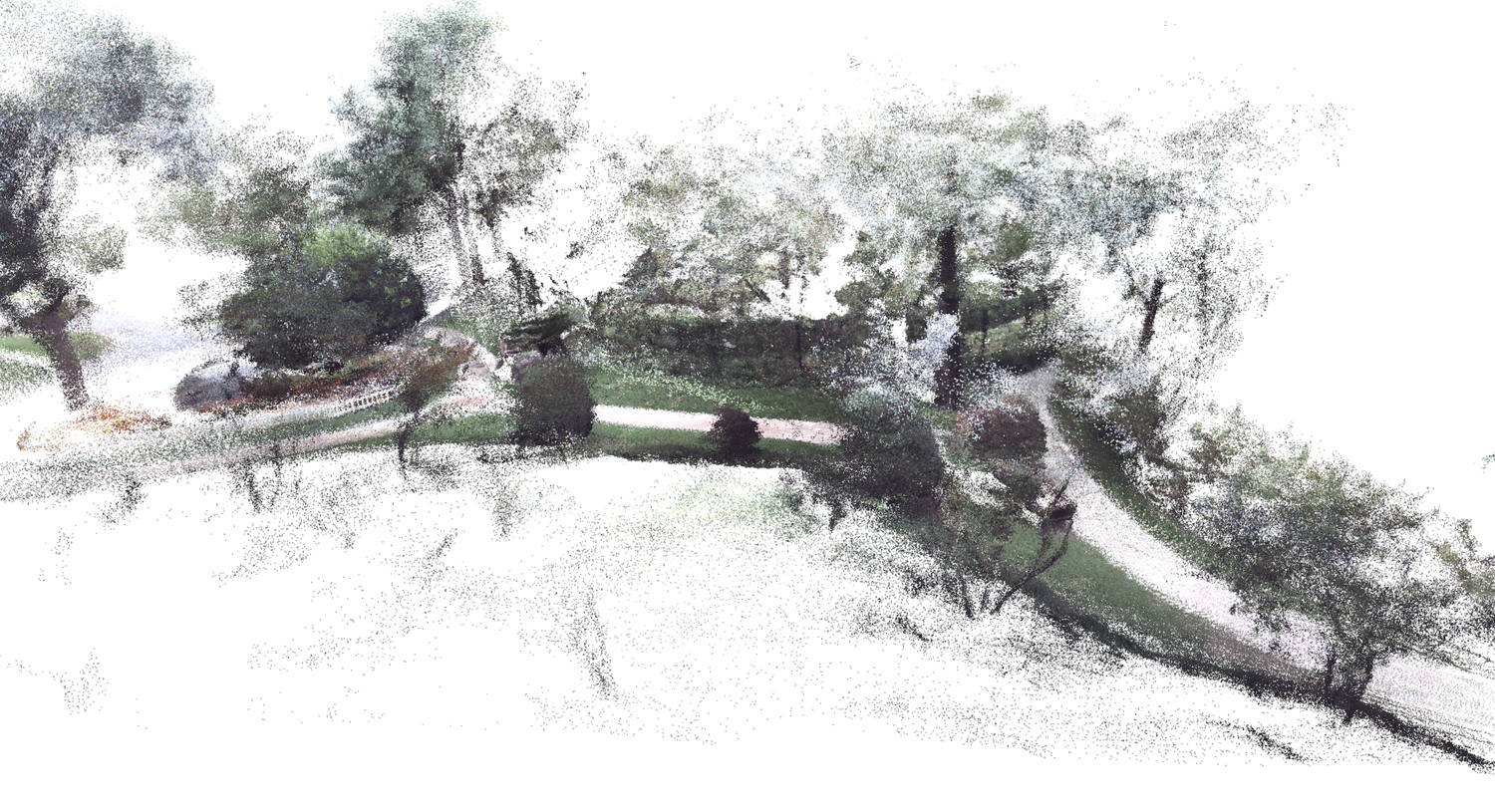}& \includegraphics[width=0.18\textwidth,height=0.09\textheight]{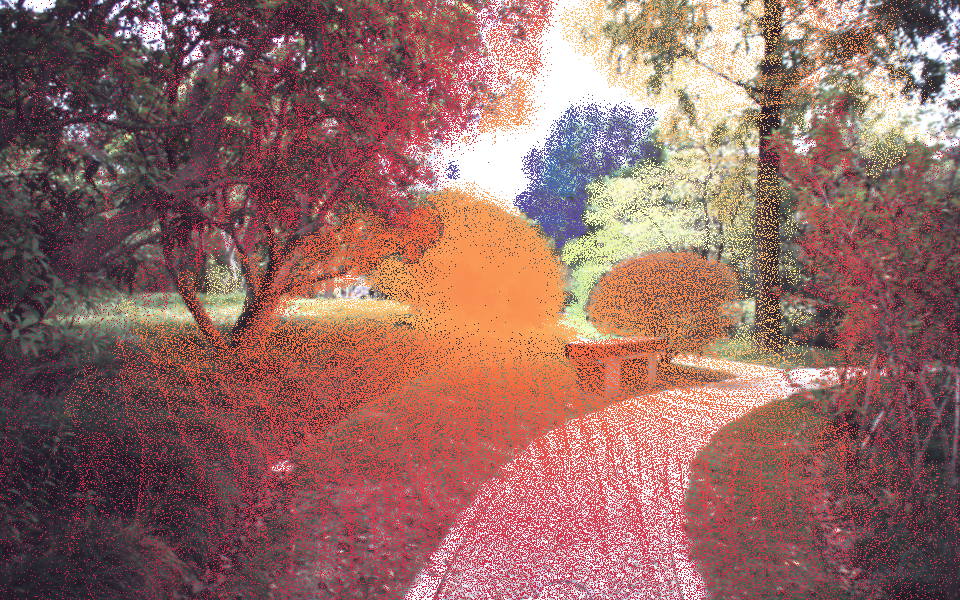}& \includegraphics[width=0.18\textwidth,height=0.09\textheight]{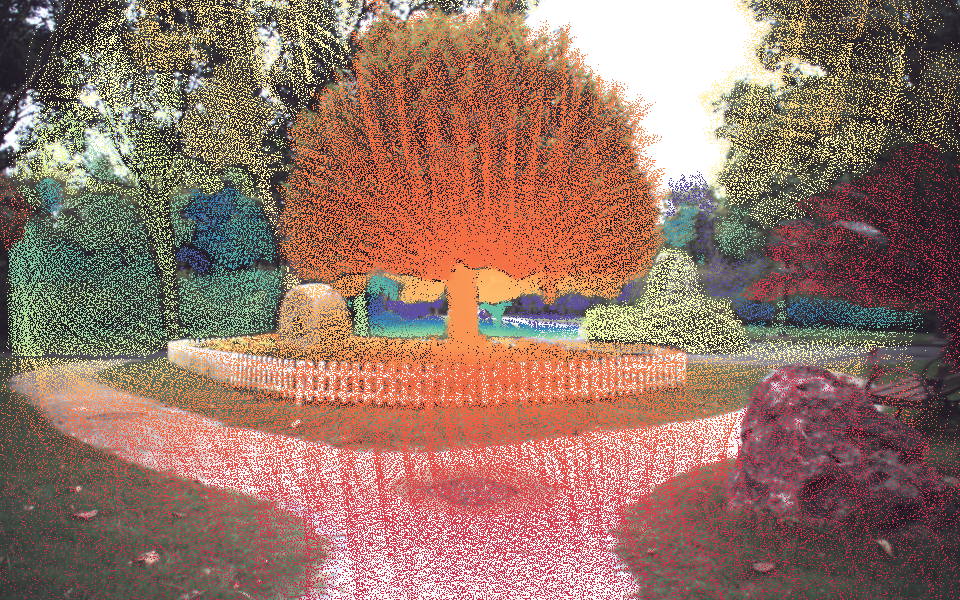} \\ 
\shortstack{\textbf{1008-01}}& \includegraphics[width=0.18\textwidth,height=0.09\textheight]{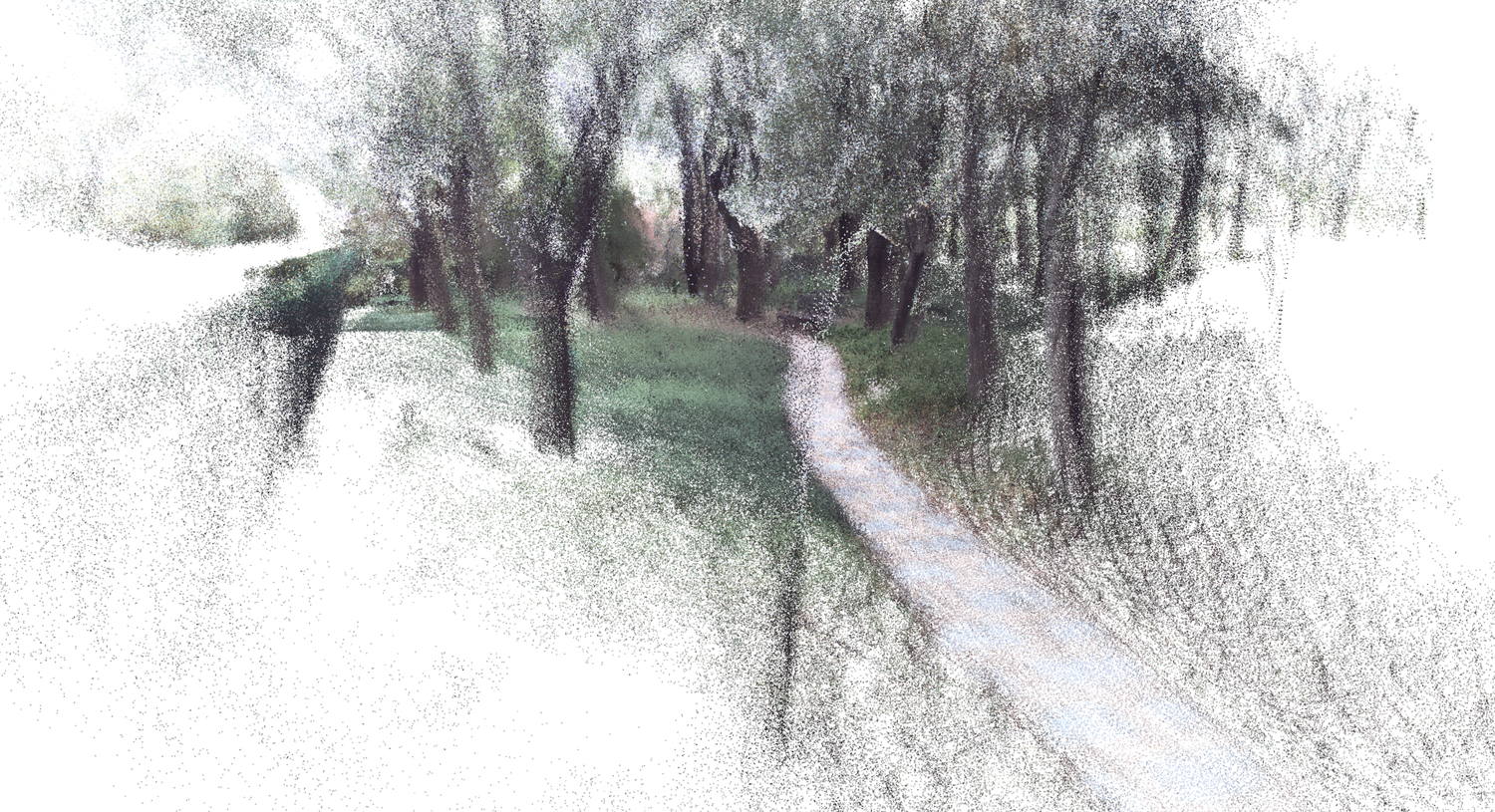}& \includegraphics[width=0.18\textwidth,height=0.09\textheight]{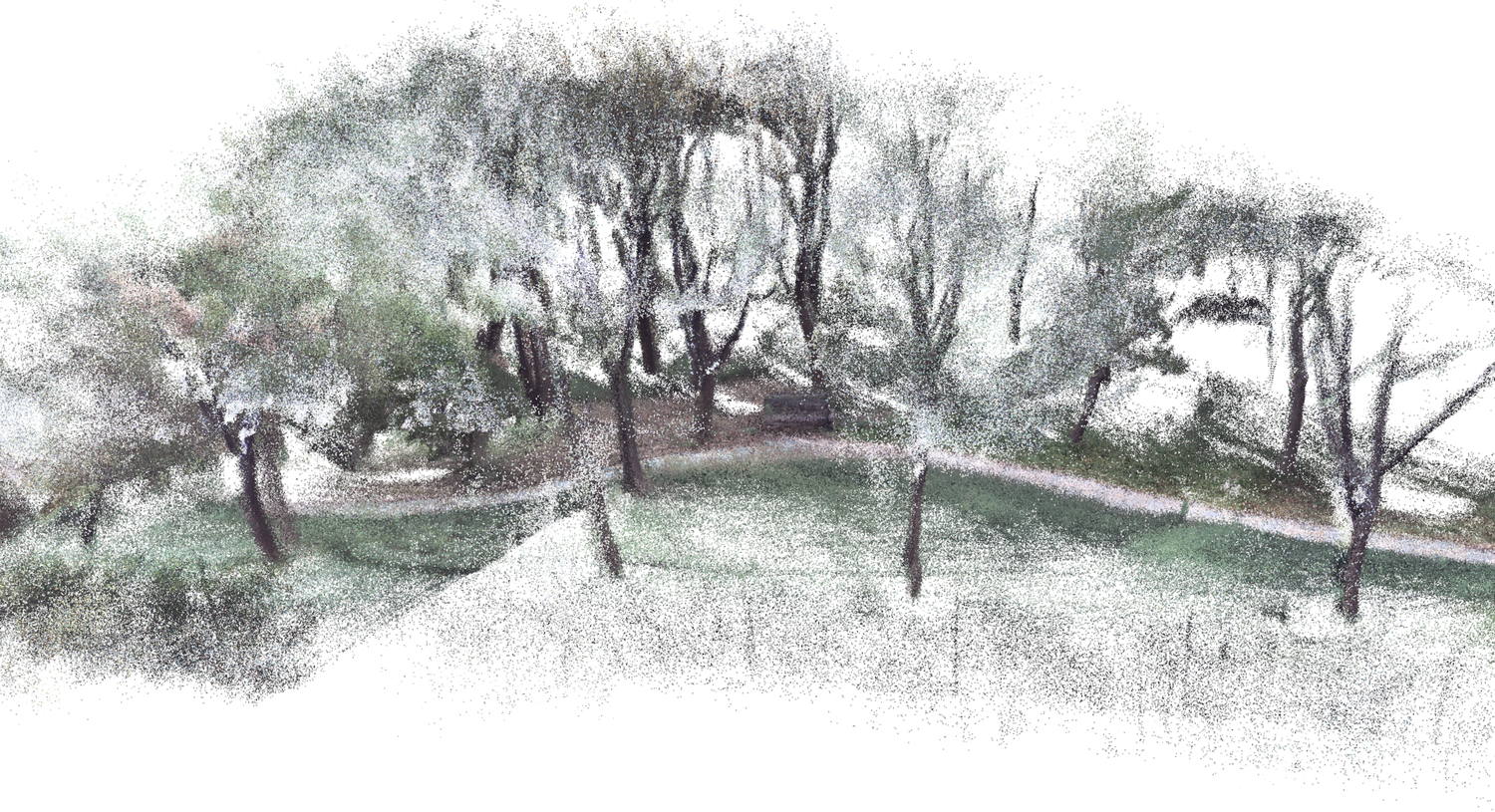}& \includegraphics[width=0.18\textwidth,height=0.09\textheight]{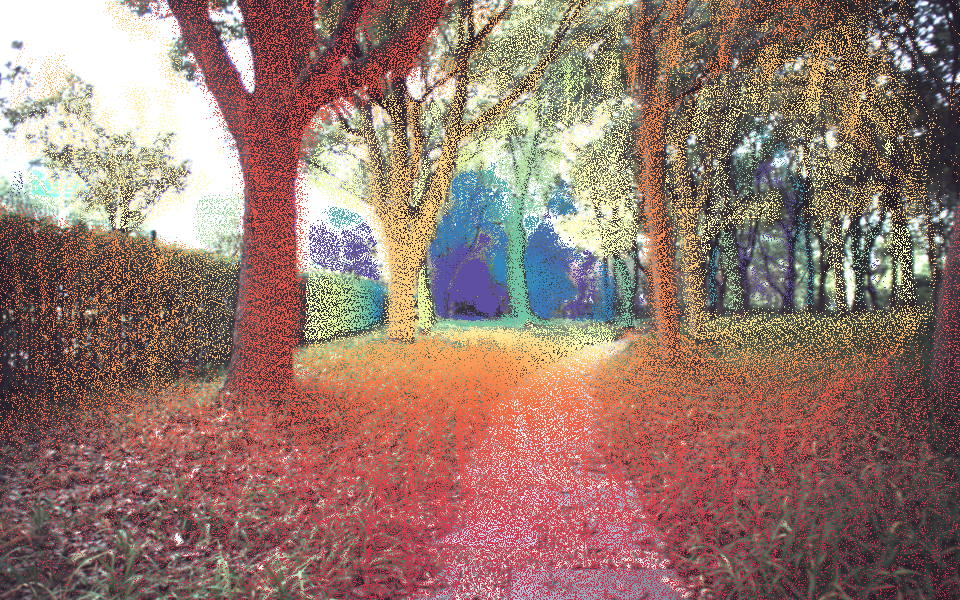}& \includegraphics[width=0.18\textwidth,height=0.09\textheight]{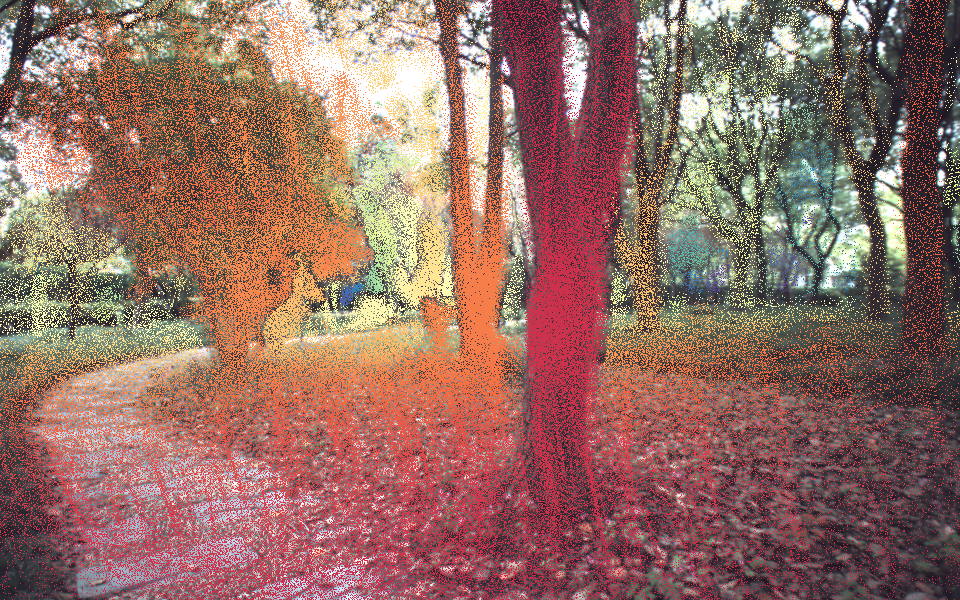} \\ 
\shortstack{\textbf{1018-00}}& \includegraphics[width=0.18\textwidth,height=0.09\textheight]{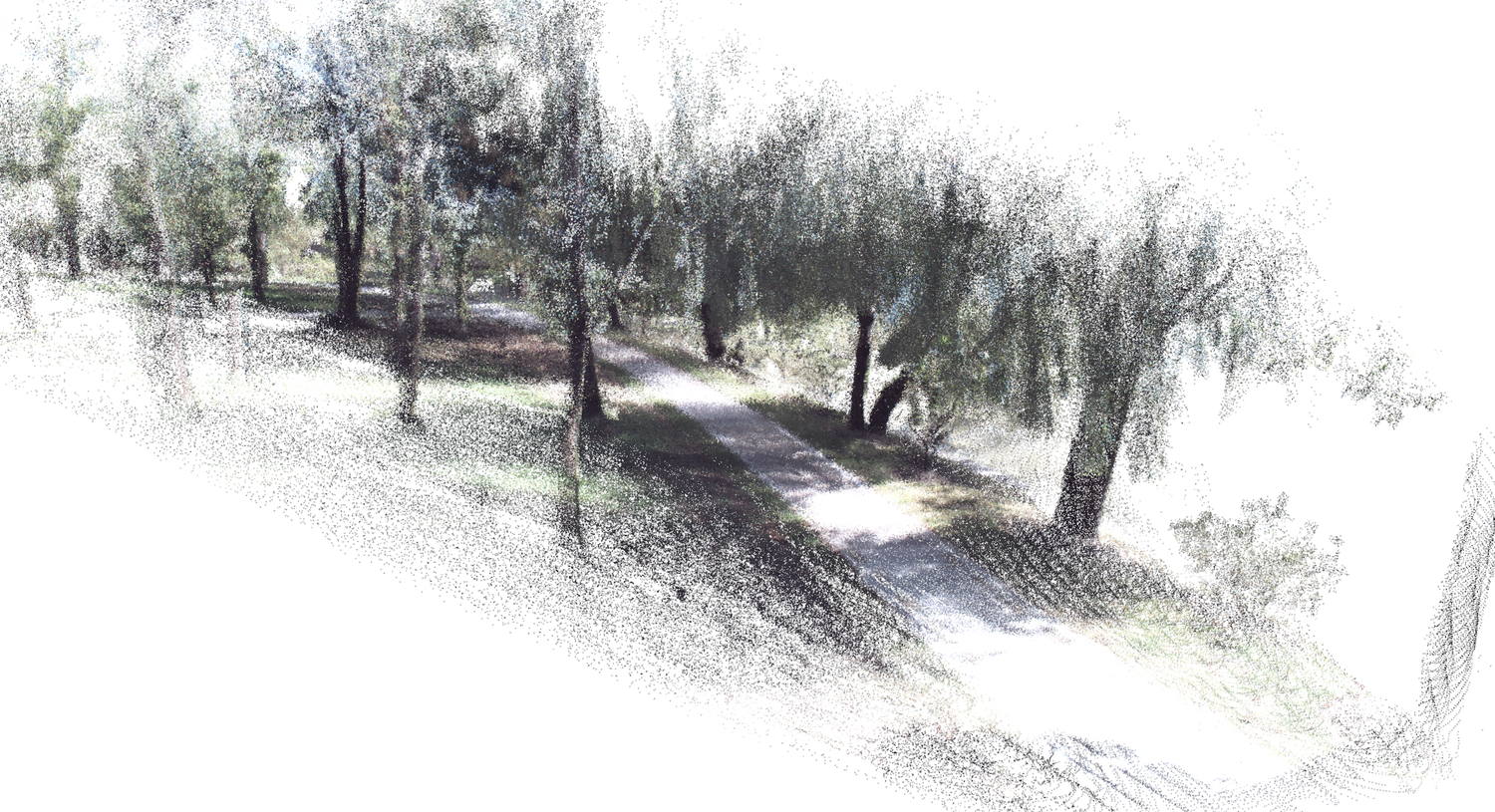}& \includegraphics[width=0.18\textwidth,height=0.09\textheight]{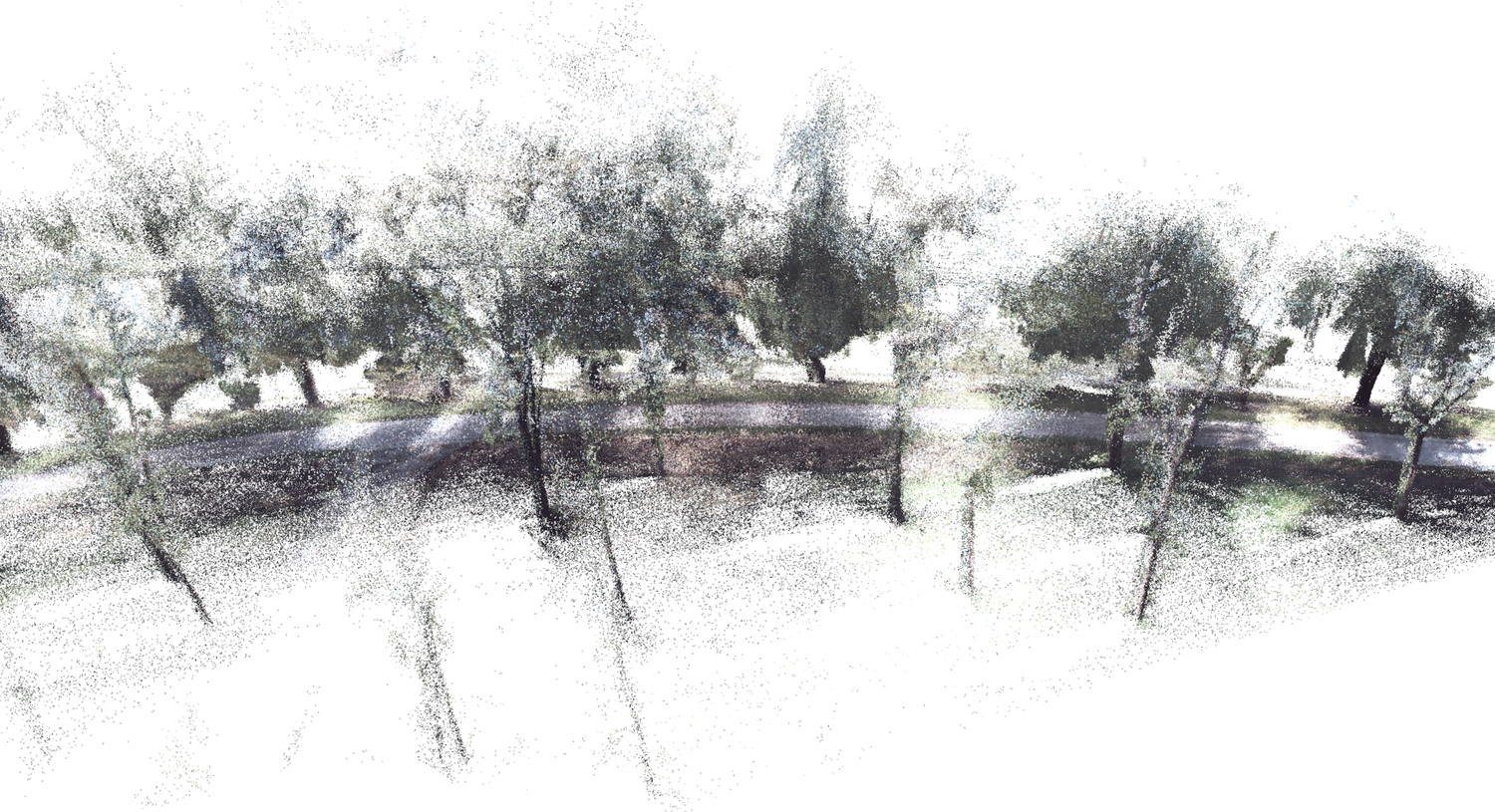}& \includegraphics[width=0.18\textwidth,height=0.09\textheight]{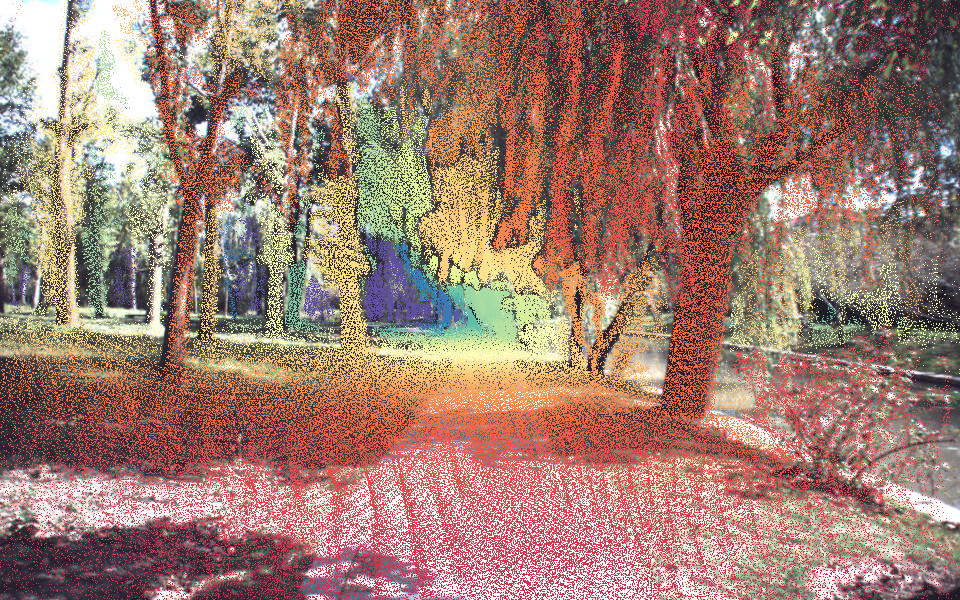}& \includegraphics[width=0.18\textwidth,height=0.09\textheight]{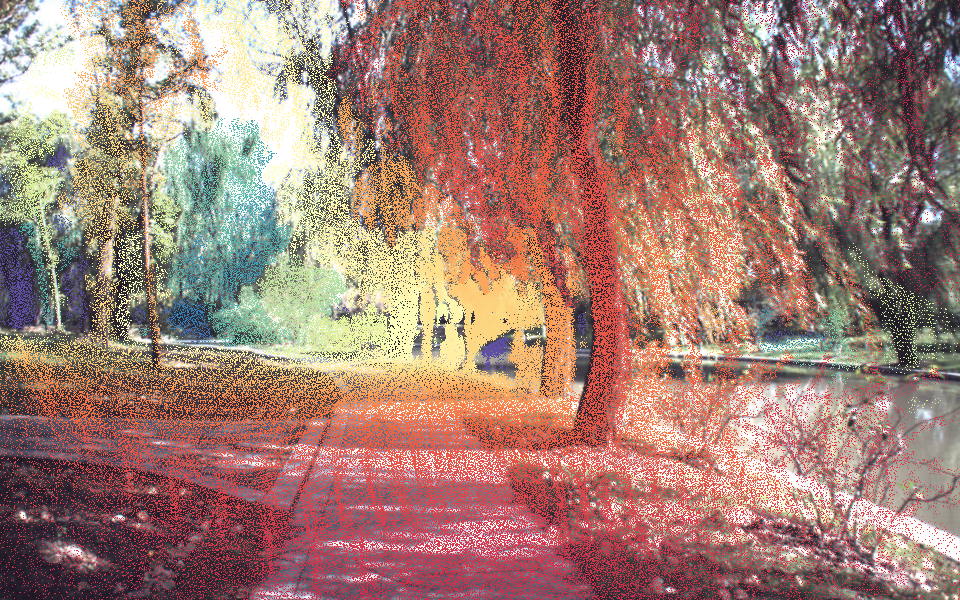} \\ 
\shortstack{\textbf{1005-00}}& \includegraphics[width=0.18\textwidth,height=0.09\textheight]{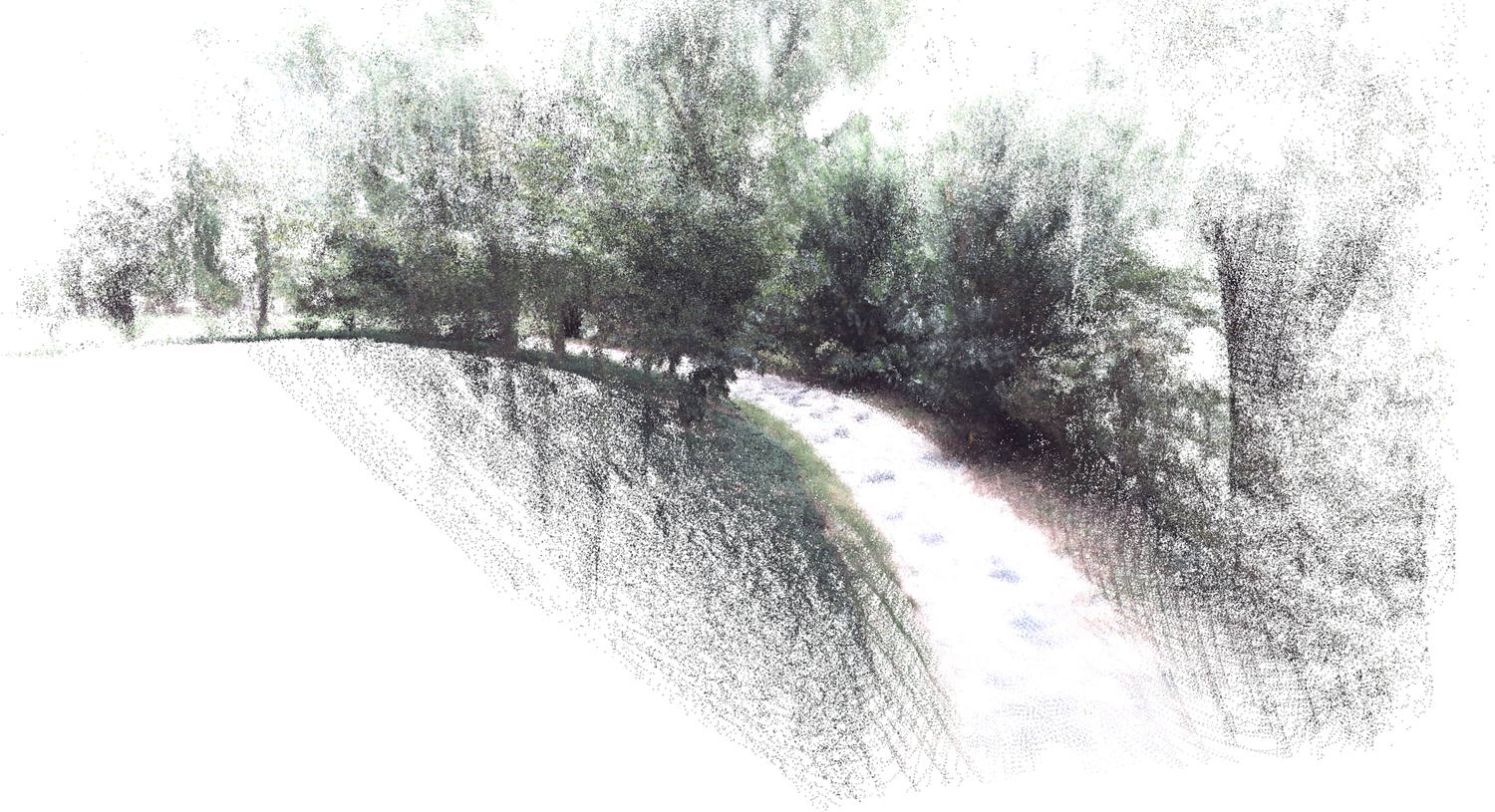}& \includegraphics[width=0.18\textwidth,height=0.09\textheight]{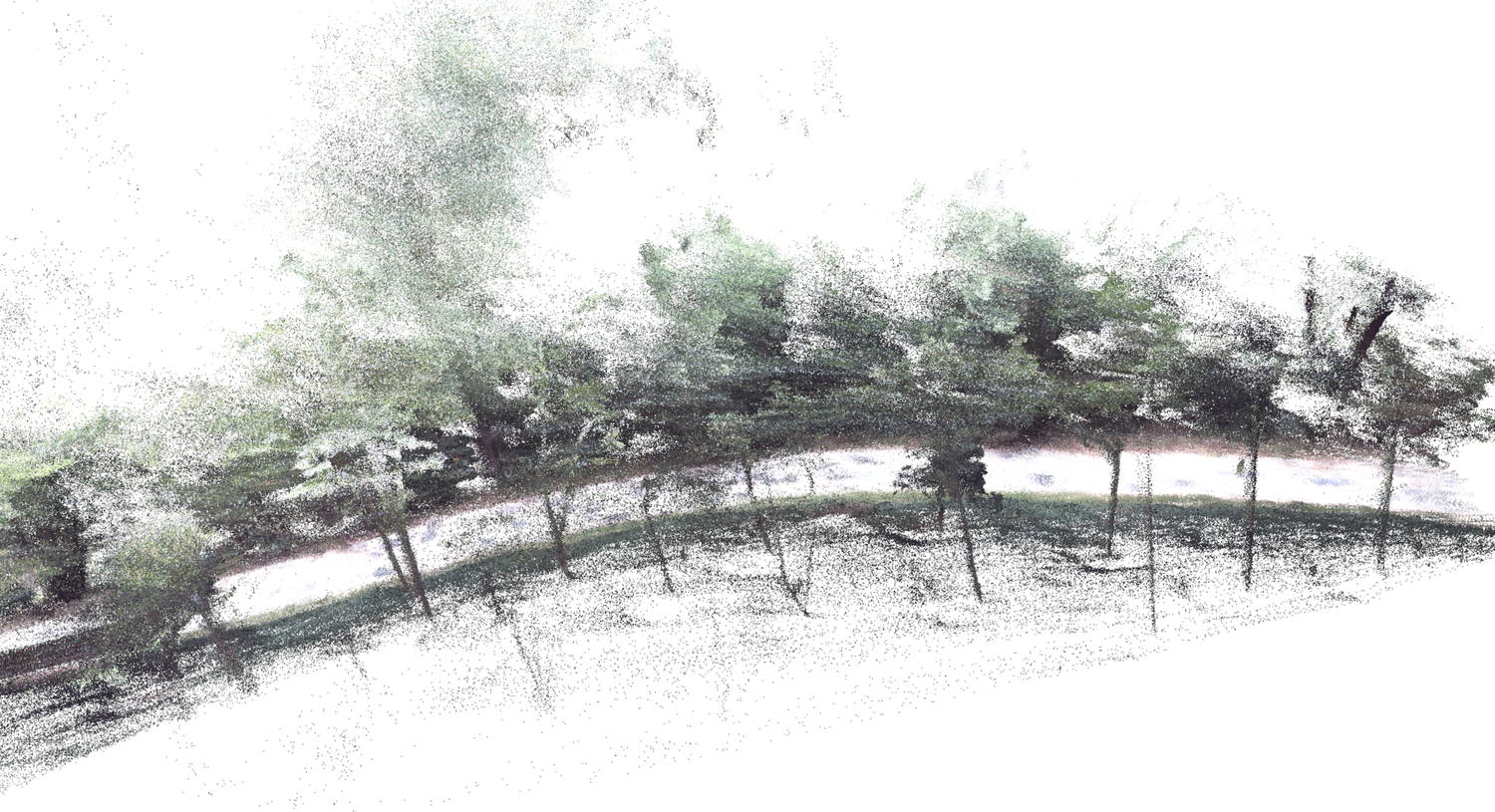}& \includegraphics[width=0.18\textwidth,height=0.09\textheight]{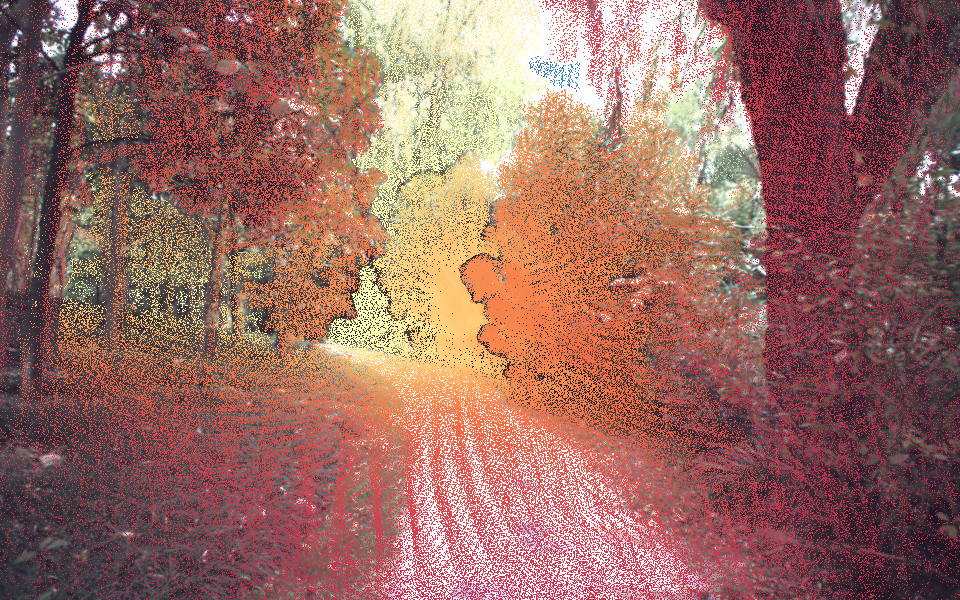}& \includegraphics[width=0.18\textwidth,height=0.09\textheight]{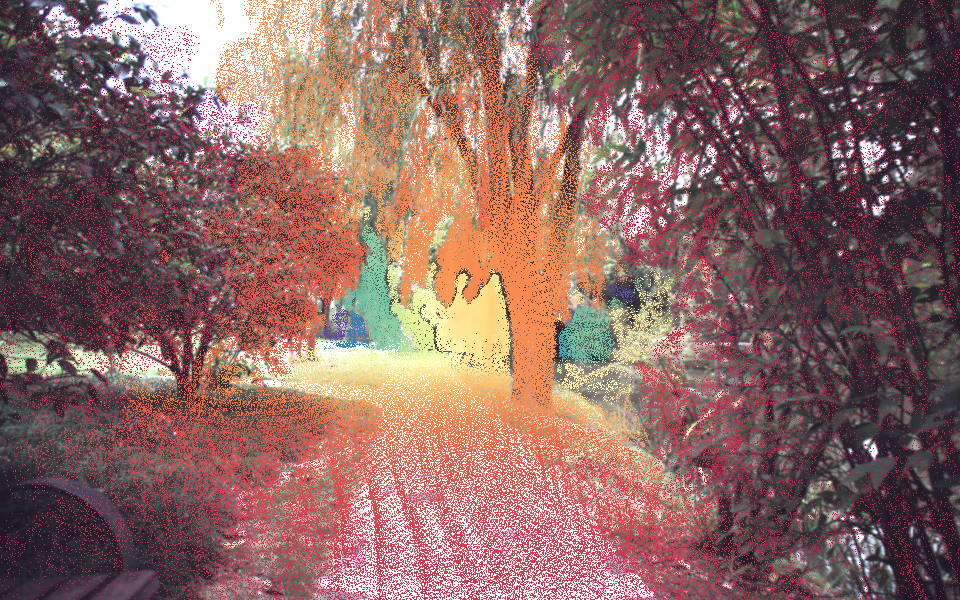} \\ 
\shortstack{\textbf{1005-01}}& \includegraphics[width=0.18\textwidth,height=0.09\textheight]{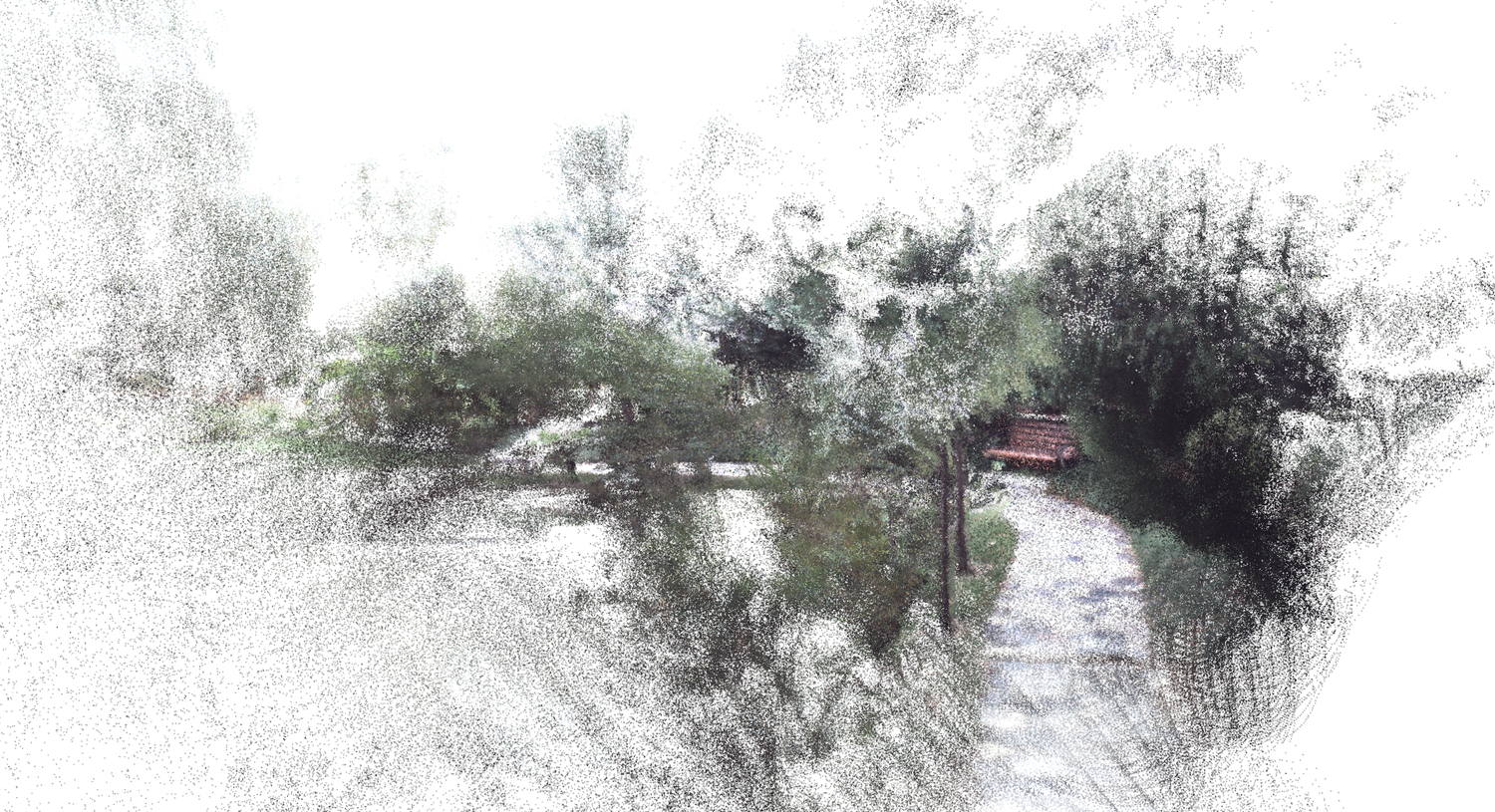}& \includegraphics[width=0.18\textwidth,height=0.09\textheight]{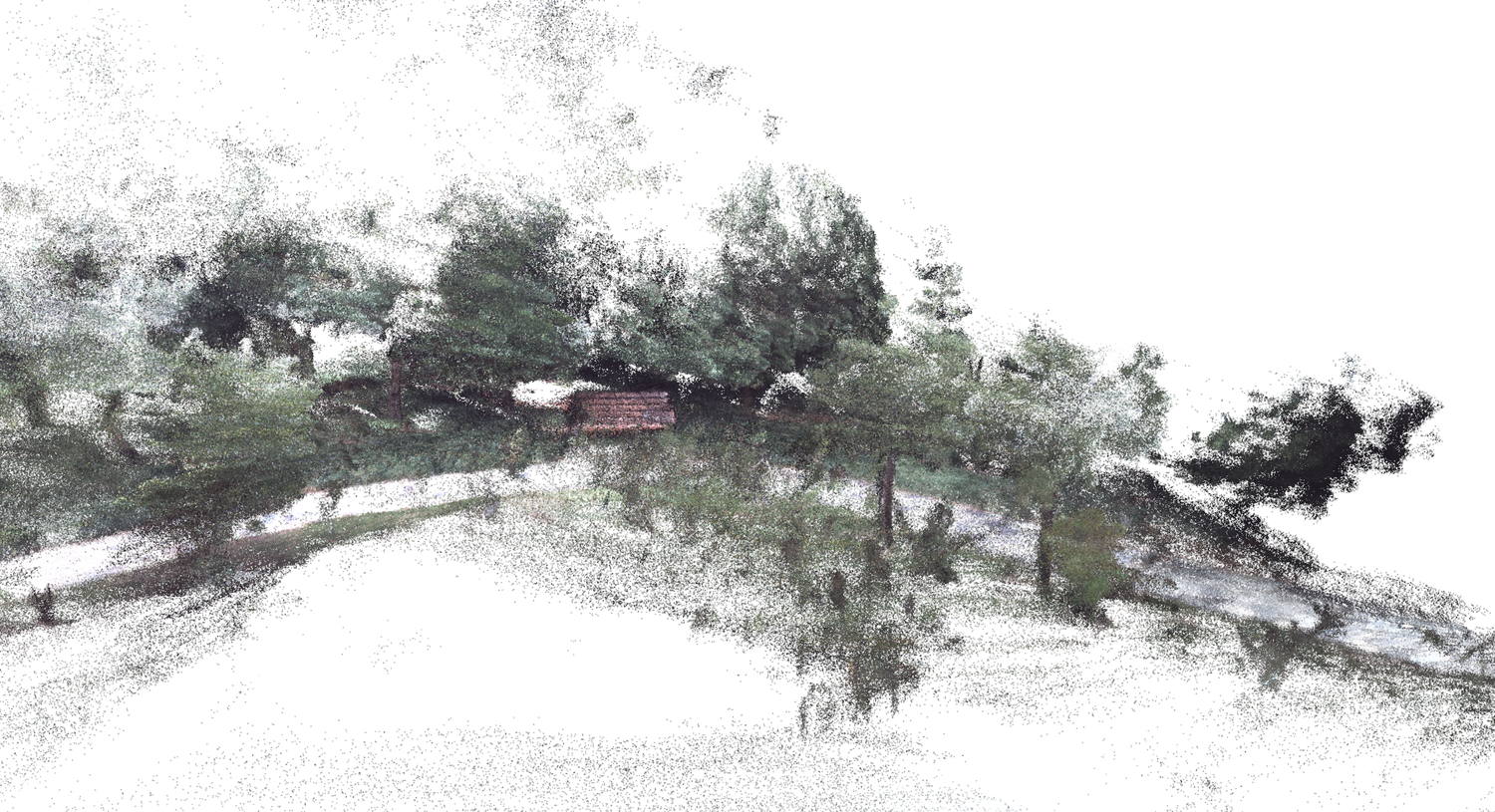}& \includegraphics[width=0.18\textwidth,height=0.09\textheight]{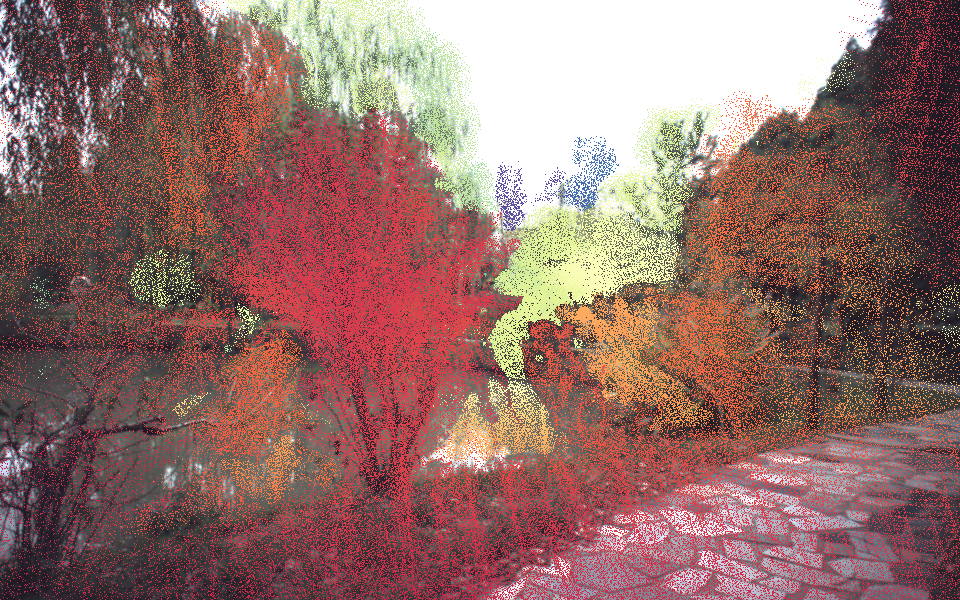}& \includegraphics[width=0.18\textwidth,height=0.09\textheight]{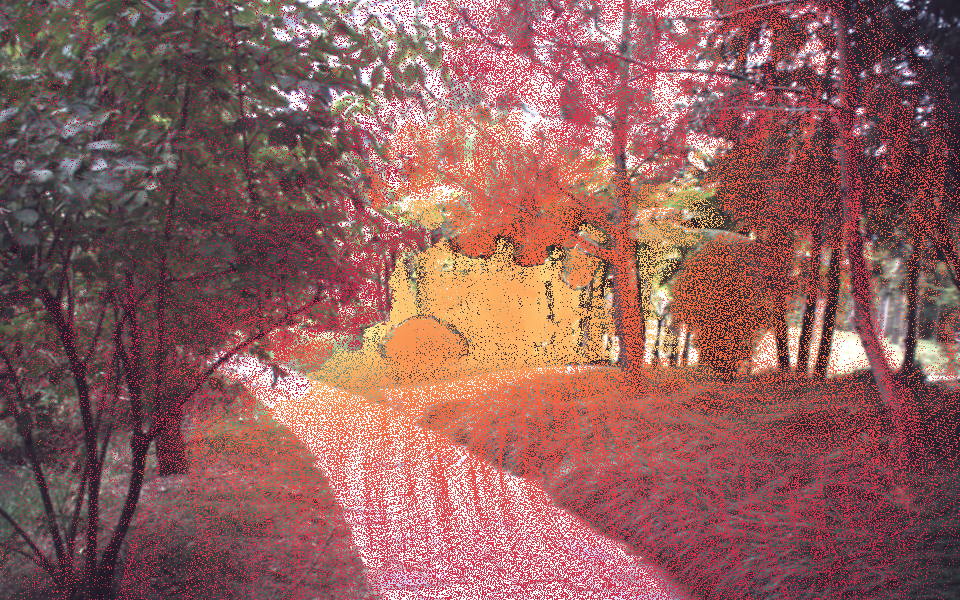} \\ 
\shortstack{\textbf{1005-05}}& \includegraphics[width=0.18\textwidth,height=0.09\textheight]{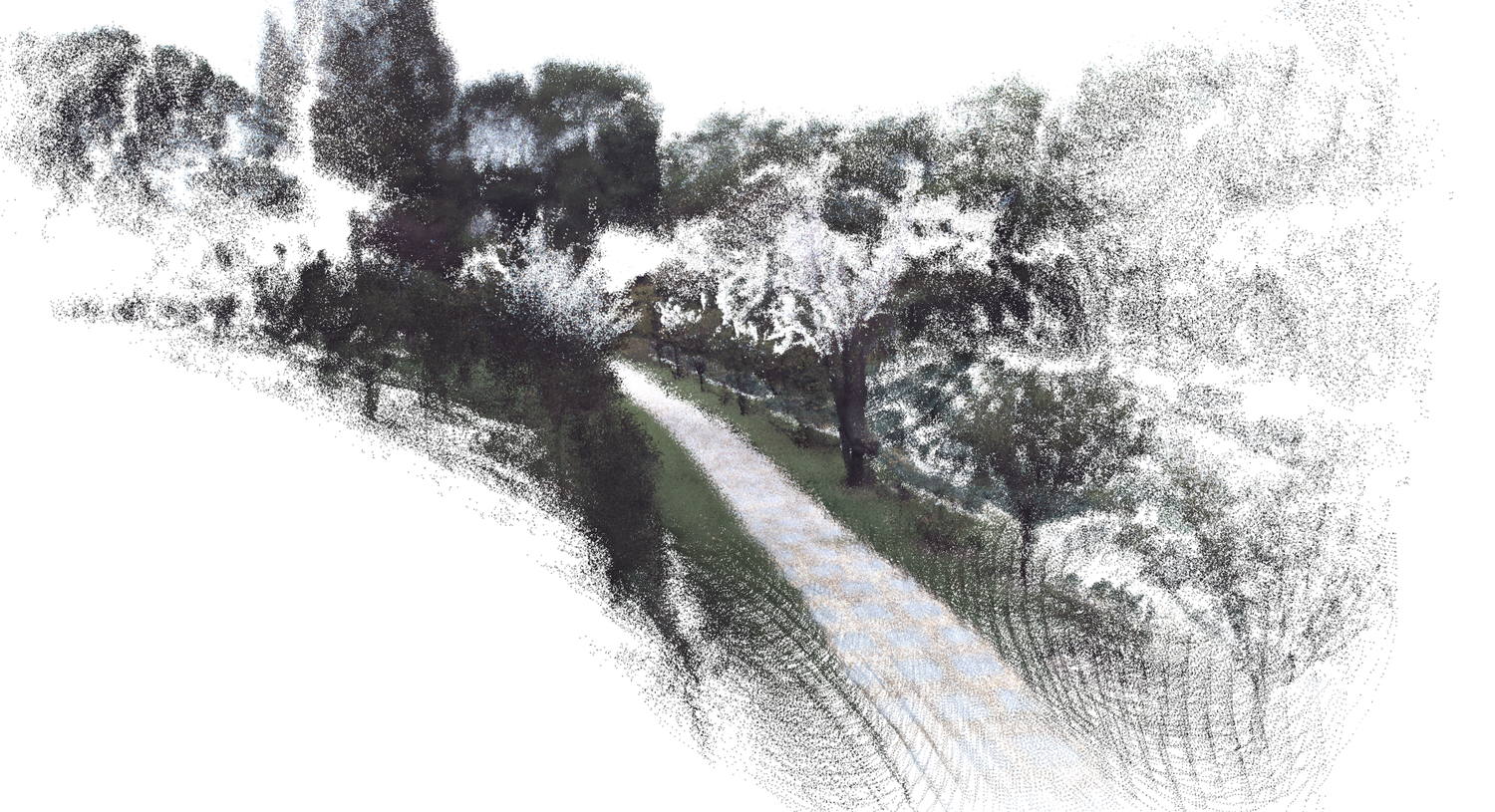}& \includegraphics[width=0.18\textwidth,height=0.09\textheight]{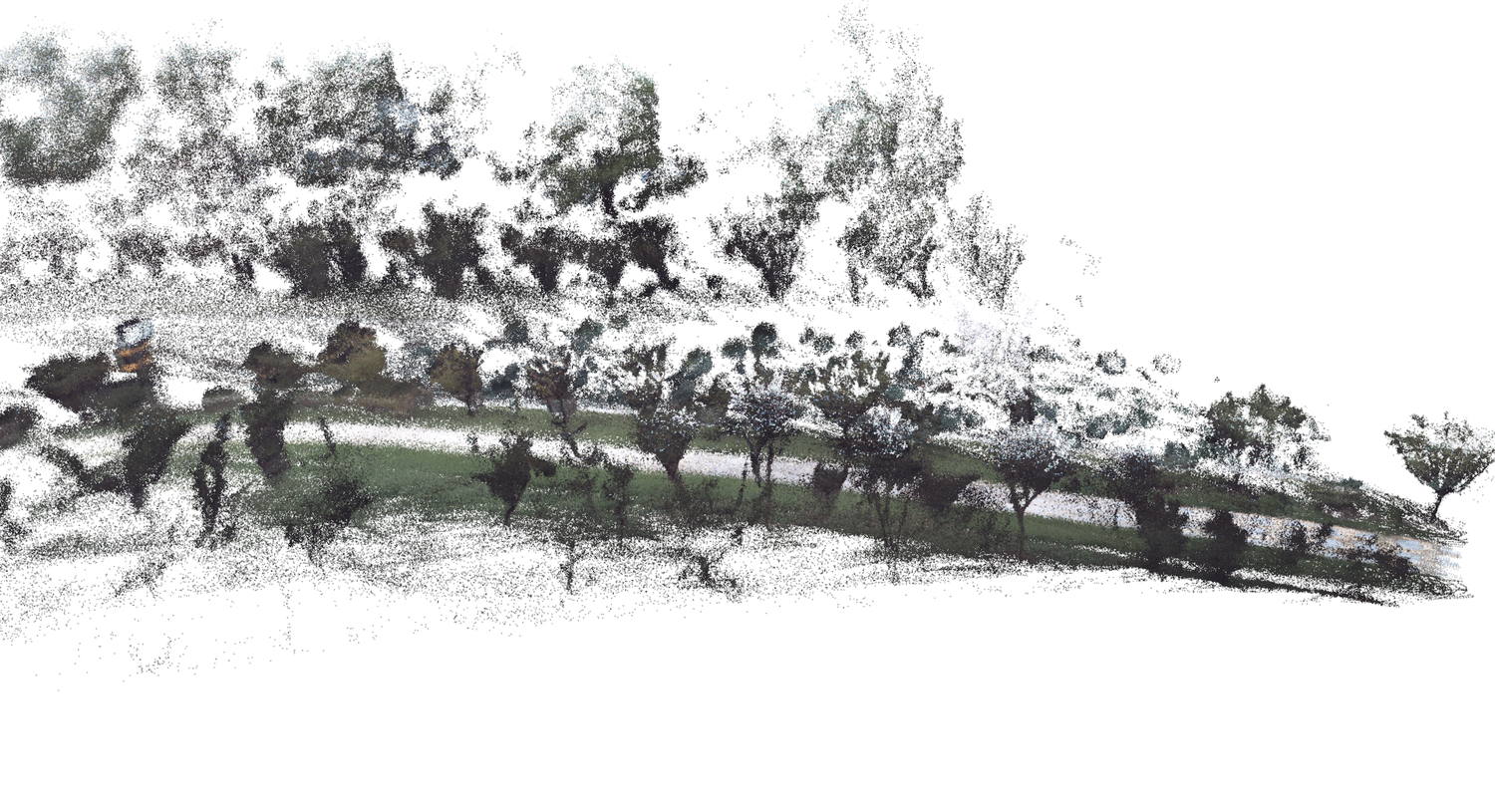}& \includegraphics[width=0.18\textwidth,height=0.09\textheight]{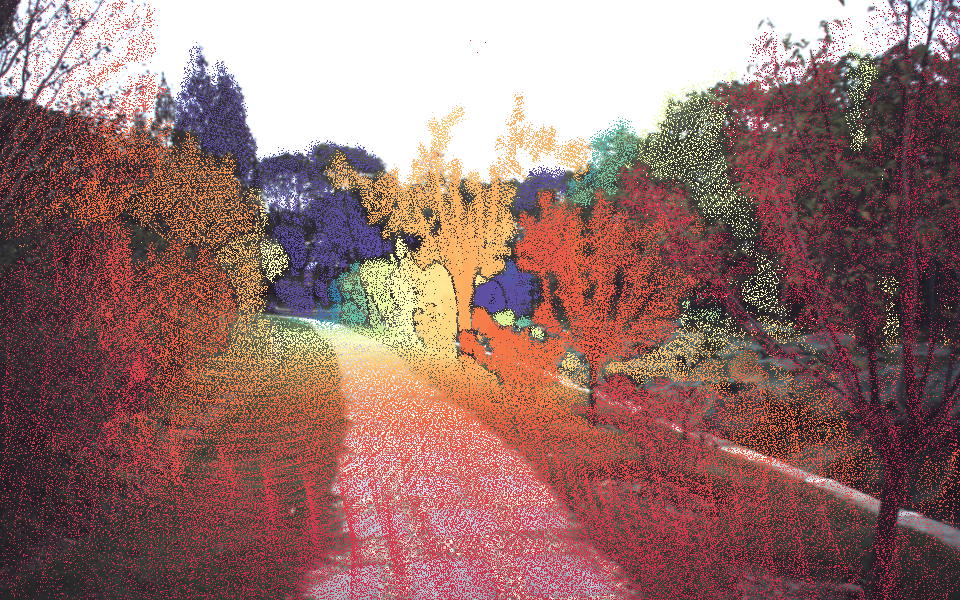}& \includegraphics[width=0.18\textwidth,height=0.09\textheight]{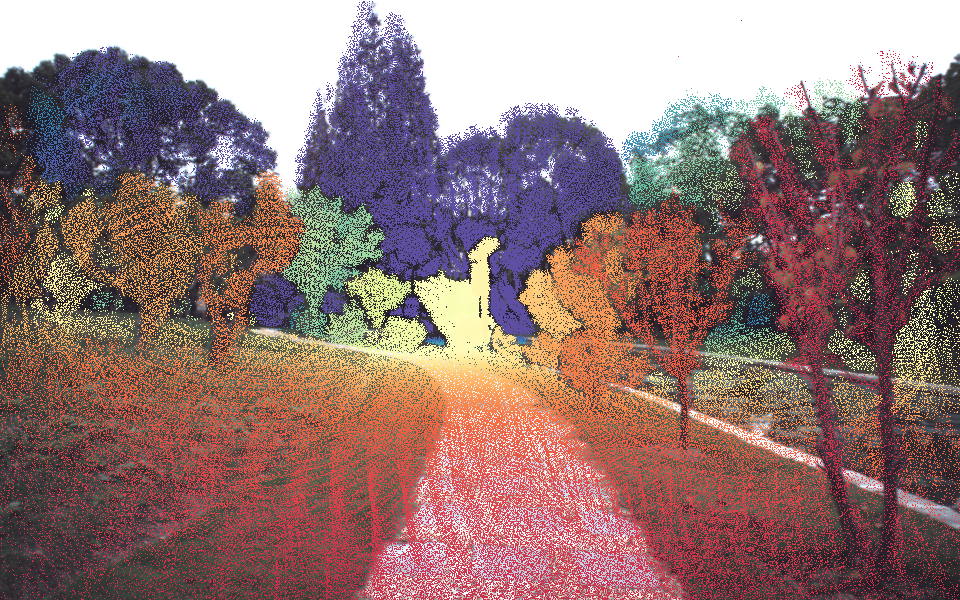} \\ 
\shortstack{\textbf{1005-07}}& \includegraphics[width=0.18\textwidth,height=0.09\textheight]{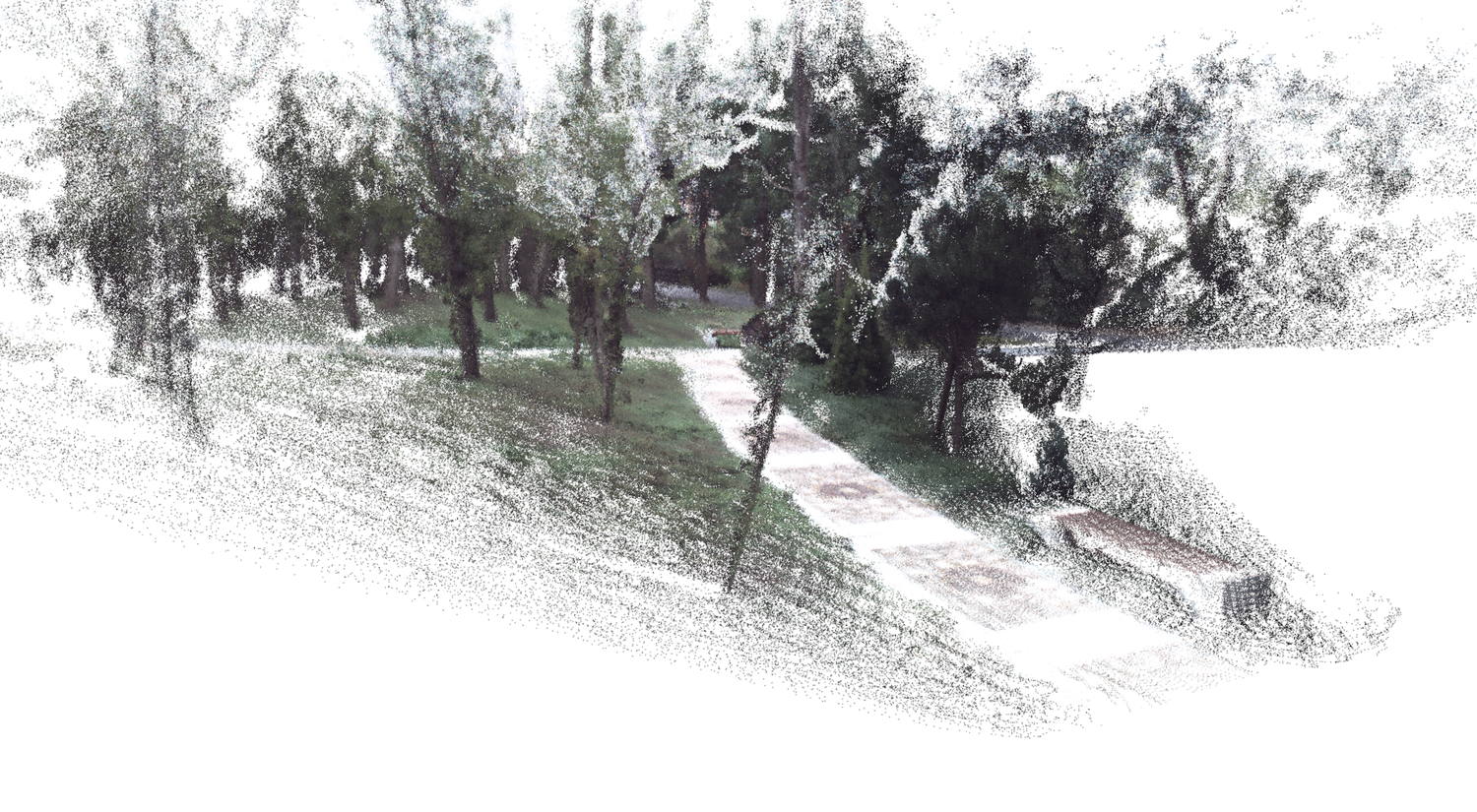}& \includegraphics[width=0.18\textwidth,height=0.09\textheight]{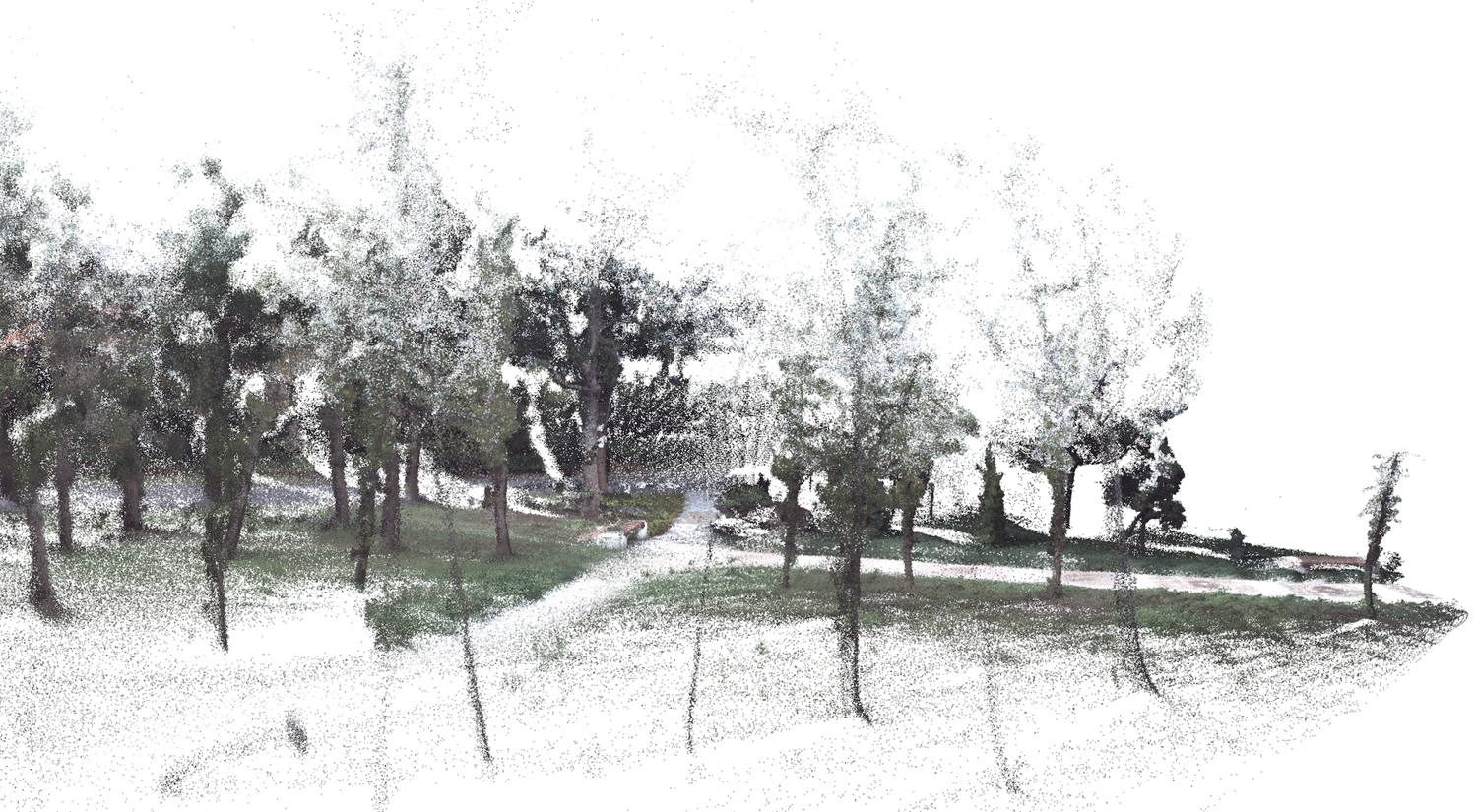}& \includegraphics[width=0.18\textwidth,height=0.09\textheight]{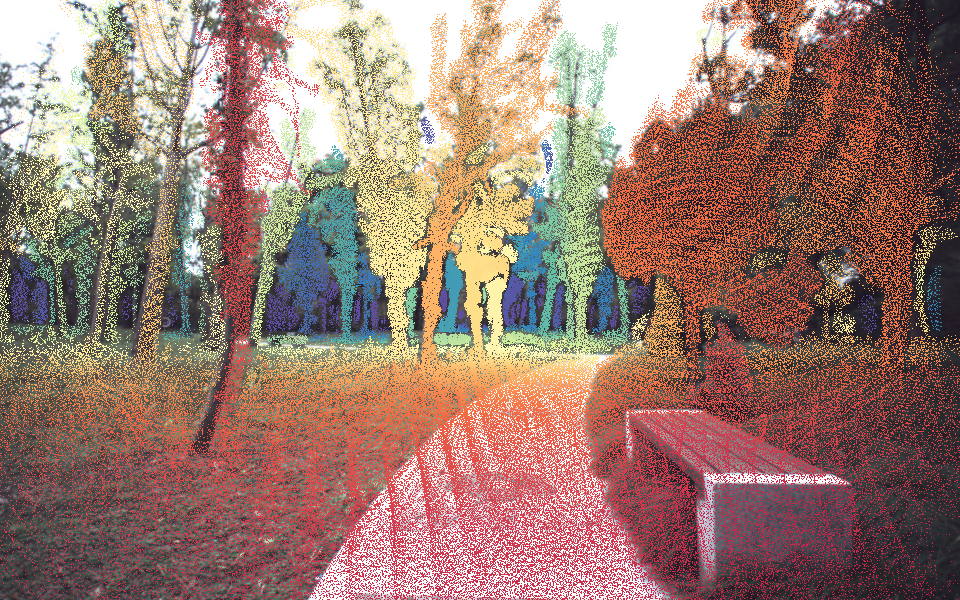}& \includegraphics[width=0.18\textwidth,height=0.09\textheight]{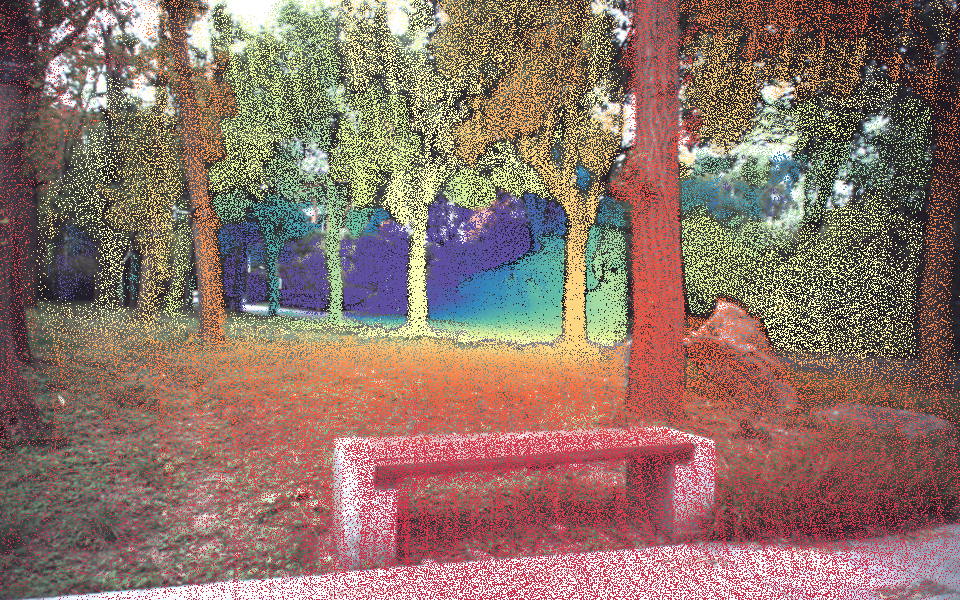} \\ 
\shortstack{\textbf{1008-03}}& \includegraphics[width=0.18\textwidth,height=0.09\textheight]{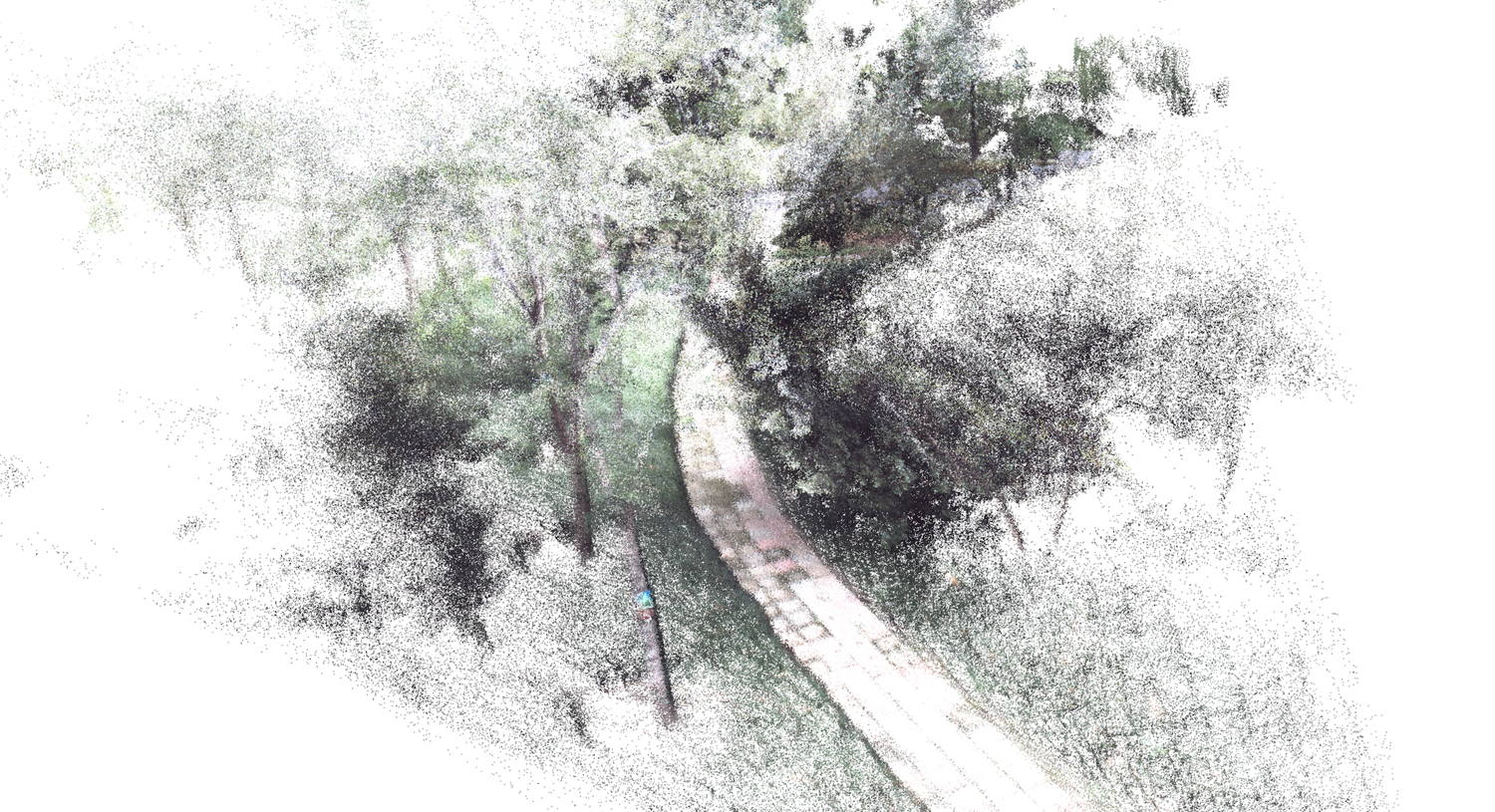}& \includegraphics[width=0.18\textwidth,height=0.09\textheight]{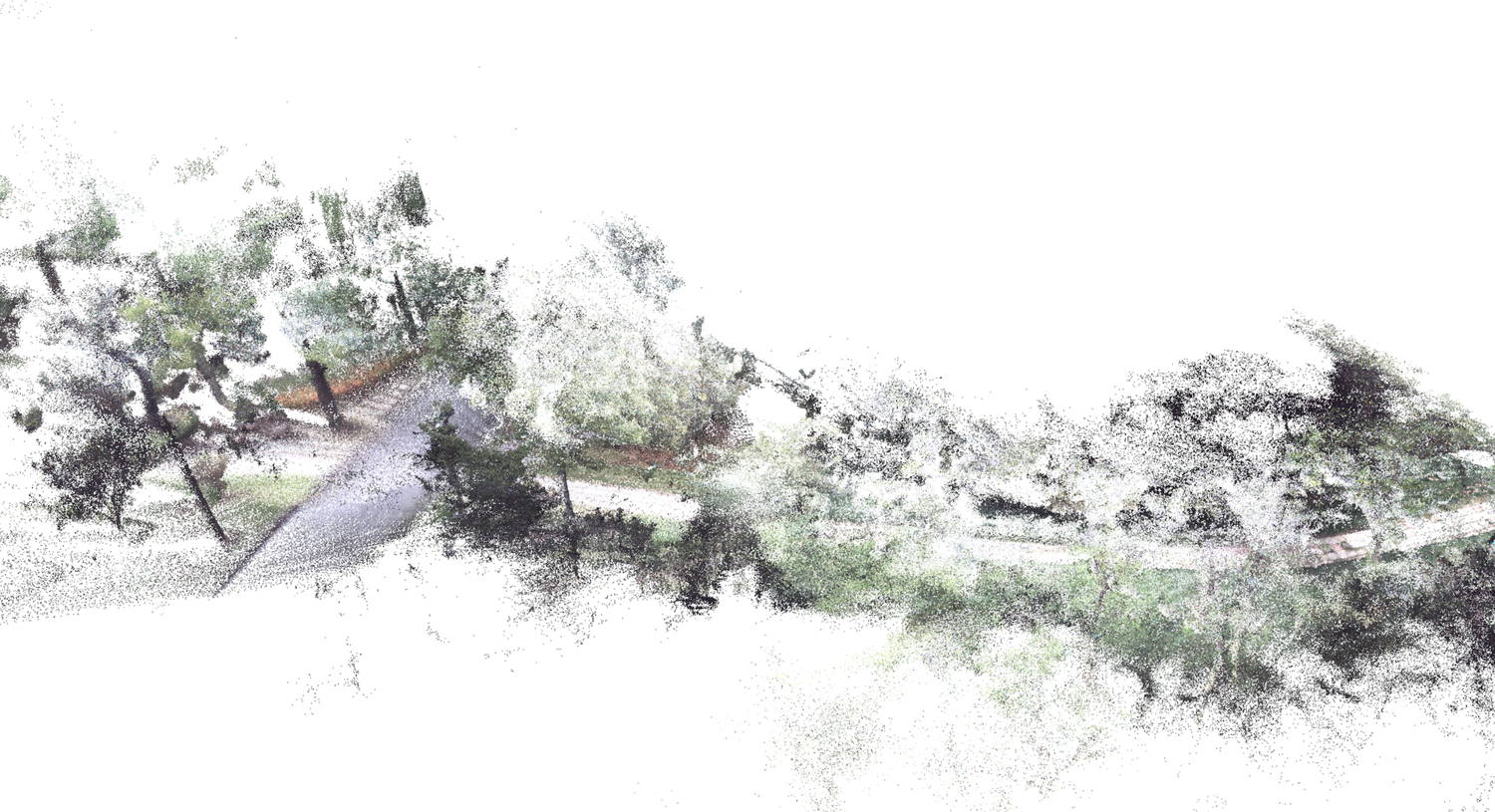}& \includegraphics[width=0.18\textwidth,height=0.09\textheight]{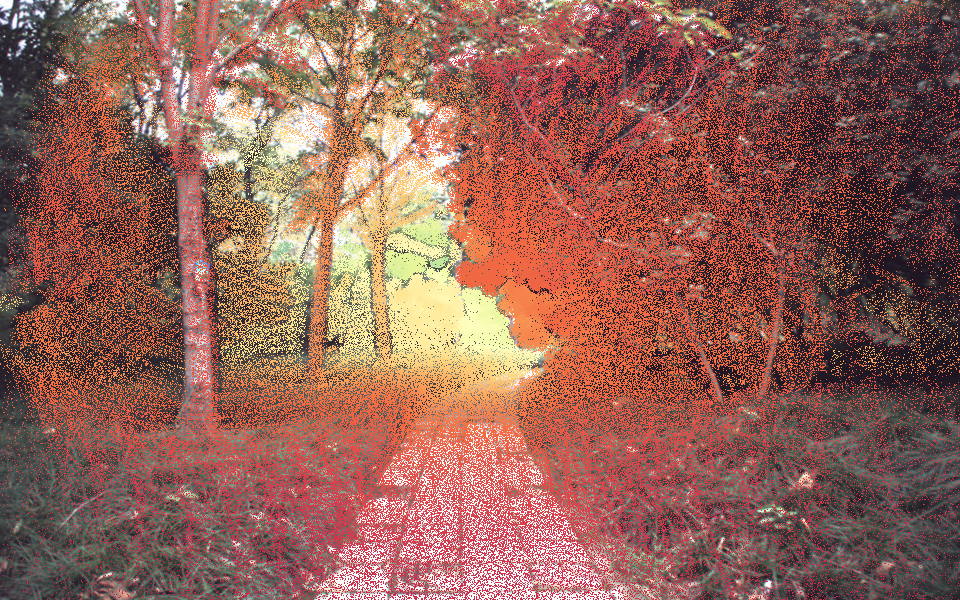}& \includegraphics[width=0.18\textwidth,height=0.09\textheight]{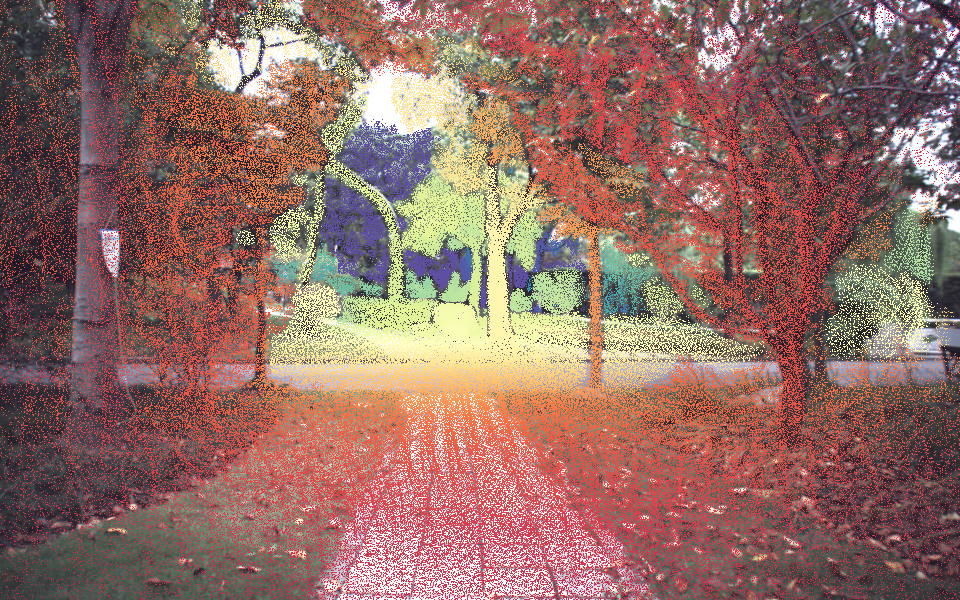} \\ 
\shortstack{\textbf{1018-13}}& \includegraphics[width=0.18\textwidth,height=0.09\textheight]{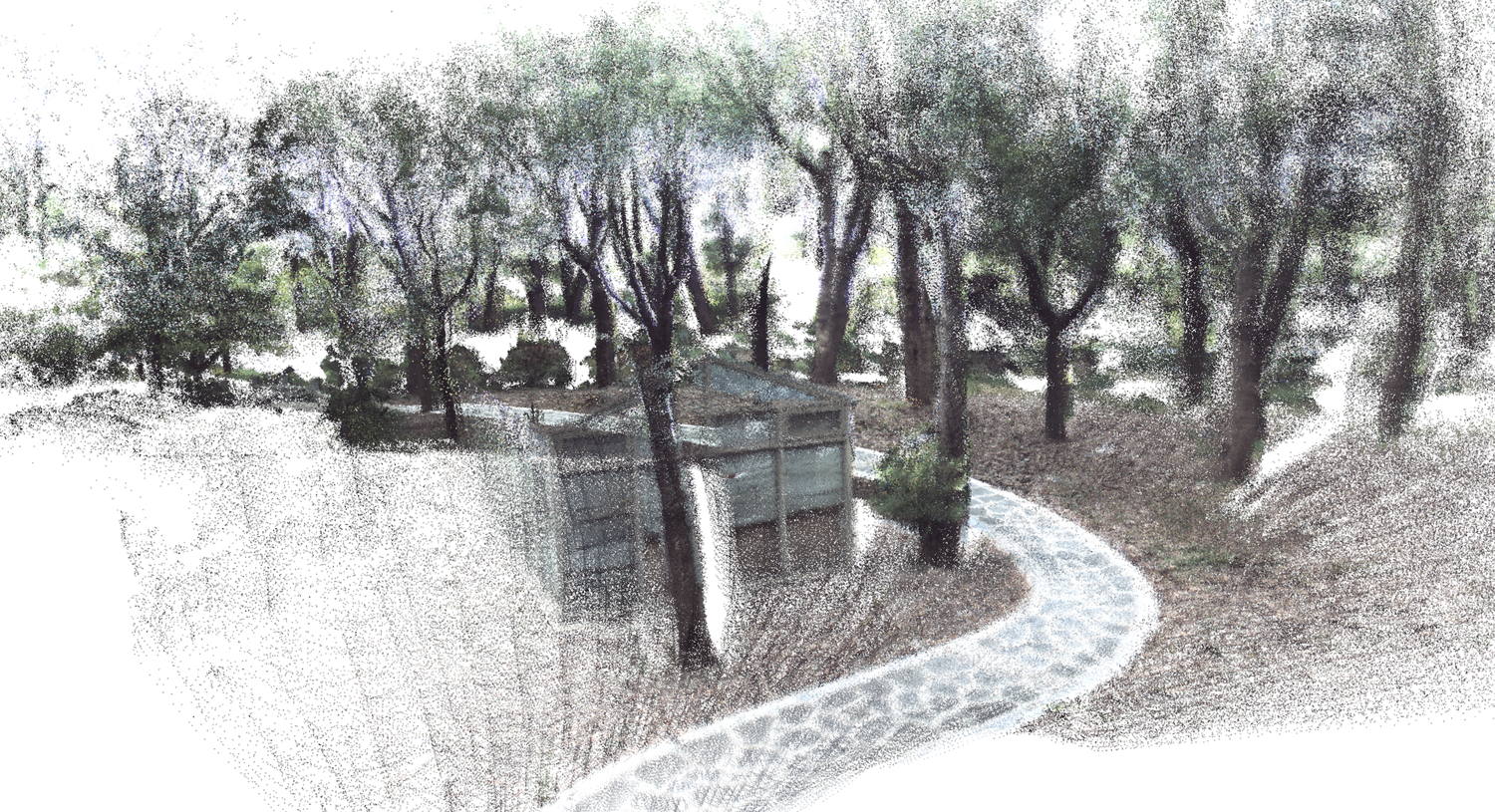}& \includegraphics[width=0.18\textwidth,height=0.09\textheight]{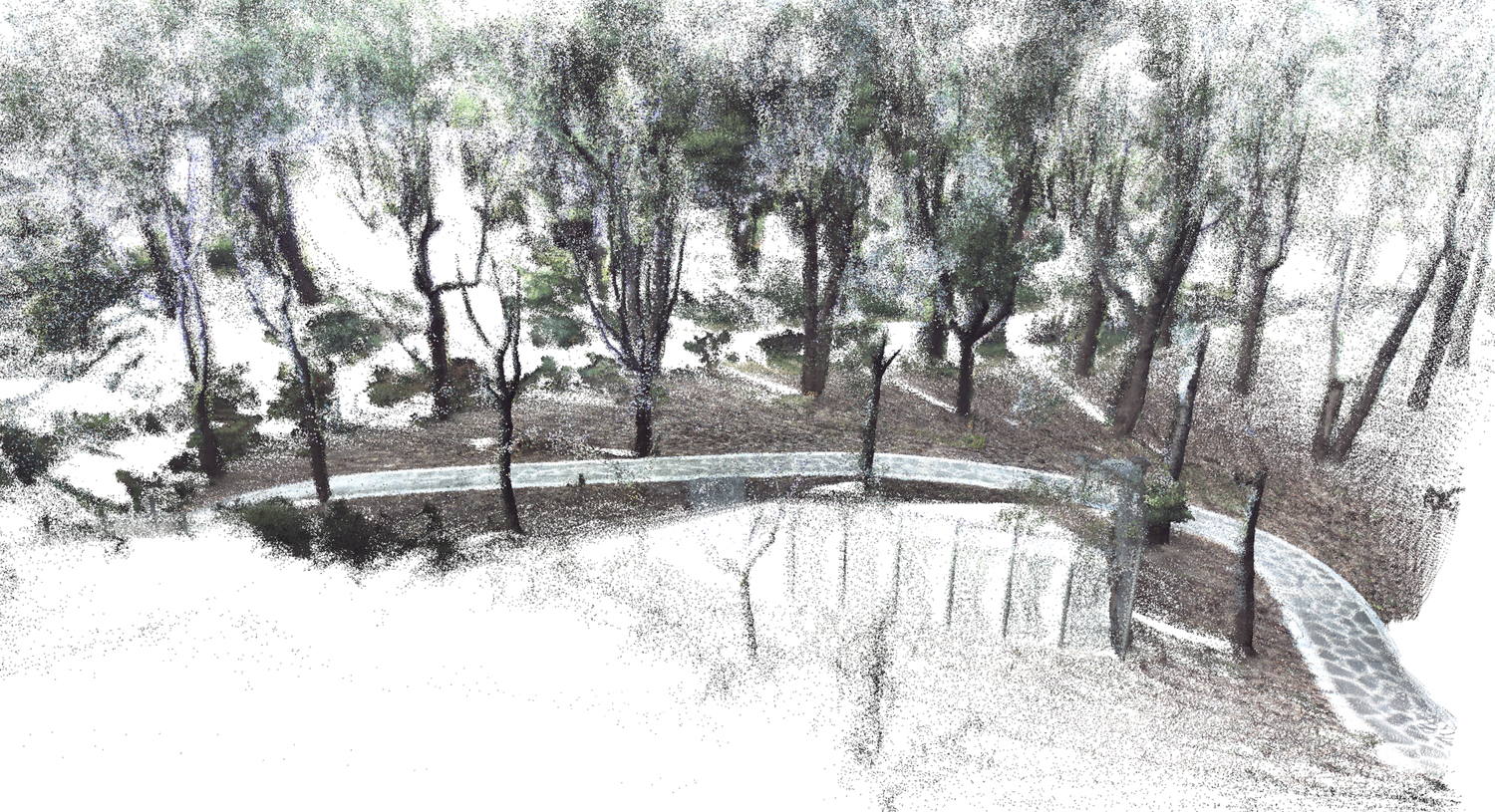}& \includegraphics[width=0.18\textwidth,height=0.09\textheight]{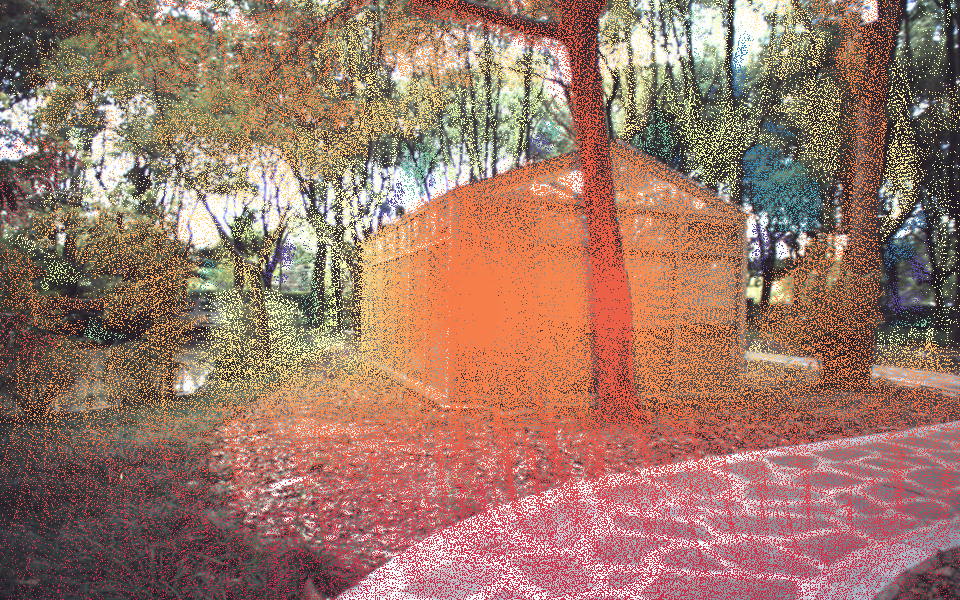}& \includegraphics[width=0.18\textwidth,height=0.09\textheight]{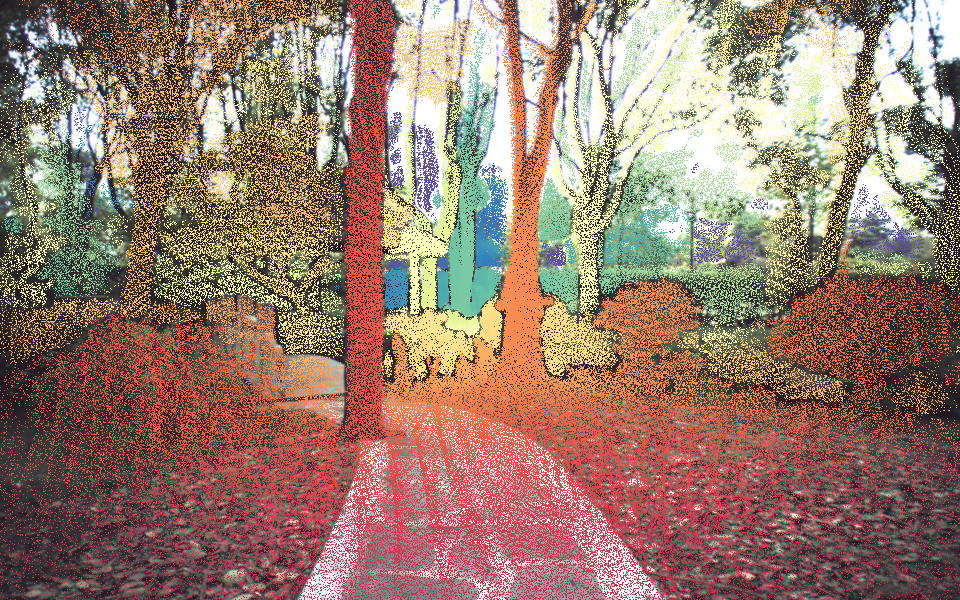} \\ 
\end{tabular}
\endgroup
\caption{Reference outputs for the ten BotanicGarden benchmark sequences.
Columns 1--2 show two 3D views of the colorized reference point clouds; columns 3--4 show
two colorized reference depth.}
\label{fig:supp-reference-point-cloud-and-depth-examples}
\end{figure*}

\endgroup

\clearpage
\end{document}